\documentclass{article}
\usepackage{paperstyle}
\usepackage{times}

\usepackage[utf8]{inputenc}
\usepackage[T1]{fontenc}
\usepackage{float}
\usepackage{hyperref}
\usepackage{url}
\usepackage{booktabs}
\usepackage{amsfonts}
\usepackage{microtype}
\usepackage{xcolor}
\usepackage{colortbl}
\usepackage{tabularx}
\usepackage{amsmath}
\usepackage{amssymb}
\usepackage{cleveref}
\usepackage{wrapfig}
\usepackage{graphicx,xspace}
\usepackage{enumitem}
\usepackage{bm}
\usepackage{tikz}

\def\vn{{\bm{n}}}
\def\vx{{\bm{x}}}
\def\vy{{\bm{y}}}
\def\vz{{\bm{z}}}
\newcommand{\E}{{\mathbb E}}
\newcommand{\Ncal}{{\mathcal N}}
\newcommand{\Id}{{\mathrm{Id}}}
\newcommand{\dd}{\texttt{d}}
\renewcommand{\eqref}[1]{\textup{(\ref{#1})}}

\newcommand{\BAMFT}{\textup{BAM}\kern-0.05em\raisebox{1.5ex}{\scalebox{0.75}{$\star$}}\xspace}
\newcommand{\RAMFT}{\textup{RAM}\kern-0.05em\raisebox{1.5ex}{\scalebox{0.75}{$\star$}}\xspace}

\definecolor{BAMRow}{RGB}{246,231,233}

\newlength{\BAMPaperImageWidth}

\newcommand{\BAMPaperMissing}{%
  \begin{minipage}[c][\BAMPaperImageWidth][c]{\BAMPaperImageWidth}\centering --\end{minipage}%
}

\newcommand{\BAMPaperZoom}[3]{%
  \begin{tikzpicture}[x=\BAMPaperImageWidth,y=\BAMPaperImageWidth,baseline=(current bounding box.center)]
    \node[anchor=south west,inner sep=0] at (0,0) {\includegraphics[width=\BAMPaperImageWidth,height=\BAMPaperImageWidth]{#1}};
    \draw[yellow,line width=.45pt] (#2,#3) rectangle ({#2+.25},{#3+.125});
    \begin{scope}
      \clip (0,-.54) rectangle (1,-.04);
      \node[anchor=south west,inner sep=0] at ({-4*#2},{-.54-4*#3}) {\includegraphics[width=4\BAMPaperImageWidth,height=4\BAMPaperImageWidth]{#1}};
    \end{scope}
    \draw[yellow,line width=.45pt] (0,-.54) rectangle (1,-.04);
    \pgfresetboundingbox
    \useasboundingbox (0,-.54) rectangle (1,1);
  \end{tikzpicture}%
}

\title{BAM! Bayesian Anything Model: a foundation model for generative computational imaging}

\author{%
  Alessio~Spagnoletti\textsuperscript{1,$*$} \quad
  Charlesquin~Kemajou~Mbakam\textsuperscript{2,$*$} \quad
  Jonathan~Spence\textsuperscript{2} \quad
  Andrés~Almansa\textsuperscript{1} \quad
  Marcelo~Pereyra\textsuperscript{2} \\[1mm]
  {\normalfont\footnotesize
  \textsuperscript{1}Laboratoire MAP5, UMR 8145, Université Paris Cité, CNRS \quad
  \textsuperscript{2}Heriot-Watt University, School of Mathematical and Computer Sciences \& Maxwell Institute for Mathematical Sciences \quad
  \textsuperscript{$*$}Equal contribution\endgraf}
}

\begin{document}
\maketitle

\begin{abstract}
Generative models are transforming Bayesian computational imaging, yet the field still lacks physics-aware foundation models. Current practice falls into two camps. Large foundation image models are deployed as plug-and-play priors with zero-shot approximate likelihood guidance, which introduces significant bias and computational cost. Physics-aware generative models avoid this bias, but each is tied to a specific dataset, task and instrument. We introduce BAM (Bayesian Anything Model), a lightweight foundation model for few-step, physics-aware posterior sampling that generalises robustly to unseen data and tasks, zero-shot or with minimal finetuning. BAM upgrades the operator-conditioned Reconstruct Anything Model (RAM) backbone \citep{Terris2025ReconstructAM} into a conditional flow map, so instrument physics is specified at inference time rather than fixed during training. BAM has just 36M parameters and is pre-trained jointly on large image corpora and libraries of forward operators. A single network then draws posterior samples in a few steps, with no likelihood approximation and no guidance weights to tune. Across linear inverse problems on FFHQ, AFHQ, LSUN, DIV2K and the K\"ohler camera-shake benchmark, BAM outperforms in just $3$ steps both specialised models and leading zero-shot methods in sample quality, at a fraction of their computational cost. BAM gives the community an accessible entry point to generative computational imaging, lowers the economic and environmental cost of training imaging models, and opens a new path for research on physics-aware Bayesian computational imaging. Official page: \url{https://bayesian-anything-model.github.io/}
\end{abstract}

\begin{figure}[H]
  \centering
  \includegraphics[width=\linewidth]{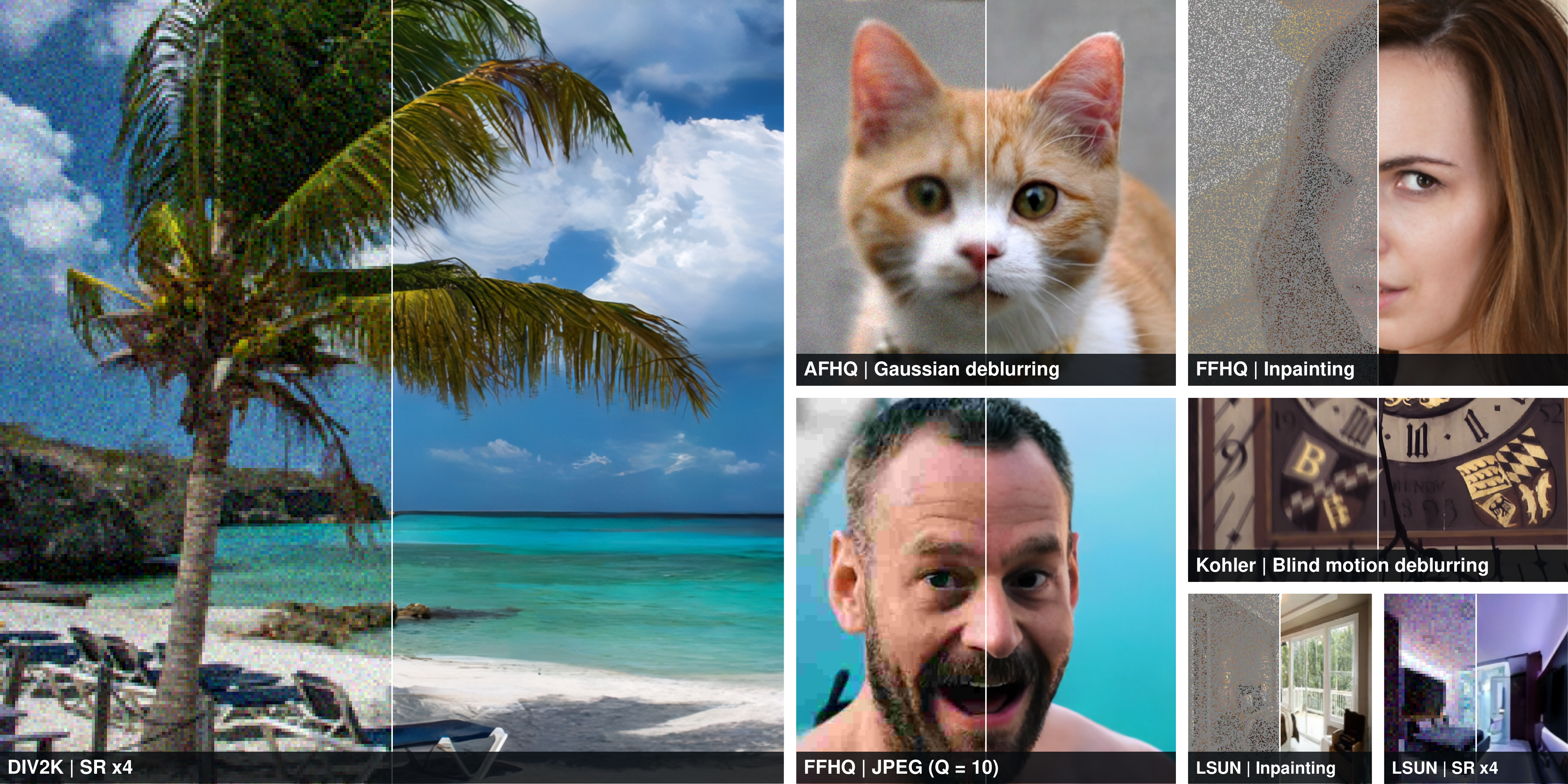}
\caption{\textbf{BAM! One small network, many imaging problems, few steps.}
Each tile pairs an observation (left) with a BAM! posterior sample (right).
A single 36M-parameter network, with the forward operator supplied at inference time, covers
$\times4$ super-resolution (DIV2K, LSUN), Gaussian deblurring (AFHQ), inpainting (FFHQ, LSUN), non-linear JPEG restoration at $Q=10$ (FFHQ) and blind motion deblurring (K\"ohler), all in $3$ steps. Settings are given in Section~\ref{sec:experimental-setup}.}
  \label{fig:teaser}
\end{figure}

\section{Introduction}
We consider imaging problems involving an unknown image $x^\star \in \mathbb R^d$ and a measurement $y \in \mathbb R^m$, related through the observation model $y = A x^\star + \sigma_y w$, where $w$ is additive Gaussian noise and the forward operator $A \in \mathbb R^{m \times d}$ and the noise level $\sigma_y > 0$ are known at inference time. We are interested in problems where estimating $x^\star$ from $y$ is ill-posed or ill-conditioned. Additional information is then required to reduce the uncertainty about $x^\star$ and deliver meaningful solutions. Within the Bayesian framework, this is achieved by modelling $x^\star$ as a realization of $\bm x \sim p(\bm x)$, the so-called prior distribution, and $y$ as a realisation from $p(\vy\mid x^\star)$. Prior and observed information are then combined in the posterior distribution $p(\bm x \mid y, A, \sigma_y) = p(y \mid \bm x, A, \sigma_y)\, p(\bm x) / p(y \mid A, \sigma_y)$. This posterior underpins Bayesian inference about $\bm x$, from point estimators such as the posterior mean to uncertainty quantification through posterior variances and credible regions, all of which can be approximated from posterior samples.

\begin{wrapfigure}{r}{0.37\textwidth}
    \centering
    \vspace{-0.6em}
    \includegraphics[width=\linewidth]{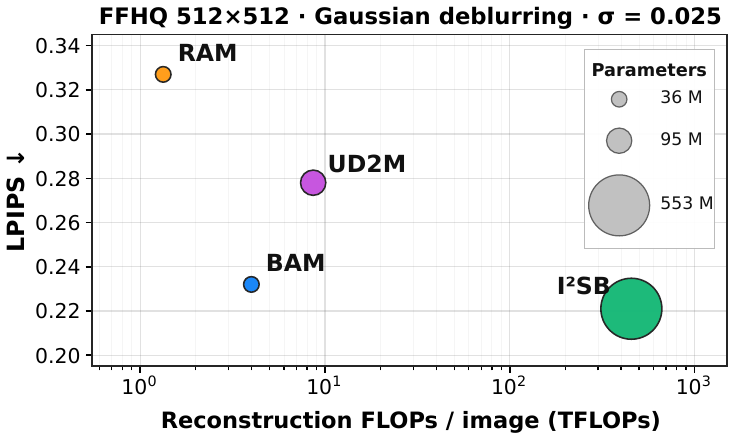}
    \caption{Pareto frontier drawn by SOTA methods compared to BAM.}
    \label{fig:pareto}
    \vspace{-0.8em}
\end{wrapfigure}

State-of-the-art Bayesian imaging methods learn from representative data via plug-and-play (PnP) or end-to-end strategies. PnP methods plug a pre-trained denoiser or generative prior into an iterative scheme that enforces the likelihood~\citep{Venkatakrishnan2013a,romano2017little,Kamilov2023PnP}, and zero-shot posterior samplers extend this idea beyond point
estimates~\citep{laumont2022pnpula,kawar2022denoising,chung2023dps}. With modern diffusion model (DM) priors~\citep{Ho2020Denoising,song2021sde}, this flexibility carries three costs: a biased likelihood approximation with tunable guidance weights, hundreds of neural function evaluations (NFEs), and priors that, even in latent space,
reach billions of parameters~\citep{Podell2023SDXLIL}. Distilled consistency and flow-map models~\citep{song2023consistency,Luo2023LatentCM,boffi2025selfdistill} cut sampling to a few NFEs, and zero-shot solvers built on distilled latent models now set the state of the art~\citep{Garber_2025_CVPR,Spagnoletti_2025_ICCV,spagnoletti2026consistencyregularisedgradientflows},
but the biased likelihood and large prior remain. End-to-end methods instead learn the reconstruction directly, typically with physics-aware unrolled or equilibrium networks~\citep{Monga2021Unrolling,Gilton2021DEQ}
trained per operator, often by fine-tuning a pre-trained denoiser, returning point estimates. Conditional diffusion and bridge models sample the posterior~\citep{Liu2023I2SBIS,zhao2024cosign,Mbakam2025LearningFP} but are mostly task-specific and as costly as the underlying DM. Since $y$ is highly informative about $\bm x$, posterior sampling should need far fewer parameters than unconditional generation, yet lightweight networks that generalize across operators remain largely unexplored. A notable exception is the Reconstruct Anything Model (RAM)~\citep{Terris2025ReconstructAM}, an
operator-conditioned DRUNet that solves many linear inverse problems with one network, but returns a point estimate of $\bm x$ not posterior samples.

We introduce the Bayesian Anything Model (BAM), a lightweight foundation model for few-step, physics-aware posterior sampling that bridges these two camps. BAM is a conditional flow map that transports a Gaussian reference to the posterior $p(\bm x \mid y, A, \sigma_y)$, with the instrument physics specified at inference time rather than fixed during training. Pre-trained jointly on large image corpora and libraries of forward operators, a single model draws posterior samples in a few steps, with no likelihood approximation and no guidance weights to tune. Our contributions are as follows:
\begin{itemize}[leftmargin=*,itemsep=2pt,topsep=3pt]
    \item \textbf{A flow-map training paradigm.} We propose a recipe that upgrades an unfolded reconstruction network into such a flow map, via Lagrangian self-distillation on a normalized stochastic interpolant of the measurement.
    \item \textbf{A lightweight foundation model.} Built on the 36M-parameter RAM backbone, a UNet with unrolled physics-aware updates, BAM learns this map across operators, noise levels and datasets, delivering high-quality posterior samples (\Cref{fig:teaser}) and generalizing to new data and tasks.
    \item \textbf{State-of-the-art quality at a fraction of the cost.} On FFHQ, AFHQ, LSUN, DIV2K and the K\"ohler benchmark, BAM outperforms specialized models and leading zero-shot methods in sample quality in just three steps, advancing the quality--cost Pareto frontier (\Cref{fig:pareto}).
\end{itemize}
BAM lowers the barrier to generative computational imaging, cuts the economic and environmental cost of training imaging models, and opens a new direction for physics-aware Bayesian imaging.

\newpage
\section{Background}\label{sec:background}
\paragraph{Diffusion models.}
DMs generate samples from $\rho_0$ by learning to reverse a
process that gradually corrupts data with Gaussian noise. The noising process is
$\vx_t=\alpha_t\vx_0+\sigma_t\vz$, where $\vx_0\sim\rho_0$ and
$\vz\sim\Ncal(0,\sigma_d^2\Id)$ are independent, and the schedules satisfy
$\alpha_0=\sigma_1=1$ and $\alpha_1=\sigma_0=0$. Each marginal $\rho_t$, the law
of $\vx_t$, is therefore a rescaled, Gaussian-smoothed version of the data
distribution. Sampling reverses this process through the probability-flow
ODE~\citep{song2021sde}
\begin{equation}\label{eq:grad-flow}
  \dot{x}_t = f_t x_t - \tfrac{1}{2} g_t^2 \nabla \log \rho_t(x_t),
  \qquad
  f_t = \frac{\dot{\alpha}_t}{\alpha_t}, \quad
  g_t^2 = 2\sigma_d^2\,\alpha_t\sigma_t \frac{\dd}{\dd t}\Big(\frac{\sigma_t}{\alpha_t}\Big).
\end{equation}
Integrated backwards in time, its solutions transport $\rho_1$ onto $\rho_t$ for
every $t\in[0,1]$. The associated integrated flow $X_{t,0}$ sends each state to
its endpoint at $t=0$, so that $(X_{t,0})_\#\rho_t=\rho_0$. DMs learn only the
score $\nabla\log\rho_t$, which makes them easy to train but slow to sample. A
standard DM solver needs hundreds of NFEs to
integrate~\eqref{eq:grad-flow} for each sample.

\paragraph{Flow matching.}
Flow matching (FM) learns the velocity field of the probability-flow ODE
directly, rather than through the
score~\citep{lipman2023flow,liu2023flow,albergo2023stochastic}. The drift that
transports the marginals $(\rho_t)_{t\in[0,1]}$ is the average velocity of the
noising paths through each state,
\begin{equation}\label{eq:background-fm-drift}
  b_t(x)=\E\big[\dot\alpha_t\vx_0+\dot\sigma_t\vz \,\big|\, \vx_t=x\big],
\end{equation}
so the probability flow reads $\dot x_t=b_t(x_t)$. The drift $b_t$ minimises
$\E\|v(\vx_t)-\dot\alpha_t\vx_0-\dot\sigma_t\vz\|^2$ over functions $v$, so it
can be learned by least-squares regression on $(\vx_0,\vz)$ pairs without
simulating ODE trajectories. FM typically uses
$\alpha_t=1-t$ and $\sigma_t=t$, for which
$b_t(x)=\E[\vz-\vx_0\mid\vx_t=x]$. By Tweedie's formula,
$b_t(x)=f_t x-\tfrac12 g_t^2\nabla\log\rho_t(x)$, the r.h.s.
of~\eqref{eq:grad-flow}. So FM and DMs learn the same ODE and differ only in
parametrization, loss weighting and choice of solver~\citep{gao2025diffusion}.

\paragraph{Consistency models.}
Consistency models (CMs) dramatically accelerate sampling by learning the
integrated flow $X_{t,0}$ directly, mapping a noisy state to the endpoint of its
trajectory in a single network evaluation. Training enforces two properties of
this map. The first is \emph{self-consistency}: all points on the same
trajectory of~\eqref{eq:grad-flow} share an endpoint, so
$X_{t,0}(x_t)=X_{s,0}(x_s)$. The second is the boundary condition
$X_{0,0}(x)=x$. CMs can be distilled from a pre-trained DM or trained directly
from data~\citep{song2023consistency,songdhariwal2023improved,boffi2025selfdistill}.

\paragraph{Flow maps.}
Flow maps generalize CMs to jumps between any two
times~\citep{boffi2025selfdistill}. The flow map $X_{t,s}$ moves a point along
its trajectory of~\eqref{eq:grad-flow} from time $t$ to time $s\le t$, i.e.\
$X_{t,s}(x_t)=x_s$. Thus $X_{t,t}(x)=x$, $X_{t,0}$ is the CM map, and $X_{1,0}$
transports pure noise to data. The parametrization
\begin{equation}\label{eq:background-affine-flow-map}
  X_{t,s}(x)=x+(s-t)\,v_{t,s}(x)
\end{equation}
enforces $X_{t,t}(x)=x$ by construction, and $v_{t,s}$ is the average velocity
along the trajectory between $t$ and $s$. On the so-called diagonal ($t = s$), it reduces to the FM
drift:
\begin{equation}\label{eq:background-tangent-condition}
  \partial_s X_{t,s}(x)\big|_{s=t}=v_{t,t}(x)=b_t(x).
\end{equation}
Under mild regularity, flow maps verify the following
characterizations~\citep[Prop.~2.2]{boffi2025selfdistill}:
\begin{equation}\label{eq:flowmap-characterizations}
  \underbrace{\partial_s X_{t,s}(x)=b_s\big(X_{t,s}(x)\big)}_{\text{Lagrangian}},
  \qquad
  \underbrace{\partial_t X_{t,s}(x)+DX_{t,s}(x)\,b_t(x)=0}_{\text{Eulerian}},
  \qquad
  \underbrace{X_{u,s}\circ X_{t,u}=X_{t,s}}_{\text{semigroup}}.
\end{equation}
for $s\le u\le t$, where $DX_{t,s}$ denotes the Jacobian. The three characterizations are equivalent, as each determines the flow map uniquely. With $s=0$, the
semigroup property is exactly CM self-consistency, and the Eulerian equation is
its infinitesimal form. The Lagrangian form avoids the Jacobian, which makes it
cheaper and more stable to train.

For a network
$X^\theta_{t,s}(x)=x+(s-t)\,v^\theta_{t,s}(x)$, the tangent and Lagrangian
conditions give two losses:
\begin{equation}\label{eq:background-losses}
  \mathcal L_b(\theta)=\E\big\|v^\theta_{\bm t,\bm t}(\vx_{\bm t})-\dot\alpha_{\bm t}\vx_0-\dot\sigma_{\bm t}\vz\big\|^2,
  \qquad
  \mathcal L_{\mathrm{LSD}}(\theta)=\E\big\|\partial_s X^\theta_{\bm t,\bm s}(\vx_{\bm t})-v^{\theta^-}_{\bm s,\bm s}\big(X^\theta_{\bm t,\bm s}(\vx_{\bm t})\big)\big\|^2.
\end{equation}
The times $\bm s\le\bm t$ are drawn at random, and $\theta^-$ is an exponential
moving average (EMA) of $\theta$ that serves as a detached teacher. $\mathcal L_b$
is the FM loss on the diagonal, whereas the Lagrangian self-distillation loss
$\mathcal L_{\mathrm{LSD}}$ propagates that drift to finite jumps.

We use this formulation to train BAM, our lightweight foundation model for few-step, physics-aware posterior sampling, which transfers to unseen data and tasks zero-shot or with minimal finetuning.

\newpage
\section{BAM! Bayesian Anything Model}\label{sec:method}
\paragraph{Model.} BAM is designed for imaging problems of the form $y = A x^\star + \sigma_y w$, where we assume that the unknown image $x^\star$ is a realization of a r.v. $\bm x \sim p(\bm x)$, the additive noise $w$ is a realization of $\bm w \sim \mathcal N(\bm 0, \mathrm{Id})$, and the forward operator $A \in \mathbb R^{m \times d}$ and noise level $\sigma_y > 0$ are known at inference time. The goal is to draw accurate samples from the posterior $p(\bm x \mid y, A, \sigma_y)$ in a small number of network evaluations. Moreover, rather than train a separate model for each problem or narrow problem class, BAM amortizes over a wide family of operators and noise levels encountered in practice, which are randomized during training. A single model therefore serves the entire family, requiring no retraining, or at most light finetuning, when the forward model changes. Finally, following \citet{Terris2025ReconstructAM}, BAM is deliberately lightweight, compatible with modest local hardware.

To this end, BAM is built as a lightweight few-step conditional flow map targeting the posterior $p(\bm x \mid y, A, \sigma_y)$. Feeding $(y, \mathcal{A}, \sigma_y, t, s)$ directly as inputs to a small network is numerically fragile, as $\sigma_y$ and $(t,s)$ span several orders of magnitude while the spectrum of $\mathcal{A}$ is fixed, so the network receives inputs whose relative scales vary inconsistently across the training range. We therefore condition on a rescaled measurement, defined through the stochastic interpolant
\begin{equation}\label{eq:measurement-interpolant}
    \bm y_\sigma = \alpha_\sigma A \bm x + \varsigma_\sigma \bm w,
\end{equation}
with schedule $\alpha_\sigma$ and $\varsigma_\sigma$, respectively decreasing and increasing in $\sigma$, satisfying $\varsigma_\sigma / \alpha_\sigma = \sigma$. Hence $y$ is a realization of $\bm y_\sigma$ at $\sigma = \sigma_y$, up to the known factor $\alpha_\sigma$, and we condition on $ y_\sigma := \alpha_\sigma y$.

Using the linear interpolant of \cref{sec:background}, we construct a flow map to transport $\rho_1 = \mathcal N(\bm 0, \sigma_d^2 \Id)$ to $\rho_0 = p(\bm x \mid y, A, \sigma_y)$. We parametrize this flow map as
\begin{equation}\label{eq:bam-map}
    \bm X^\theta_{t,s}(x \mid y_\sigma, \alpha_\sigma A) = x + (s-t)\, \bm v_\theta(x, t, s, y_\sigma, \alpha_\sigma A),
    \qquad 0 \le s \le t \le 1,
\end{equation}
so that the identity boundary condition $\bm X^\theta_{t,t} (x \mid y_\sigma, \alpha_\sigma A)= x$ holds $\forall x \in \mathbb{R}^d$ by construction. The velocity must then satisfy two conditions. On the diagonal, the tangent condition requires $\bm v_\theta(x, t, t, y_\sigma, \alpha_\sigma A) = \bm b_t(x \mid y_\sigma, \alpha_\sigma A)$, the conditional probability-flow drift. Off the diagonal, backwards jumps from $t$ to $s$ are governed by the Lagrangian condition of \cref{sec:background}, applied conditionally for posterior sampling. Together these two conditions characterize the conditional flow map and lead to BAM's two main training objectives. Note that $\sigma_y$ is not supplied to the network in \eqref{eq:bam-map}. It enters only through the rescaling \eqref{eq:measurement-interpolant}, so $\bm v_\theta$ acts partially blindly w.r.t. $\sigma_y$. Since $\sigma_y$ is easily estimated from $y$, dropping the explicit conditioning costs little and removes an embedding pathway, keeping the architecture lightweight.

The network $\bm v_\theta$ can be realized by any physics-aware architecture for linear imaging problems, notably deep unfolding architectures, by conditioning on the augmented measurement $[ x_t\,;\, y_\sigma]$ and the stacked forward operator $[(1-t)\mathrm{Id}\,;\, \alpha_\sigma A]$ supplied at inference time. For BAM we adopt a RAM backbone, which embeds the unfolding philosophy in a DRUNet, and assign its two noise-level embeddings to input $t$ and $s$ (see \cref{app:architecture,app:ablation-architecture} for architecture details and ablations).

\paragraph{Training.} As mentioned previously, BAM relies on the following two flow map training objectives. On the diagonal ($t=s$), the velocity is fitted to the interpolant slope by minimizing
\begin{equation}\label{eq:loss-diag}
    \mathcal L_b(\theta) = \mathbb E\Big[\big\| \bm v_\theta(\bm x_t, \bm t, \bm t, \bm y_\sigma, \bm A) - (\bm z - \bm x_0) \big\|_2^2\Big].
\end{equation}
Off diagonal ($s < t$), holding the flow's starting point $(x_t, t)$ and the conditioning $(y_\sigma, \sigma A)$ fixed, the endpoint derivative of \eqref{eq:bam-map} is
\begin{equation}\label{eq:endpoint-derivative}
    \partial_s \bm X^\theta_{t,s} = \bm v_\theta(x_t, t, s, y_\sigma, \alpha_\sigma A) + (s-t)\, \partial_s \bm v_\theta(x_t, t, s, y_\sigma, \alpha_\sigma A),
\end{equation}
which we compare with the diagonal velocity of the EMA teacher at the transported point, i.e.,
\begin{equation}\label{eq:loss-lsd}
    \mathcal L_{\mathrm{LSD}}(\theta) = \mathbb E\Big[\big\| \partial_s \bm X^\theta_{\bm t,\bm s} - \mathrm{sg}\big[\bm v_{\theta^-}(\hat{\bm x}_{\bm t,\bm s}, \bm s, \bm s, \bm y_\sigma, \bm A)\big] \big\|_2^2\Big]\,, \quad \hat{\bm x}_{\bm t, \bm s} := \bm X^\theta_{\bm t,\bm s}(\bm x_t \mid \bm y_\sigma, \bm \sigma \bm A)\,.
\end{equation}
BAM is trained on the flow objective $\mathcal L_b + \lambda \mathcal L_{\mathrm{LSD}}$, with the expectations taken over $\bm x_{\bm t} = (1-t) \bm x + t \bm z$ with $\bm x \sim p_{\bm x}$ and $\bm z \sim \mathcal N(\bm 0, \mathrm{Id})$ independent of $\bm x$, $\bm y \sim N(\bm A \bm x, \sigma^2 \mathrm{Id})$ with $\bm A \sim p_{\bm A}$ independent of $(\bm x, \bm z)$, and $\bm t \sim \mathcal{U}[0,1]$ and $\bm s|\bm t$ from \eqref{eq:endpoint-sampling}. Rather than draw $\sigma$ independently, we tie it to $\bm t$ during training through $\sigma(t) = \sigma_{\max}\gamma t / (1-(1-\gamma)t)$, which increases from $0$ to $\sigma_{\max}$ with shape controlled by $\gamma > 0$. This concentrates the capacity of the network on the regime where $\bm y_\sigma$ and $\bm x_t$ carry broadly comparable amounts of noise, a challenging regime where the flow must bend to take both $y_\sigma$ and $x_t$ into account. Note that $\mathrm{sg}$ in \eqref{eq:loss-lsd} stops gradients through the complete teacher evaluation, including its input, and $\partial_s \bm X^\theta_{t,s}$ is evaluated by an automatic-differentiation Jacobian--vector product that retains the graph needed to optimize it.

The above flow objective is paired with two auxiliary terms, targeting perceptual quality and contrast bias, which act near $s \approx 0$. More precisely, the full objective for BAM is
\begin{equation}\label{eq:total-loss}
    \mathcal L(\theta) = \mathbb E\Big[\lambda_b \ell_b + \lambda_L \ell_{\mathrm{LSD}}
    + g(s)\big\{\lambda_p \ell_{\mathrm{LPIPS}}(\hat{\bm x}_{t,s}, \bm x_0) + \lambda_c \ell_{\mathrm{ctr}}(\hat{\bm x}_{t,s}, \bm x_0)\big\}\Big],
\end{equation}
where $g(s) = \exp(-4s)$ assigns weight to the auxiliary losses only when $s$ is small, $\ell_{\mathrm{LPIPS}}$ is the squeeze-based perceptual loss of \citet{Zhang2018TheUE}, applied to the clipped output, and $\ell_{\mathrm{ctr}}$ compares intensity histograms to penalize contrast bias. The latter provides robustness when training across datasets with differing image statistics, and we set $\lambda_c = 0$ after pre-training. No adversarial loss is used at any stage. \Cref{app:training,app:ablation-adversarial,app:ablation-contrast}, detail the loss functions, their optimization and ablations.

\paragraph{Posterior sampling.} Since BAM is trained on the locus where the noise in $\bm y_\sigma$ and in $\bm x_t$ is balanced, each sampling step should pair the current iterate $\bm x_t$ with a draw of the interpolant $\bm y_\sigma$ at the matching level $\sigma$. For $\sigma > \sigma_y$, such a draw consistent with $y$ is obtained by adding the missing noise,
\begin{equation}\label{eq:inference-interpolant}
    \bm y_\sigma = \alpha_\sigma y + \sqrt{\varsigma_\sigma^2 - \alpha_\sigma^2 \sigma_y^2}\; \bm \varepsilon,
    \qquad \bm \varepsilon \sim \mathcal N(\bm 0, \mathrm{Id}),
\end{equation}
which has noise variance $\varsigma_\sigma^2$. We hold $\bm \varepsilon$ fixed across steps. Sampling follows a decreasing schedule $1 = t_1 > t_2 > t_3$, each step noising the current iterate to level $t_k$ and mapping it back to $s=0$,
\begin{align}
    \bm x_{t_k} &= (1-t_k)\hat{\bm x}^{(k-1)} + t_k \bm z^{(k)},
    \qquad \bm z^{(k)} \sim \mathcal N(\bm 0, \sigma_d^2 \mathrm{Id}), \label{eq:sampler-renoise}\\
    \hat{\bm x}^{(k)} &= \bm x_{t_k} - t_k\, \bm v_\theta(\bm x_{t_k}, t_k, 0, \bm y_{\sigma_k}, \sigma_k A), \label{eq:sampler-predict}
\end{align}
where $\sigma_k=\sigma(t_k)$ and $t_1 = 1$, so that $\bm x_{t_1} = \bm z^{(1)}$ is pure noise and \eqref{eq:sampler-renoise} requires no previous iterate.

When $\sigma_y$ is small, the intermediate level satisfies $\sigma_2 \in (\sigma_y, \sigma_1)$ and the final step is taken at $\sigma_3 = \sigma_y$, where no re-noising of the measurement is required. When $\sigma_y$ is large, ending at $\sigma_y$ would leave a long final jump and thus introduce some sampling bias. Hence, we instead set $\sigma_2 = \sigma_y$ and take the last step at $\sigma_3 < \sigma_y$, drawing the interpolant by a Brownian bridge between $\bm A \hat{\bm x}^{(2)}$ and the rescaled measurement. Each prediction costs one network evaluation and applies $\bm X^\theta_{t_k,0}$ rather than composing consecutive maps. Randomness enters through the initialization, \eqref{eq:inference-interpolant}, and the re-noising, so independent runs give independent posterior samples.

\section{Experiments}\label{sec:experimental-setup}

\subsection{Setup}
\paragraph{Datasets.}
BAM is trained on 4KLSDB~\citep{zhu2026_4klsdb}, bicubically downsampled to 2K resolution, together with HQ-50K~\citep{yang2023hq50k}, DIV2K~\citep{agustsson2017ntire}, FFHQ-512~\citep{Karras2018ASG}, and the LSDIR~\citep{li2023lsdir} images above 1K resolution. Pixel values are scaled to $[0,1]$.\footnote{Internally, the network rescales images to $[-1,1]$ and the noise level to $2\sigma_y$.} We evaluate on held-out test sets of $300$
LSUN Bedroom~\citep{yu2015lsun} images, $64$ FFHQ and $64$ AFHQ
cat~\citep{choi2020stargan} images, and between $64$ and $100$ DIV2K images depending on the method, at noise levels $\sigma_y\in\{0.025,0.05\}$.

\paragraph{Metrics.} We report PSNR ($\uparrow$) for pixel-level fidelity, vgg LPIPS~\citep{Zhang2018TheUE} ($\downarrow$) for perceptual similarity, and CMMD~\citep{jayasumana2024rethinking} ($\downarrow$) and FID~\citep{heusel2017gans} ($\downarrow$) for distributional similarity to the ground truth.
FID is unstable for small test sets and is reported for comparability with prior work; CMMD, designed to address this, should be considered the more reliable of the two.

\paragraph{Training.}
Starting from the public RAM checkpoint\footnote{\url{https://github.com/matthieutrs/ram}},
we train the baseline BAM for $170$ hours on 8 H100 GPUs, jointly on all inverse problems. These are bicubic SR$\times4$ and $\times8$, isotropic Gaussian blur with $\sigma_\text{blur}\in[1,5]$, motion blur with random
$61\times61$ kernels of intensity $0.5$\footnote{\url{https://github.com/LeviBorodenko/motionblur}; see Appendix~\ref{app:motion}.}, compressed sensing at a $25\%$ rate, inpainting with $80\%$ of pixels masked at random, and demosaicing, all with noise levels $\sigma_y\in[0,0.05]$. The fine-tuned variant, \BAMFT, continues from this baseline for $16$ hours on 2 H100 GPUs on each target dataset, still across all problems (see~\cref{app:training}).

\paragraph{Baselines.}

We compare BAM and \BAMFT with zero-shot methods (LATINO, LATINO-PRO, TReg) and training-based methods (RAM, SILO, UD2M, I$^2$SB), described in Appendix~\ref{app:compared-methods}. UD2M and I$^2$SB are pixel-space diffusion models limited to lower resolutions, so we evaluate them only on FFHQ-512. SILO, UD2M and I$^2$SB use one checkpoint per dataset--problem pair (although UD2M can be trained over operator families), and RAM is evaluated before and after finetuning. All are fine-tuned with their public code, at the evaluated noise levels, for the cumulative runtime of one BAM dataset finetuning, whereas a single BAM model covers all noise levels.

\subsection{Results}

\paragraph{Reconstruction quality.}
Table~\ref{tab:main-restoration} compares all methods on DIV2K and FFHQ at $\sigma_y=0.05$. finetuning improves the LPIPS of three-step BAM on FFHQ for all five problems at both noise levels, but slightly degrades it on DIV2K (\cref{tab:compact-FFHQ,tab:compact-DIV2K}), indicating that the extended training set helps the network generalize to diverse, high-resolution data, while more focused finetuning on a specific distribution such as FFHQ improves results there. Tables~\ref{tab:compact-AFHQ}--\ref{tab:compact-LSUN} in Appendix~\ref{app:additional-results} report all problems on which BAM is trained, and include the out-of-distribution datasets AFHQ and LSUN bedroom, where finetuning noticeably improves results and the baseline BAM remains competitive with the other methods. Perceptually, BAM stands out: across the four datasets, five problems and two noise levels reported in Appendix~\ref{app:additional-results}, a BAM variant attains the best LPIPS in $38$ of $40$ settings and the best CMMD in $36$ of $40$, the exceptions being FFHQ deblurring (I$^2$SB) and one super-resolution setting each on AFHQ (SILO) and LSUN (UD2M). Representative restorations are shown in \cref{fig:main-DIV2K,fig:main-FFHQ,fig:main-AFHQ}: BAM restores sharp and realistic textures such as fur, hair and foliage while remaining consistent with the measurements, whereas RAM, trained as an MMSE estimator, often attains higher PSNR at the cost of smoother textures, and zero-shot and latent-space methods exhibit noise artifacts or hallucinate content that is inconsistent with the observation.

\begin{table}[t]
\centering
\caption{\textbf{DIV2K and FFHQ restoration at $\sigma_y=0.05$.} PSNR (dB; $\uparrow$), LPIPS ($\downarrow$), CMMD ($\downarrow$), and FID ($\downarrow$). BAM variants use thre e steps; \BAMFT and \RAMFT denote finetuning. Bold (underline) mark the best (second-best) reported values for each dataset and problem; -- denotes an unavailable result.}
\label{tab:main-restoration}
\label{tab:main-DIV2K}
\label{tab:main-FFHQ}
\begingroup
\fontsize{8}{9}\selectfont
\setlength{\tabcolsep}{3pt}
\begin{tabularx}{\linewidth}{@{}l r *{8}{>{\raggedleft\arraybackslash}X}@{}}
\toprule
Method & NFEs & \multicolumn{4}{c}{DIV2K} & \multicolumn{4}{c}{FFHQ} \\
\cmidrule(lr){3-6} \cmidrule(lr){7-10}
& & PSNR$\uparrow$ & LPIPS$\downarrow$ & CMMD$\downarrow$ & FID$\downarrow$ & PSNR$\uparrow$ & LPIPS$\downarrow$ & CMMD$\downarrow$ & FID$\downarrow$ \\
\midrule
\multicolumn{10}{@{}l}{\textbf{Gaussian deblurring}} \\[-1pt]
\rowcolor{BAMRow} BAM & 3 & 24.48 & \textbf{0.35} & \underline{0.06} & \textbf{34.4} & 29.63 & \underline{0.25} & 0.10 & \underline{45.0} \\
\rowcolor{BAMRow} \BAMFT & 3 & 24.44 & \textbf{0.35} & \textbf{0.04} & \underline{35.6} & 30.16 & \underline{0.25} & \underline{0.08} & 51.6 \\
RAM & 1 & \underline{25.63} & 0.43 & 0.44 & 52.6 & \underline{30.54} & 0.35 & 1.15 & 94.7 \\
\RAMFT & 1 & \textbf{26.00} & \underline{0.42} & 0.34 & 47.0 & \textbf{31.53} & 0.34 & 1.07 & 92.4 \\
LATINO-PRO & 65 & 23.55 & 0.54 & 0.49 & 71.9 & 23.40 & 0.46 & 1.17 & 128.3 \\
SILO & 200 & 17.20 & 0.62 & 1.84 & 298.8 & 24.92 & 0.38 & 0.25 & 91.8 \\
I$^2$SB & 50 & -- & -- & -- & -- & 30.24 & \textbf{0.24} & \textbf{0.02} & \textbf{44.2} \\
UD2M & 12 & -- & -- & -- & -- & 29.30 & 0.29 & 0.72 & 67.3 \\
\midrule
\multicolumn{10}{@{}l}{\textbf{Super-resolution}} \\[-1pt]
\rowcolor{BAMRow} BAM & 3 & 24.30 & \textbf{0.36} & \textbf{0.08} & \textbf{39.6} & 28.84 & \textbf{0.27} & \textbf{0.19} & \textbf{50.4} \\
\rowcolor{BAMRow} \BAMFT & 3 & 23.56 & \underline{0.37} & \underline{0.09} & \underline{41.1} & 29.37 & \textbf{0.27} & \textbf{0.19} & \underline{62.4} \\
RAM & 1 & \underline{25.79} & 0.43 & 0.26 & 74.1 & \textbf{31.15} & 0.36 & 1.02 & 107.1 \\
\RAMFT & 1 & \textbf{25.88} & 0.42 & 0.27 & 51.7 & \underline{31.01} & 0.35 & 1.52 & 93.0 \\
LATINO-PRO & 65 & 22.82 & 0.54 & 0.59 & 95.3 & 23.18 & 0.53 & 0.90 & 146.3 \\
SILO & 200 & 16.53 & 0.64 & 2.04 & 308.4 & 25.04 & 0.37 & \underline{0.20} & 80.7 \\
I$^2$SB & 50 & -- & -- & -- & -- & 26.82 & 0.39 & 1.02 & 148.8 \\
UD2M & 12 & -- & -- & -- & -- & 28.16 & \underline{0.31} & 0.30 & 82.2 \\
\midrule
\multicolumn{10}{@{}l}{\textbf{Inpainting}} \\[-1pt]
\rowcolor{BAMRow} BAM & 3 & \underline{26.61} & \textbf{0.27} & \textbf{0.02} & \textbf{20.5} & 31.50 & \underline{0.23} & \underline{0.08} & \textbf{32.1} \\
\rowcolor{BAMRow} \BAMFT & 3 & 25.66 & \textbf{0.27} & \underline{0.03} & \underline{21.0} & \underline{31.57} & \textbf{0.20} & \textbf{0.05} & \underline{32.4} \\
RAM & 1 & 23.52 & 0.41 & 0.30 & 66.9 & 25.55 & 0.39 & 1.53 & 126.5 \\
\RAMFT & 1 & \textbf{27.61} & \underline{0.32} & 0.18 & 35.5 & \textbf{32.40} & 0.29 & 1.12 & 56.5 \\
LATINO-PRO & 65 & 11.25 & 0.71 & 2.01 & 282.4 & 13.44 & 0.79 & 2.63 & 305.0 \\
SILO & 200 & 12.92 & 0.73 & 2.91 & 382.4 & 17.46 & 0.52 & 1.00 & 139.4 \\
I$^2$SB & 50 & -- & -- & -- & -- & 18.89 & 0.58 & 1.93 & 174.2 \\
UD2M & 12 & -- & -- & -- & -- & 29.29 & 0.25 & 0.21 & 45.54 \\
\bottomrule
\end{tabularx}
\endgroup
\end{table}

\paragraph{Computation/quality trade-off.} \Cref{fig:pareto} reports image quality (LPIPS), computational cost (floating-point operations to restore a $512\times 512$ input) and model size (number of parameters) for \BAMFT and the state-of-the-art baselines in a single graph. We observe that \BAMFT advances the Pareto frontier of the baselines, attaining better quality at a lower cost, and the same conclusion holds across metrics, inverse problems and datasets (Table~\ref{tab:main-restoration} and Appendix~\ref{app:additional-results}). BAM is also robust to post-training quantization: converting it to INT8 under TensorRT,  together with a TensorRT-friendly reformulation of the physics-aware operations, cuts the latency on a $512\times512$ image from $281$\,ms to $80$\,ms on an NVIDIA A40 GPU ($3.5\times$ faster), with only a moderate loss in accuracy (PSNR $-1.3$\,dB, LPIPS $0.400\to0.427$) and \emph{without any fine-tuning after quantization} (Appendix~\ref{app:int8}). Quantization-aware fine-tuning is expected to narrow the remaining gap.

\begin{figure}[h]
\centering
\scriptsize
\setlength{\tabcolsep}{1pt}
\begin{tabular}{@{}cccccc@{}}
\shortstack{Ground\\truth} &
\shortstack{Filtered\\backprojection} &
\shortstack{BAM\\sample} &
\shortstack{BAM\\mean} &
\shortstack{$4\times4$ std.\\($N=64$)} &
\shortstack{$4\times4$ residual}\\[2pt]
\includegraphics[width=.16\linewidth]{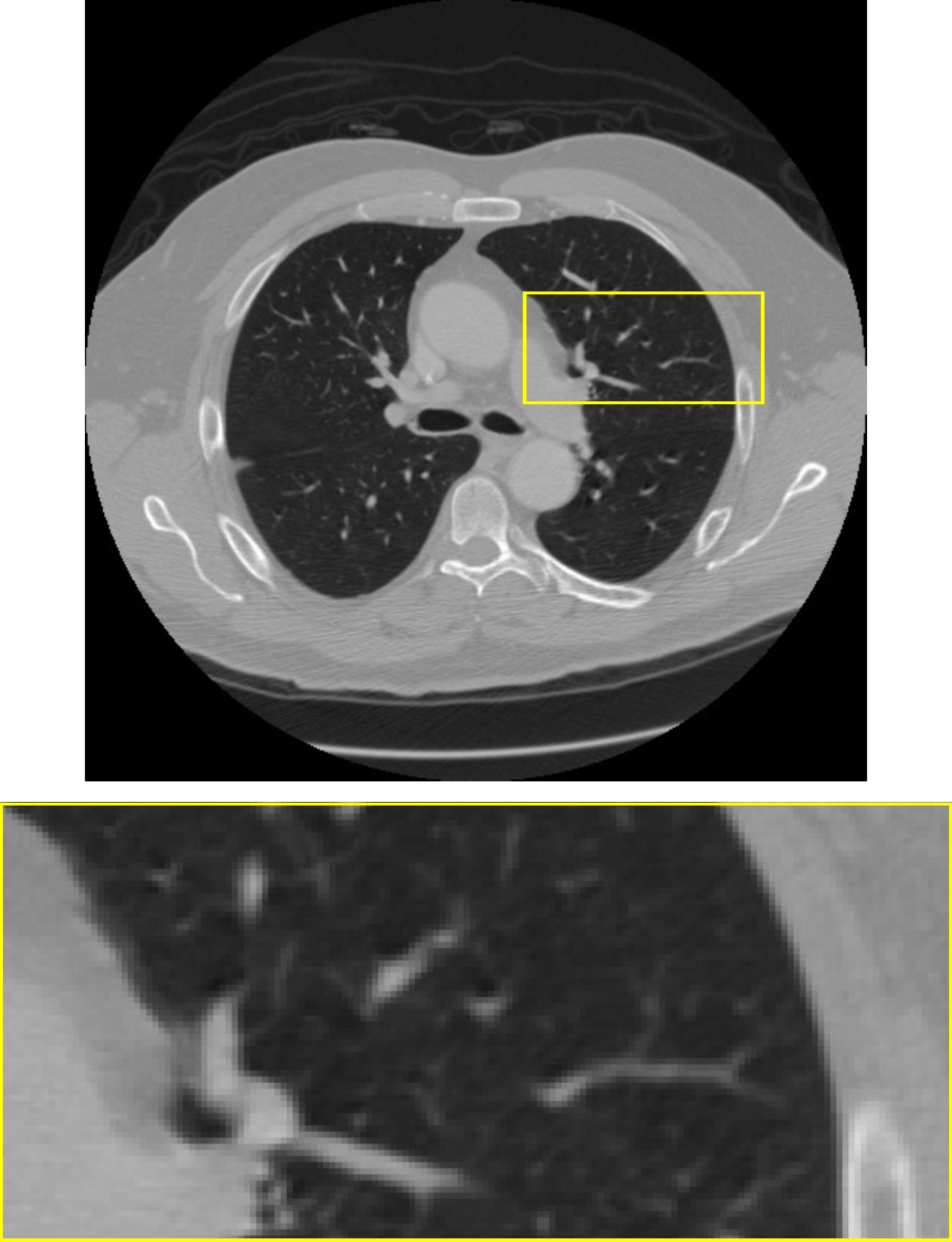} &
\includegraphics[width=.16\linewidth]{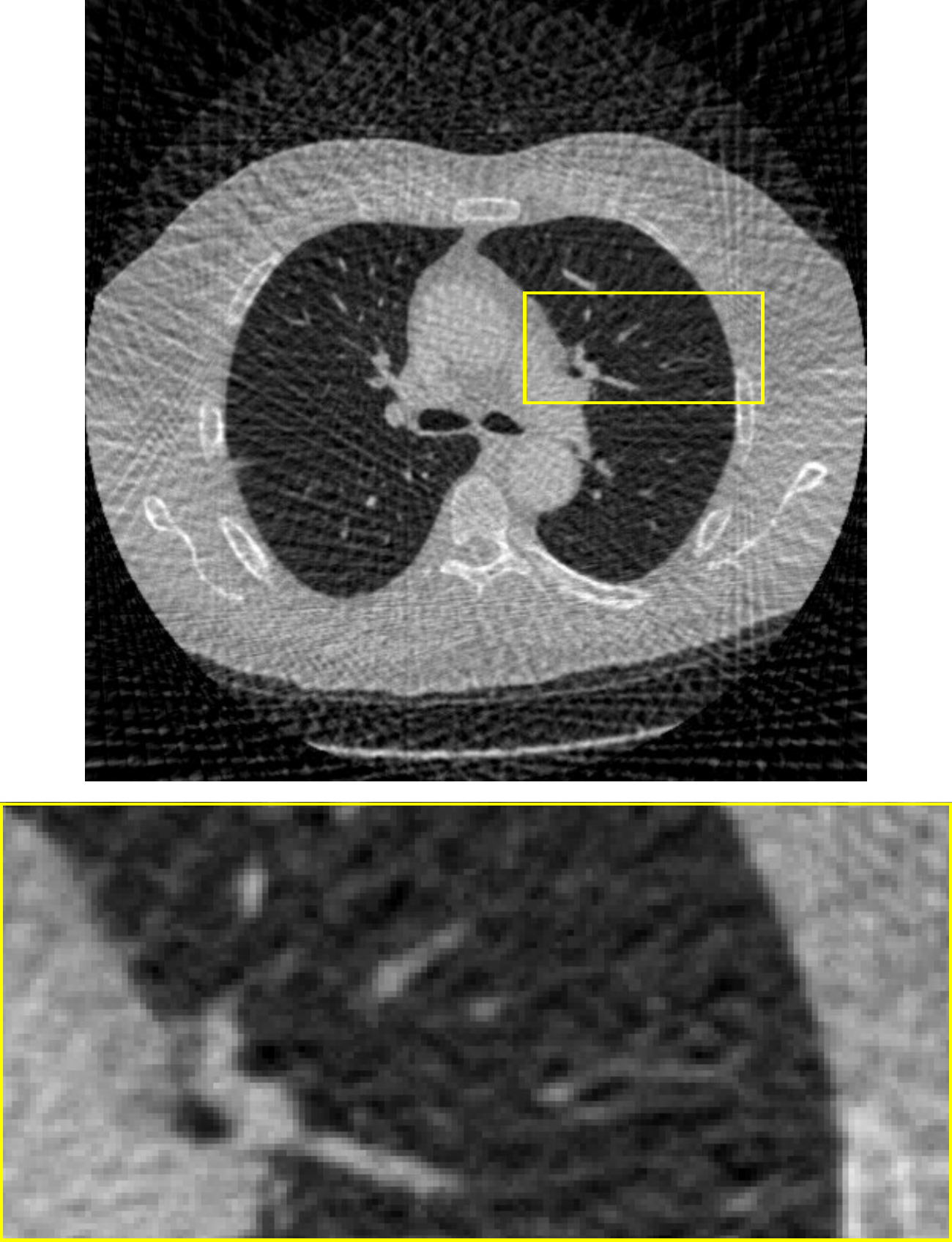} &
\includegraphics[width=.16\linewidth]{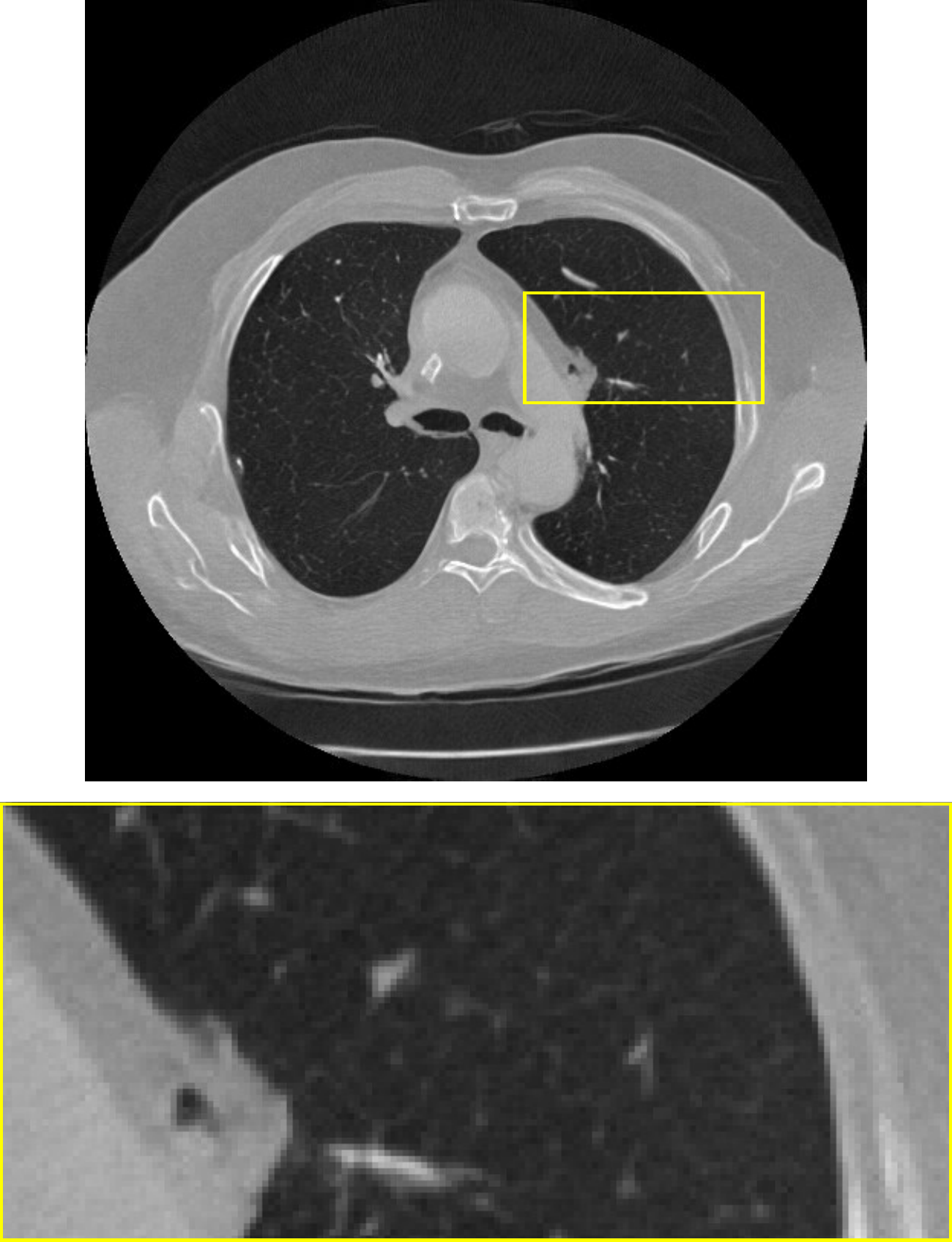} &
\includegraphics[width=.16\linewidth]{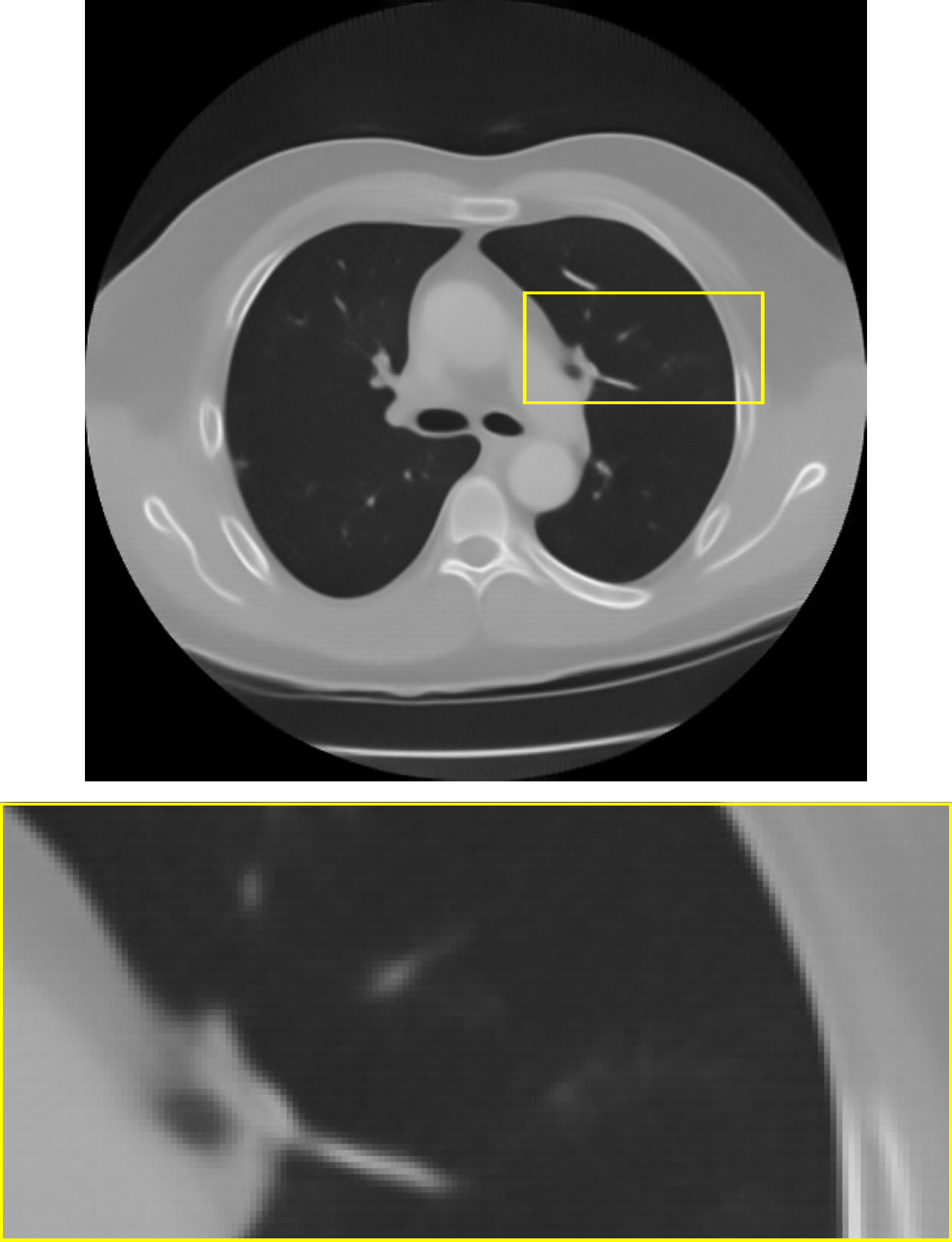} &
\includegraphics[width=.16\linewidth]{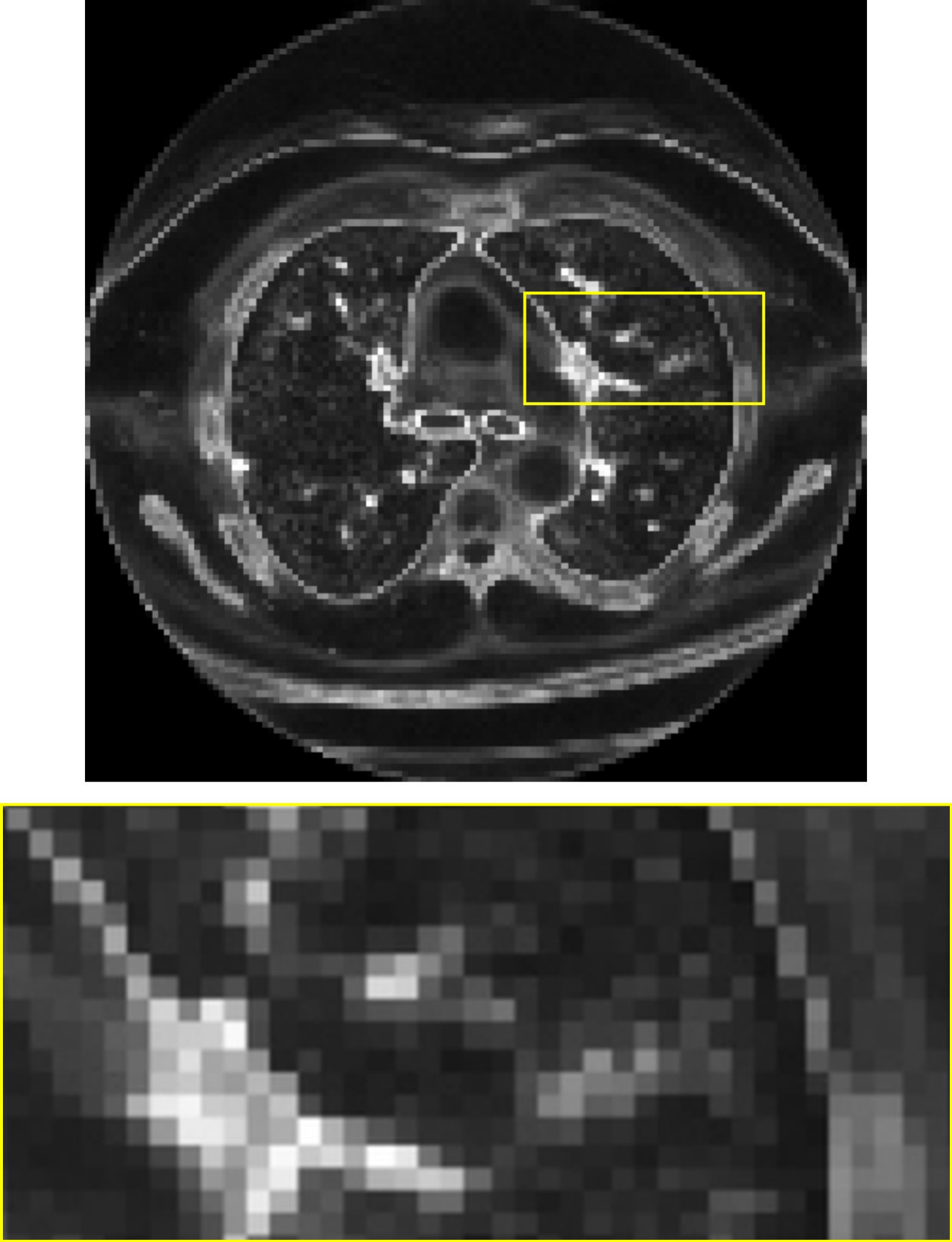} &
\includegraphics[width=.16\linewidth]{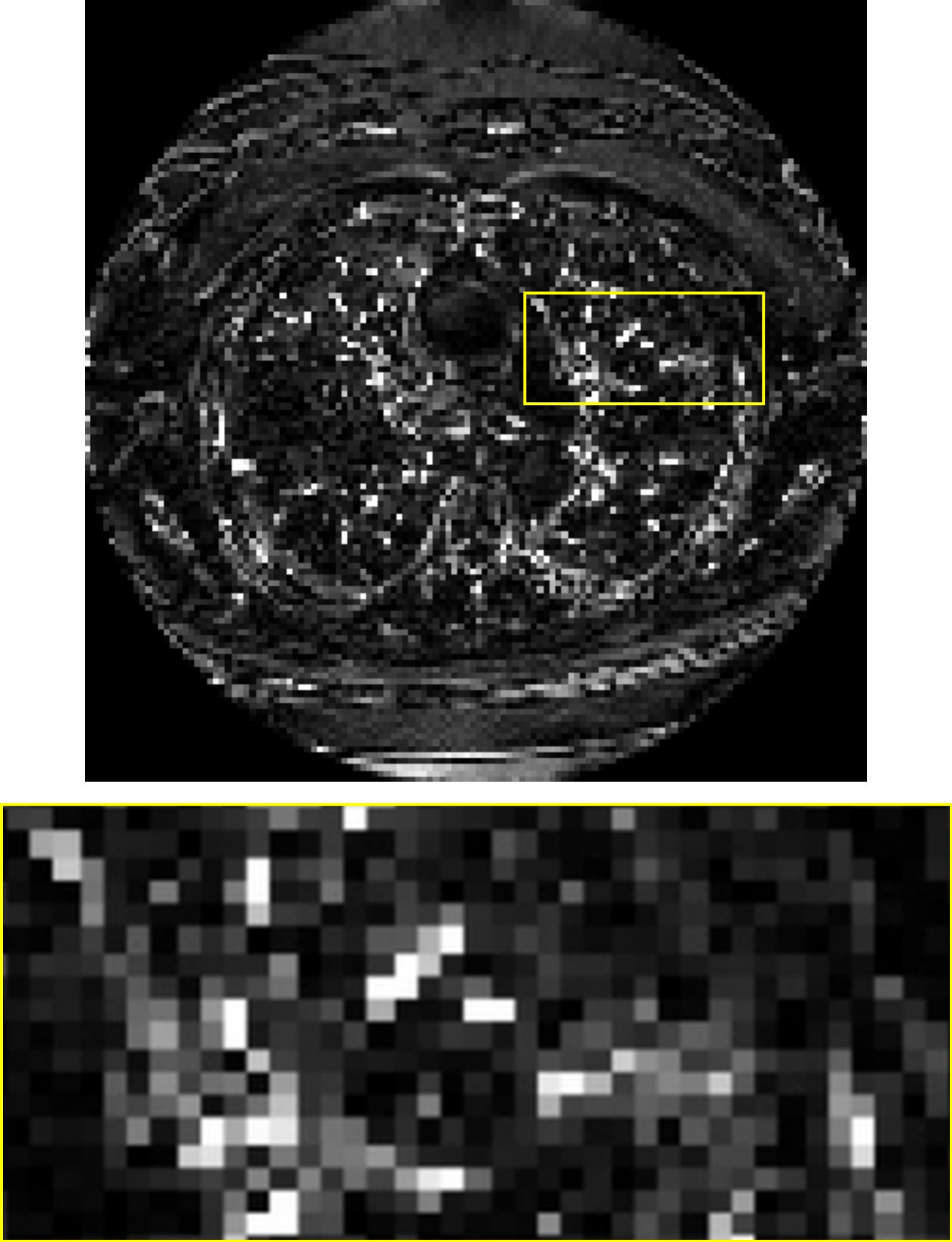}
\end{tabular}
\caption{CT reconstruction and $4\times4$-block uncertainty and residual maps. Left to right: ground truth, reconstruction $\mathcal A^{\dagger}\vy$, one BAM sample, posterior mean, standard deviation over 64 draws, and residual $|\mathrm{GT}-\mathrm{mean}|$. Yellow boxes and $4\times$ strips show the same lung region in every panel.}
\label{fig:ct-section-summary}
\end{figure}

\paragraph{Other experiments.}
Appendix~\ref{app:ct} extends BAM to sparse-view CT on LIDC-IDRI chest slices with $51$ parallel-beam projections, following the setting of~\citet{Terris2025ReconstructAM}, a single-channel modality far from the natural images seen during pre-training. Figure~\ref{fig:ct-section-summary} shows the ground truth, the adjoint reconstruction $\mathcal A^{\dagger}\vy$, a BAM sample, and the $4\times4$-block standard-deviation and residual maps. After finetuning on CT, three-step BAM samples recover fine lung vessels that the RAM estimate smooths out. Moreover, the pixelwise standard deviation over $64$ draws concentrates on anatomical edges and vessels, and is of the same order of magnitude as the error of the empirical posterior mean, so BAM provides a spatial uncertainty map at no extra training cost.

Appendix~\ref{app:motion} turns to blind motion deblurring with spatially varying kernels. Following~\citet{Terris2025ReconstructAM}, we estimate $\bm A$ with the pre-trained Kernel Predictor Network (KPN) of~\citet{carbajal2023blind} and solve the resulting inverse problem. Although BAM has never been trained on such operators, it outperforms both RAM and the original unrolled PnP architecture of~\citet{carbajal2023blind} on the K\"ohler dataset~\citep{kohler2012recording} (see \Cref{fig:motion-three-settings}). Fine-tuning BAM jointly with the KPN, as in the state of the art, is a natural next step towards a generative counterpart of their method.

Appendix~\ref{app:jpeg} reports a preliminary exploration of non-linear imaging problems, namely the restoration of noisy, JPEG-compressed images. Zero-shot and unrolled methods generalize poorly here, as they require a quadratic approximation of the log-likelihood. After finetuning on this problem, BAM instead generates the missing details and regularizes noisy areas. This approach has limitations, and extending BAM to more general degradation models is a natural direction for future work.

Appendix~\ref{app:single-problem} reports a preliminary assessment of BAM from a distortion-perception viewpoint, based on repeated draws for a fixed problem. Drawing multiple realizations from a single observation $\vy$ and averaging them gives a Monte Carlo estimate of the posterior mean, which we compare with the MMSE estimator produced by RAM. The empirical mean exceeds the PSNR of \RAMFT on four of the five problems, the exception being super-resolution. The draws are therefore diverse enough to be perceptually sharp individually, while their average remains an accurate posterior mean estimate.

\begin{figure}[h]
\centering
\begingroup
\setlength{\BAMPaperImageWidth}{.120\linewidth}
\setlength{\tabcolsep}{1pt}
\scriptsize
\begin{tabular}{@{}cccccccc@{}}
\toprule
\textbf{Observation} & \textbf{BAM} & \textbf{\BAMFT} & \textbf{RAM} & \textbf{\RAMFT} & \textbf{SILO} & \textbf{TReg} & \textbf{GT} \\
\midrule
\multicolumn{8}{@{}l}{\textbf{Gaussian deblurring} $\sigma_y=0.05$} \\[1pt]
\BAMPaperZoom{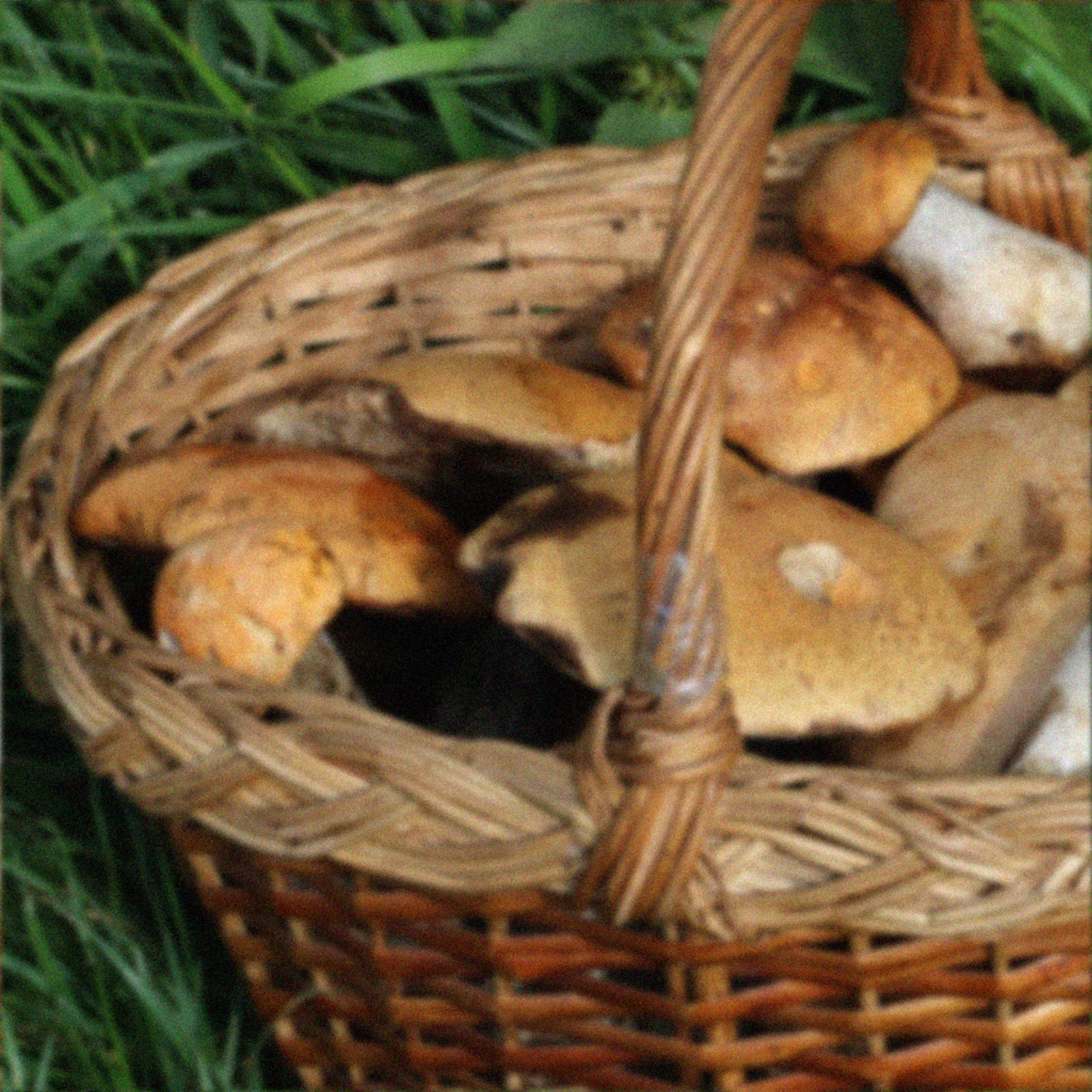}{0.08}{0.28} & \BAMPaperZoom{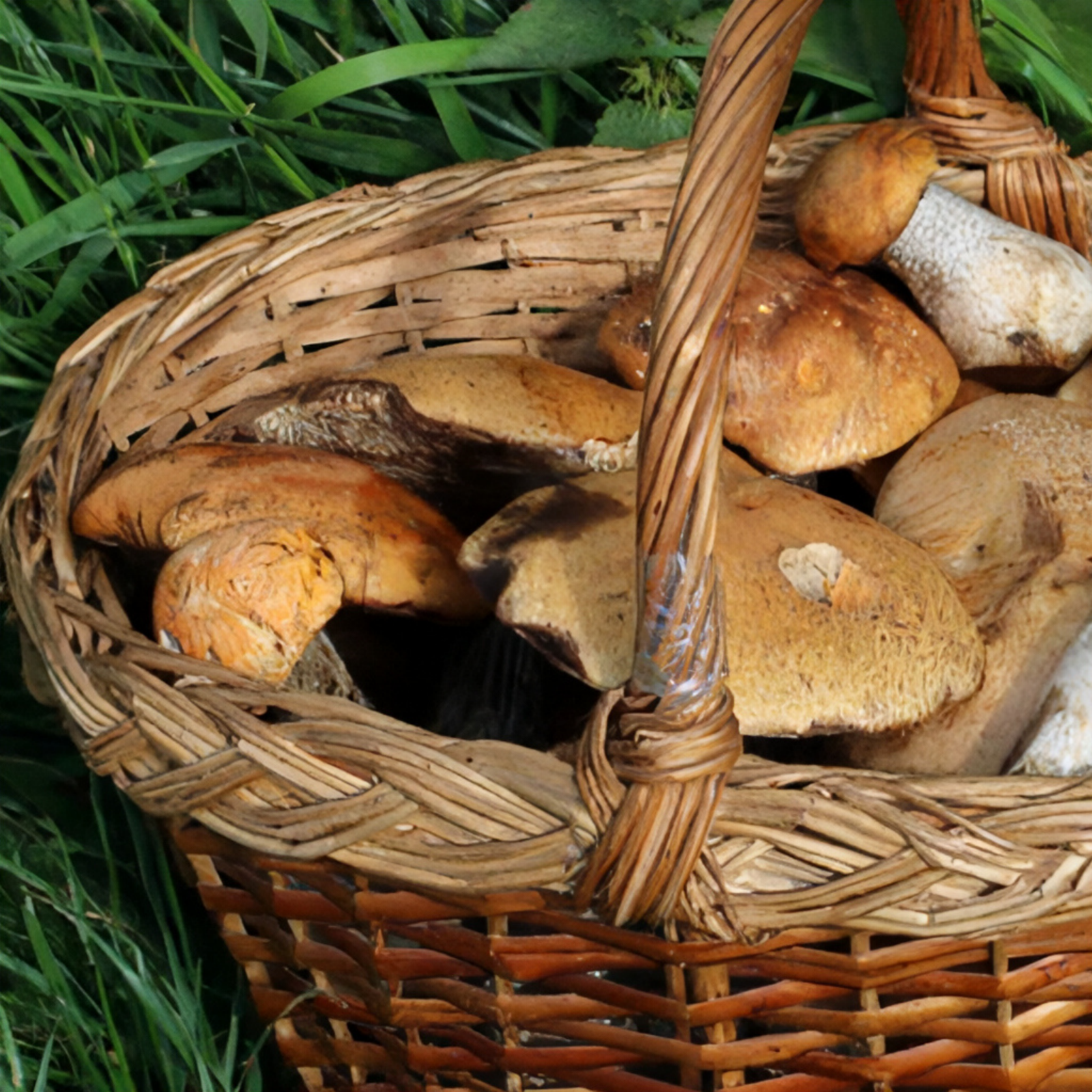}{0.08}{0.28} & \BAMPaperZoom{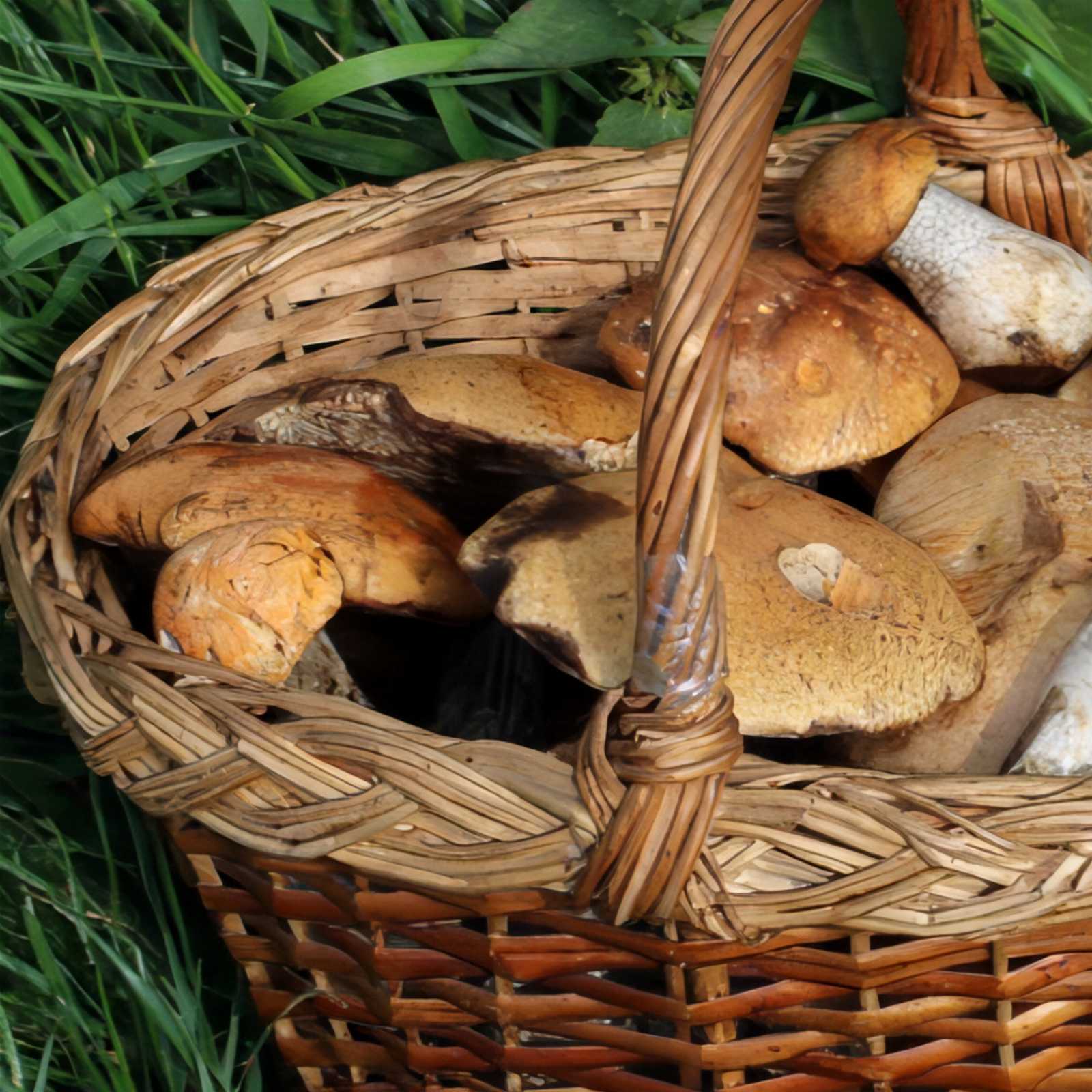}{0.08}{0.28} & \BAMPaperZoom{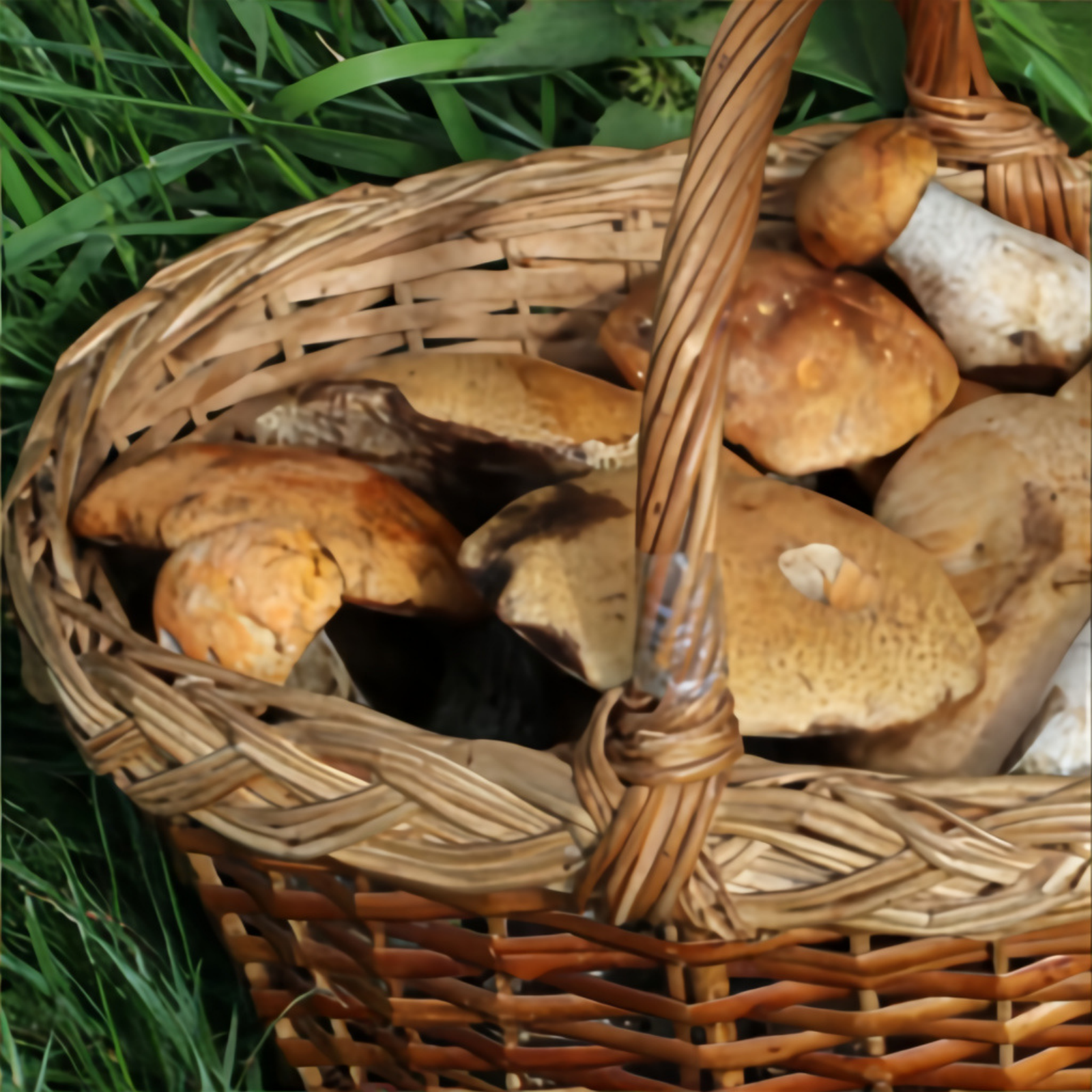}{0.08}{0.28} & \BAMPaperZoom{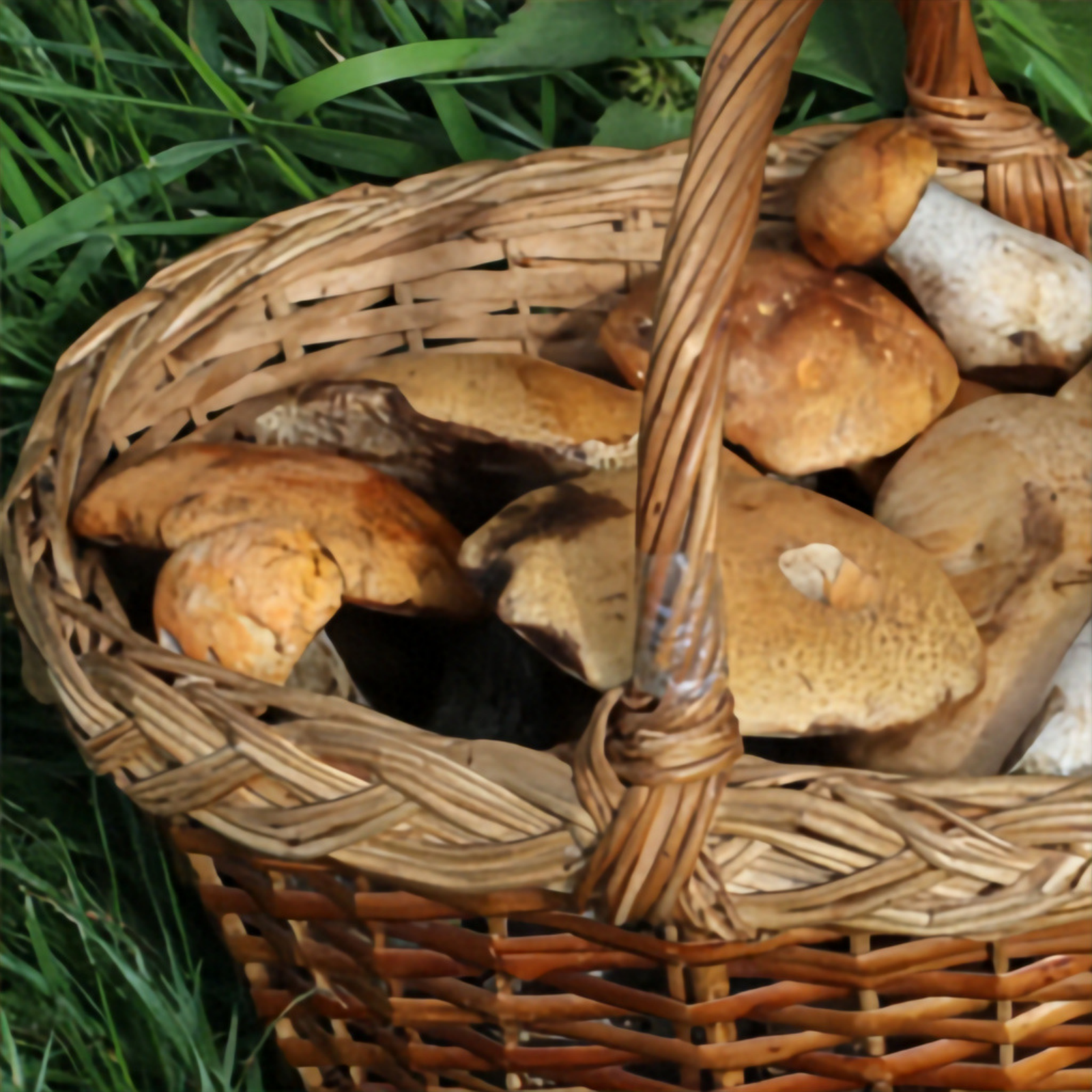}{0.08}{0.28} & \BAMPaperZoom{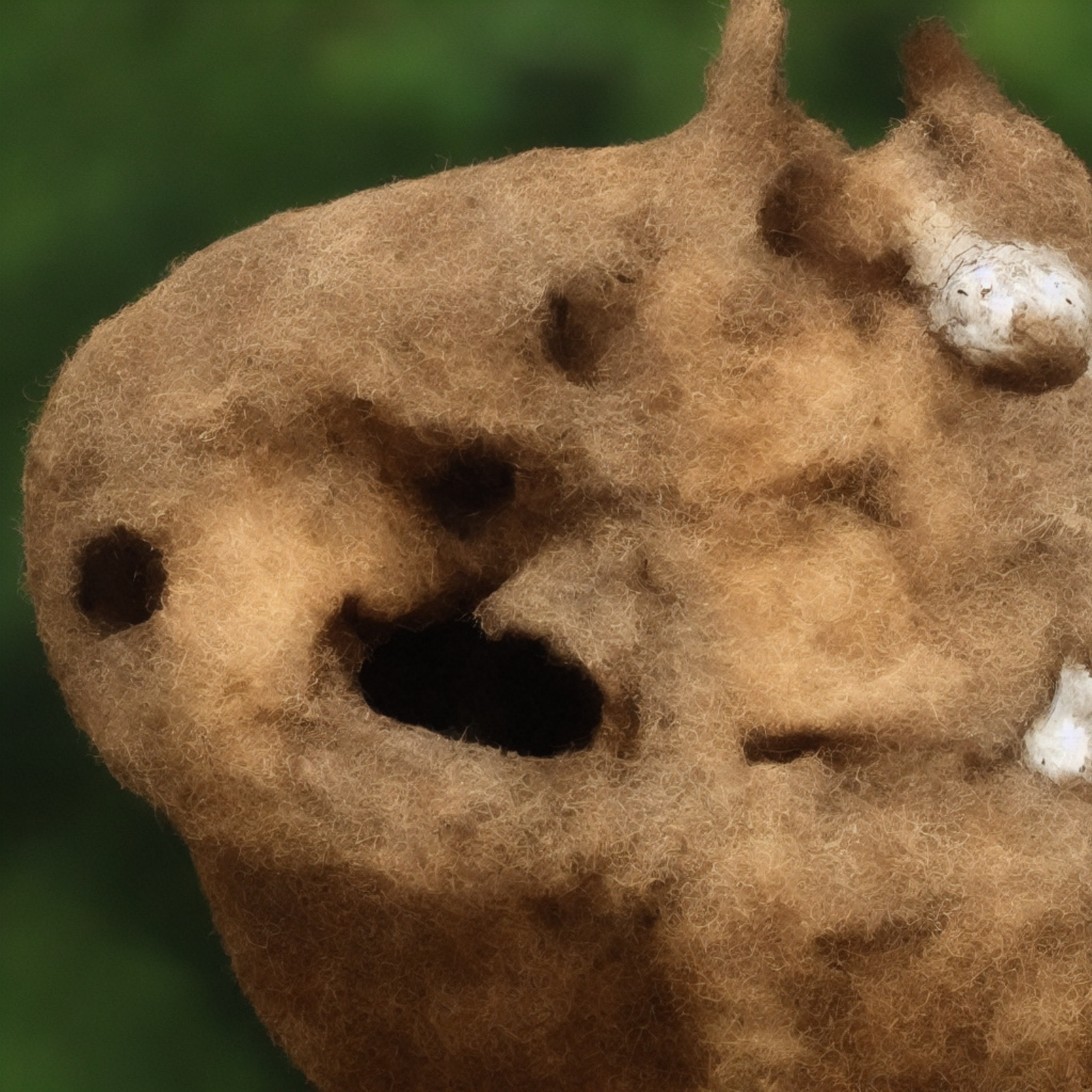}{0.08}{0.28} & \BAMPaperZoom{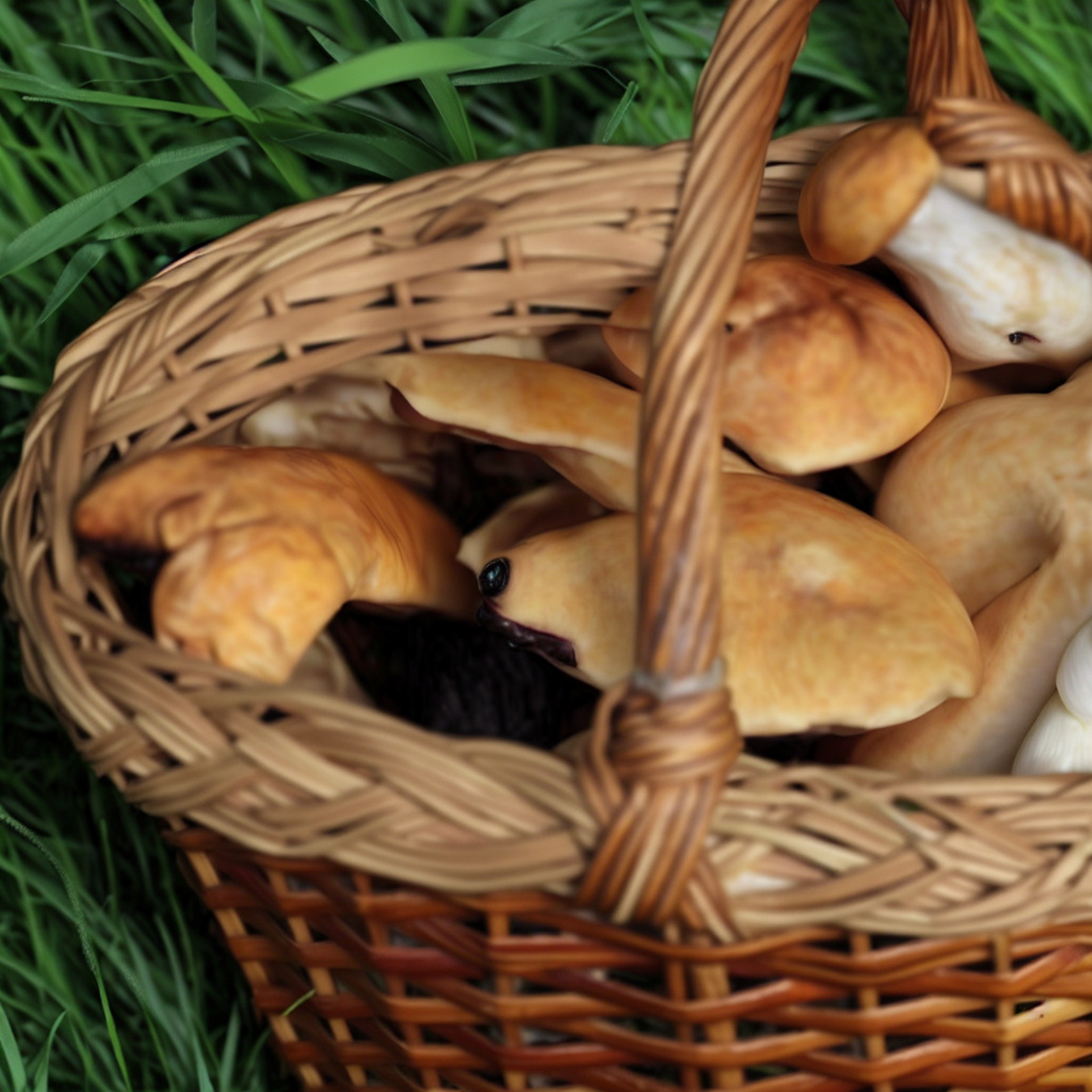}{0.08}{0.28} & \BAMPaperZoom{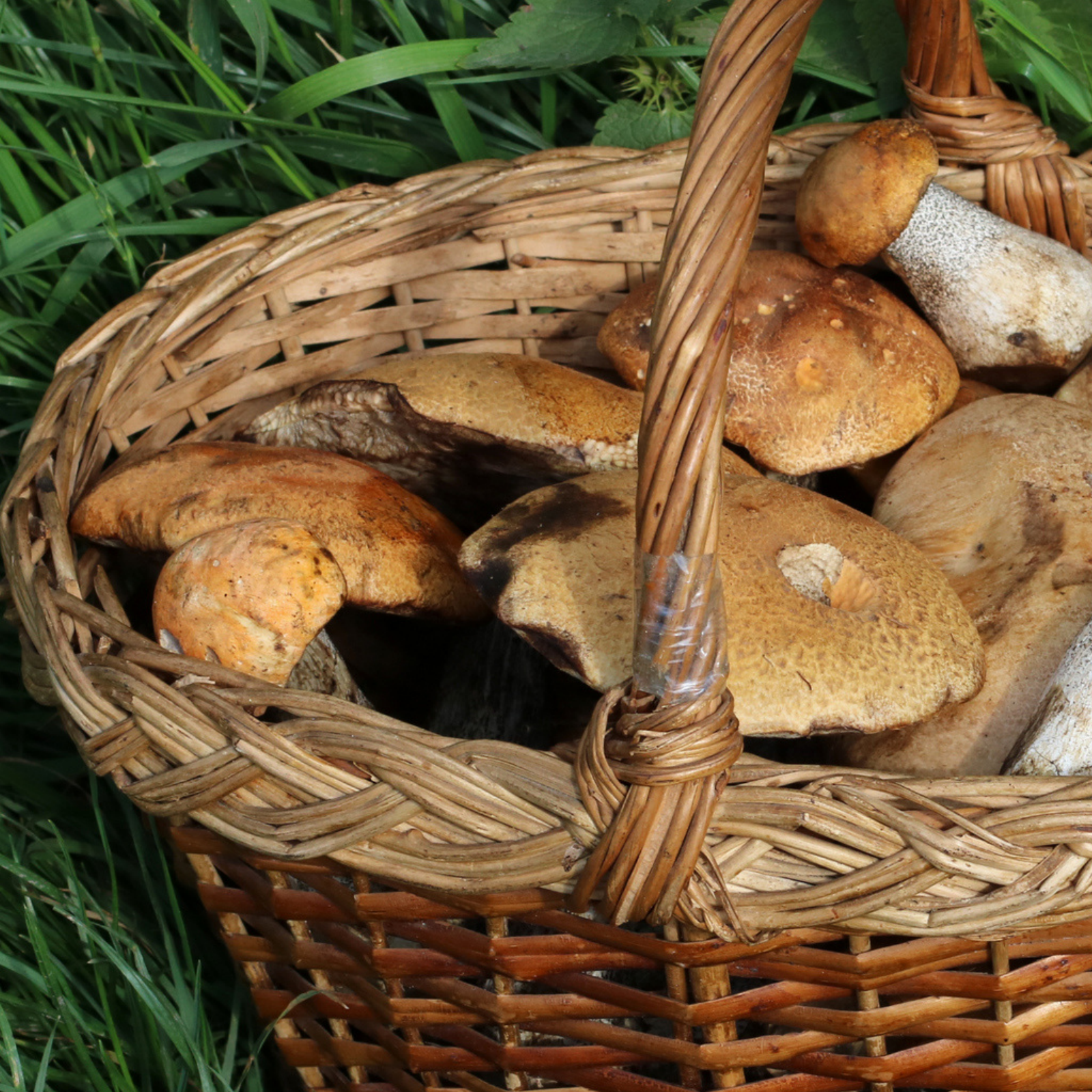}{0.08}{0.28} \\[3pt]
\multicolumn{8}{@{}l}{\textbf{SR $\times4$} $\sigma_y=0.05$} \\[1pt]
\BAMPaperZoom{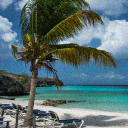}{0.46}{0.54} & \BAMPaperZoom{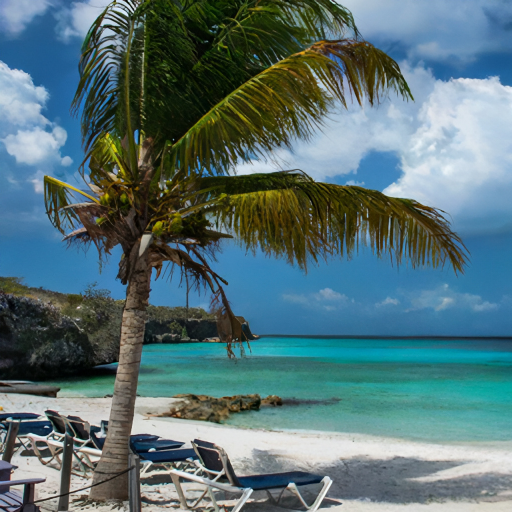}{0.46}{0.54} & \BAMPaperZoom{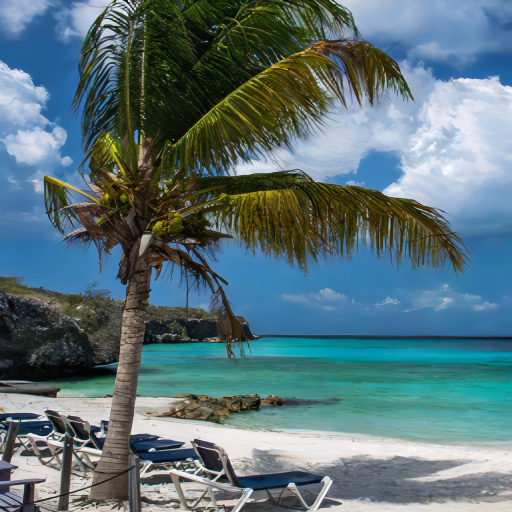}{0.46}{0.54} & \BAMPaperZoom{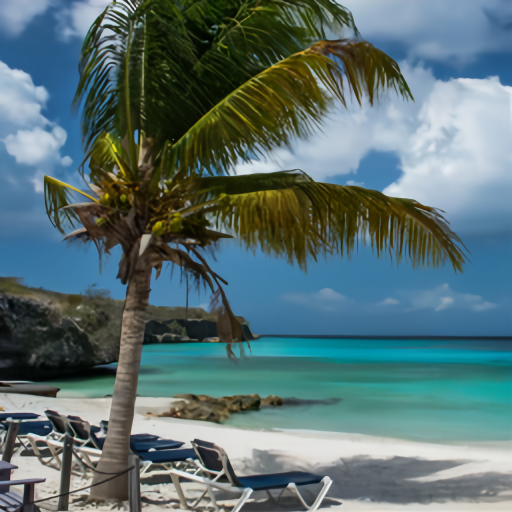}{0.46}{0.54} & \BAMPaperZoom{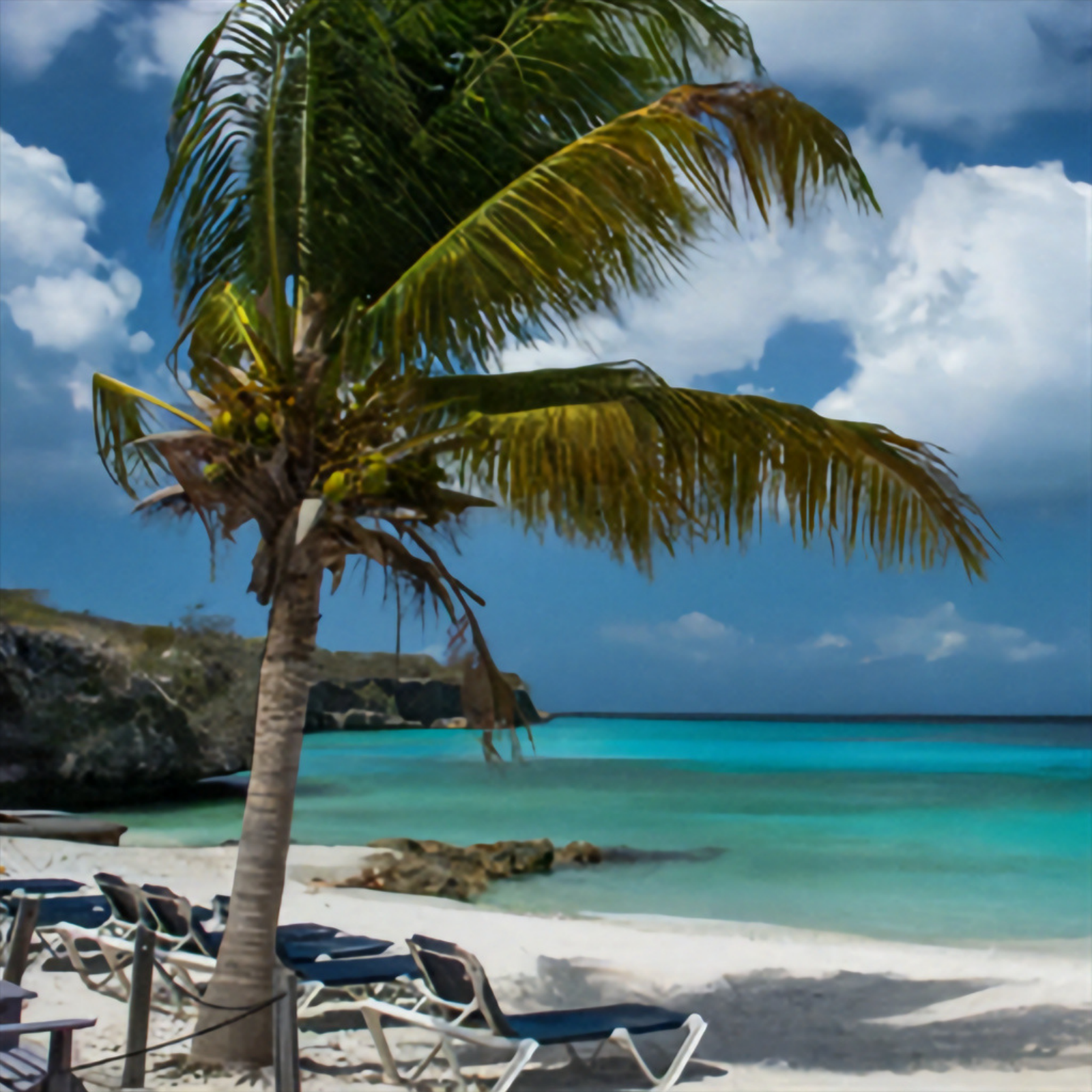}{0.46}{0.54} & \BAMPaperZoom{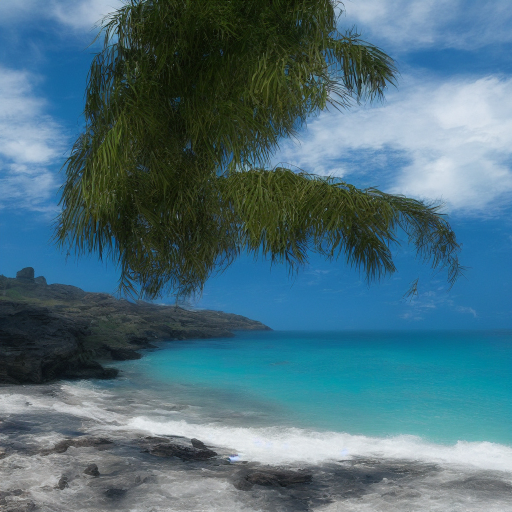}{0.46}{0.54} & \BAMPaperZoom{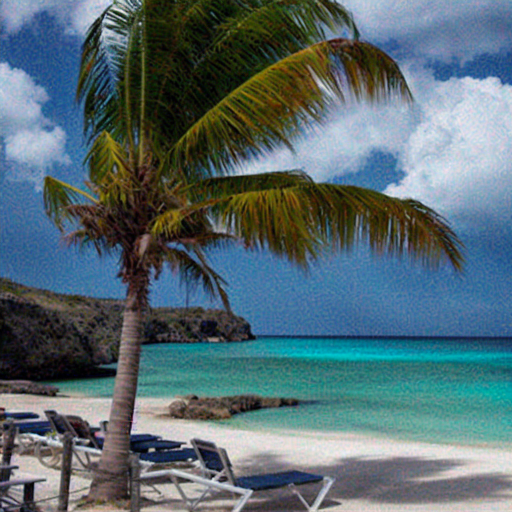}{0.46}{0.54} & \BAMPaperZoom{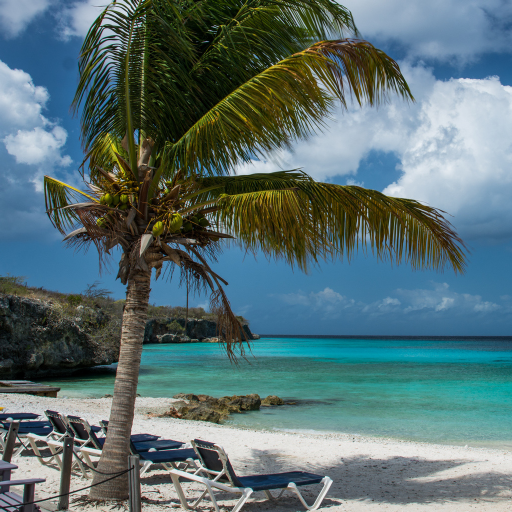}{0.46}{0.54} \\[3pt]
\multicolumn{8}{@{}l}{\textbf{Inpainting} $\sigma_y=0.05$} \\[1pt]
\BAMPaperZoom{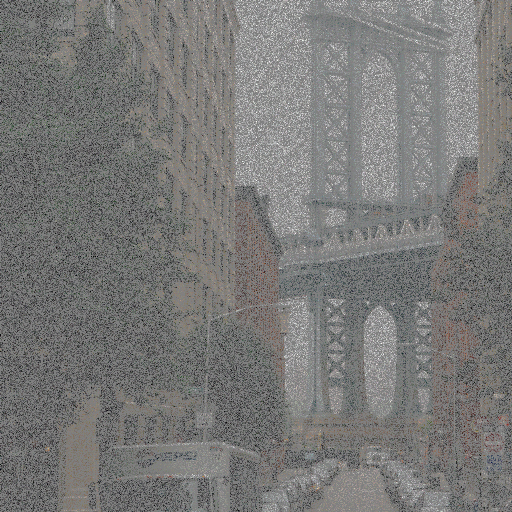}{0.53}{0.48} & \BAMPaperZoom{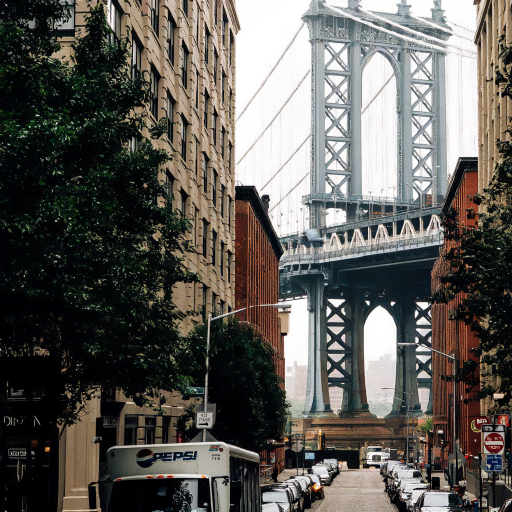}{0.53}{0.48} & \BAMPaperZoom{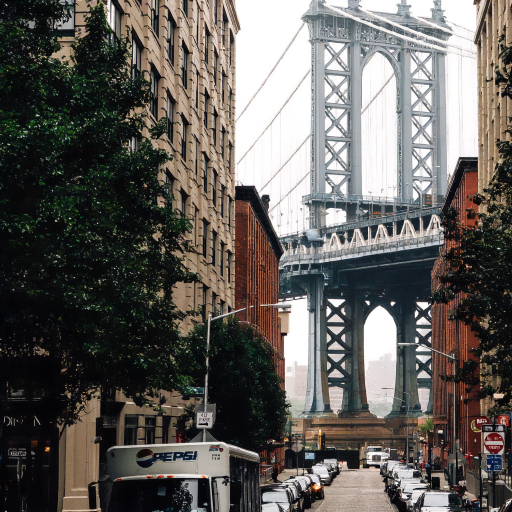}{0.53}{0.48} & \BAMPaperZoom{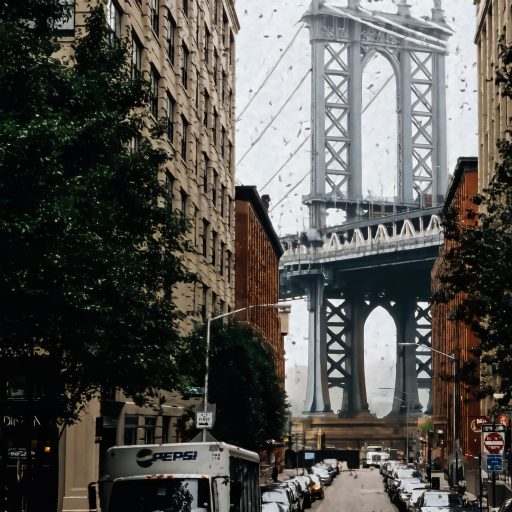}{0.53}{0.48} & \BAMPaperZoom{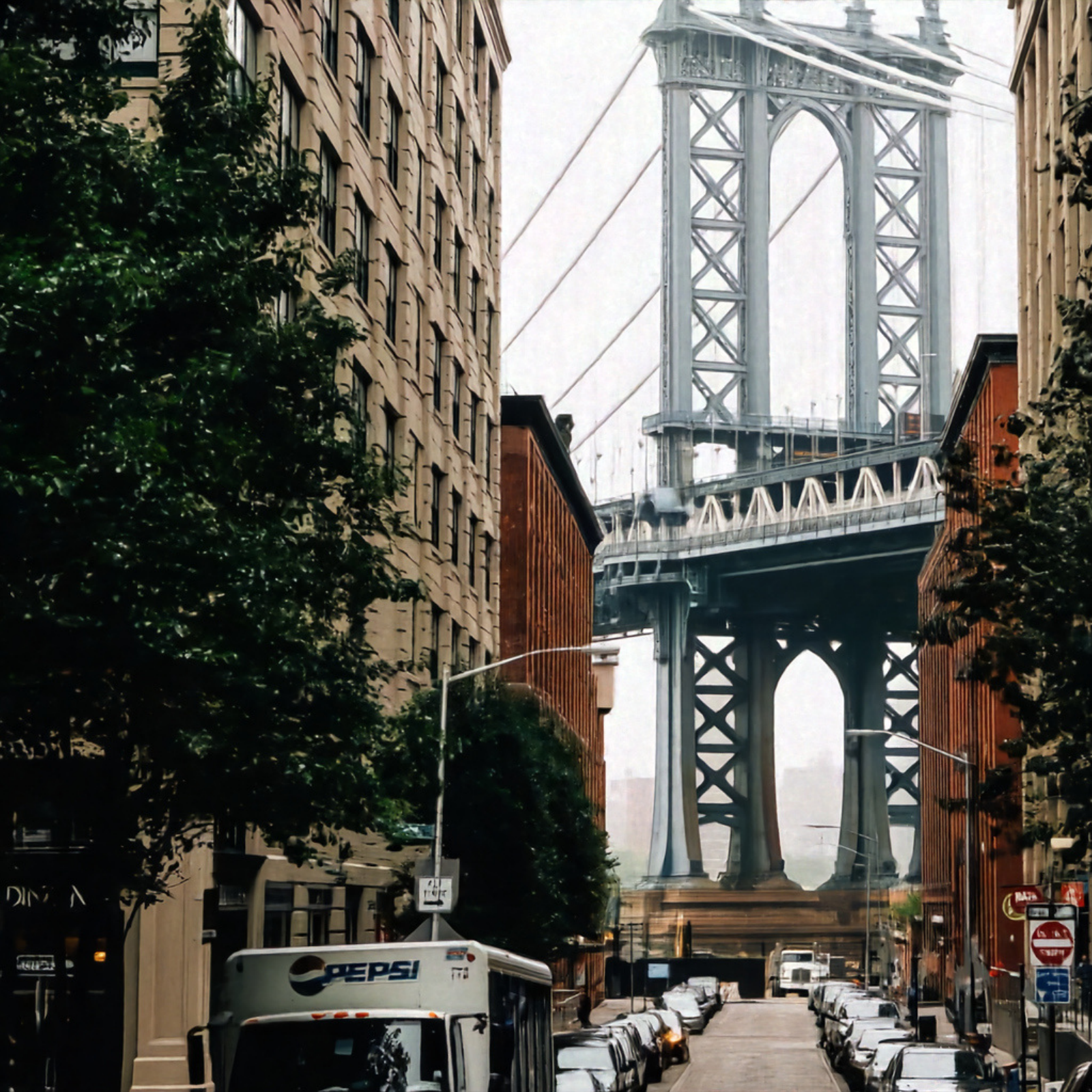}{0.53}{0.48} & \BAMPaperZoom{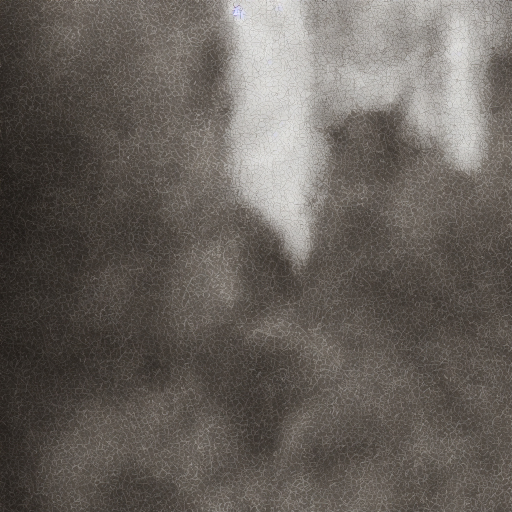}{0.53}{0.48} & \BAMPaperZoom{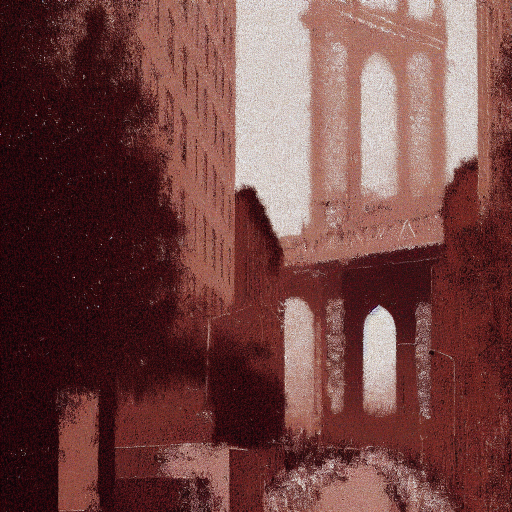}{0.53}{0.48} & \BAMPaperZoom{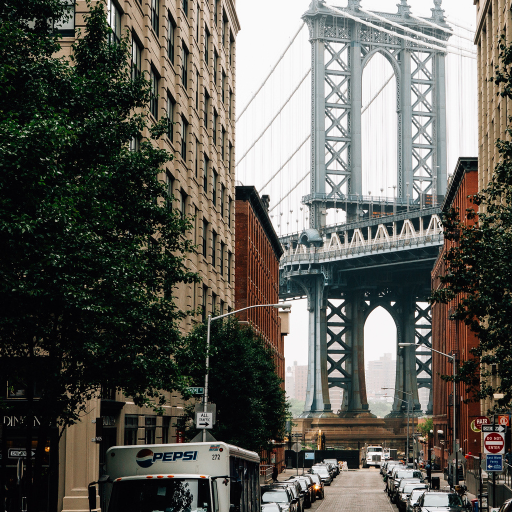}{0.53}{0.48} \\[3pt]
\bottomrule
\end{tabular}
\endgroup
\caption{\textbf{DIV2K restorations} at $\sigma_y = 0.05$, one example per problem. BAM and \BAMFT use three steps. Strips show $4\times$ zoom. More problems and comparisons in Figure~\ref{fig:compact-DIV2K}.}
\label{fig:main-DIV2K}
\end{figure}

\begin{figure}[h]
\centering
\begingroup
\setlength{\BAMPaperImageWidth}{.106\linewidth}
\setlength{\tabcolsep}{1pt}
\scriptsize
\begin{tabular}{@{}ccccccccc@{}}
\toprule
\textbf{Observation} & \textbf{BAM} & \textbf{\BAMFT} & \textbf{RAM} & \textbf{\RAMFT} & \textbf{SILO} & \textbf{UD2M} & \textbf{I2SB} & \textbf{GT} \\
\midrule
\multicolumn{9}{@{}l}{\textbf{FFHQ --- Gaussian deblurring}, $\sigma_y=0.05$} \\[1pt]
\BAMPaperZoom{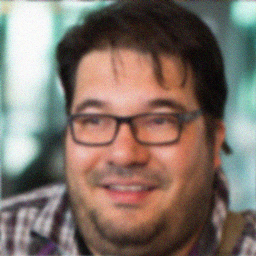}{0.27}{0.45} & \BAMPaperZoom{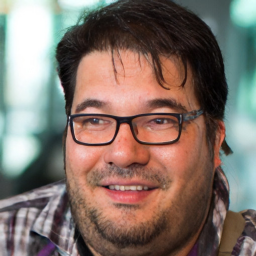}{0.27}{0.45} & \BAMPaperZoom{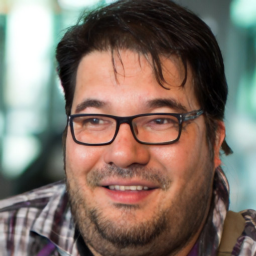}{0.27}{0.45} & \BAMPaperZoom{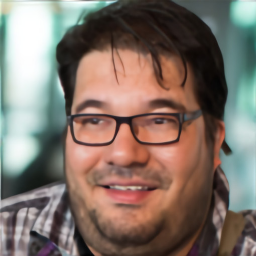}{0.27}{0.45} & \BAMPaperZoom{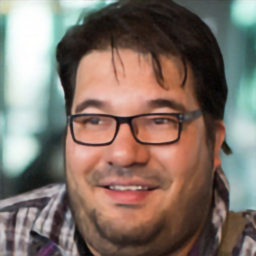}{0.27}{0.45} & \BAMPaperZoom{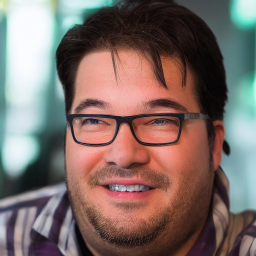}{0.27}{0.45} & \BAMPaperZoom{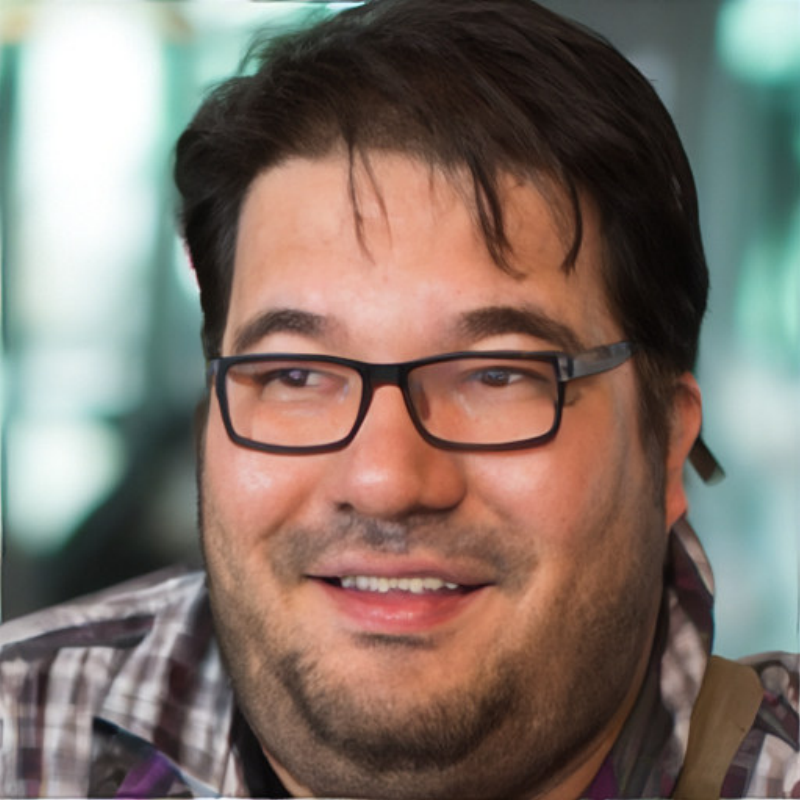}{0.27}{0.45} & \BAMPaperZoom{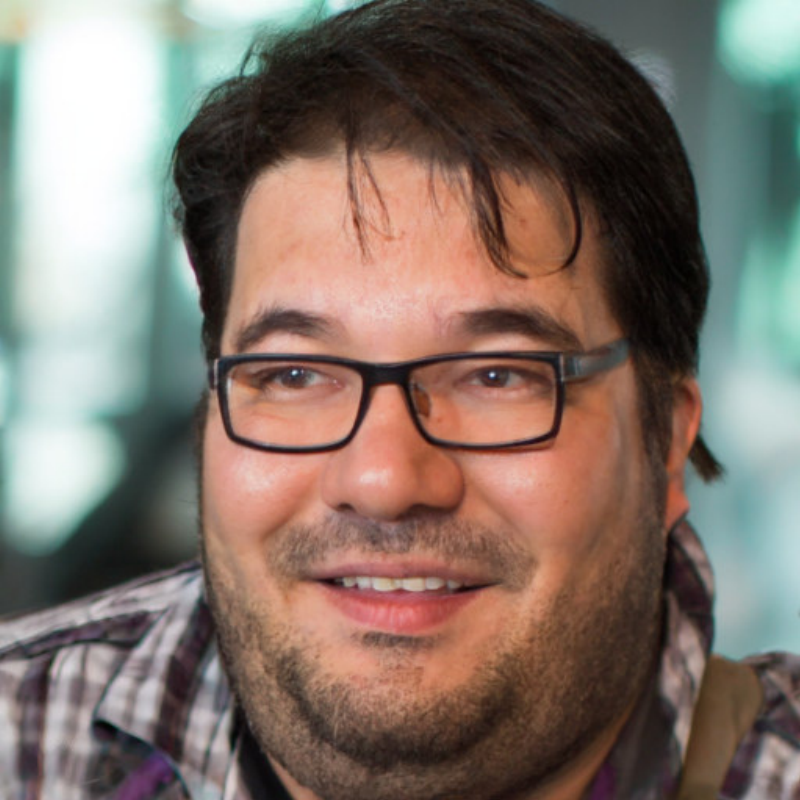}{0.27}{0.45} & \BAMPaperZoom{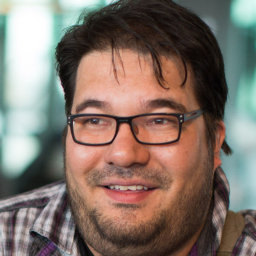}{0.27}{0.45} \\[3pt]
\multicolumn{9}{@{}l}{\textbf{FFHQ --- SR $\times4$}, $\sigma_y=0.05$} \\[1pt]
\BAMPaperZoom{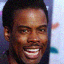}{0.34}{0.48} & \BAMPaperZoom{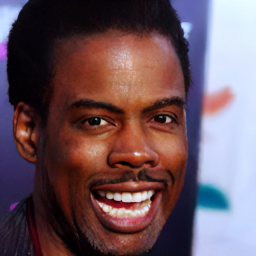}{0.34}{0.48} & \BAMPaperZoom{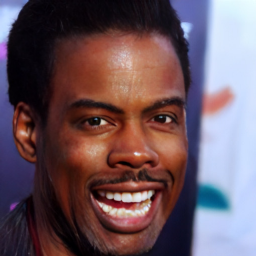}{0.34}{0.48} & \BAMPaperZoom{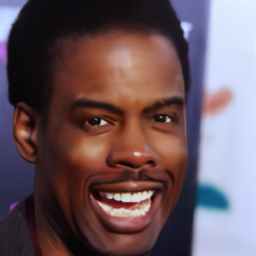}{0.34}{0.48} & \BAMPaperZoom{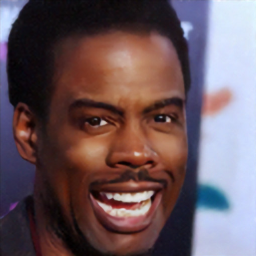}{0.34}{0.48} & \BAMPaperZoom{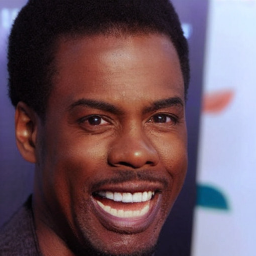}{0.34}{0.48} & \BAMPaperZoom{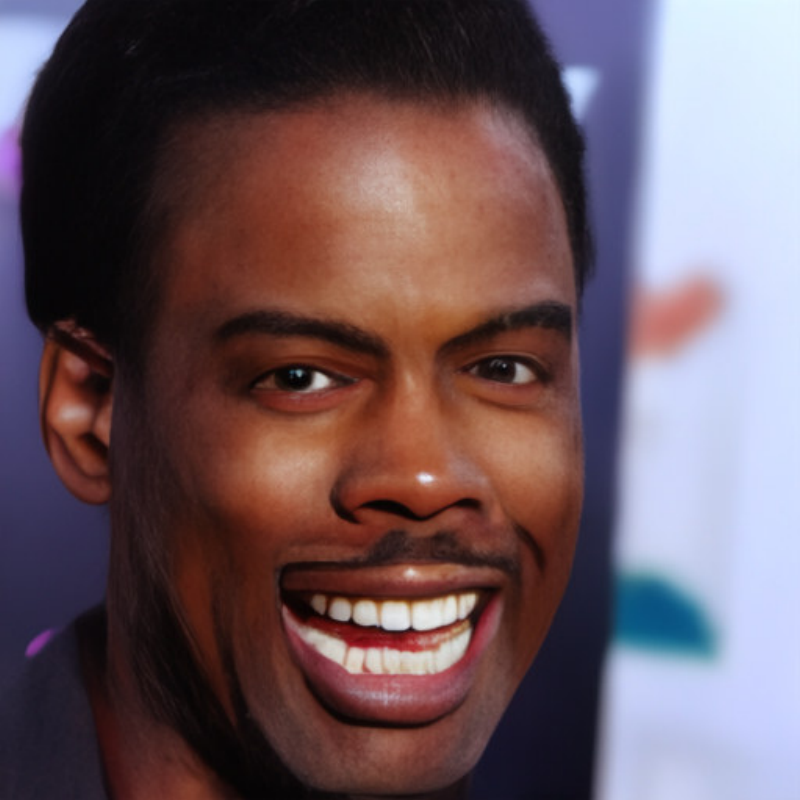}{0.34}{0.48} & \BAMPaperZoom{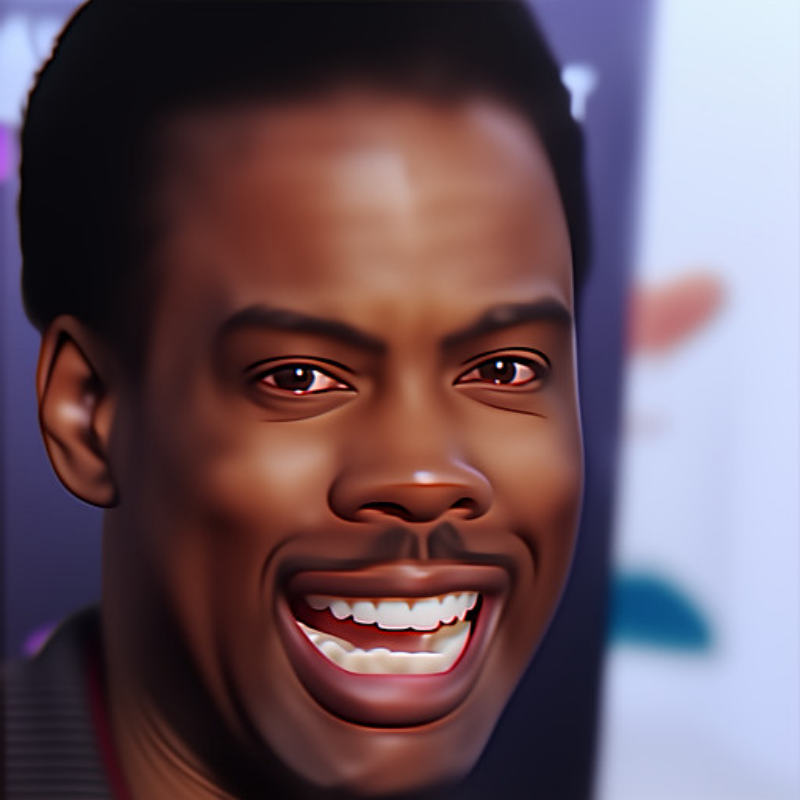}{0.34}{0.48} & \BAMPaperZoom{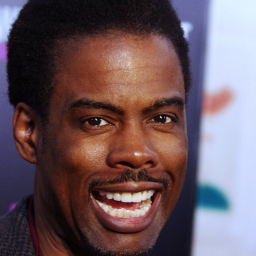}{0.34}{0.48} \\[3pt]
\multicolumn{9}{@{}l}{\textbf{FFHQ --- JPEG, $Q=10$}, pre-compression $\sigma_\text{JPEG}=0.01$} \\[1pt]
\BAMPaperZoom{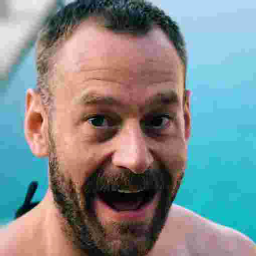}{0.39}{0.46} & \BAMPaperZoom{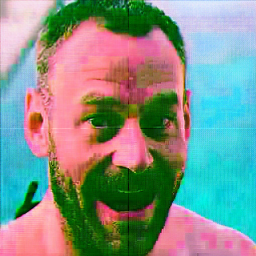}{0.39}{0.46} & \BAMPaperZoom{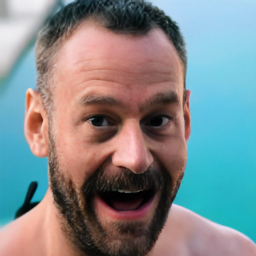}{0.39}{0.46} & \BAMPaperZoom{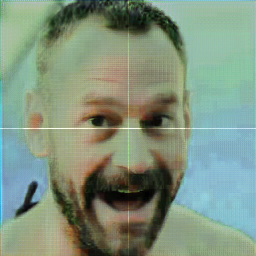}{0.39}{0.46} & \BAMPaperZoom{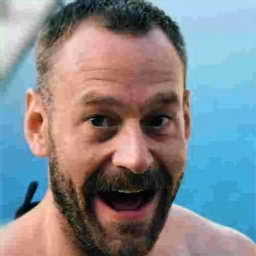}{0.39}{0.46} & \BAMPaperZoom{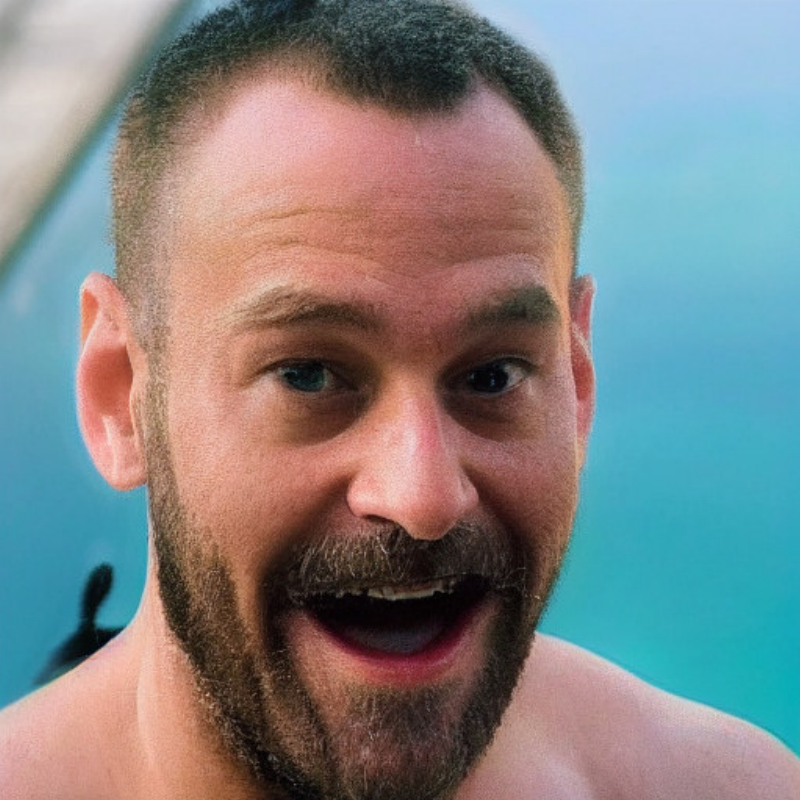}{.39}{.46} & \BAMPaperZoom{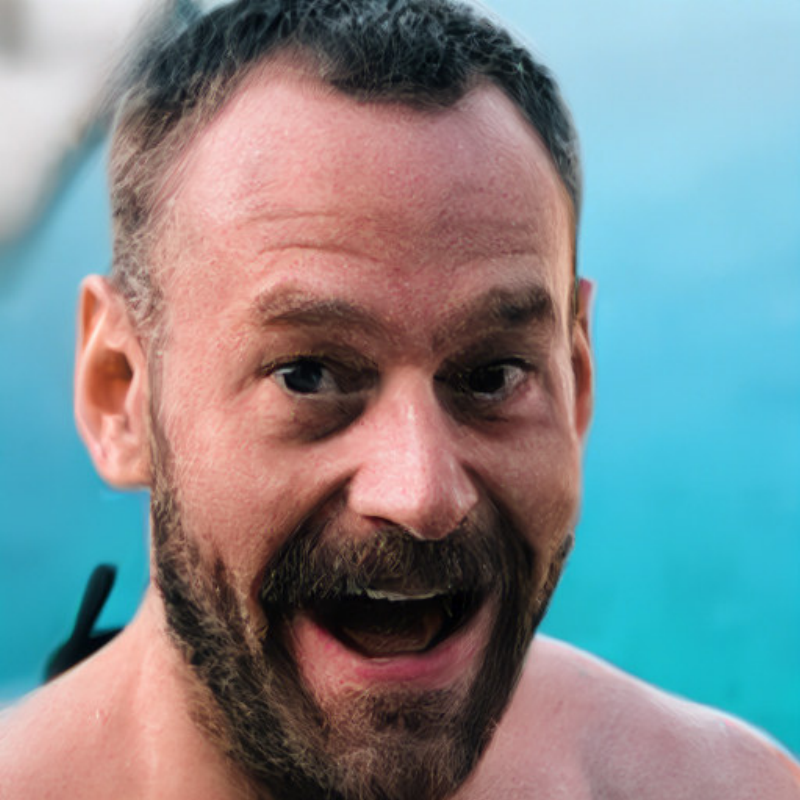}{0.39}{0.46} & \BAMPaperMissing & \BAMPaperZoom{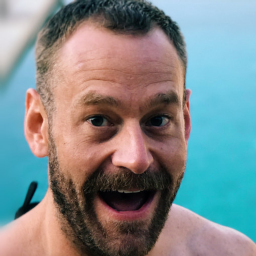}{0.39}{0.46} \\[3pt]
\bottomrule
\end{tabular}
\endgroup
\caption{\textbf{FFHQ restorations.} Deblurring and super-resolution at $\sigma_y = 0.05$; JPEG at $Q = 10$ with pre-compression noise $\sigma_{\mathrm{JPEG}} = 0.01$. BAM and \BAMFT use three steps. Strips show $4\times$ zoom; -- marks unavailable results. More problems and comparisons in Figure~\ref{fig:compact-FFHQ}.}
\label{fig:main-FFHQ}
\end{figure}

\begin{figure}[h]
\centering
\begingroup
\setlength{\BAMPaperImageWidth}{.093\linewidth}
\setlength{\tabcolsep}{1pt}
\scriptsize
\begin{tabular}{@{}cccccccccc@{}}
\toprule
\textbf{Observation} & \textbf{BAM} & \textbf{\BAMFT} & \textbf{RAM} & \textbf{\RAMFT} & \textbf{LATINO} & \textbf{L-PRO} & \textbf{TReg} & \textbf{SILO} & \textbf{GT} \\
\midrule
\multicolumn{10}{@{}l}{\textbf{AFHQ --- Inpainting}, $\sigma_y=0.05$} \\[1pt]
\BAMPaperZoom{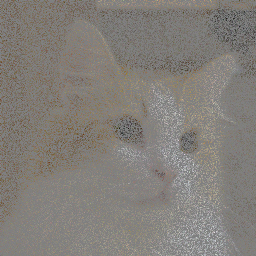}{0.37}{0.44} & \BAMPaperZoom{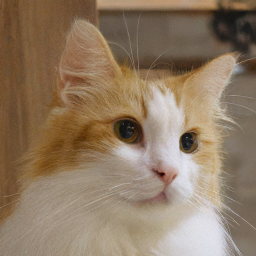}{0.37}{0.44} & \BAMPaperZoom{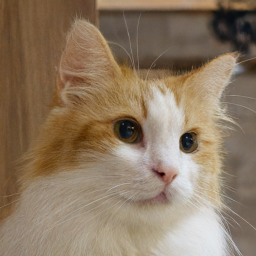}{0.37}{0.44} & \BAMPaperZoom{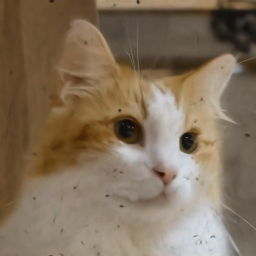}{0.37}{0.44} & \BAMPaperZoom{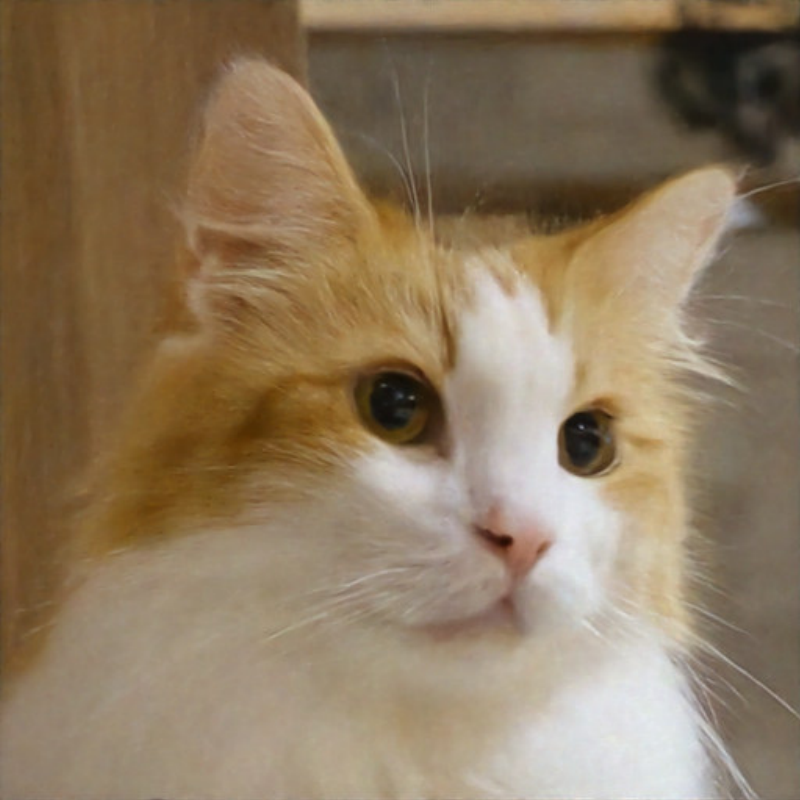}{0.37}{0.44} & \BAMPaperZoom{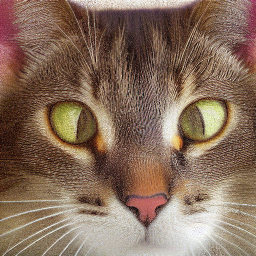}{0.37}{0.44} & \BAMPaperZoom{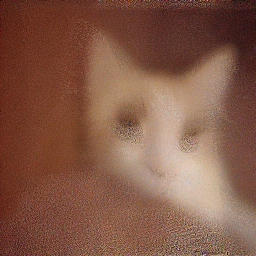}{0.37}{0.44} & \BAMPaperZoom{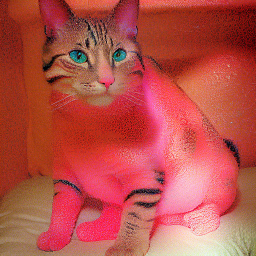}{0.37}{0.44} & \BAMPaperZoom{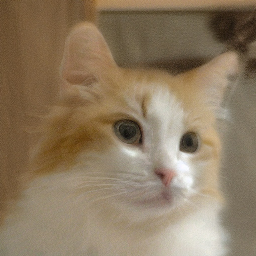}{0.37}{0.44} & \BAMPaperZoom{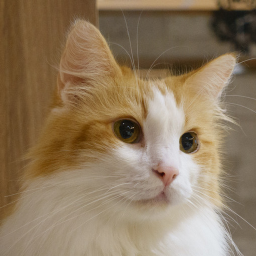}{0.37}{0.44} \\[3pt]
\multicolumn{10}{@{}l}{\textbf{AFHQ --- Demosaicing}, $\sigma_y=0.05$} \\[1pt]
\BAMPaperZoom{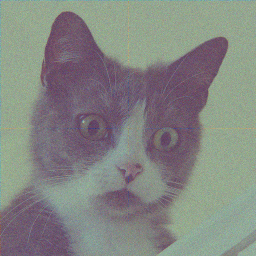}{0.24}{0.45} & \BAMPaperZoom{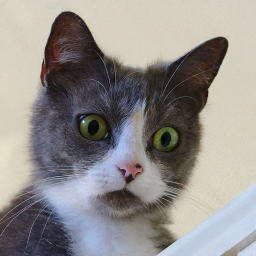}{0.24}{0.45} & \BAMPaperZoom{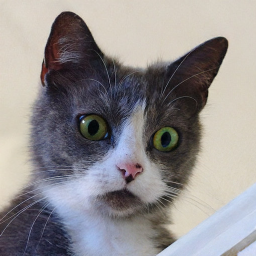}{0.24}{0.45} & \BAMPaperZoom{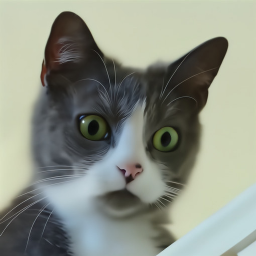}{0.24}{0.45} & \BAMPaperZoom{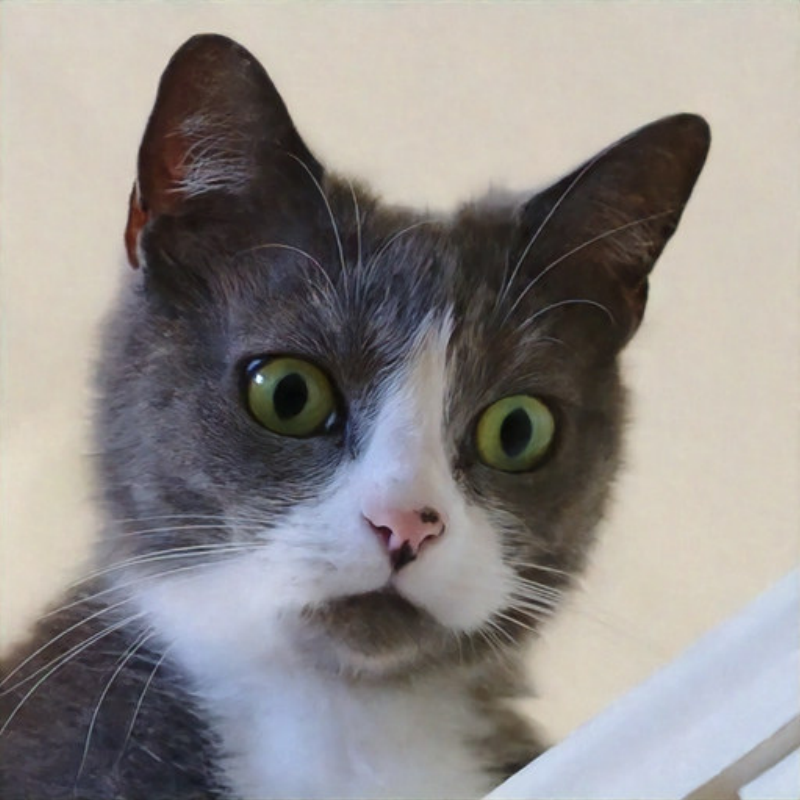}{0.24}{0.45} & \BAMPaperZoom{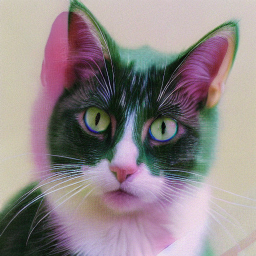}{0.24}{0.45} & \BAMPaperZoom{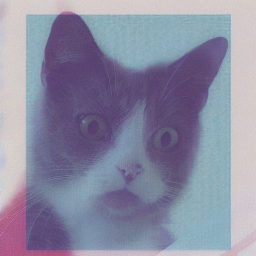}{0.24}{0.45} & \BAMPaperZoom{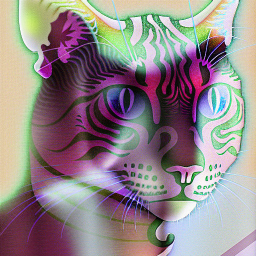}{0.24}{0.45} & \BAMPaperZoom{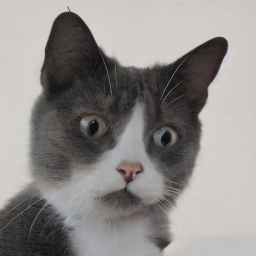}{0.24}{0.45} & \BAMPaperZoom{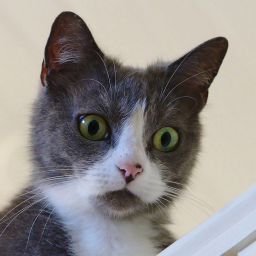}{0.24}{0.45} \\[3pt]
\multicolumn{10}{@{}l}{\textbf{AFHQ --- Compressed sensing}, $\sigma_y=0.05$} \\[1pt]
\BAMPaperZoom{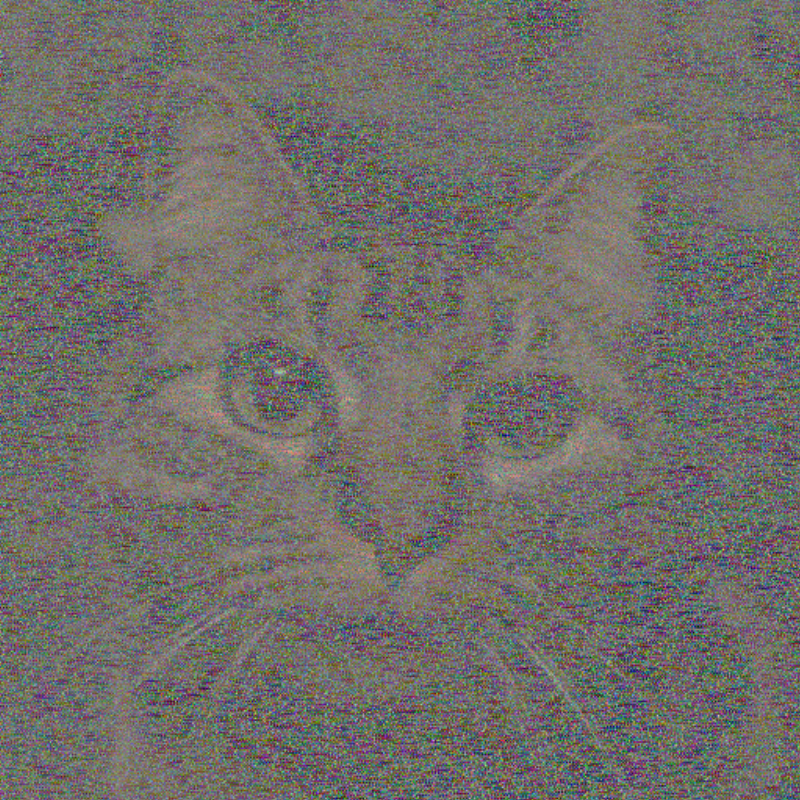}{0.22}{0.46} & \BAMPaperZoom{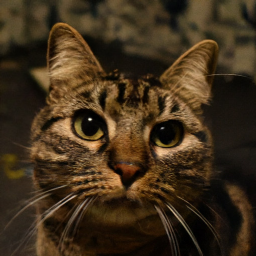}{0.22}{0.46} & \BAMPaperZoom{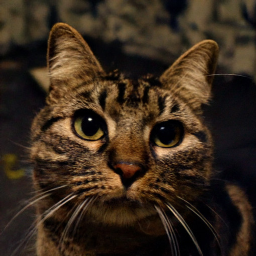}{0.22}{0.46} & \BAMPaperZoom{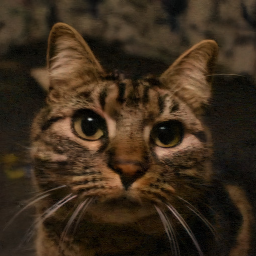}{0.22}{0.46} & \BAMPaperZoom{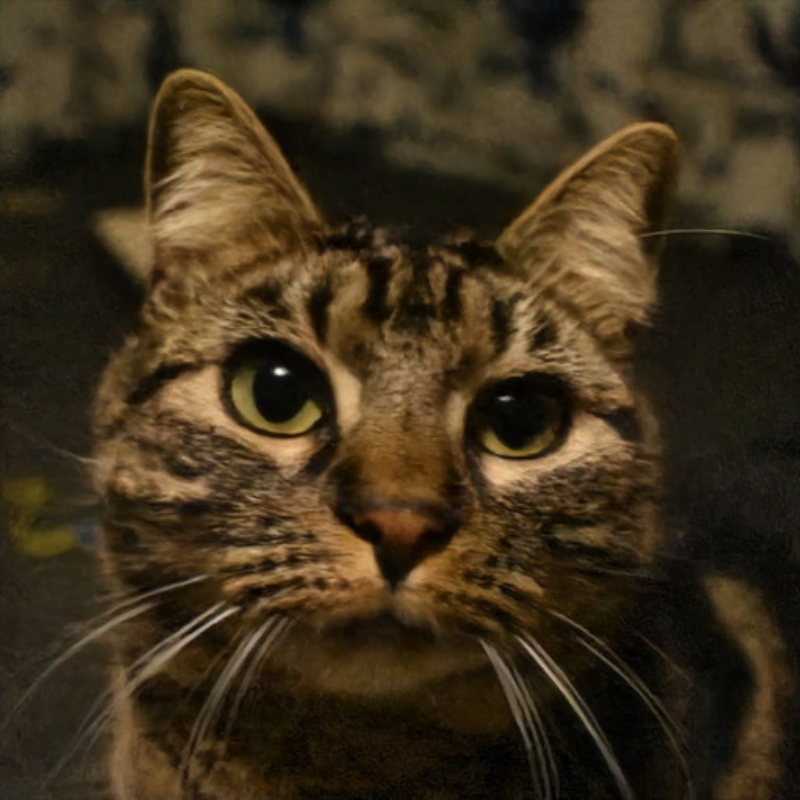}{0.22}{0.46} & \BAMPaperZoom{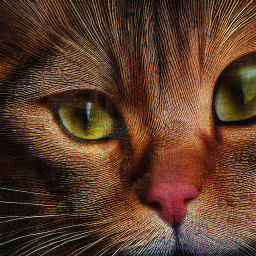}{0.22}{0.46} & \BAMPaperZoom{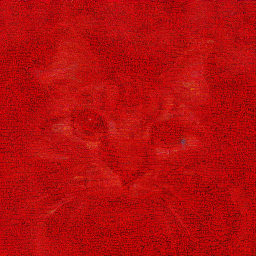}{0.22}{0.46} & \BAMPaperZoom{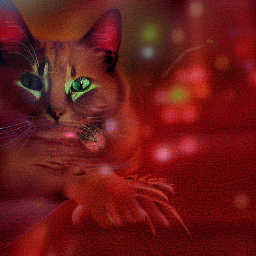}{0.22}{0.46} & \BAMPaperMissing & \BAMPaperZoom{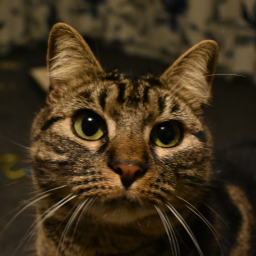}{0.22}{0.46} \\[3pt]
\bottomrule
\end{tabular}
\endgroup
\caption{\textbf{AFHQ restorations} at $\sigma_y = 0.05$: inpainting, demosaicing, compressed sensing. BAM and \BAMFT use three steps. Strips show $4\times$ zoom. For compressed sensing, the observation column shows $A^\dagger y$; -- marks unavailable results. More examples in Figure~\ref{fig:compact-AFHQ}.}
\label{fig:main-AFHQ}
\end{figure}

\section{Conclusion and limitations}
We introduced BAM, a 36M-parameter physics-aware generative foundation model for Bayesian computational imaging that generalizes robustly to unseen data and tasks, zero-shot or with minimal finetuning. BAM upgrades the operator-conditioned RAM backbone into a conditional flow map or consistency model, pre-trained jointly on large image corpora and libraries of forward operators. The forward operator is an input to the network, so instrument physics is specified at inference time. A single lightweight network then draws posterior samples without likelihood approximations, guidance weights or task-specific retraining. Across linear inverse problems on FFHQ, AFHQ, LSUN, DIV2K and Köhler, BAM outperformed both specialised models and leading zero-shot methods in just $3$ steps, at a fraction of their computational cost.

BAM has four main limitations. It handles only additive Gaussian noise and linear or mildly nonlinear forward operators. It is non-blind, i.e., the instrument model must be supplied, so blind problems rely on an external estimate of the operator. It does not detect model misspecification, so under strong distribution shift it can return unreliable posteriors without warning. Finally, posterior sample quality was assessed empirically, and formal guarantees for the learned conditional flow map remain open.

BAM brings generative Bayesian imaging within reach of the global computational imaging community. The natural next applications are scientific and medical imaging, such as MRI, CT, microscopy and astronomy, where forward operators are known and ground truth is scarce. In these domains, posterior samples support Bayesian decision theory, optimal estimation, uncertainty quantification and experimental design, adding significant value to existing point estimation approaches. On the methodological side, the priorities are formal convergence and accuracy guarantees, non-Gaussian noise models, nonlinear and partially unknown forward models, larger operator libraries, automatic out-of-distribution detection, and more sophisticated backbones. The community can then fine-tune BAM instead of training from scratch, widening access and reducing the economic and environmental cost of deep learning for imaging.

\section{Acknowledgments}

MP gratefully acknowledges support from UKRI Engineering and Physical Sciences Research Council (EPSRC) (EP/Z534481/1). AS and AA acknowledge support from the France 2030 research program on artificial intelligence via the PEPR PDE-AI grant (ANR-23-PEIA-0004).
The authors heavily rely on the \href{https://nova.mi.parisdescartes.fr/}{Nova HPC platform} of UFR Math-Info and MAP5 lab at Université Paris Cité. We are grateful to Azedine Mani and to Arnaud Meunier from the IT support unit for their technical help in maintaining the cluster nodes and assisting with software environment configuration. We acknowledge the use of the HWU high-performance computing facility (DMOG) and associated support services in the completion of this work. Additional HPC resources provided by \href{https://www.genci.fr/en/institut-du-developpement-et-des-ressources-en-informatique-scientifique-idris}{GENCI-IDRIS Jean-Zay} (Grants 
2024-AD011014557R3 and 2026-AD011017639).
We are grateful to Jean-Zay's IT support team for their assistance.

\bibliography{references}

\begin{thebibliography}{45}
\providecommand{\natexlab}[1]{#1}
\providecommand{\url}[1]{\texttt{#1}}
\expandafter\ifx\csname urlstyle\endcsname\relax
  \providecommand{\doi}[1]{doi: #1}\else
  \providecommand{\doi}{doi: \begingroup \urlstyle{rm}\Url}\fi

\bibitem[Agustsson \& Timofte(2017)Agustsson and Timofte]{agustsson2017ntire}
Eirikur Agustsson and Radu Timofte.
\newblock {NTIRE} 2017 challenge on single image super-resolution: Dataset and study.
\newblock In \emph{Proceedings of the IEEE Conference on Computer Vision and Pattern Recognition (CVPR) Workshops}, pp.\  126--135, 2017.
\newblock \href{https://openaccess.thecvf.com/content_cvpr_2017_workshops/w12/html/Agustsson_NTIRE_2017_Challenge_CVPR_2017_paper.html}{PDF}.

\bibitem[Albergo et~al.(2025)Albergo, Boffi, and Vanden-Eijnden]{albergo2023stochastic}
Michael~S Albergo, Nicholas~M Boffi, and Eric Vanden-Eijnden.
\newblock Stochastic interpolants: A unifying framework for flows and diffusions.
\newblock \emph{Journal of Machine Learning Research}, 26\penalty0 (209):\penalty0 1--80, 2025.
\newblock \href{https://arxiv.org/abs/2303.08797}{arXiv:2303.08797}.
\newblock \href{https://jmlr.org/papers/v26/23-1605.html}{PDF}.

\bibitem[Boffi et~al.(2025)Boffi, Albergo, and Vanden-Eijnden]{boffi2025selfdistill}
Nicholas~M. Boffi, Michael~S. Albergo, and Eric Vanden-Eijnden.
\newblock How to build a consistency model: Learning flow maps via self-distillation.
\newblock In \emph{Advances in Neural Information Processing Systems ({NeurIPS})}, 2025.
\newblock \href{https://arxiv.org/abs/2505.18825}{arXiv:2505.18825}.
\newblock \href{https://openreview.net/forum?id=Di5apl8HSH}{PDF}.

\bibitem[Carbajal et~al.(2023)Carbajal, Vitoria, Lezama, and Mus{\'e}]{carbajal2023blind}
Guillermo Carbajal, Patricia Vitoria, Jos{\'e} Lezama, and Pablo Mus{\'e}.
\newblock Blind motion deblurring with pixel-wise kernel estimation via kernel prediction networks.
\newblock \emph{IEEE Transactions on Computational Imaging}, 9:\penalty0 928--943, 2023.
\newblock \href{https://doi.org/10.1109/TCI.2023.3322012}{doi:10.1109/TCI.2023.3322012}.
\newblock \href{https://arxiv.org/abs/2308.02947}{arXiv:2308.02947}.

\bibitem[Choi et~al.(2020)Choi, Uh, Yoo, and Ha]{choi2020stargan}
Yunjey Choi, Youngjung Uh, Jaejun Yoo, and Jung-Woo Ha.
\newblock {StarGAN v2}: Diverse image synthesis for multiple domains.
\newblock In \emph{Proceedings of the IEEE/CVF Conference on Computer Vision and Pattern Recognition (CVPR)}, pp.\  8188--8197, 2020.
\newblock \href{https://arxiv.org/abs/1912.01865}{arXiv:1912.01865}.

\bibitem[Chung et~al.(2023)Chung, Kim, McCann, Klasky, and Ye]{chung2023dps}
Hyungjin Chung, Jeongsol Kim, Michael~T. McCann, Marc~L. Klasky, and Jong~Chul Ye.
\newblock Diffusion posterior sampling for general noisy inverse problems.
\newblock In \emph{The Eleventh International Conference on Learning Representations ({ICLR})}, 2023.
\newblock \href{https://arxiv.org/abs/2209.14687}{arXiv:2209.14687}.

\bibitem[Elata et~al.(2025)Elata, Chung, Ye, Michaeli, and Elad]{Elata2025InvFusion}
Noam Elata, Hyungjin Chung, Jong~Chul Ye, Tomer Michaeli, and Michael Elad.
\newblock {InvFusion}: Bridging supervised and zero-shot diffusion for inverse problems.
\newblock In \emph{Advances in Neural Information Processing Systems (NeurIPS)}, 2025.
\newblock \href{https://arxiv.org/abs/2504.01689}{arXiv:2504.01689}.

\bibitem[Gao et~al.(2025)Gao, Hoogeboom, Heek, Bortoli, Murphy, and Salimans]{gao2025diffusion}
Ruiqi Gao, Emiel Hoogeboom, Jonathan Heek, Valentin~De Bortoli, Kevin~Patrick Murphy, and Tim Salimans.
\newblock Diffusion models and gaussian flow matching: Two sides of the same coin.
\newblock In \emph{The Fourth Blogpost Track at ICLR 2025}, 2025.
\newblock \href{https://openreview.net/forum?id=C8Yyg9wy0s}{PDF}.

\bibitem[Garber \& Tirer(2025)Garber and Tirer]{Garber_2025_CVPR}
Tomer Garber and Tom Tirer.
\newblock Zero-shot image restoration using few-step guidance of consistency models (and beyond).
\newblock In \emph{Proceedings of the Computer Vision and Pattern Recognition Conference (CVPR)}, pp.\  2398--2407, June 2025.
\newblock \href{https://arxiv.org/abs/2412.20596}{arXiv:2412.20596}.

\bibitem[Gilton et~al.(2021)Gilton, Ongie, and Willett]{Gilton2021DEQ}
Davis Gilton, Gregory Ongie, and Rebecca Willett.
\newblock Deep equilibrium architectures for inverse problems in imaging.
\newblock \emph{IEEE Transactions on Computational Imaging}, 7:\penalty0 1123--1133, 2021.
\newblock \href{https://doi.org/10.1109/TCI.2021.3118944}{doi:10.1109/TCI.2021.3118944}.
\newblock \href{https://arxiv.org/abs/2102.07944}{arXiv:2102.07944}.

\bibitem[Heusel et~al.(2017)Heusel, Ramsauer, Unterthiner, Nessler, and Hochreiter]{heusel2017gans}
Martin Heusel, Hubert Ramsauer, Thomas Unterthiner, Bernhard Nessler, and Sepp Hochreiter.
\newblock {GAN}s trained by a two time-scale update rule converge to a local {N}ash equilibrium.
\newblock In \emph{Advances in Neural Information Processing Systems}, volume~30, pp.\  6626--6637, 2017.
\newblock \href{https://arxiv.org/abs/1706.08500}{arXiv:1706.08500}.

\bibitem[Ho et~al.(2020)Ho, Jain, and Abbeel]{Ho2020Denoising}
Jonathan Ho, Ajay Jain, and Pieter Abbeel.
\newblock Denoising diffusion probabilistic models.
\newblock In \emph{Advances in Neural Information Processing Systems}, volume~33, pp.\  6840--6851, 2020.
\newblock \href{https://arxiv.org/abs/2006.11239}{arXiv:2006.11239}.

\bibitem[Jayasumana et~al.(2024)Jayasumana, Ramalingam, Veit, Glasner, Chakrabarti, and Kumar]{jayasumana2024rethinking}
Sadeep Jayasumana, Srikumar Ramalingam, Andreas Veit, Daniel Glasner, Ayan Chakrabarti, and Sanjiv Kumar.
\newblock Rethinking {FID}: Towards a better evaluation metric for image generation.
\newblock In \emph{Proceedings of the IEEE/CVF Conference on Computer Vision and Pattern Recognition (CVPR)}, pp.\  9307--9315, 2024.
\newblock \href{https://arxiv.org/abs/2401.09603}{arXiv:2401.09603}.

\bibitem[Kamilov et~al.(2023)Kamilov, Bouman, Buzzard, and Wohlberg]{Kamilov2023PnP}
Ulugbek~S. Kamilov, Charles~A. Bouman, Gregery~T. Buzzard, and Brendt Wohlberg.
\newblock Plug-and-play methods for integrating physical and learned models in computational imaging: Theory, algorithms, and applications.
\newblock \emph{IEEE Signal Processing Magazine}, 40\penalty0 (1):\penalty0 85--97, 2023.
\newblock \href{https://doi.org/10.1109/MSP.2022.3199595}{doi:10.1109/MSP.2022.3199595}.
\newblock \href{https://arxiv.org/abs/2203.17061}{arXiv:2203.17061}.

\bibitem[Karras et~al.(2019)Karras, Laine, and Aila]{Karras2018ASG}
Tero Karras, Samuli Laine, and Timo Aila.
\newblock A style-based generator architecture for generative adversarial networks.
\newblock In \emph{Proceedings of the IEEE/CVF Conference on Computer Vision and Pattern Recognition (CVPR)}, pp.\  4396--4405, 2019.
\newblock \href{https://arxiv.org/abs/1812.04948}{arXiv:1812.04948}.

\bibitem[Kawar et~al.(2022)Kawar, Elad, Ermon, and Song]{kawar2022denoising}
Bahjat Kawar, Michael Elad, Stefano Ermon, and Jiaming Song.
\newblock Denoising diffusion restoration models.
\newblock In \emph{Advances in Neural Information Processing Systems}, volume~35, pp.\  23593--23606, 2022.
\newblock \href{https://arxiv.org/abs/2201.11793}{arXiv:2201.11793}.

\bibitem[Kim et~al.(2023)Kim, Lai, Liao, Murata, Takida, Uesaka, He, Mitsufuji, and Ermon]{kim2023consistency}
Dongjun Kim, Chieh-Hsin Lai, Wei-Hsiang Liao, Naoki Murata, Yuhta Takida, Toshimitsu Uesaka, Yutong He, Yuki Mitsufuji, and Stefano Ermon.
\newblock Consistency trajectory models: Learning probability flow ode trajectory of diffusion.
\newblock \emph{arXiv preprint arXiv:2310.02279}, 2023.

\bibitem[Kim et~al.(2025)Kim, Park, Chung, and Ye]{Kim2023RegularizationBT}
Jeongsol Kim, Geon~Yeong Park, Hyungjin Chung, and Jong~Chul Ye.
\newblock Regularization by texts for latent diffusion inverse solvers.
\newblock In \emph{The Thirteenth International Conference on Learning Representations}, 2025.
\newblock \href{https://arxiv.org/abs/2311.15658}{arXiv:2311.15658}.
\newblock \href{https://openreview.net/forum?id=TtUh0TOlGX}{PDF}.

\bibitem[K{\"o}hler et~al.(2012)K{\"o}hler, Hirsch, Mohler, Sch{\"o}lkopf, and Harmeling]{kohler2012recording}
Rolf K{\"o}hler, Michael Hirsch, Betty Mohler, Bernhard Sch{\"o}lkopf, and Stefan Harmeling.
\newblock Recording and playback of camera shake: Benchmarking blind deconvolution with a real-world database.
\newblock In \emph{European Conference on Computer Vision}, pp.\  27--40. Springer, 2012.

\bibitem[Laumont et~al.(2022)Laumont, De~Bortoli, Almansa, Delon, Durmus, and Pereyra]{laumont2022pnpula}
R{\'e}mi Laumont, Valentin De~Bortoli, Andr{\'e}s Almansa, Julie Delon, Alain Durmus, and Marcelo Pereyra.
\newblock Bayesian imaging using plug \& play priors: When {Langevin} meets {Tweedie}.
\newblock \emph{SIAM Journal on Imaging Sciences}, 15\penalty0 (2):\penalty0 701--737, 2022.
\newblock \href{https://doi.org/10.1137/21M1406349}{doi:10.1137/21M1406349}.
\newblock \href{https://arxiv.org/abs/2103.04715}{arXiv:2103.04715}.

\bibitem[Li et~al.(2023)Li, Zhang, Liang, Cao, Liu, Gong, Zhang, Tang, Liu, Demandolx, Ranjan, Timofte, and Van~Gool]{li2023lsdir}
Yawei Li, Kai Zhang, Jingyun Liang, Jiezhang Cao, Ce~Liu, Rui Gong, Yulun Zhang, Hao Tang, Yun Liu, Denis Demandolx, Rakesh Ranjan, Radu Timofte, and Luc Van~Gool.
\newblock {LSDIR}: A large scale dataset for image restoration.
\newblock In \emph{Proceedings of the IEEE/CVF Conference on Computer Vision and Pattern Recognition (CVPR) Workshops}, pp.\  1775--1787, 2023.

\bibitem[Lipman et~al.(2023)Lipman, Chen, Ben-Hamu, Nickel, and Le]{lipman2023flow}
Yaron Lipman, Ricky T.~Q. Chen, Heli Ben-Hamu, Maximilian Nickel, and Matthew Le.
\newblock Flow matching for generative modeling.
\newblock In \emph{International Conference on Learning Representations (ICLR)}, 2023.
\newblock \href{https://arxiv.org/abs/2210.02747}{arXiv:2210.02747}.
\newblock \href{https://openreview.net/forum?id=PqvMRDCJT9t}{PDF}.

\bibitem[Liu et~al.(2023{\natexlab{a}})Liu, Vahdat, Huang, Theodorou, Nie, and Anandkumar]{Liu2023I2SBIS}
Guan-Horng Liu, Arash Vahdat, De-An Huang, Evangelos~A. Theodorou, Weili Nie, and Anima Anandkumar.
\newblock I2sb: Image-to-image schr{\"o}dinger bridge.
\newblock In \emph{Proceedings of the 40th International Conference on Machine Learning (ICML)}, volume 202 of \emph{Proceedings of Machine Learning Research}, pp.\  22042--22062. PMLR, 2023{\natexlab{a}}.
\newblock \href{https://arxiv.org/abs/2302.05872}{arXiv:2302.05872}.

\bibitem[Liu et~al.(2023{\natexlab{b}})Liu, Gong, and Liu]{liu2023flow}
Xingchao Liu, Chengyue Gong, and Qiang Liu.
\newblock Flow straight and fast: Learning to generate and transfer data with rectified flow.
\newblock In \emph{International Conference on Learning Representations (ICLR)}, 2023{\natexlab{b}}.
\newblock \href{https://arxiv.org/abs/2209.03003}{arXiv:2209.03003}.
\newblock \href{https://openreview.net/forum?id=XVjTT1nw5z}{PDF}.

\bibitem[Luo et~al.(2023)Luo, Tan, Huang, Li, and Zhao]{Luo2023LatentCM}
Simian Luo, Yiqin Tan, Longbo Huang, Jian Li, and Hang Zhao.
\newblock Latent consistency models: Synthesizing high-resolution images with few-step inference, 2023.
\newblock \href{https://arxiv.org/abs/2310.04378}{arXiv:2310.04378}.

\bibitem[Mbakam et~al.(2025)Mbakam, Spence, and Pereyra]{Mbakam2025LearningFP}
Charlesquin~Kemajou Mbakam, Jonathan Spence, and Marcelo Pereyra.
\newblock Learning few-step posterior samplers by unfolding and distillation of diffusion models.
\newblock \emph{Transactions on Machine Learning Research}, 2025.
\newblock ISSN 2835-8856.
\newblock \href{https://arxiv.org/abs/2507.02686}{arXiv:2507.02686}.
\newblock \href{https://openreview.net/forum?id=oGCfD8YKN2}{PDF}.

\bibitem[Monga et~al.(2021)Monga, Li, and Eldar]{Monga2021Unrolling}
Vishal Monga, Yuelong Li, and Yonina~C. Eldar.
\newblock Algorithm unrolling: Interpretable, efficient deep learning for signal and image processing.
\newblock \emph{IEEE Signal Processing Magazine}, 38\penalty0 (2):\penalty0 18--44, 2021.
\newblock \href{https://doi.org/10.1109/MSP.2020.3016905}{doi:10.1109/MSP.2020.3016905}.
\newblock \href{https://arxiv.org/abs/1912.10557}{arXiv:1912.10557}.

\bibitem[Noble et~al.(2026)Noble, Quintana, Aubin, and Chadebec]{noble2026fastfaithfulphotorealisticdiffusionbased}
Maxence Noble, Gonzalo~Iñaki Quintana, Benjamin Aubin, and Clément Chadebec.
\newblock Fast, faithful and photorealistic diffusion-based image super-resolution with enhanced flow map models, 2026.
\newblock \href{https://arxiv.org/abs/2601.16660}{arXiv:2601.16660}.

\bibitem[Podell et~al.(2024)Podell, English, Lacey, Blattmann, Dockhorn, M{\"u}ller, Penna, and Rombach]{Podell2023SDXLIL}
Dustin Podell, Zion English, Kyle Lacey, Andreas Blattmann, Tim Dockhorn, Jonas M{\"u}ller, Joe Penna, and Robin Rombach.
\newblock Sdxl: Improving latent diffusion models for high-resolution image synthesis.
\newblock In \emph{The Twelfth International Conference on Learning Representations (ICLR)}, 2024.
\newblock \href{https://arxiv.org/abs/2307.01952}{arXiv:2307.01952}.

\bibitem[Raphaeli et~al.(2025)Raphaeli, Man, and Elad]{Raphaeli_2025_ICCV}
Ron Raphaeli, Sean Man, and Michael Elad.
\newblock {SILO}: Solving inverse problems with latent operators.
\newblock In \emph{Proceedings of the IEEE/CVF International Conference on Computer Vision (ICCV)}, pp.\  10570--10580, October 2025.
\newblock \href{https://arxiv.org/abs/2501.11746}{arXiv:2501.11746}.
\newblock \href{https://openaccess.thecvf.com/content/ICCV2025/html/Raphaeli_SILO_Solving_Inverse_Problems_with_Latent_Operators_ICCV_2025_paper.html}{PDF}.

\bibitem[Romano et~al.(2017)Romano, Elad, and Milanfar]{romano2017little}
Yaniv Romano, Michael Elad, and Peyman Milanfar.
\newblock The little engine that could: Regularization by denoising (red).
\newblock \emph{SIAM Journal on Imaging Sciences}, 10\penalty0 (4):\penalty0 1804--1844, 2017.
\newblock \href{https://doi.org/10.1137/16M1102884}{doi:10.1137/16M1102884}.
\newblock \href{https://arxiv.org/abs/1611.02862}{arXiv:1611.02862}.

\bibitem[Rombach et~al.(2022)Rombach, Blattmann, Lorenz, Esser, and Ommer]{rombach2022ldm}
Robin Rombach, Andreas Blattmann, Dominik Lorenz, Patrick Esser, and Bj\"orn Ommer.
\newblock High-resolution image synthesis with latent diffusion models.
\newblock In \emph{Proceedings of the IEEE/CVF Conference on Computer Vision and Pattern Recognition (CVPR)}, pp.\  10684--10695, June 2022.
\newblock \href{https://arxiv.org/abs/2112.10752}{arXiv:2112.10752}.

\bibitem[Song \& Dhariwal(2024)Song and Dhariwal]{songdhariwal2023improved}
Yang Song and Prafulla Dhariwal.
\newblock Improved techniques for training consistency models.
\newblock In \emph{International Conference on Learning Representations (ICLR)}, 2024.
\newblock \href{https://arxiv.org/abs/2310.14189}{arXiv:2310.14189}.

\bibitem[Song et~al.(2021)Song, Sohl-Dickstein, Kingma, Kumar, Ermon, and Poole]{song2021sde}
Yang Song, Jascha Sohl-Dickstein, Diederik~P. Kingma, Abhishek Kumar, Stefano Ermon, and Ben Poole.
\newblock Score-based generative modeling through stochastic differential equations.
\newblock In \emph{International Conference on Learning Representations (ICLR)}, 2021.
\newblock \href{https://arxiv.org/abs/2011.13456}{arXiv:2011.13456}.

\bibitem[Song et~al.(2023)Song, Dhariwal, Chen, and Sutskever]{song2023consistency}
Yang Song, Prafulla Dhariwal, Mark Chen, and Ilya Sutskever.
\newblock Consistency models.
\newblock In Andreas Krause, Emma Brunskill, Kyunghyun Cho, Barbara Engelhardt, Sivan Sabato, and Jonathan Scarlett (eds.), \emph{Proceedings of the 40th International Conference on Machine Learning}, volume 202 of \emph{Proceedings of Machine Learning Research}, pp.\  32211--32252. PMLR, 23--29 Jul 2023.
\newblock \href{https://arxiv.org/abs/2303.01469}{arXiv:2303.01469}.
\newblock \href{https://proceedings.mlr.press/v202/song23a.html}{PDF}.

\bibitem[Spagnoletti et~al.(2025)Spagnoletti, Prost, Almansa, Papadakis, and Pereyra]{Spagnoletti_2025_ICCV}
Alessio Spagnoletti, Jean Prost, Andr\'es Almansa, Nicolas Papadakis, and Marcelo Pereyra.
\newblock Latino-pro: Latent consistency inverse solver with prompt optimization.
\newblock In \emph{Proceedings of the IEEE/CVF International Conference on Computer Vision (ICCV)}, pp.\  19597--19607, October 2025.
\newblock \href{https://arxiv.org/abs/2503.12615}{arXiv:2503.12615}.

\bibitem[Spagnoletti et~al.(2026)Spagnoletti, Wang, Pereyra, and Akyildiz]{spagnoletti2026consistencyregularisedgradientflows}
Alessio Spagnoletti, Tim Y.~J. Wang, Marcelo Pereyra, and O.~Deniz Akyildiz.
\newblock Consistency regularised gradient flows for inverse problems, 2026.
\newblock \href{https://arxiv.org/abs/2605.07907}{arXiv:2605.07907}.

\bibitem[Tachella et~al.(2025)Tachella, Terris, Hurault, Wang, Chen, Nguyen, Song, Davies, Davy, Dong, Escande, Hertrich, Hu, Liaudat, Laurent, Levac, Massias, Moreau, Modrzyk, Monroy, Neumayer, Scanvic, Sarron, Sechaud, Schramm, Vo, and Weiss]{Tachella2025DeepInverseAP}
Juli{\'a}n Tachella, Matthieu Terris, Samuel Hurault, Andrew Wang, Dongdong Chen, Minh-Hai Nguyen, Maxime Song, Thomas Davies, Leo Davy, Jonathan Dong, Paul Escande, Johannes Hertrich, Zhiyuan Hu, Tob{\'i}as~I. Liaudat, Nils Laurent, Brett Levac, Mathurin Massias, Thomas Moreau, Thibaut Modrzyk, Brayan Monroy, Sebastian Neumayer, J{\'e}r{\'e}my Scanvic, Florian Sarron, Victor Sechaud, Georg Schramm, Romain Vo, and Pierre Weiss.
\newblock Deepinverse: A python package for solving imaging inverse problems with deep learning.
\newblock \emph{Journal of Open Source Software}, 10\penalty0 (115):\penalty0 8923, 2025.
\newblock \href{https://doi.org/10.21105/joss.08923}{doi:10.21105/joss.08923}.
\newblock \href{https://arxiv.org/abs/2505.20160}{arXiv:2505.20160}.

\bibitem[Terris et~al.(2026)Terris, Hurault, Song, and Tachella]{Terris2025ReconstructAM}
Matthieu Terris, Samuel Hurault, Maxime Song, and Juli{\'a}n Tachella.
\newblock Reconstruct anything model: a lightweight foundation model for computational imaging.
\newblock In \emph{International Conference on Learning Representations (ICLR)}, 2026.
\newblock \href{https://arxiv.org/abs/2503.08915}{arXiv:2503.08915}.

\bibitem[Venkatakrishnan et~al.(2013)Venkatakrishnan, Bouman, and Wohlberg]{Venkatakrishnan2013a}
Singanallur~V Venkatakrishnan, Charles~A Bouman, and Brendt Wohlberg.
\newblock {Plug-and-Play priors for model based reconstruction}.
\newblock In \emph{2013 IEEE Global Conference on Signal and Information Processing}, pp.\  945--948. IEEE, dec 2013.
\newblock ISBN 978-1-4799-0248-4.
\newblock \href{https://doi.org/10.1109/GlobalSIP.2013.6737048}{doi:10.1109/GlobalSIP.2013.6737048}.
\newblock \href{http://brendt.wohlberg.net/publications/pdf/venkatakrishnan-2013-plugandplay2.pdf}{PDF}.

\bibitem[Yang et~al.(2023)Yang, Chen, Tan, Liu, Chu, Bao, Yuan, Hua, and Yu]{yang2023hq50k}
Qinhong Yang, Dongdong Chen, Zhentao Tan, Qiankun Liu, Qi~Chu, Jianmin Bao, Lu~Yuan, Gang Hua, and Nenghai Yu.
\newblock {HQ-50K}: A large-scale, high-quality dataset for image restoration.
\newblock \emph{arXiv preprint arXiv:2306.05390}, 2023.

\bibitem[Yu et~al.(2015)Yu, Seff, Zhang, Song, Funkhouser, and Xiao]{yu2015lsun}
Fisher Yu, Ari Seff, Yinda Zhang, Shuran Song, Thomas Funkhouser, and Jianxiong Xiao.
\newblock {LSUN}: Construction of a large-scale image dataset using deep learning with humans in the loop.
\newblock \emph{arXiv preprint arXiv:1506.03365}, 2015.

\bibitem[Zhang et~al.(2018)Zhang, Isola, Efros, Shechtman, and Wang]{Zhang2018TheUE}
Richard Zhang, Phillip Isola, Alexei~A. Efros, Eli Shechtman, and Oliver Wang.
\newblock The unreasonable effectiveness of deep features as a perceptual metric.
\newblock In \emph{Proceedings of the IEEE/CVF Conference on Computer Vision and Pattern Recognition (CVPR)}, pp.\  586--595, 2018.
\newblock \href{https://arxiv.org/abs/1801.03924}{arXiv:1801.03924}.

\bibitem[Zhao et~al.(2024)Zhao, Song, and Shen]{zhao2024cosign}
Jiankun Zhao, Bowen Song, and Liyue Shen.
\newblock {CoSIGN}: Few-step guidance of {ConSIstency} model to solve general {INverse} problems.
\newblock In \emph{Computer Vision -- {ECCV} 2024}, volume 15106 of \emph{Lecture Notes in Computer Science}, pp.\  108--126. Springer, 2024.
\newblock \href{https://doi.org/10.1007/978-3-031-73195-2_7}{doi:10.1007/978-3-031-73195-2\_7}.
\newblock \href{https://arxiv.org/abs/2407.12676}{arXiv:2407.12676}.

\bibitem[Zhu et~al.(2026)Zhu, Huang, Xu, Li, Wu, Bai, Wu, Paul, and Tu]{zhu2026_4klsdb}
Zihao Zhu, Kuan-Ru Huang, Zhaoming Xu, Renjie Li, Bo~Wu, Ruizheng Bai, Mingyang Wu, Sayak Paul, and Zhengzhong Tu.
\newblock {4KLSDB}: A large-scale dataset for {4K} image restoration and text-to-image.
\newblock In \emph{Proceedings of the IEEE/CVF Conference on Computer Vision and Pattern Recognition (CVPR) Workshops}, 2026.
\newblock arXiv:2605.24762.

\end{thebibliography}
\bibliographystyle{paperstyle}

\appendix
\newpage
\section{Compared methods and related restoration models}\label{app:compared-methods}

We describe the methods used in our numerical comparisons and discuss two closely related approaches. The main distinctions concern the reconstruction target, the way the forward operator enters the model, and the computation required at inference time. The experimental adaptation of each baseline follows Section~\ref{sec:experimental-setup}.

\subsection{Methods used in the numerical comparisons}

\paragraph{RAM.}
The Reconstruct Anything Model~\citep{Terris2025ReconstructAM} is a non-iterative restoration network built on a DRUNet backbone. It incorporates the acquisition model through a proximal initialization and multiscale Krylov subspace modules, which combine learned features with applications of the forward and adjoint operators. Together with noise conditioning, this construction allows one network to address several imaging problems and to adapt to unseen operators. RAM is trained as a point estimator: under a squared-error objective, its population target is the conditional mean $\E[\vx\mid\vy,\mathcal A]$. It therefore provides a natural reference for reconstruction accuracy and for the empirical mean of repeated BAM samples. BAM retains this lightweight, physics-aware backbone but learns a conditional flow map with a random source, rather than a deterministic observation-to-image regressor.

\paragraph{TReg.}
Regularization by Texts (TReg)~\citep{Kim2023RegularizationBT} uses a pre-trained text-to-image latent DM (LDM)~\citep{rombach2022ldm} to resolve ambiguities in inverse problems. Its reverse diffusion procedure alternates data-consistency optimization with latent updates regularized by the text-conditioned clean-image estimate. An adaptive negation mechanism updates the null-text embedding used in classifier-free guidance, suppressing concepts inconsistent with the current reconstruction. The physical operator enters the inference-time optimization rather than the diffusion backbone. TReg thus provides a zero-shot, text-regularized comparison, with iterative diffusion and optimization costs that require hundreds of NFEs.

\paragraph{LATINO.}
LAtent consisTency INverse sOlver (LATINO)~\citep{Spagnoletti_2025_ICCV} is a zero-shot PnP inverse solver that uses a pre-trained latent consistency model (LCM)~\citep{Luo2023LatentCM} as an image prior. Its stochastic autoencoding construction combines a forward noising step with fast generative restoration, and incorporates the measurements through a proximal data-fidelity update in image space. This separates the supplied acquisition model from the learned prior and avoids differentiating through the generative network or the autoencoder for measurement conditioning. The resulting solver requires few neural evaluations, but still alternates a general-purpose latent generator with external data-consistency updates. BAM instead learns the measurement- and operator-conditioned map itself.

\paragraph{LATINO-PRO.}
LATINO-PRO~\citep{Spagnoletti_2025_ICCV} extends LATINO by estimating the text conditioning from the observation. It uses an empirical Bayesian formulation that maximizes the marginal likelihood of the measurements with respect to the prompt embedding, alternating prompt updates with stochastic reconstruction. This can correct an incomplete or misleading prompt, at the cost of additional computation for its calibration. We compare both variants to distinguish the contribution of the fast inverse solver from that of adapting the text-conditioned prior.

\paragraph{UD2M.}
The Unfolded and Distilled Diffusion Model (UD2M)~\citep{Mbakam2025LearningFP} converts a pre-trained diffusion prior into a few-step conditional sampler by unfolding the LATINO Langevin algorithm. Its trainable blocks retain explicit likelihood-dependent proximal updates and stochastic generative steps. A supervised distillation objective, inspired by consistency trajectory models~\citep{kim2023consistency} and combining distortion, perceptual and adversarial terms, adapts the unfolded network for posterior sampling. Importantly, UD2M supports joint training over families of likelihoods and specialization to the supplied measurement model at inference time. Relative to BAM, its central construction unfolds and distills an existing sampling algorithm, rather than implementing a conditional image flow map with a lightweight restoration backbone.

\paragraph{SILO.}
SILO (Solving Inverse Problems with Latent Operators)~\citep{Raphaeli_2025_ICCV} retains a pre-trained LDM prior and learns a surrogate for the degradation in latent space. The surrogate predicts the latent representation of the measurements, allowing likelihood guidance to operate without repeatedly decoding and re-encoding images. Sampling still differentiates through the latent denoiser and learned operator, and requires hundreds of NFEs. SILO is consequently a PnP method with an additional operator-learning stage, rather than a fully zero-shot solver or a directly trained conditional diffusion model. Its measurement conditioning relies on the learned latent surrogate, whereas BAM receives the physical operator through the forward and adjoint actions in its architecture.

\paragraph{I$^2$SB.}
I$^2$SB (Image-to-Image Schr\"odinger Bridge)~\citep{Liu2023I2SBIS} learns a diffusion bridge between clean images and their degraded counterparts. Given paired endpoints, its tractable bridge marginals permit simulation-free training, and reconstruction starts from the degraded image rather than unstructured Gaussian noise. This exploits the spatial information already present in the observation and supports stochastic image restoration. In its standard formulation, the degradation is represented by the training pairs and the degraded endpoint; an arbitrary forward operator is not supplied to the network as in BAM. Our comparison therefore contrasts operator-conditioned flow maps with a learned image-to-image diffusion bridge adapted to each evaluated restoration task.

\subsection{Related operator-conditioned and flow-map approaches}

\paragraph{InvFusion.}
InvFusion~\citep{Elata2025InvFusion} is a supervised, operator-conditioned diffusion posterior sampler. Its feature degradation layers apply the forward operator and its pseudo-inverse to internal activations and fuse the resulting measurement information through joint attention. Its main benchmarks emphasize strongly underdetermined problems with large null spaces: patch inpainting retains less than $10\%$ of the patches, while strided motion blur combines blurring with subsampling. In these settings, substantial image content must be generated in directions unconstrained by the measurements. Our emphasis is on a lightweight sampler across restoration problems where fidelity to the available measurements remains central, including noisy deblurring and demosaicing. Its HDiT backbone, with feature widths reaching $1536$ channels, joint-attention blocks, and a reported $63$-NFE sampling configuration, also entails a substantially heavier inference procedure than BAM's $36$M-parameter, three-step model. These architectural and sampling differences motivate our focus on compact conditional flow maps.

\paragraph{FlowMapSR.}
FlowMapSR~\citep{noble2026fastfaithfulphotorealisticdiffusionbased} is particularly close to our work in its use of flow maps for few-step restoration. It learns a transport from low-resolution to high-resolution image latents using paired data, and studies Lagrangian, Eulerian and Semigroup formulations, enhanced with perceptual training, positive-negative prompting and adversarial fine-tuning. The distinction is the source of the transport: FlowMapSR starts from the encoded low-resolution image and maps it to a high-resolution reconstruction. For a fixed source latent and guidance configuration, this flow defines a single output; it does not transport independent random inputs into a posterior distribution for that fixed observation. Moreover, the model is not conditioned on a supplied forward operator or noise model, and the same network is used across upscaling factors without degradation-guided mechanisms. Thus, despite the shared flow-map machinery, FlowMapSR addresses perceptual super-resolution through an image-to-image map, rather than operator-conditioned posterior sampling. BAM instead retains a random source while conditioning on the observation and acquisition physics.

\section{BAM! Architecture and modifications relative to RAM}\label{app:architecture}

 BAM builds on the Reconstruct Anything Model (RAM)~\citep{Terris2025ReconstructAM}, a lightweight network designed for a wide range of linear imaging inverse problems. It retains RAM's physics-aware encoder--decoder structure while adapting its conditioning mechanism to our flow formulation in \eqref{eq:bam-map}. In RAM, an initial reconstruction is obtained from $A^\top y$ and refined through a proximal step that enforces consistency with the measurements. Krylov-based features also provide information about the forward operator throughout the network.\par
To adapt RAM to our flow formulation, we additionally condition the network on the current flow state $x_t$ and the flow times $(s,t)$. In the final measurement-conditioned residual block, we denote by $\bm h_{y_\sigma}$ the physics-aware features extracted from the measurement $y_\sigma$, and by $\bm h_{x_t}$ the features associated with $x_t$ and $(s,t)$. These two representations are combined through a simple learned fusion step defined as follows
\begin{equation}
    \bm h_{x_t y_\sigma} = C_{1\times1}\left([ \theta_y\bm h_{y_\sigma}, \theta_x\bm h_{x_t}]\right),
\end{equation}
where $\theta_y$ and $\theta_x$ are learned scalar gains that control the contribution of each representation, and $C_{1\times1}$ denotes a learned $1\times1$ convolution. This allows the prediction to account for both the observed measurements and the current state of the flow. Apart from this additional conditioning mechanism, the RAM backbone is kept largely unchanged. Finally, the network output is negated to obtain the flow prediction $\bm v_\theta$.

\section{Training objectives and implementation details}\label{app:training}

\paragraph{Time sampling and branch selection.}

Training images are represented in $[-1,1]$. At each update, we draw a source time uniformly from $(10^{-4},1)$, shared across the minibatch, and independent Gaussian image and measurement noises. The general training loop can sample an operator from a configured bank at each minibatch; restricting this bank gives training or finetuning for a particular problem. The conditioning pair is always constructed using the noiseless forward projection and~\eqref{eq:measurement-interpolant}.

The first $5000$ updates use the diagonal main objective. Subsequently, finetuning chooses the off-diagonal branch with probability $p_D=0.25$, provided $t>\delta$, where $\delta=10^{-4}$. For that branch,
\begin{equation}\label{eq:endpoint-sampling}
    \bm s=\bm U^4(t-\delta),\qquad \bm U\sim\mathcal U(0,1),
\end{equation}
which gives greater sampling density near the clean endpoint. Otherwise, $s=t$ and the diagonal branch is used. The expected main objective is $(1-p_D)\lambda_b\mathcal L_b+p_D\lambda_L\mathcal L_{\mathrm{LSD}}$ after the warm-up. All squared errors are averaged over pixels, channels, and minibatch elements in the implementation.

The diagonal coefficient is $\lambda_b=10^{-2}$; the off-diagonal coefficient is ramped to $\lambda_L=10^{-2}$. LPIPS is introduced after $100$ updates with limiting weight $\lambda_p=10^{-1}$. For a term with activation update $k_0$ and limiting weight $\lambda$, the ramp is
\[
    w(k)=\mathbf 1_{\{k\geq k_0\}}\,
       \frac{\lambda}{1+\exp[-0.1(k-k_0)]}.
\]
The baseline contrast term starts after $5000$ updates and is multiplied, like LPIPS, by $g(s)=\exp(-4s)$. finetuning sets its coefficient to zero. Most of these parameters are inspired by previous works~\citep{boffi2025selfdistill,Mbakam2025LearningFP,noble2026fastfaithfulphotorealisticdiffusionbased}.

\paragraph{Teacher and differentiation.}
The EMA decay is set to $0.9999$. For LSD, the flow map and its endpoint derivative are computed jointly using a Jacobian--vector product (JVP) along the all-ones time direction. The transported image is passed directly to the teacher without clipping, and the teacher output is detached, while gradients are preserved through the student derivative. In our implementation, the source and endpoint noise maps provided to BAM are detached when converted to scalar values. As a result, the JVP captures the explicit differentiable dependence on the endpoint while treating these scalar noise levels as fixed. This defines the practical derivative used to enforce the LSD objective in~\eqref{eq:loss-lsd}.

\paragraph{Perceptual and contrast objectives}
Let $h(x)=(x+\bm 1)/2$ map normalized images to the $[0,1]$ range. We define the perceptual loss as
\begin{equation}\label{eq:appendix-lpips}
    \ell_{\mathrm{LPIPS}}(\hat{\bm x},\bm x_0) = \operatorname{LPIPS}_{\mathrm{squeeze}} \bigl(h(\hat{\bm x}),h(\bm x_0)\bigr).
\end{equation}
The loss is evaluated on the clipped clean prediction produced by the selected training branch. In this way, the LPIPS gradients are applied to the same velocity prediction optimized by the corresponding main loss.

For contrast adjustment, let $S_k(x)$ be the per-channel standard deviation in non-overlapping $k\times k$ windows, with $k\in\{16,32,64\}$, and let $S_{\mathrm{global}}(x)$ be the standard deviation over the full spatial extent of each channel. The set $\mathcal K$ contains these window sizes when they fit within the image, together with the global statistic. We use
\begin{equation}\label{eq:appendix-contrast}
    \ell_{\mathrm{ctr}}(\hat{\bm x}_{\mathrm{raw}},\bm x_0)
    =\frac{1}{|\mathcal K|}\sum_{k\in\mathcal K}
      \operatorname{mean}\!\left[
       \left(\frac{[S_k(\hat{\bm x}_{\mathrm{raw}})
         -(1+m)S_k(\bm x_0)]_+}{S_k(\bm x_0)+c}\right)^2\right],
\end{equation}
where $[u]_+=\max\{u,0\}$, $m=0.01$, and $c=0.02$. The mean averages windows, channels, and examples. This one-sided penalty acts on excessive contrast relative to the reference, with a small margin and a denominator floor to prevent nearly constant regions from dominating. It is evaluated before clipping so that saturation cannot hide excessive output amplitudes. We use it during baseline training across datasets and remove it during finetuning.

\paragraph{Optimization.}
The active flow-matching/LSD path uses AdamW with effective learning rate $10^{-5}$, weight decay $10^{-3}$, momentum parameters $(0.9,0.999)$, and gradient-norm clipping at $1.5$.

\section{Ablation studies}\label{app:ablations}

\newcommand{\BAMAblationZoomRect}[3]{%
  \begin{tikzpicture}[x=\BAMPaperImageWidth,y=\BAMPaperImageWidth,baseline=(current bounding box.center)]
    \node[anchor=south west,inner sep=0] at (0,0) {\includegraphics[width=\BAMPaperImageWidth]{#1}};
    \draw[yellow,line width=.45pt] (#2,#3) rectangle ({#2+.25},{#3+.125});
    \begin{scope}
      \clip (0,-.54) rectangle (1,-.04);
      \node[anchor=south west,inner sep=0] at ({-4*#2},{-.54-4*#3}) {\includegraphics[width=4\BAMPaperImageWidth]{#1}};
    \end{scope}
    \draw[yellow,line width=.45pt] (0,-.54) rectangle (1,-.04);
    \pgfresetboundingbox
    \useasboundingbox (0,-.54) rectangle (1,1.3113);
  \end{tikzpicture}%
}

\newcommand{\BAMAblationZoomRectyty}[3]{%
  \begin{tikzpicture}[
    x=\BAMPaperImageWidth,
    y=\BAMPaperImageWidth,
    baseline=(current bounding box.center)
  ]
    \node[anchor=south west,inner sep=0] at (0,0)
    {\includegraphics[width=\BAMPaperImageWidth]{#1}};
    \pgfmathsetmacro{\yselect}{#3-0.45}
    \draw[yellow,line width=.45pt]
      (#2,\yselect)
      rectangle
      ({#2+0.25},{\yselect+0.125});
    \begin{scope}
      \clip (0,-.54) rectangle (1,-.04);
      \node[anchor=south west,inner sep=0]
      at ({-4*#2},{-.54-4*\yselect})
      {\includegraphics[width=4\BAMPaperImageWidth]{#1}};
    \end{scope}
    \draw[yellow,line width=.45pt]
      (0,-.54) rectangle (1,-.04);
    \pgfresetboundingbox
    \useasboundingbox (0,-.54) rectangle (1,1);
  \end{tikzpicture}%
}

We ablate the architectural and training choices that distinguish the final BAM recipe. Unless stated otherwise, the comparisons below modify only the component under study while keeping the remaining training setup unchanged.

\subsection{Effect of Noise-Aware Conditioning ($y_\sigma$ vs. $y$)}\label{app:ablation-y-sigma}
\begin{wrapfigure}{r}{0.49\linewidth}
    \vspace{-0.8\baselineskip}
    \centering
    \includegraphics[width=\linewidth]{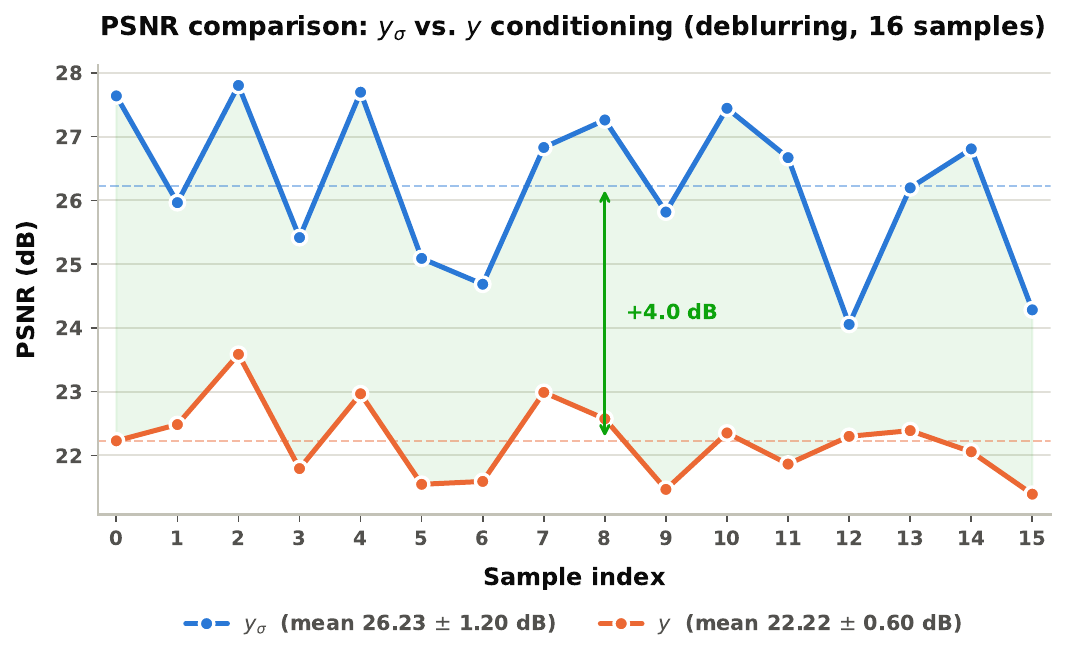}
    \caption{\textbf{$y$ vs. $y_\sigma$ ablation.} Per-sample PSNR for conditioning on $y_\sigma$ and $y$ on the Gaussian deblurring task. Dashed lines indicate the mean PSNR for each variant, while the shaded region shows the per-sample improvement.}
    \label{fig:ablation-yt-y-psnr}
    \vspace{-0.8\baselineskip}
\end{wrapfigure}
\begin{figure}[H]
\centering
\begingroup
\setlength{\BAMPaperImageWidth}{.19\linewidth}
\setlength{\tabcolsep}{2pt}
\scriptsize
\begin{tabular}{@{}cccc@{}}
\toprule
\textbf{Observation} & \textbf{BAM with $y$} & \textbf{BAM with $y_\sigma$} & \textbf{GT} \\
\midrule
\BAMAblationZoomRectyty{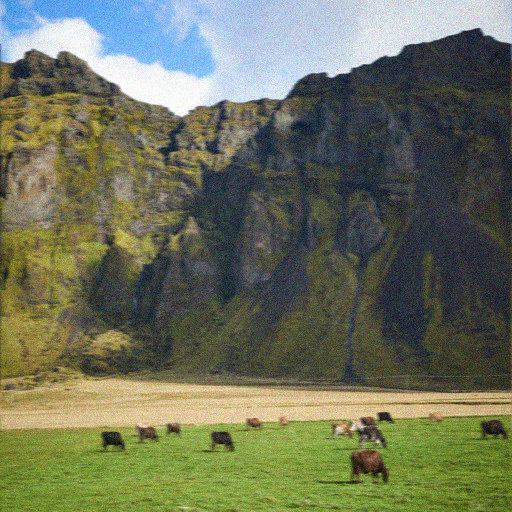}{0.285}{0.927} &
\BAMAblationZoomRectyty{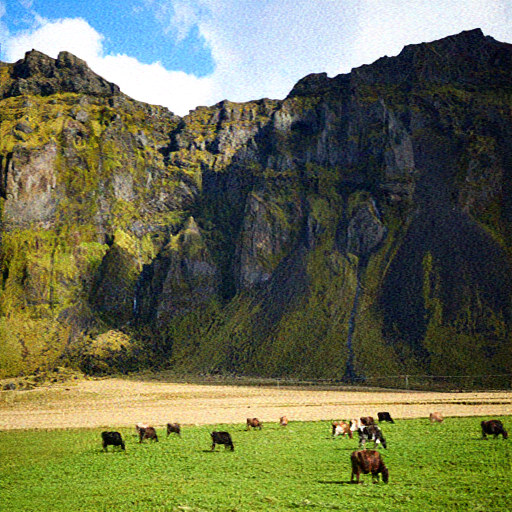}{0.285}{0.927} &
\BAMAblationZoomRectyty{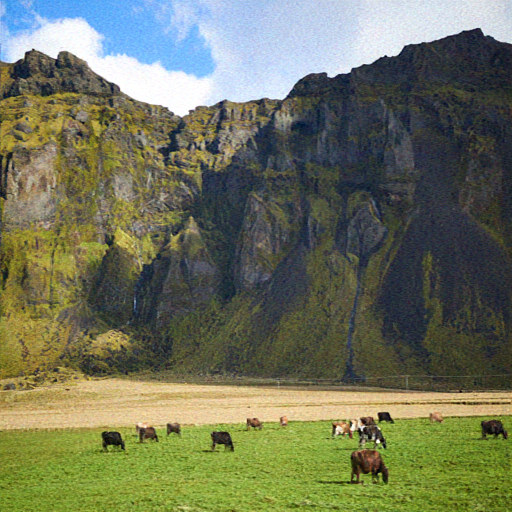}{0.285}{0.927} &\BAMAblationZoomRectyty{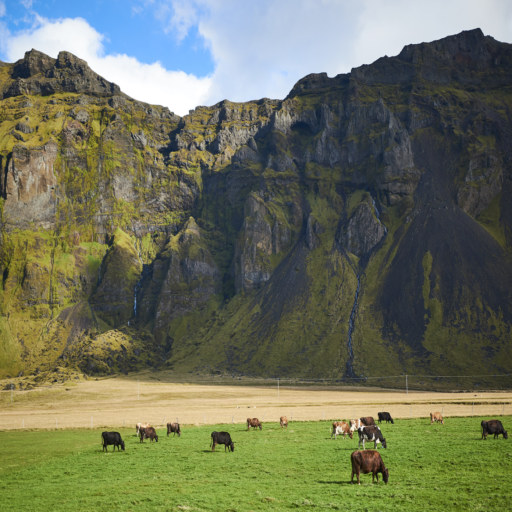}{0.285}{0.927} \\[3pt]
\bottomrule
\end{tabular}
\endgroup
\caption{\textbf{Qualitative $y$ vs. $y_\sigma$ ablation on LSDIR Gaussian deblurring.} The observation, reconstruction obtained by conditioning BAM on the original measurement $y$, reconstruction obtained with noise-aware conditioning $y_\sigma$, and ground truth, for Gaussian deblurring with noise level $\sigma_y=0.05$. Strips below show $4\times$ linear magnification.}
\label{fig:ablation-yt-vs-y-qualitative}
\end{figure}

In this section, we study the effect of the measurement used to condition BAM during training and sampling. In particular, we compare conditioning on the original measurement $y$ with conditioning on a noise-aware measurement $y_\sigma$.
We evaluate both variants on Gaussian deblurring with noise level $\sigma_y=0.05$ using 16 LSDIR test images. As shown in \Cref{fig:ablation-yt-y-psnr}, conditioning on $y_\sigma$ consistently improves reconstruction quality. The mean PSNR increase from $22.22 \pm 0.60\,$dB with$y$ to $26.23 \pm 1.20\,$dB with $y_\sigma$, corresponding to an improvement of approximately $4.0 \,$dB. This improvement is also visible in the qualitative results shown in \Cref{fig:ablation-yt-vs-y-qualitative}, where noise-aware conditioning produces a reconstruction that is visually closer to the ground truth.  Overall, these results highlight the benefit of using $y_\sigma$ as the conditioning in BAM.
\vspace{2em}
\subsection{Why the modified RAM architecture matters}\label{app:ablation-architecture}

\begin{wrapfigure}{r}{0.49\linewidth}
    \vspace{-0.8\baselineskip}
    \centering
    \includegraphics[width=\linewidth]{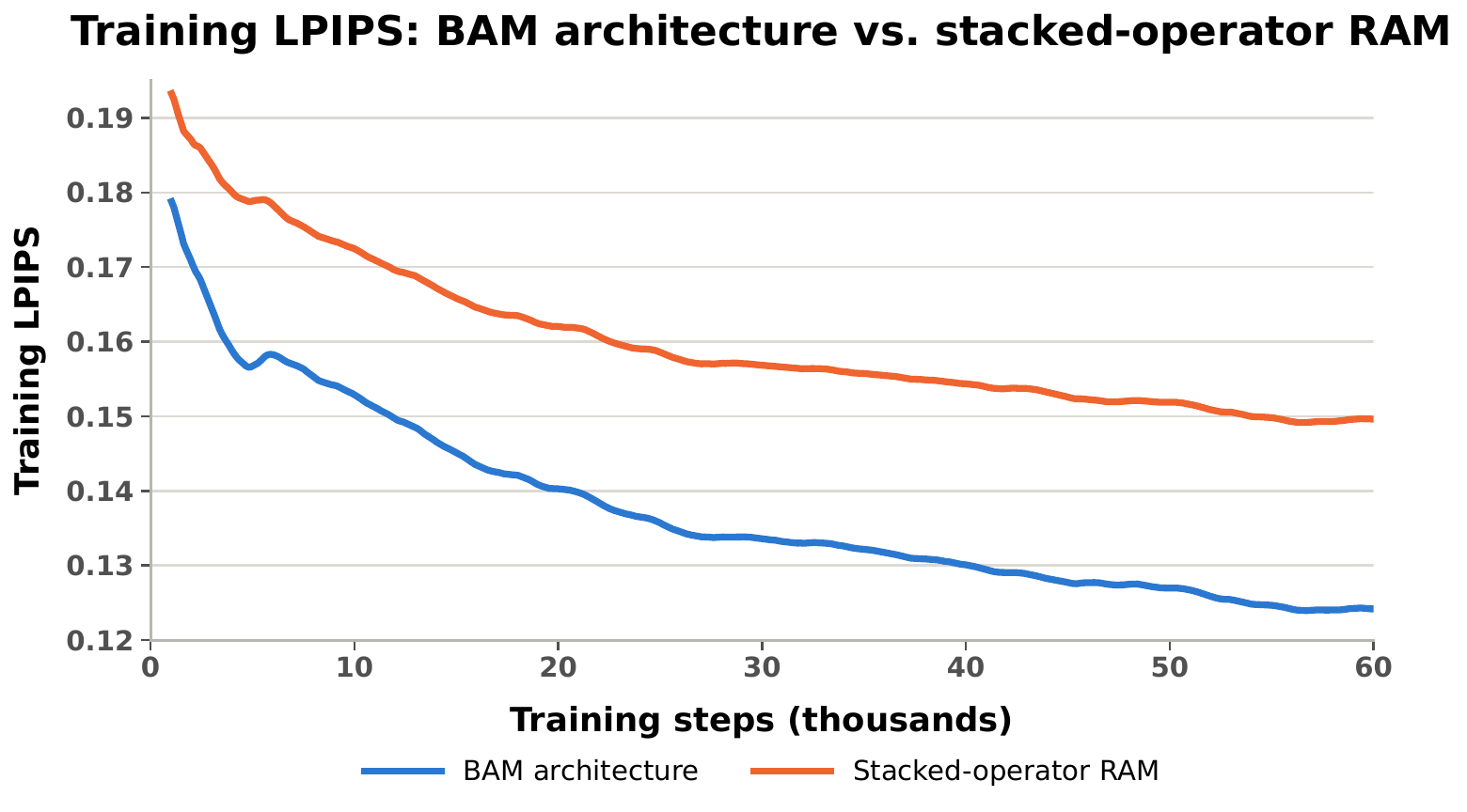}
    \caption{\textbf{Architecture ablation.} Smoothed training LPIPS for the BAM architecture and the base RAM design with stacked operators%
    We show the first $60\,$k training steps. Lower is better.}
    \label{fig:ablation-architecture-lpips}
    \vspace{-0.8\baselineskip}
\end{wrapfigure}
The BAM backbone retains RAM's multiscale operator conditioning, but replaces the final stacked conditioning block by separate measurement and image-state branches before fusion (Section~\ref{app:architecture}). Figure~\ref{fig:ablation-architecture-lpips} isolates the effect of this modification. Both models are trained on the full mixed training set using eight H100 GPUs. We report the first $60\,$k training steps, over which the architecture used by BAM reaches a substantially lower LPIPS than the original stacked-operator RAM design and maintains the gap throughout training.

\begin{figure}[H]
\centering
\begingroup
\setlength{\BAMPaperImageWidth}{.19\linewidth}
\setlength{\tabcolsep}{2pt}
\scriptsize
\begin{tabular}{@{}cccc@{}}
\toprule
\textbf{Observation} & \textbf{Stacked RAM} & \textbf{BAM architecture} & \textbf{GT} \\
\midrule
\BAMAblationZoomRect{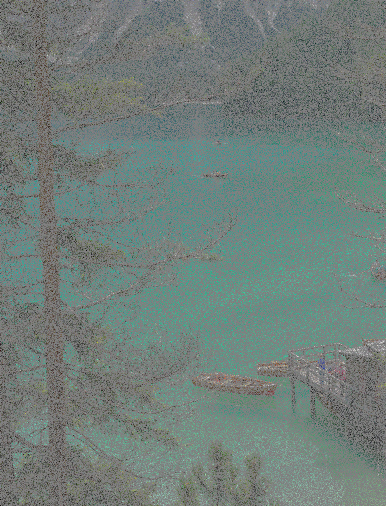}{0.285}{0.927} &
\BAMAblationZoomRect{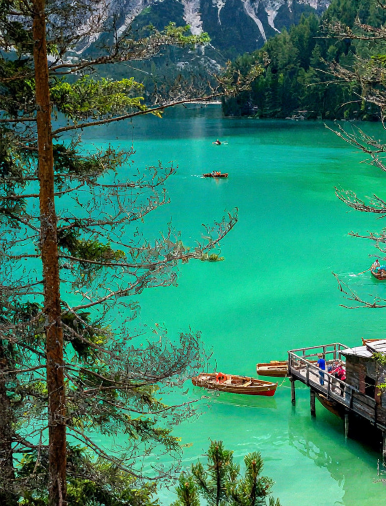}{0.285}{0.927} &
\BAMAblationZoomRect{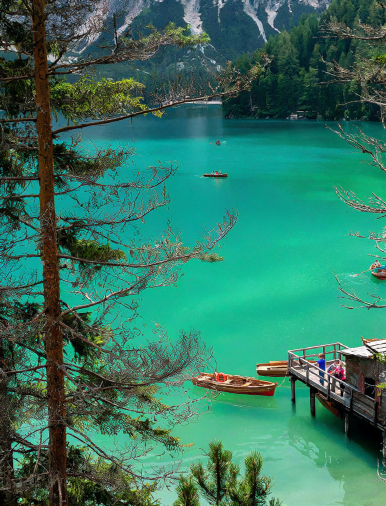}{0.285}{0.927} &
\BAMAblationZoomRect{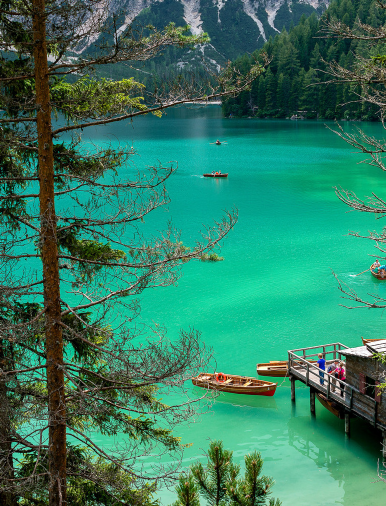}{0.285}{0.927} \\[3pt]
\bottomrule
\end{tabular}
\endgroup
\caption{\textbf{Qualitative architecture ablation on DIV2K inpainting.} The observation, reconstruction obtained with the original stacked-operator RAM design, reconstruction obtained with the BAM architecture, and ground truth. Strips below show $4\times$ linear magnification.}
\label{fig:ablation-architecture-qualitative}
\end{figure}

The difference is also visible qualitatively on difficult inverse problems such as inpainting. Figure~\ref{fig:ablation-architecture-qualitative} compares the two backbones on the same DIV2K example. The modified architecture better preserves fine structures and suppresses reconstruction artifacts, illustrating the importance of preserving a dedicated image-state pathway rather than relying exclusively on the stacked operator in the last feature update.
\vspace{-0.5em}

\subsection{Why the contrast loss matters}\label{app:ablation-contrast}

\begin{figure}[H]
\centering
\begingroup
\setlength{\BAMPaperImageWidth}{.19\linewidth}
\setlength{\tabcolsep}{2pt}
\scriptsize
\begin{tabular}{@{}cccc@{}}
\toprule
\textbf{Observation} & \textbf{Without contrast loss} & \textbf{With contrast loss} & \textbf{GT} \\
\midrule
\BAMPaperZoom{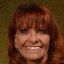}{0.30}{0.37} &
\BAMPaperZoom{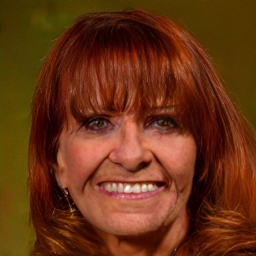}{0.30}{0.37} &
\BAMPaperZoom{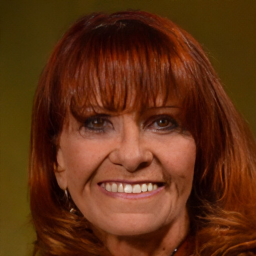}{0.30}{0.37} &
\BAMPaperZoom{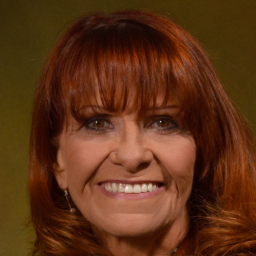}{0.30}{0.37} \\[3pt]
\bottomrule
\end{tabular}
\endgroup
\caption{\textbf{Contrast-loss ablation on FFHQ.} Removing the contrast regularizer can produce visibly over-saturated reconstructions. The auxiliary term stabilizes output contrast during heterogeneous multi-dataset, multi-operator training. Strips below show $4\times$ linear magnification.}
\label{fig:ablation-contrast}
\end{figure}

The contrast term in~\eqref{eq:appendix-contrast} is introduced to prevent excessive local and global contrast during mixed-dataset pre-training. Without it, we observe a recurrent tendency toward over-saturated outputs, as illustrated in Figure~\ref{fig:ablation-contrast}. This effect is plausibly caused by the heterogeneity of the joint training distribution: the constituent datasets have different image statistics and effective intensity scales, while the different forward operators induce conditioning signals with different dynamic ranges. A one-sided contrast penalty provides a simple way to regularize this scale mismatch without forcing the reconstruction toward a specific global histogram. Consistently with this interpretation, we remove the term during dataset-specific fine-tuning.

\subsection{Why an adversarial loss is not required}\label{app:ablation-adversarial}

\begin{wrapfigure}{r}{0.49\linewidth}
    \vspace{-0.8\baselineskip}
    \centering
    \includegraphics[width=\linewidth]{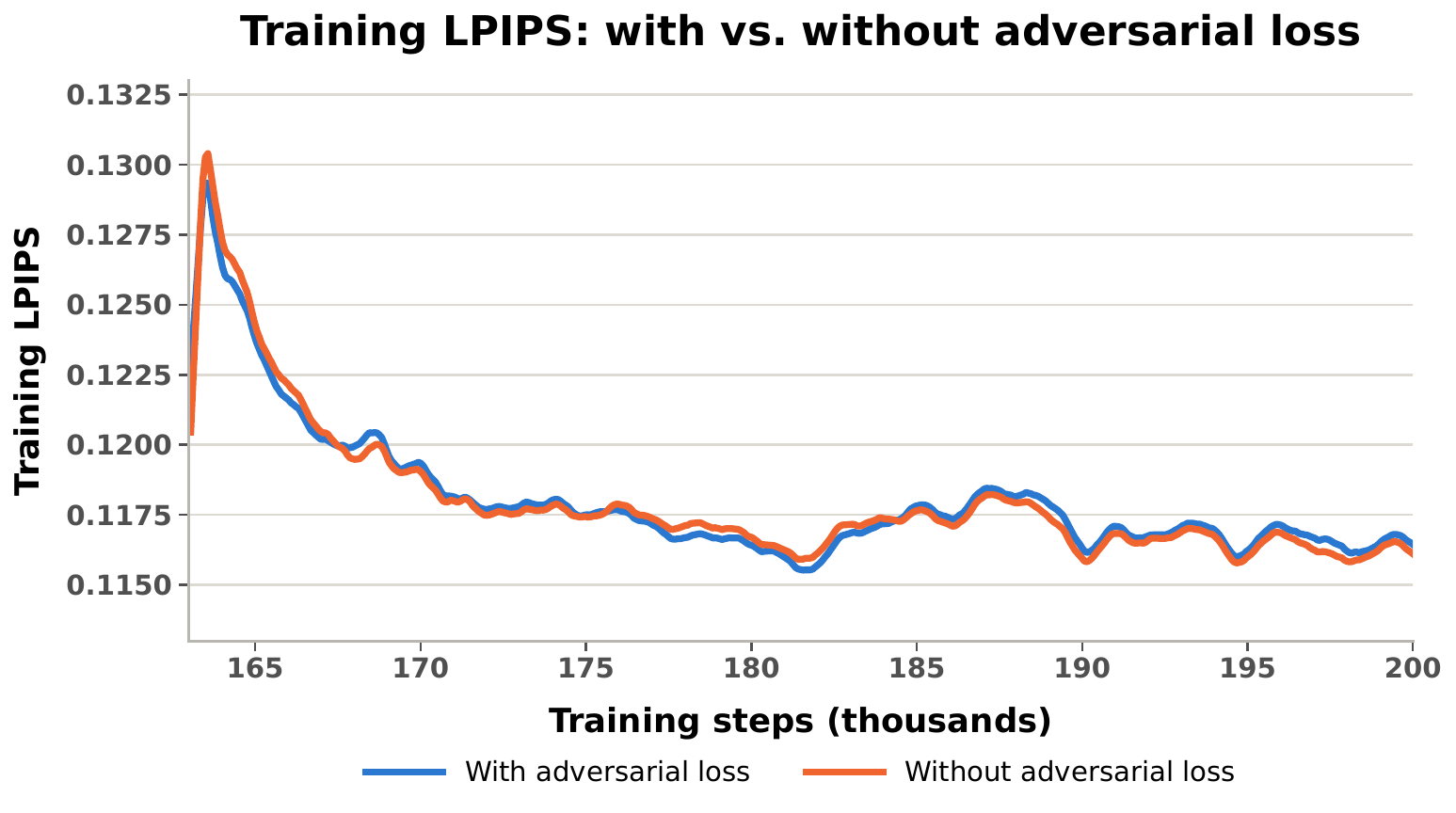}
    \caption{\textbf{Adversarial-loss ablation.} Smoothed training LPIPS during FFHQ super-resolution fine-tuning with and without a discriminator loss. Lower is better.}
    \label{fig:ablation-adversarial}
    \vspace{-0.8\baselineskip}
\end{wrapfigure}
We also tested augmenting the objective with a discriminator loss during FFHQ super-resolution fine-tuning, a common practice also used by~\citet{noble2026fastfaithfulphotorealisticdiffusionbased} and UD2M~\citep{Mbakam2025LearningFP}. In particular, we adopt the same discriminator architecture used to train UD2M. Figure~\ref{fig:ablation-adversarial} shows that the training LPIPS trajectories with and without the adversarial term are nearly indistinguishable over the observed window. In particular, the discriminator does not provide a measurable improvement in the perceptual training criterion used by the final model. We therefore omit adversarial training from the final BAM objective, avoiding the additional discriminator, optimization instability, and computational overhead.

\subsection{Number of sampling steps}\label{app:ablation-steps}

We compare BAM with one, two, and three network evaluations on the same 64 FFHQ test images for Gaussian deblurring, $4\times$ super-resolution, inpainting, demosaicing, and compressed sensing. We use the baseline all-problem model without dataset-specific fine-tuning. The runs use the same clean image, sampled observations and sampler randomness.

Tables~\ref{tab:ablation-steps-low-noise} and~\ref{tab:ablation-steps-high-noise} report PSNR, LPIPS, CMMD, and FID. Moving from one to two steps improves PSNR and LPIPS for every problem at both noise levels. Three steps give the highest PSNR in all ten settings. The perceptual trend is not uniformly monotonic: at $\sigma_y=0.05$, two steps give lower LPIPS than three steps for super-resolution, inpainting, demosaicing, and compressed sensing. Figures~\ref{fig:ablation-steps-low-noise} and~\ref{fig:ablation-steps-high-noise} show matched examples.

\begin{table}[H]
\centering
\caption{\textbf{BAM sampling-step ablation on FFHQ at $\sigma_y=0.025$.} Results over 64 images per problem. PSNR ($\uparrow$), LPIPS ($\downarrow$), CMMD ($\downarrow$), and FID ($\downarrow$). Each step requires one network evaluation. Bold marks the best available value per problem and metric; -- denotes an unavailable value.}
\label{tab:ablation-steps-low-noise}
\begingroup
\fontsize{8}{9}\selectfont
\setlength{\tabcolsep}{4pt}
\begin{tabularx}{\linewidth}{@{}l r *{4}{>{\raggedleft\arraybackslash}X}@{}}
\toprule
Method & NFEs & PSNR $\uparrow$ & LPIPS $\downarrow$ & CMMD $\downarrow$ & FID $\downarrow$ \\
\midrule
\multicolumn{6}{@{}l}{\textbf{Gaussian deblurring}} \\
BAM 1-step & 1 & 26.56 & 0.323 & 0.334 & 54.38 \\
BAM 2-step & 2 & 29.68 & 0.249 & 0.149 & 42.07 \\
BAM 3-step & 3 & \textbf{30.63} & \textbf{0.232} & \textbf{0.089} & \textbf{40.63} \\
\midrule
\multicolumn{6}{@{}l}{\textbf{SR $\times4$}} \\
BAM 1-step & 1 & 26.52 & 0.344 & 0.386 & 58.85 \\
BAM 2-step & 2 & 28.73 & 0.263 & 0.162 & 48.04 \\
BAM 3-step & 3 & \textbf{29.99} & \textbf{0.249} & \textbf{0.128} & \textbf{45.48} \\
\midrule
\multicolumn{6}{@{}l}{\textbf{Inpainting}} \\
BAM 1-step & 1 & 26.68 & 0.367 & 0.274 & 46.95 \\
BAM 2-step & 2 & 32.79 & 0.180 & \textbf{0.043} & 22.98 \\
BAM 3-step & 3 & \textbf{32.90} & \textbf{0.174} & 0.045 & \textbf{21.30} \\
\midrule
\multicolumn{6}{@{}l}{\textbf{Demosaicing}} \\
BAM 1-step & 1 & 26.95 & 0.331 & 0.264 & 36.96 \\
BAM 2-step & 2 & 35.38 & 0.125 & 0.030 & 11.88 \\
BAM 3-step & 3 & \textbf{36.45} & \textbf{0.119} & \textbf{0.026} & \textbf{10.95} \\
\midrule
\multicolumn{6}{@{}l}{\textbf{Compressed sensing}} \\
BAM 1-step & 1 & 26.83 & 0.354 & 0.251 & 41.82 \\
BAM 2-step & 2 & 33.56 & 0.163 & \textbf{0.032} & 17.71 \\
BAM 3-step & 3 & \textbf{34.58} & \textbf{0.151} & 0.034 & \textbf{16.37} \\
\bottomrule
\end{tabularx}
\endgroup
\end{table}

\begin{table}[H]
\centering
\caption{\textbf{BAM sampling-step ablation on FFHQ at $\sigma_y=0.05$.} Results over 64 images per problem. PSNR ($\uparrow$), LPIPS ($\downarrow$), CMMD ($\downarrow$), and FID ($\downarrow$). Each step requires one network evaluation. Bold marks the best available value per problem and metric; -- denotes an unavailable value.}
\label{tab:ablation-steps-high-noise}
\begingroup
\fontsize{8}{9}\selectfont
\setlength{\tabcolsep}{4pt}
\begin{tabularx}{\linewidth}{@{}l r *{4}{>{\raggedleft\arraybackslash}X}@{}}
\toprule
Method & NFEs & PSNR $\uparrow$ & LPIPS $\downarrow$ & CMMD $\downarrow$ & FID $\downarrow$ \\
\midrule
\multicolumn{6}{@{}l}{\textbf{Gaussian deblurring}} \\
BAM 1-step & 1 & 26.53 & 0.324 & 0.329 & 54.55 \\
BAM 2-step & 2 & 28.97 & 0.251 & 0.158 & \textbf{43.36} \\
BAM 3-step & 3 & \textbf{29.63} & \textbf{0.250} & \textbf{0.099} & 44.98 \\
\midrule
\multicolumn{6}{@{}l}{\textbf{SR $\times4$}} \\
BAM 1-step & 1 & 26.54 & 0.346 & 0.396 & 58.14 \\
BAM 2-step & 2 & 28.58 & \textbf{0.270} & 0.239 & 50.47 \\
BAM 3-step & 3 & \textbf{28.84} & 0.272 & \textbf{0.186} & \textbf{50.42} \\
\midrule
\multicolumn{6}{@{}l}{\textbf{Inpainting}} \\
BAM 1-step & 1 & 26.66 & 0.372 & 0.276 & 46.04 \\
BAM 2-step & 2 & 31.06 & \textbf{0.214} & 0.126 & \textbf{27.89} \\
BAM 3-step & 3 & \textbf{31.50} & 0.228 & \textbf{0.078} & 32.09 \\
\midrule
\multicolumn{6}{@{}l}{\textbf{Demosaicing}} \\
BAM 1-step & 1 & 26.94 & 0.333 & 0.273 & 38.66 \\
BAM 2-step & 2 & 32.76 & \textbf{0.180} & 0.098 & \textbf{19.69} \\
BAM 3-step & 3 & \textbf{33.38} & 0.202 & \textbf{0.051} & 23.77 \\
\midrule
\multicolumn{6}{@{}l}{\textbf{Compressed sensing}} \\
BAM 1-step & 1 & 26.81 & 0.354 & 0.252 & 42.83 \\
BAM 2-step & 2 & 31.93 & \textbf{0.197} & 0.108 & \textbf{24.08} \\
BAM 3-step & 3 & \textbf{32.41} & 0.208 & \textbf{0.056} & 25.52 \\
\bottomrule
\end{tabularx}
\endgroup
\end{table}

\clearpage
\begin{figure}[H]
\centering
\begingroup
\scriptsize
\setlength{\BAMPaperImageWidth}{.175\linewidth}
\setlength{\tabcolsep}{1pt}
\begin{tabular}{@{}ccccc@{}}
\toprule
\textbf{Observation} & \textbf{BAM 1-step} & \textbf{BAM 2-step} & \textbf{BAM 3-step} & \textbf{GT} \\
\midrule
\multicolumn{5}{@{}l}{\textbf{Gaussian deblurring}, $\sigma_y=0.025$} \\[1pt]
\BAMPaperZoom{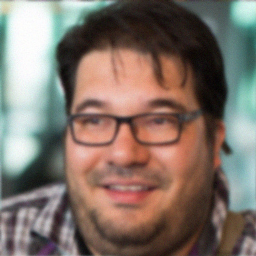}{0.27}{0.45} & \BAMPaperZoom{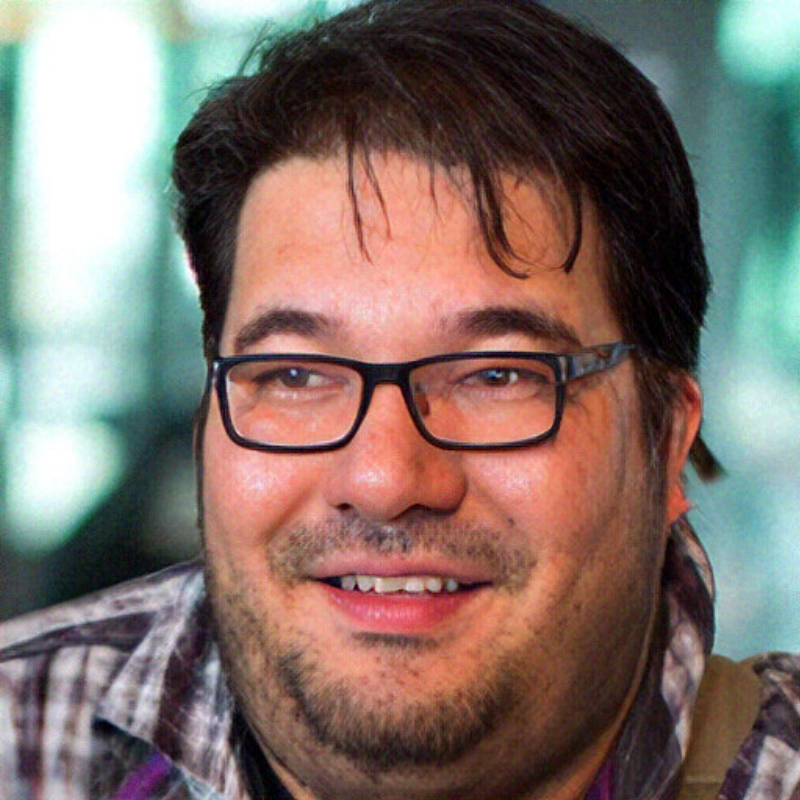}{0.27}{0.45} & \BAMPaperZoom{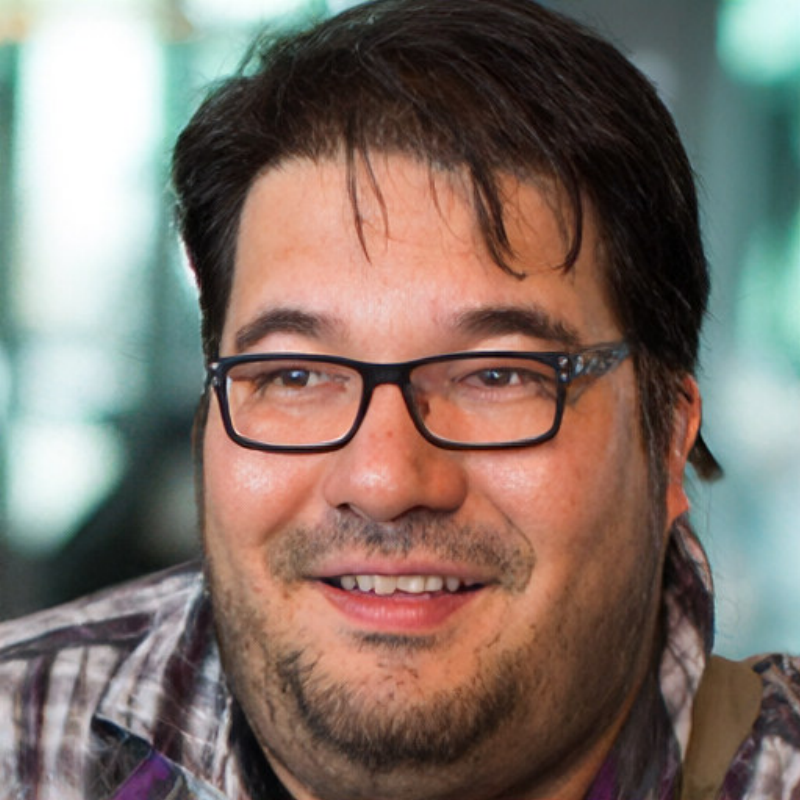}{0.27}{0.45} & \BAMPaperZoom{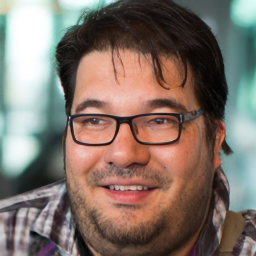}{0.27}{0.45} & \BAMPaperZoom{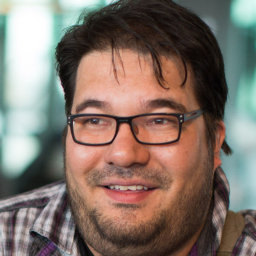}{0.27}{0.45} \\[3pt]
\multicolumn{5}{@{}l}{\textbf{SR $\times4$}, $\sigma_y=0.025$} \\[1pt]
\BAMPaperZoom{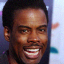}{0.34}{0.48} & \BAMPaperZoom{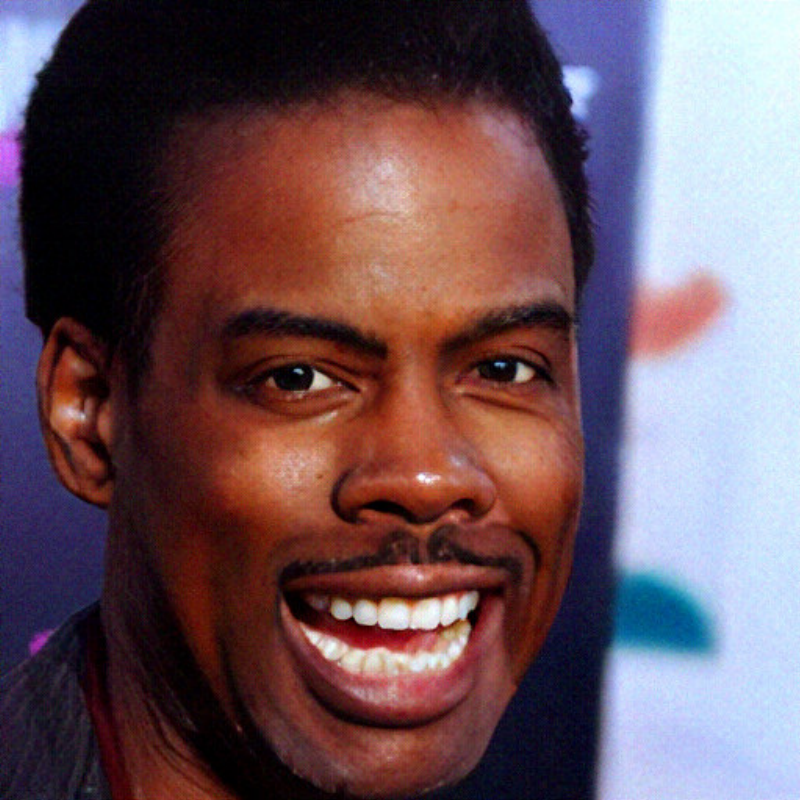}{0.34}{0.48} & \BAMPaperZoom{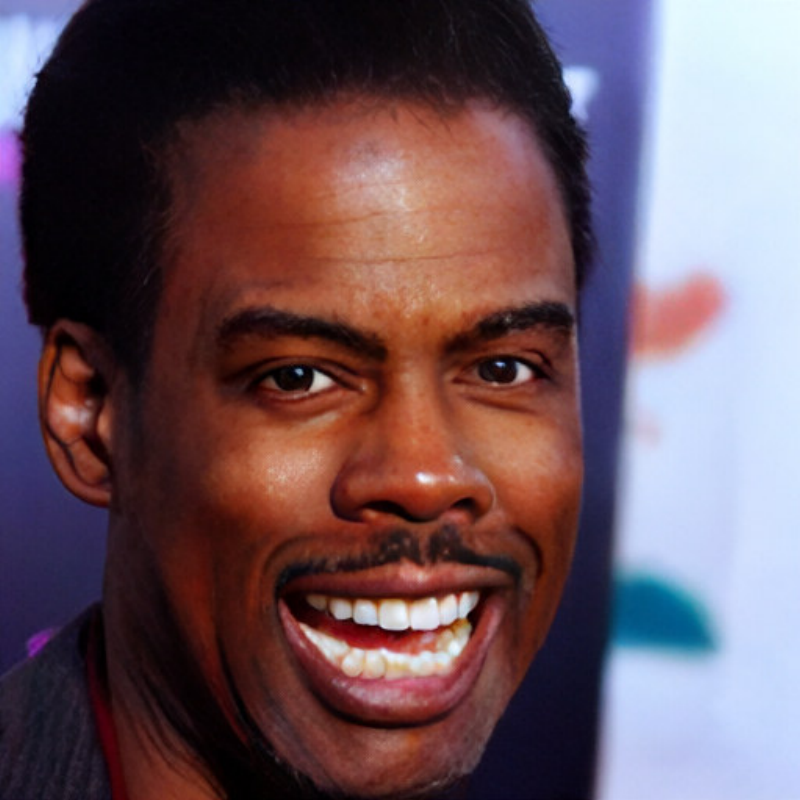}{0.34}{0.48} & \BAMPaperZoom{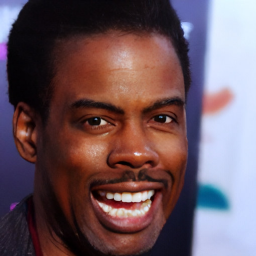}{0.34}{0.48} & \BAMPaperZoom{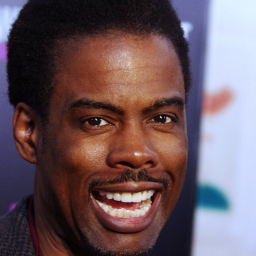}{0.34}{0.48} \\[3pt]
\multicolumn{5}{@{}l}{\textbf{Inpainting}, $\sigma_y=0.025$} \\[1pt]
\BAMPaperZoom{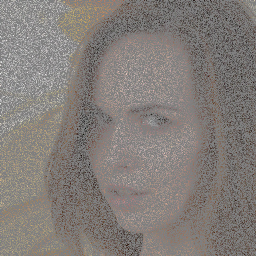}{0.36}{0.48} & \BAMPaperZoom{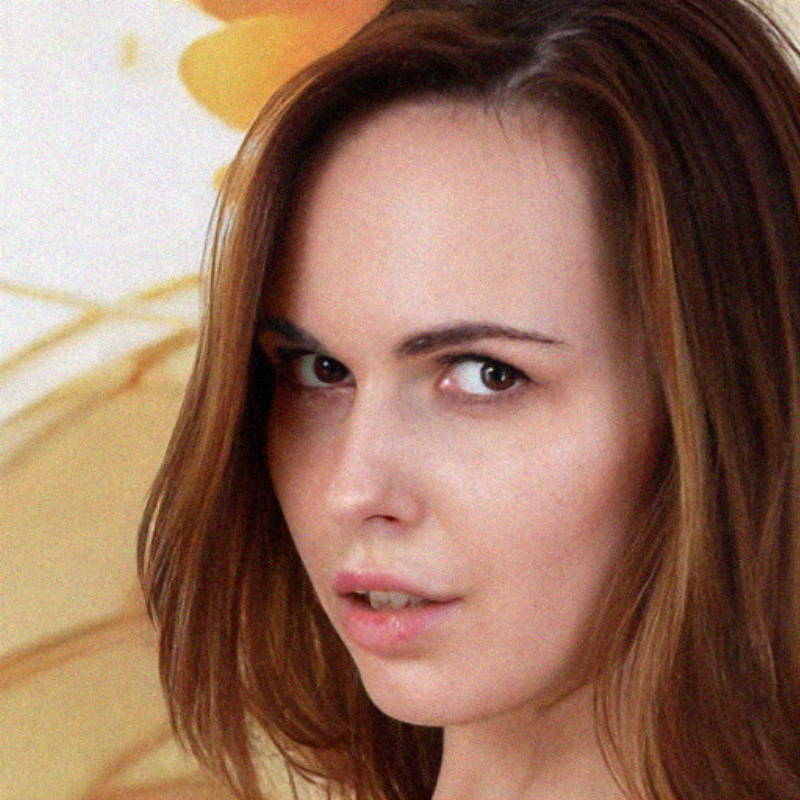}{0.36}{0.48} & \BAMPaperZoom{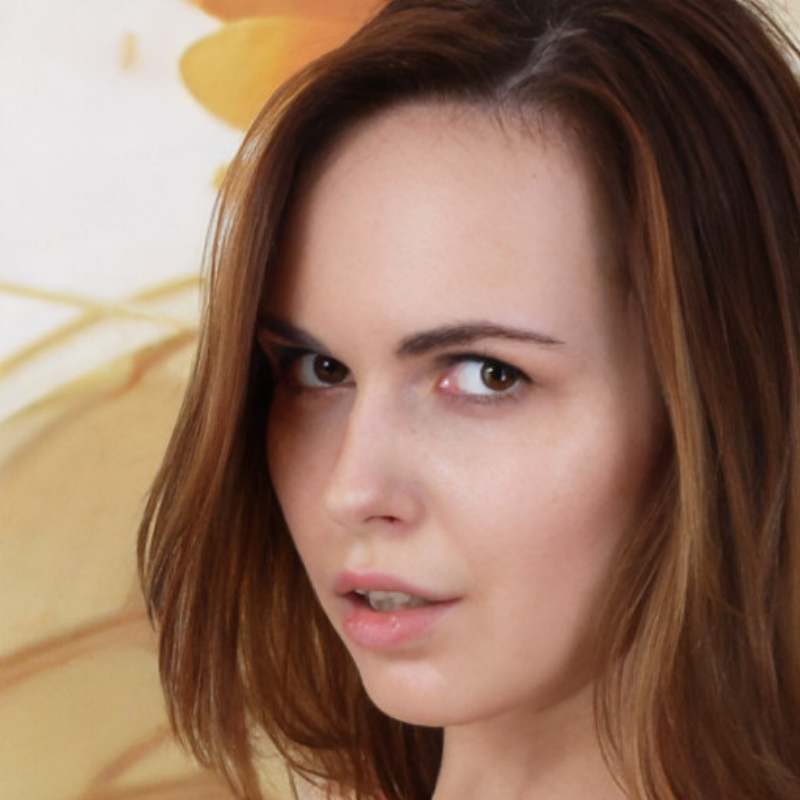}{0.36}{0.48} & \BAMPaperZoom{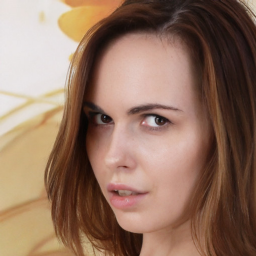}{0.36}{0.48} & \BAMPaperZoom{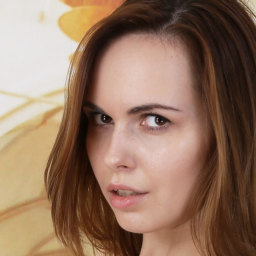}{0.36}{0.48} \\[3pt]
\multicolumn{5}{@{}l}{\textbf{Demosaicing}, $\sigma_y=0.025$} \\[1pt]
\BAMPaperZoom{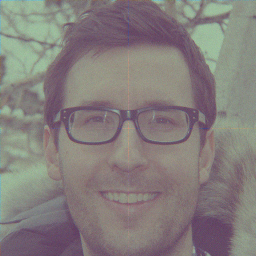}{0.27}{0.45} & \BAMPaperZoom{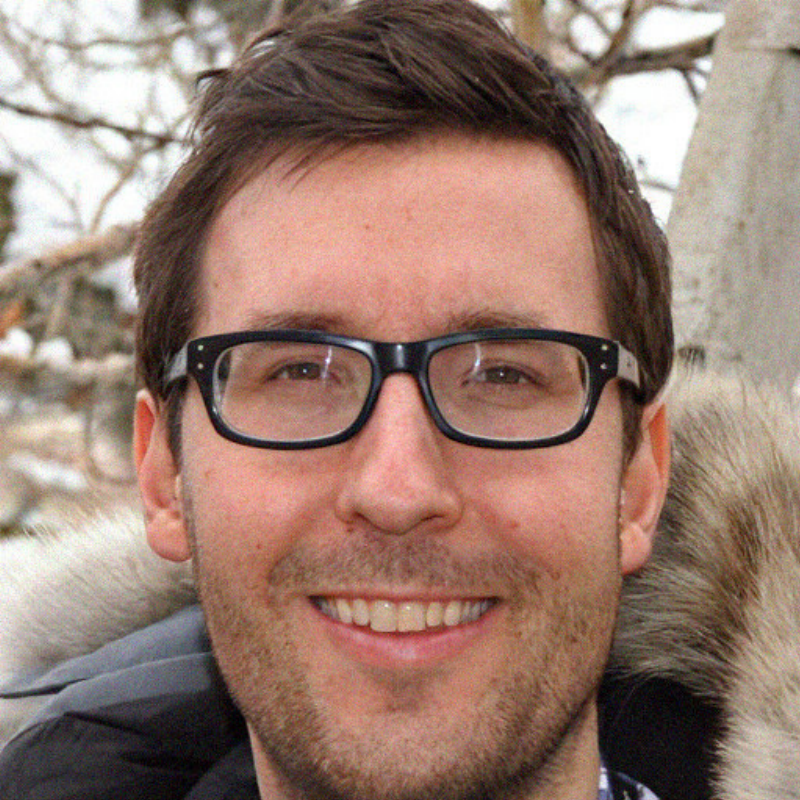}{0.27}{0.45} & \BAMPaperZoom{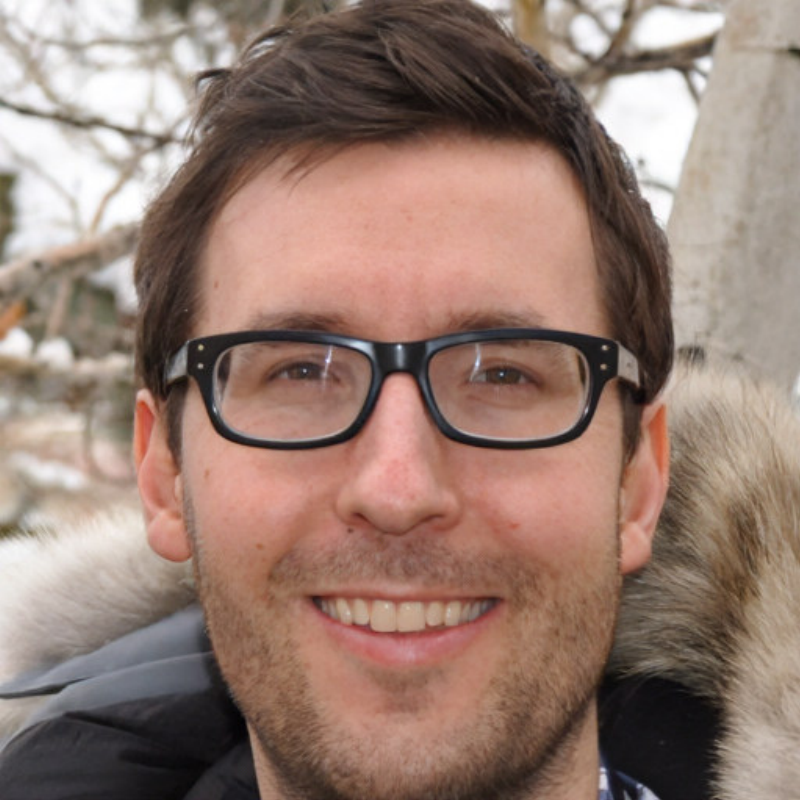}{0.27}{0.45} & \BAMPaperZoom{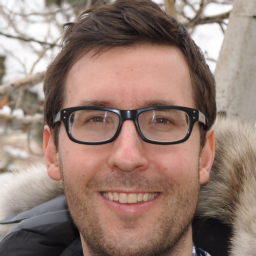}{0.27}{0.45} & \BAMPaperZoom{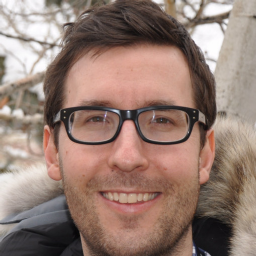}{0.27}{0.45} \\[3pt]
\multicolumn{5}{@{}l}{\textbf{Compressed sensing}, $\sigma_y=0.025$} \\[1pt]
\BAMPaperZoom{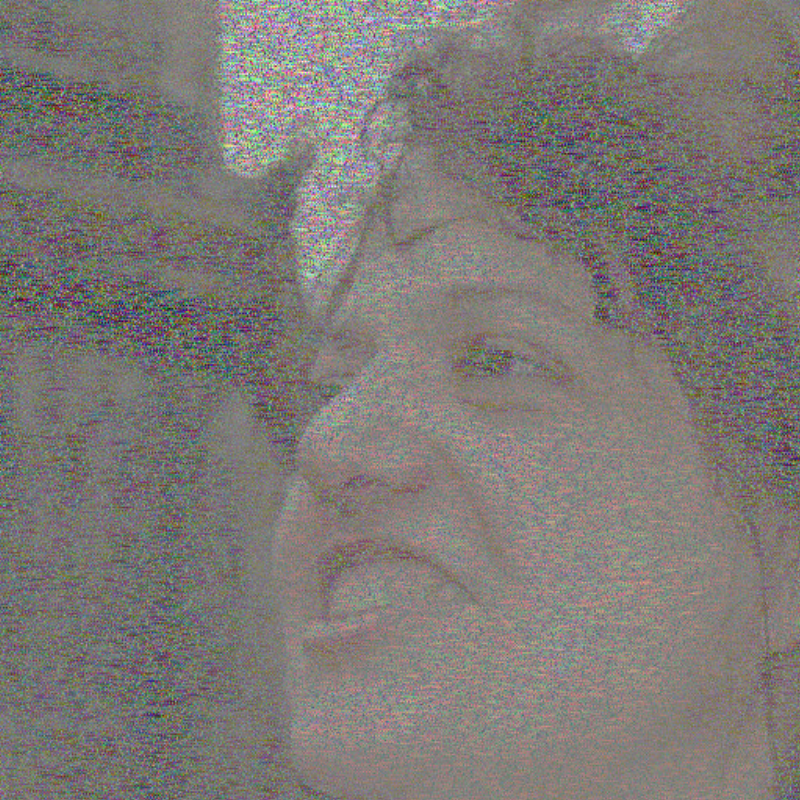}{0.45}{0.5} & \BAMPaperZoom{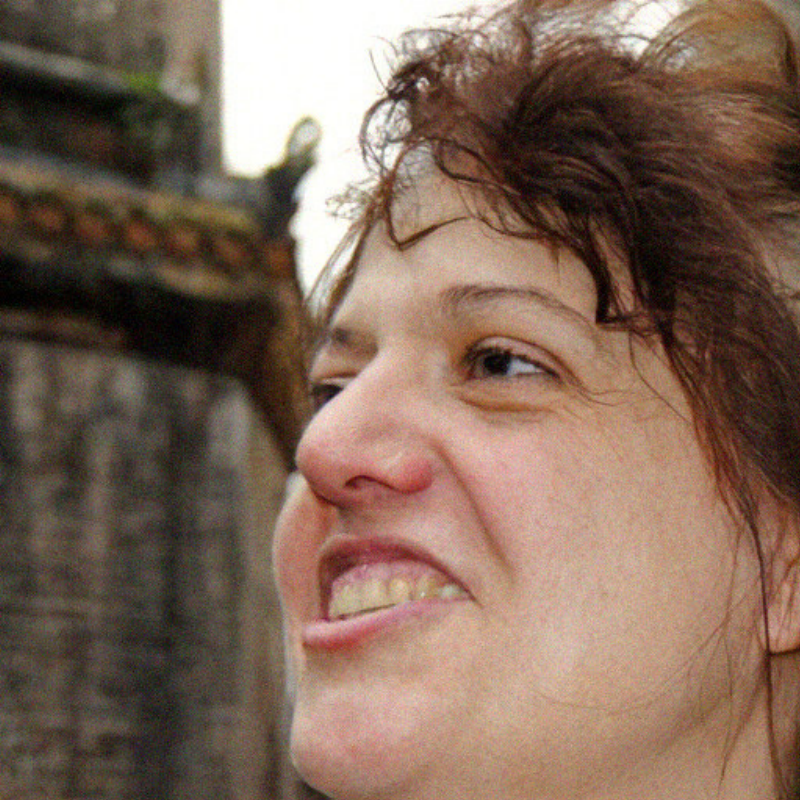}{0.45}{0.5} & \BAMPaperZoom{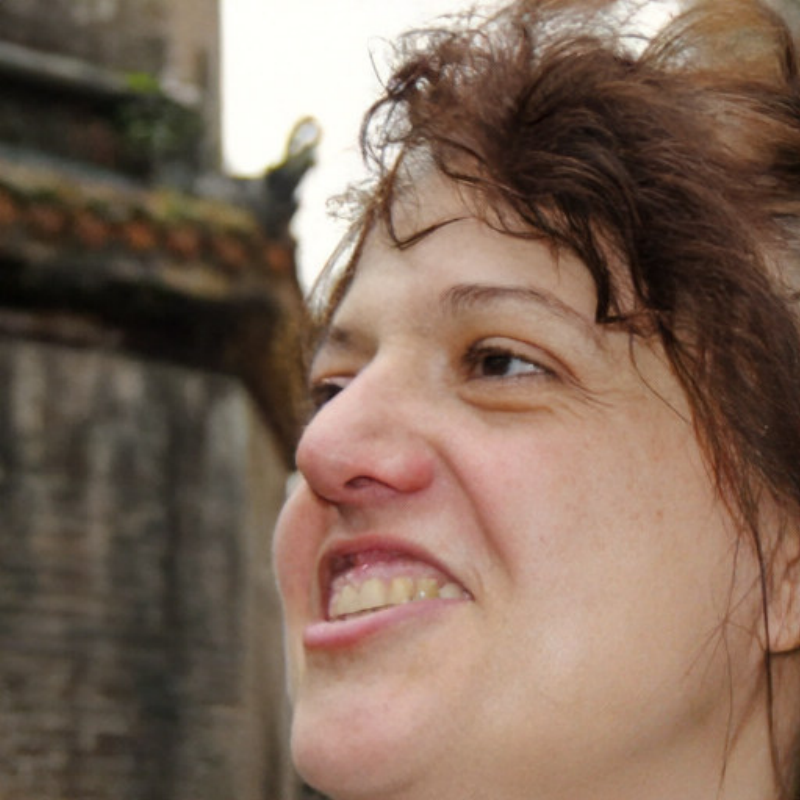}{0.45}{0.5} & \BAMPaperZoom{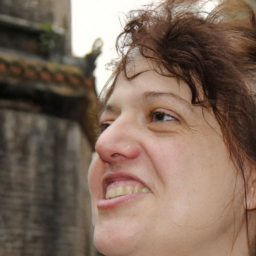}{0.45}{0.5} & \BAMPaperZoom{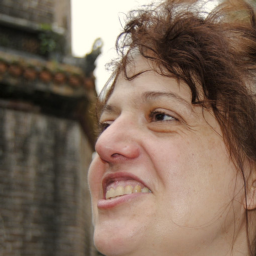}{0.45}{0.5} \\[3pt]
\bottomrule
\end{tabular}
\endgroup
\caption{\textbf{BAM sampling-step comparison on FFHQ at $\sigma_y=0.025$.} For compressed sensing, the observation column shows $A^\dagger y$. Yellow boxes select matching regions, with $4\times$ linear magnification below each image.}
\label{fig:ablation-steps-low-noise}
\end{figure}

\clearpage
\begin{figure}[H]
\centering
\begingroup
\scriptsize
\setlength{\BAMPaperImageWidth}{.175\linewidth}
\setlength{\tabcolsep}{1pt}
\begin{tabular}{@{}ccccc@{}}
\toprule
\textbf{Observation} & \textbf{BAM 1-step} & \textbf{BAM 2-step} & \textbf{BAM 3-step} & \textbf{GT} \\
\midrule
\multicolumn{5}{@{}l}{\textbf{Gaussian deblurring}, $\sigma_y=0.05$} \\[1pt]
\BAMPaperZoom{Figures/FFHQ/deb/sigma_0.05/BAM/obs/y_00062.pdf}{0.27}{0.45} & \BAMPaperZoom{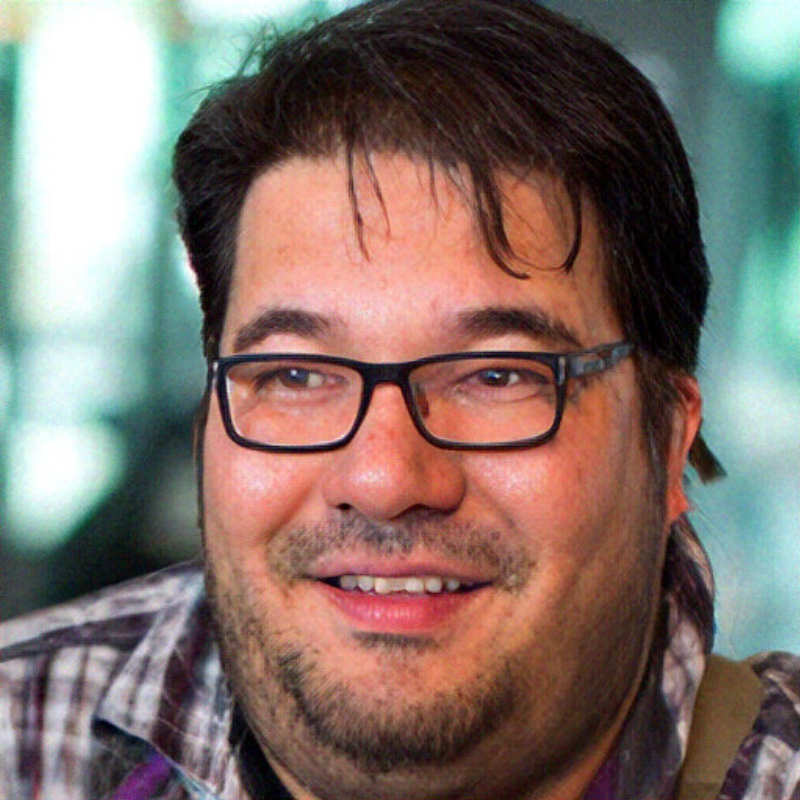}{0.27}{0.45} & \BAMPaperZoom{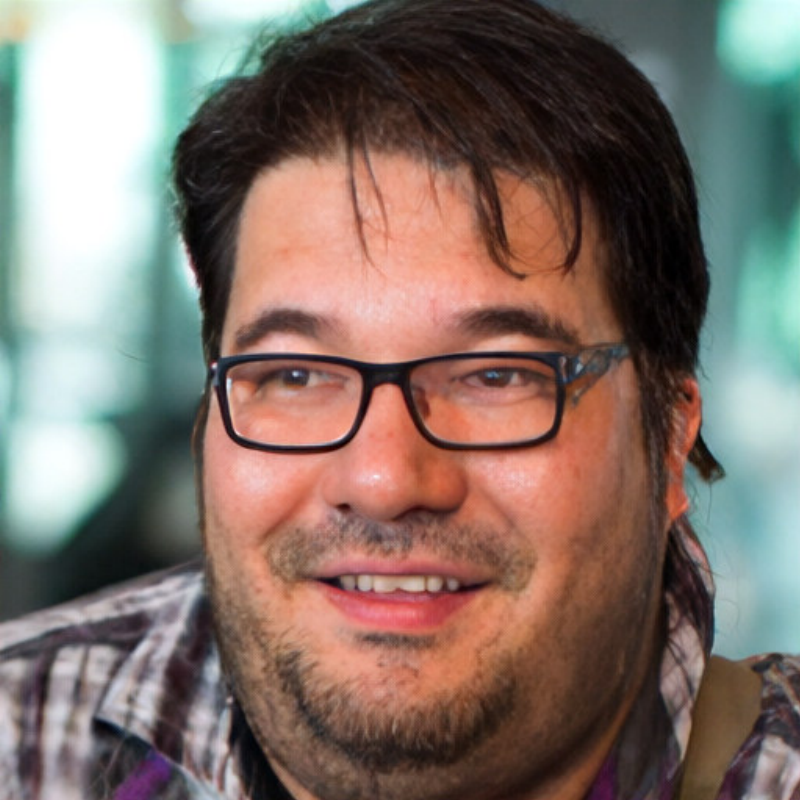}{0.27}{0.45} & \BAMPaperZoom{Figures/FFHQ/deb/sigma_0.05/BAM/3-steps/x_pred_3step_00062.pdf}{0.27}{0.45} & \BAMPaperZoom{Figures/FFHQ/deb/sigma_0.05/BAM/clean/x_clean_00062.pdf}{0.27}{0.45} \\[3pt]
\multicolumn{5}{@{}l}{\textbf{SR $\times4$}, $\sigma_y=0.05$} \\[1pt]
\BAMPaperZoom{Figures/FFHQ/sr/sigma_0.05/BAM/obs/y_00033.pdf}{0.34}{0.48} & \BAMPaperZoom{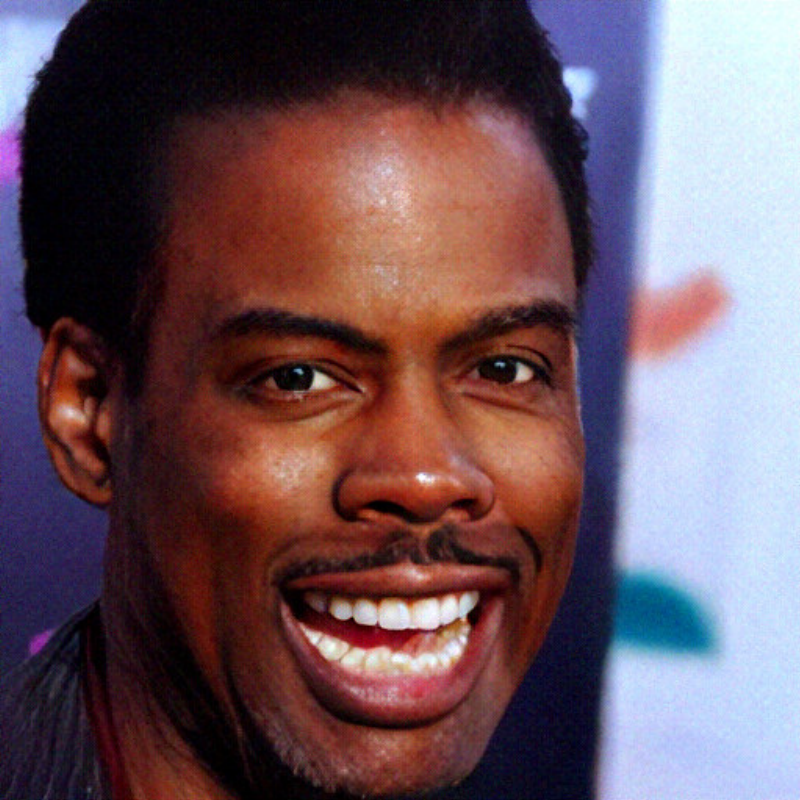}{0.34}{0.48} & \BAMPaperZoom{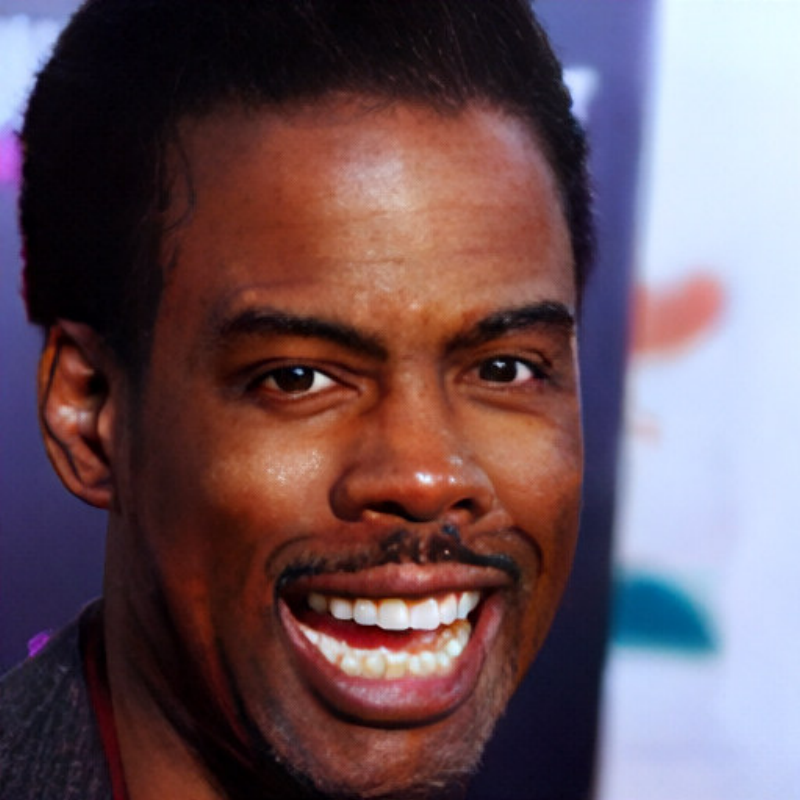}{0.34}{0.48} & \BAMPaperZoom{Figures/FFHQ/sr/sigma_0.05/BAM/3-steps/x_pred_3step_00033.pdf}{0.34}{0.48} & \BAMPaperZoom{Figures/FFHQ/sr/sigma_0.05/BAM/clean/x_clean_00033.pdf}{0.34}{0.48} \\[3pt]
\multicolumn{5}{@{}l}{\textbf{Inpainting}, $\sigma_y=0.05$} \\[1pt]
\BAMPaperZoom{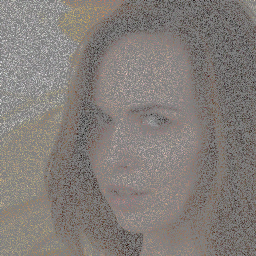}{0.36}{0.48} & \BAMPaperZoom{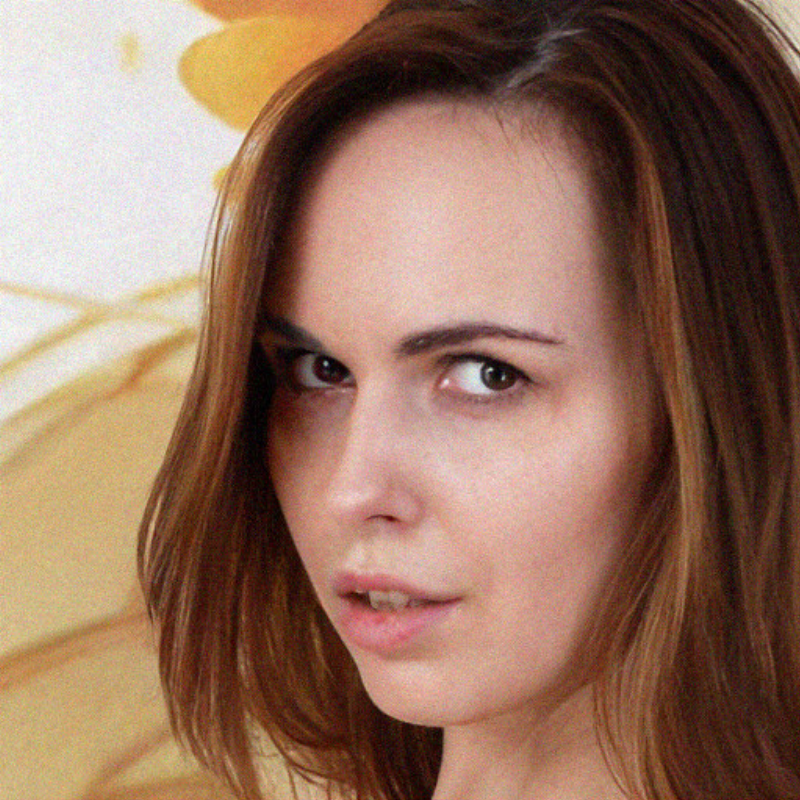}{0.36}{0.48} & \BAMPaperZoom{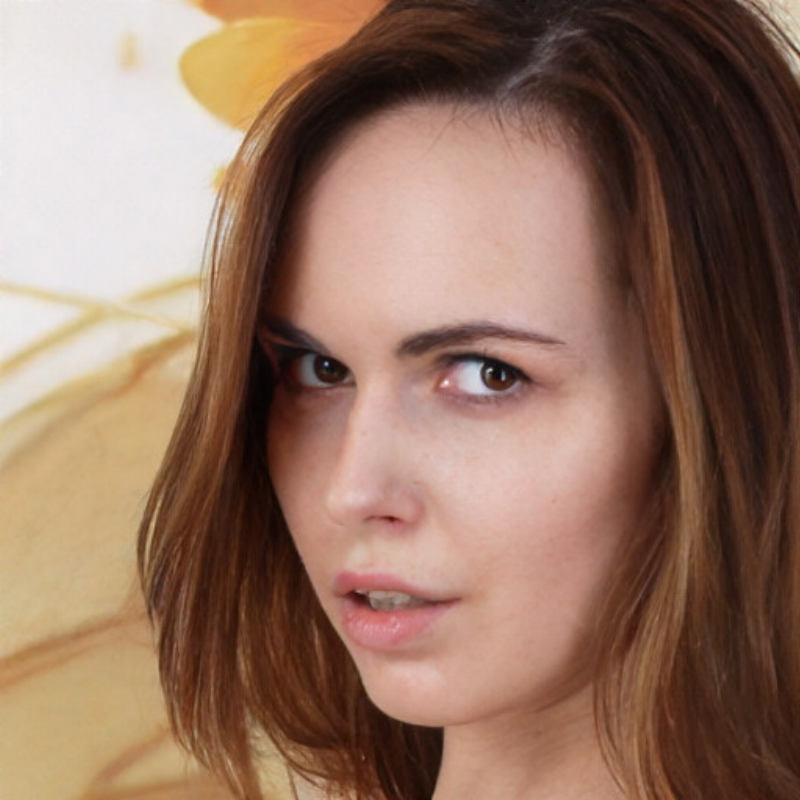}{0.36}{0.48} & \BAMPaperZoom{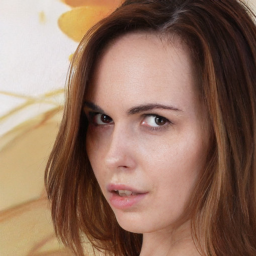}{0.36}{0.48} & \BAMPaperZoom{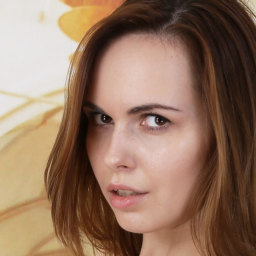}{0.36}{0.48} \\[3pt]
\multicolumn{5}{@{}l}{\textbf{Demosaicing}, $\sigma_y=0.05$} \\[1pt]
\BAMPaperZoom{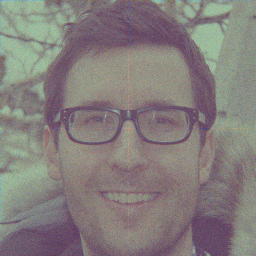}{0.27}{0.45} & \BAMPaperZoom{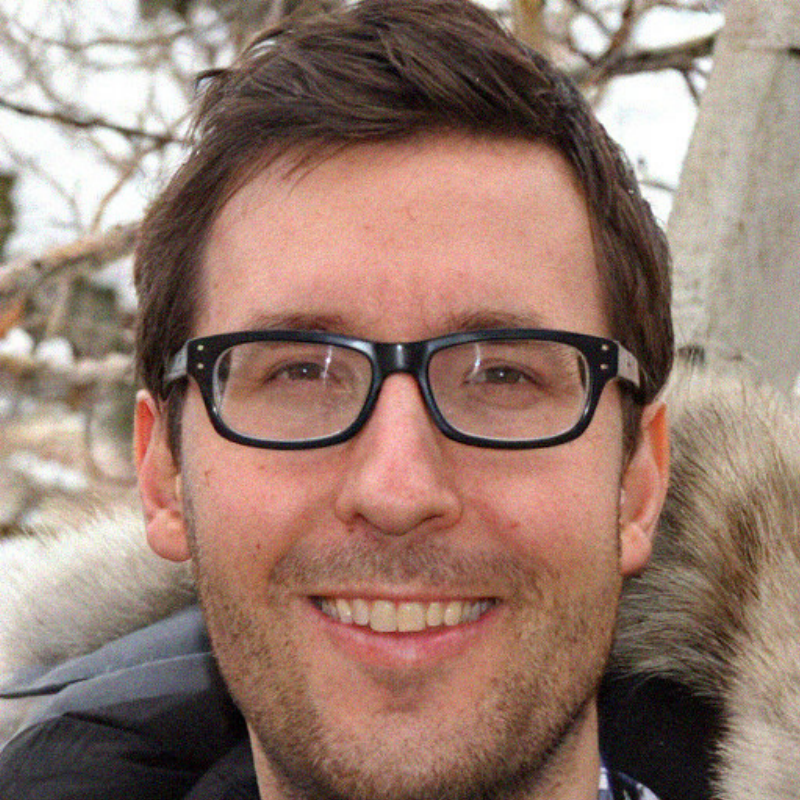}{0.27}{0.45} & \BAMPaperZoom{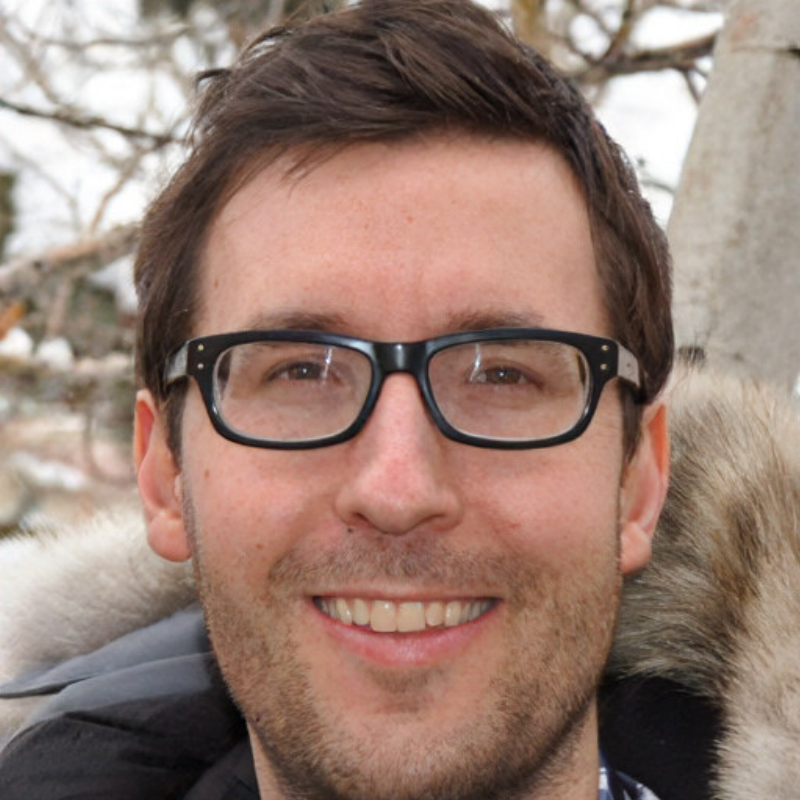}{0.27}{0.45} & \BAMPaperZoom{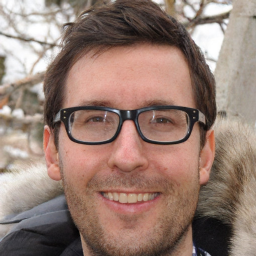}{0.27}{0.45} & \BAMPaperZoom{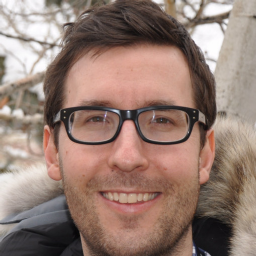}{0.27}{0.45} \\[3pt]
\multicolumn{5}{@{}l}{\textbf{Compressed sensing}, $\sigma_y=0.05$} \\[1pt]
\BAMPaperZoom{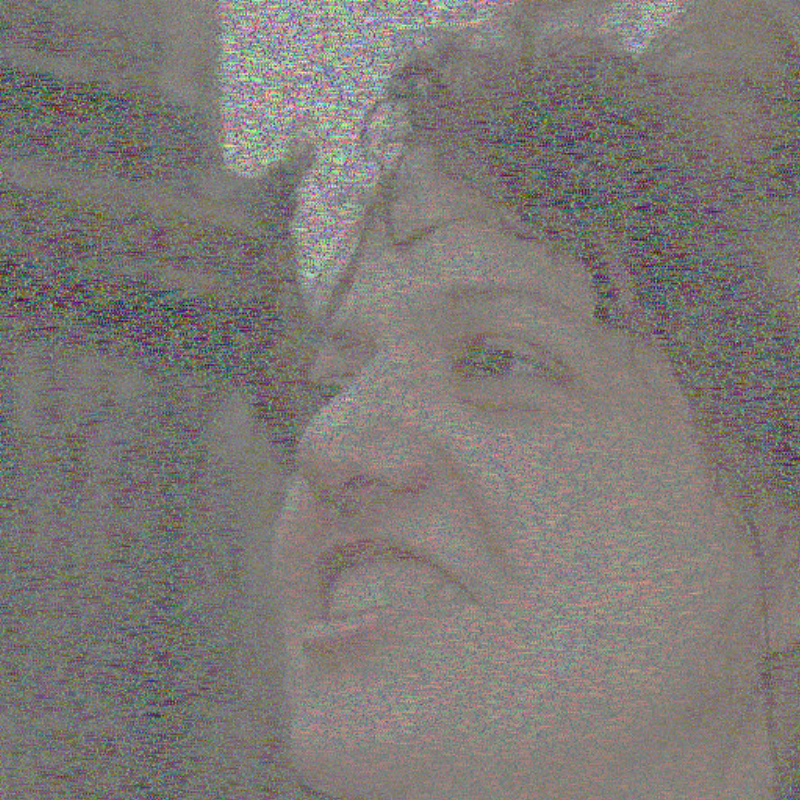}{0.45}{0.5} & \BAMPaperZoom{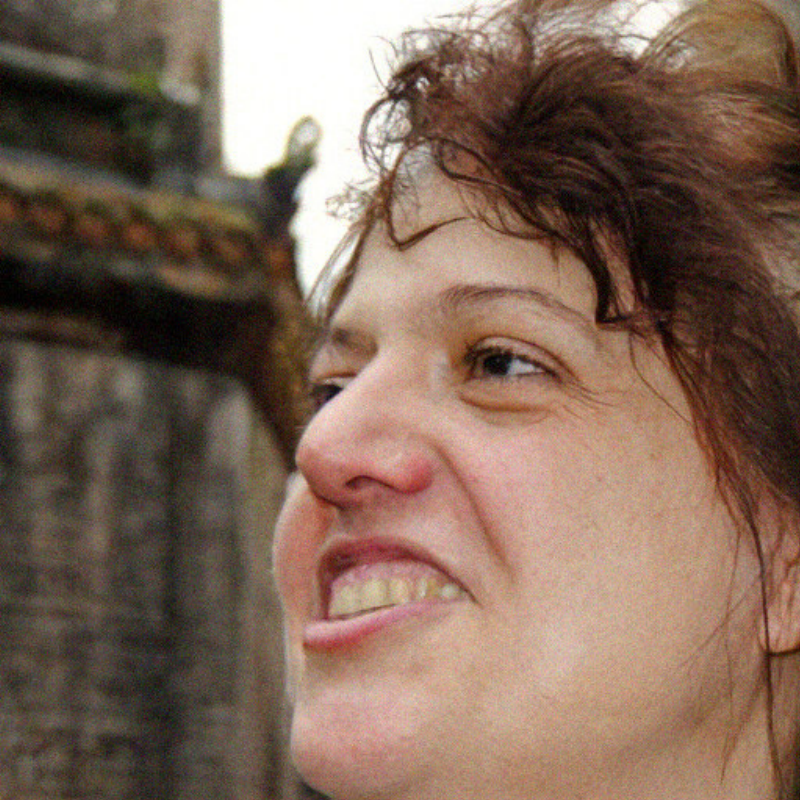}{0.45}{0.5} & \BAMPaperZoom{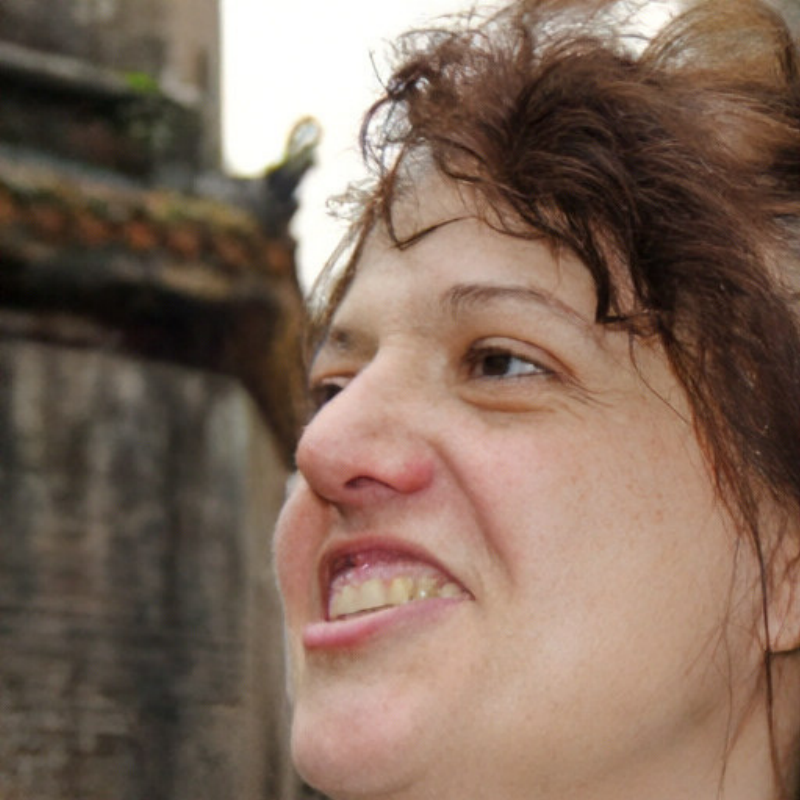}{0.45}{0.5} & \BAMPaperZoom{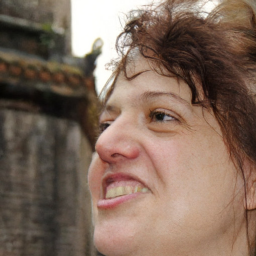}{0.45}{0.5} & \BAMPaperZoom{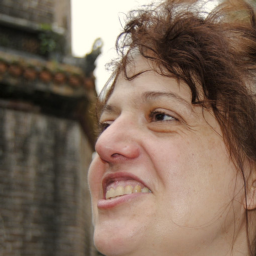}{0.45}{0.5} \\[3pt]
\bottomrule
\end{tabular}
\endgroup
\caption{\textbf{BAM sampling-step comparison on FFHQ at $\sigma_y=0.05$.} For compressed sensing, the observation column shows $A^\dagger y$. Yellow boxes select matching regions, with $4\times$ linear magnification below each image.}
\label{fig:ablation-steps-high-noise}
\end{figure}

\clearpage

\clearpage
\section{Additional experimental results}\label{app:additional-results}

\subsection{Quantitative comparisons}
The following tables show results on all five problems on which we train BAM at the two selected noise levels $\sigma_y=0.05$ and $\sigma_y=0.025$. \BAMFT and \RAMFT denote the corresponding fine-tuned models.
We do not report results for SILO in the compressed sensing setting as the original implementation of the algorithm is not compatible with this type of measurement. We limit the tasks on which we evaluate UD2M, I2SB and CoSIGN, as these models require a heavy fine-tuning process specialized for each dataset, problem, and noise level. We therefore remind that \BAMFT is instead finetuned across all problems on a specific dataset.

\begin{table}[H]
\centering
\caption{\textbf{AFHQ.} PSNR (dB; $\uparrow$), LPIPS ($\downarrow$), CMMD ($\downarrow$), and FID ($\downarrow$), with 64 images. BAM and \BAMFT use three steps. NFEs denote the number of neural function evaluations. Bold marks the best values, and underlining marks the second-best distinct value for each problem, noise level, and metric.}
\label{tab:compact-AFHQ}
\begingroup
\fontsize{8}{9}\selectfont
\setlength{\tabcolsep}{3pt}
\renewcommand{\arraystretch}{1.0}
\begin{tabularx}{\linewidth}{@{}l r *{8}{>{\raggedleft\arraybackslash}X}@{}}
\toprule
Method & NFEs & \multicolumn{4}{c}{$\sigma_y=0.025$} & \multicolumn{4}{c}{$\sigma_y=0.05$} \\
\cmidrule(lr){3-6} \cmidrule(lr){7-10}
& & PSNR$\uparrow$ & LPIPS$\downarrow$ & CMMD$\downarrow$ & FID$\downarrow$ & PSNR$\uparrow$ & LPIPS$\downarrow$ & CMMD$\downarrow$ & FID$\downarrow$ \\
\midrule

\multicolumn{10}{@{}l}{\textbf{Deblurring}} \\[-1pt]
\rowcolor{BAMRow} BAM & 3 & \underline{27.41} & \underline{0.336} & \underline{0.23} & \underline{33.52} & \underline{26.89} & \underline{0.354} & \underline{0.26} & \underline{35.62} \\
\rowcolor{BAMRow} \BAMFT & 3 & 26.85 & \textbf{0.294} & \textbf{0.18} & \textbf{23.89} & 26.41 & \textbf{0.311} & \textbf{0.19} & \textbf{24.21} \\
RAM & 1 & \textbf{27.85} & 0.417 & 1.17 & 52.11 & \textbf{27.46} & 0.444 & 1.41 & 60.40 \\
LATINO & 8 & 23.37 & 0.440 & 0.48 & 64.81 & 21.48 & 0.486 & 0.50 & 71.87 \\
LATINO-PRO & 65 & 24.62 & 0.494 & 1.44 & 67.77 & 22.63 & 0.558 & 1.60 & 92.28 \\
\bottomrule
\end{tabularx}
\endgroup
\end{table}

\clearpage
\begin{table}[H]
\centering
\caption{\textbf{DIV2K.} PSNR (dB; $\uparrow$), LPIPS ($\downarrow$), CMMD ($\downarrow$), and FID ($\downarrow$), with 64 images. BAM and \BAMFT use three steps. NFEs denote the number of neural function evaluations. Bold marks the best values, and underlining marks the second-best distinct value for each problem, noise level, and metric.}
\label{tab:compact-DIV2K}
\begingroup
\fontsize{8}{9}\selectfont
\setlength{\tabcolsep}{3pt}
\renewcommand{\arraystretch}{1.0}
\begin{tabularx}{\linewidth}{@{}l r *{8}{>{\raggedleft\arraybackslash}X}@{}}
\toprule
Method & NFEs & \multicolumn{4}{c}{$\sigma_y=0.025$} & \multicolumn{4}{c}{$\sigma_y=0.05$} \\
\cmidrule(lr){3-6} \cmidrule(lr){7-10}
& & PSNR$\uparrow$ & LPIPS$\downarrow$ & CMMD$\downarrow$ & FID$\downarrow$ & PSNR$\uparrow$ & LPIPS$\downarrow$ & CMMD$\downarrow$ & FID$\downarrow$ \\
\midrule

\multicolumn{10}{@{}l}{\textbf{Deblurring}} \\[-1pt]
\rowcolor{BAMRow} BAM & 3 & \underline{25.01} & \textbf{0.334} & \underline{0.04} & \textbf{33.51} & \underline{24.48} & \textbf{0.349} & \underline{0.06} & \textbf{34.43} \\
\rowcolor{BAMRow} \BAMFT & 3 & 24.91 & \underline{0.339} & \textbf{0.03} & \underline{33.80} & 24.44 & \underline{0.353} & \textbf{0.04} & \underline{35.59} \\
RAM & 1 & \textbf{26.08} & 0.401 & 0.37 & 41.35 & \textbf{25.63} & 0.427 & 0.44 & 52.55 \\
LATINO & 8 & 23.29 & 0.473 & 0.28 & 74.62 & 22.00 & 0.514 & 0.39 & 100.27 \\
LATINO-PRO & 65 & 24.07 & 0.454 & 0.35 & 66.02 & 23.55 & 0.536 & 0.49 & 71.90 \\
\bottomrule
\end{tabularx}
\endgroup
\end{table}

\clearpage
\begin{table}[H]
\centering
\caption{\textbf{FFHQ.} PSNR (dB; $\uparrow$), LPIPS ($\downarrow$), CMMD ($\downarrow$), and FID ($\downarrow$), with 64 images. BAM and \BAMFT use three steps. NFEs denote the number of neural function evaluations. Bold marks the best values, and underlining marks the second-best distinct value for each problem, noise level, and metric.}
\label{tab:compact-FFHQ}
\begingroup
\fontsize{8}{9}\selectfont
\setlength{\tabcolsep}{3pt}
\renewcommand{\arraystretch}{1.0}
\begin{tabularx}{\linewidth}{@{}l r *{8}{>{\raggedleft\arraybackslash}X}@{}}
\toprule
Method & NFEs & \multicolumn{4}{c}{$\sigma_y=0.025$} & \multicolumn{4}{c}{$\sigma_y=0.05$} \\
\cmidrule(lr){3-6} \cmidrule(lr){7-10}
& & PSNR$\uparrow$ & LPIPS$\downarrow$ & CMMD$\downarrow$ & FID$\downarrow$ & PSNR$\uparrow$ & LPIPS$\downarrow$ & CMMD$\downarrow$ & FID$\downarrow$ \\
\midrule

\multicolumn{10}{@{}l}{\textbf{Deblurring}} \\[-1pt]
\rowcolor{BAMRow} BAM & 3 & 30.63 & 0.232 & 0.09 & \underline{40.63} & 29.63 & 0.250 & 0.10 & \underline{44.98} \\
\rowcolor{BAMRow} \BAMFT & 3 & 31.17 & \underline{0.225} & \underline{0.08} & 47.33 & 30.16 & \underline{0.245} & \underline{0.08} & 51.61 \\
RAM & 1 & 31.07 & 0.327 & 0.94 & 83.37 & \underline{30.54} & 0.351 & 1.15 & 94.74 \\
\RAMFT & 1 & \textbf{32.31} & 0.316 & 0.85 & 82.58 & \textbf{31.53} & 0.341 & 1.07 & 92.38 \\
LATINO & 8 & 25.08 & 0.377 & 0.62 & 100.28 & 22.72 & 0.428 & 0.85 & 119.48 \\
\bottomrule
\end{tabularx}
\endgroup
\end{table}

\clearpage
\begin{table}[H]
\centering
\caption{\textbf{LSUN.} PSNR (dB; $\uparrow$), LPIPS ($\downarrow$), FID ($\downarrow$), and CMMD ($\downarrow$) with 300 test images. BAM and \BAMFT use three steps. NFEs denote the number of neural function evaluations. Bold marks the best values, and underlining marks the second-best distinct value for each problem, noise level, and metric.}
\label{tab:compact-LSUN}
\begingroup
\fontsize{8}{9}\selectfont
\setlength{\tabcolsep}{3pt}
\renewcommand{\arraystretch}{1.0}
\begin{tabularx}{\linewidth}{@{}l r *{8}{>{\raggedleft\arraybackslash}X}@{}}
\toprule
Method & NFEs & \multicolumn{4}{c}{$\sigma_y=0.025$} & \multicolumn{4}{c}{$\sigma_y=0.05$} \\
\cmidrule(lr){3-6} \cmidrule(lr){7-10}
& & PSNR$\uparrow$ & LPIPS$\downarrow$ & CMMD$\downarrow$ & FID$\downarrow$ & PSNR$\uparrow$ & LPIPS$\downarrow$ & CMMD$\downarrow$ & FID$\downarrow$ \\
\midrule

\multicolumn{10}{@{}l}{\textbf{Deblurring}} \\[-1pt]
\rowcolor{BAMRow} BAM & 3 & 25.78 & \textbf{0.154} & 1.86 & 89.35 & 25.38 & \textbf{0.160} & 2.15 & 90.52 \\
\rowcolor{BAMRow} \BAMFT & 3 & \underline{26.16} & \underline{0.155} & \textbf{1.02} & \underline{49.41} & \underline{25.79} & \textbf{0.160} & \textbf{1.26} & \underline{48.28} \\
RAM & 1 & 25.97 & 0.213 & 1.92 & 77.53 & 25.58 & 0.239 & 2.20 & 87.09 \\
\RAMFT & 1 & \textbf{27.25} & 0.304 & 1.87 & 76.90 & \textbf{26.59} & 0.333 & 2.11 & 86.54 \\
COSIGN & 2 & 24.74 & 0.215 & \underline{1.08} & \textbf{43.40} & 24.60 & \underline{0.230} & \underline{1.43} & \textbf{44.41} \\
\bottomrule
\end{tabularx}
\endgroup
\end{table}

\clearpage
\subsection{Qualitative comparisons}

\begin{figure}[H]
\centering
\begingroup
\setlength{\BAMPaperImageWidth}{.076\linewidth}
\setlength{\tabcolsep}{1pt}
\scriptsize
\begin{tabular}{@{}cccccccccc@{}}
\toprule
\textbf{Obs.} & \textbf{BAM} & \textbf{\BAMFT} & \textbf{RAM} & \textbf{\RAMFT} & \textbf{LATINO} & \textbf{L-PRO} & \textbf{TReg} & \textbf{SILO} & \textbf{GT} \\
\midrule
\multicolumn{10}{@{}l}{\textbf{AFHQ --- Gaussian deblurring}, $\sigma_y=0.025$} \\[1pt]
\BAMPaperZoom{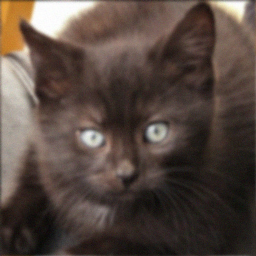}{0.25}{0.43} & \BAMPaperZoom{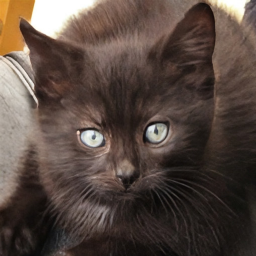}{0.25}{0.43} & \BAMPaperZoom{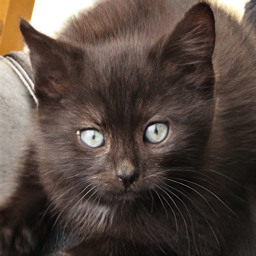}{0.25}{0.43} & \BAMPaperZoom{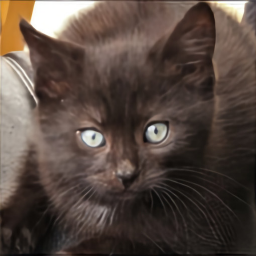}{0.25}{0.43} & \BAMPaperZoom{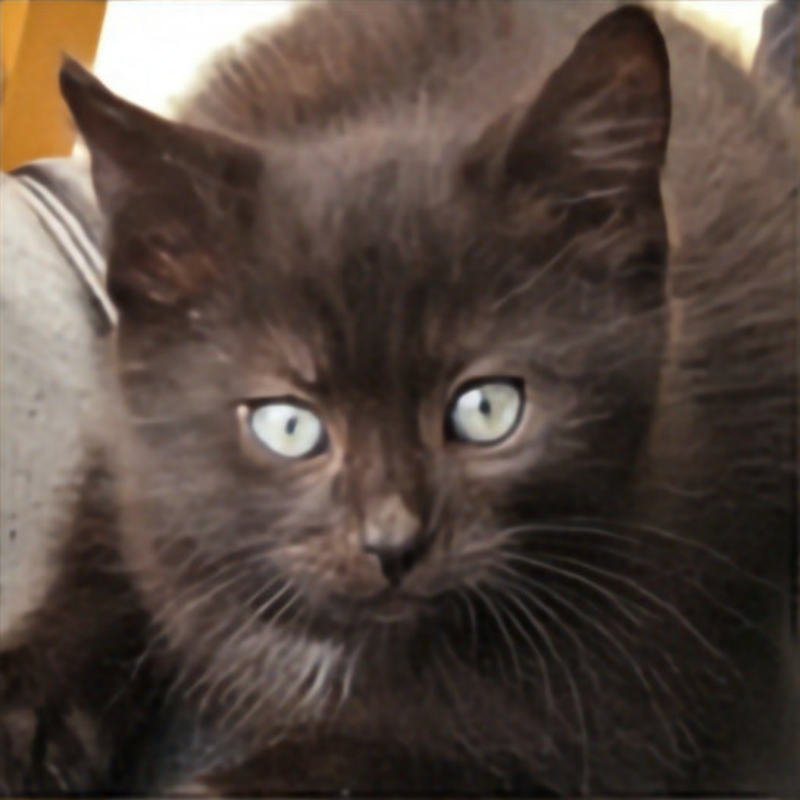}{0.25}{0.43} & \BAMPaperZoom{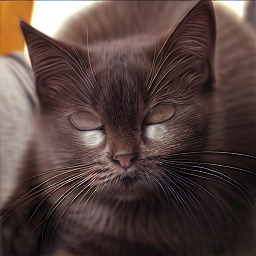}{0.25}{0.43} & \BAMPaperZoom{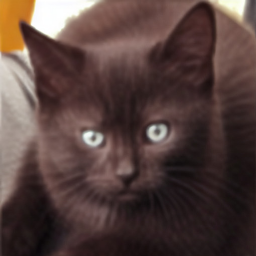}{0.25}{0.43} & \BAMPaperZoom{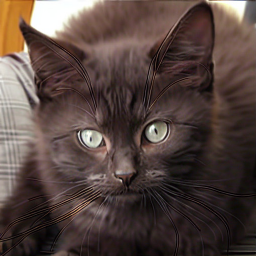}{0.25}{0.43} & \BAMPaperZoom{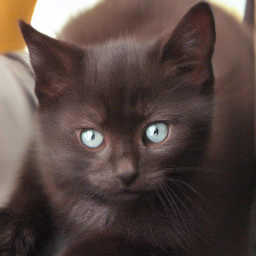}{0.25}{0.43} & \BAMPaperZoom{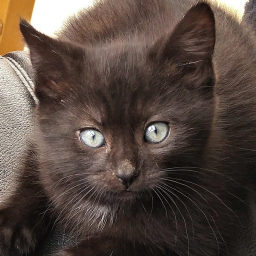}{0.25}{0.43} \\[3pt]
\multicolumn{10}{@{}l}{\textbf{AFHQ --- Gaussian deblurring}, $\sigma_y=0.05$} \\
\bottomrule
\end{tabular}
\endgroup
\caption{\textbf{AFHQ restorations.} One example per problem, repeated at both noise levels. BAM and \BAMFT use three steps. For compressed sensing, the observation column shows $A^\dagger y$. Matched yellow boxes show regions magnified $4\times$.}
\label{fig:compact-AFHQ}
\end{figure}

\clearpage
\begin{figure}[H]
\centering
\begingroup
\setlength{\BAMPaperImageWidth}{.079\linewidth}
\setlength{\tabcolsep}{1pt}
\scriptsize
\begin{tabular}{@{}cccccccccc@{}}
\toprule
\textbf{Obs.} & \textbf{BAM} & \textbf{\BAMFT} & \textbf{RAM} & \textbf{\RAMFT} & \textbf{LATINO} & \textbf{L-PRO} & \textbf{TReg} & \textbf{SILO} & \textbf{GT} \\
\midrule
\multicolumn{10}{@{}l}{\textbf{DIV2K --- Gaussian deblurring}, $\sigma_y=0.025$} \\[1pt]
\BAMPaperZoom{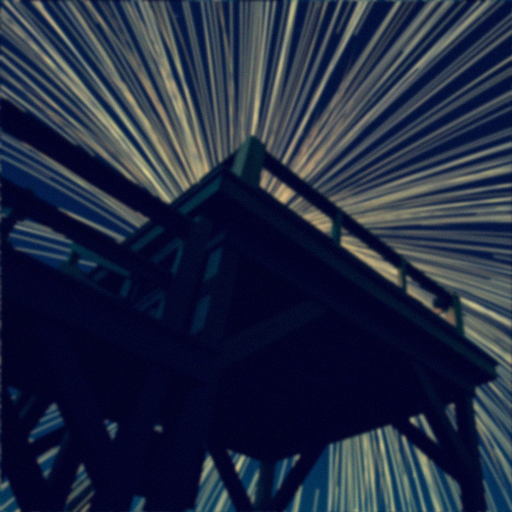}{0.4}{0.6} & \BAMPaperZoom{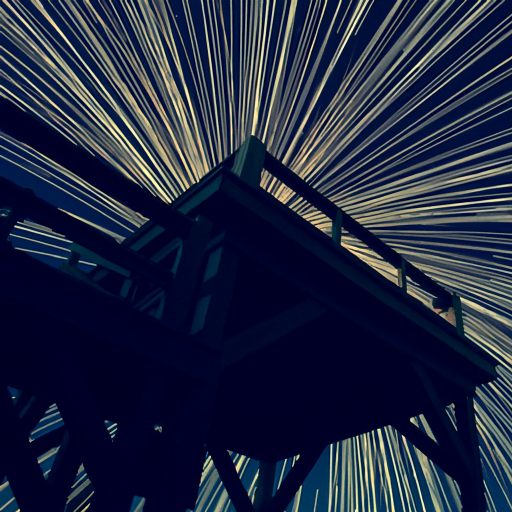}{0.4}{0.6} & \BAMPaperZoom{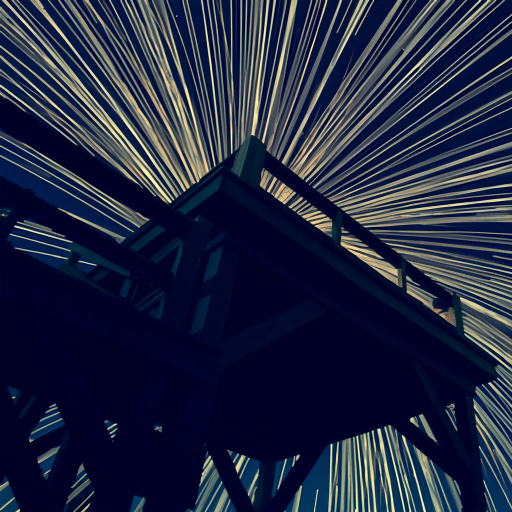}{0.4}{0.6} & \BAMPaperZoom{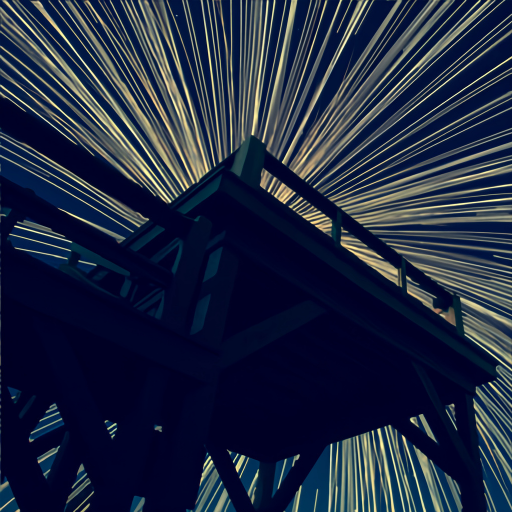}{0.4}{0.6} & \BAMPaperZoom{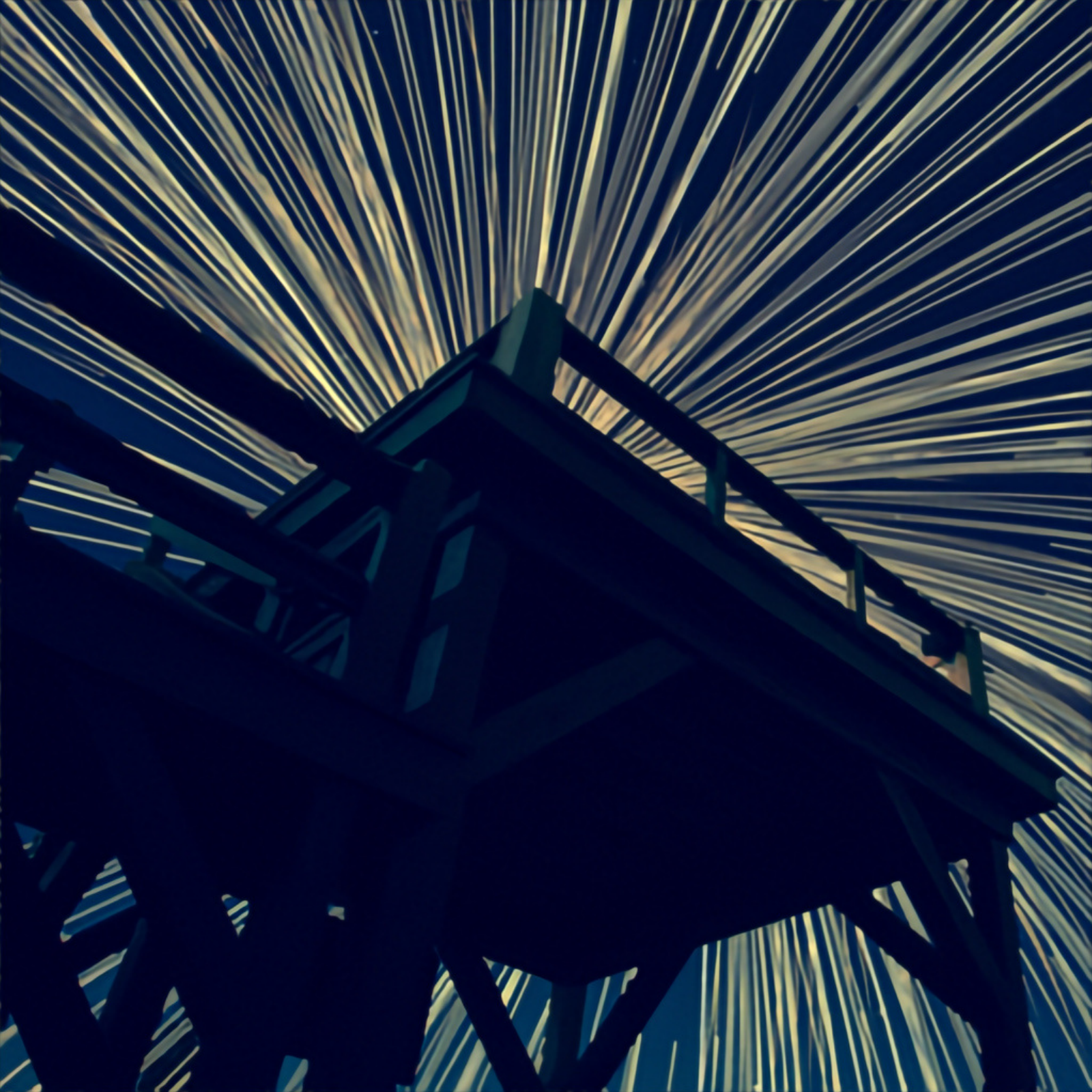}{0.4}{0.6} & \BAMPaperZoom{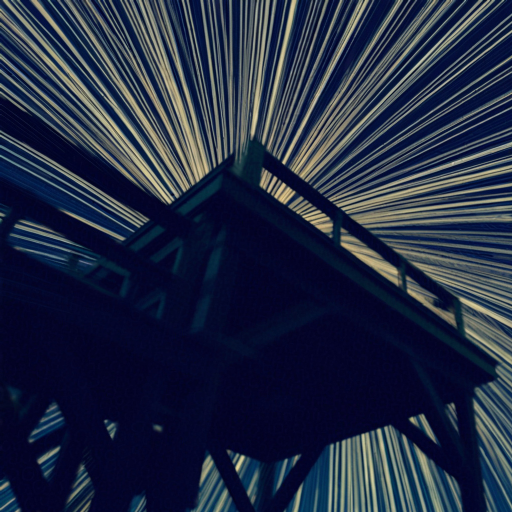}{0.4}{0.6} & \BAMPaperZoom{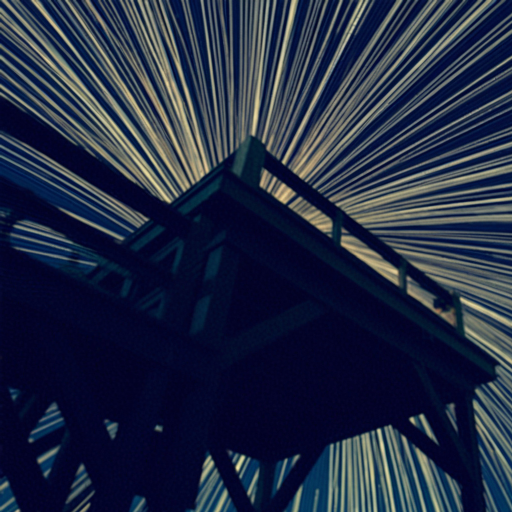}{0.4}{0.6} & \BAMPaperZoom{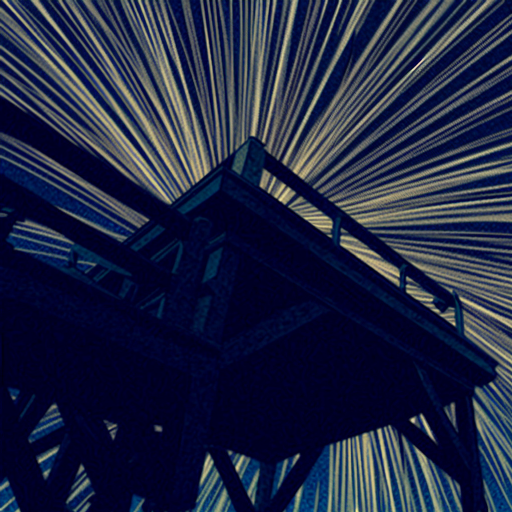}{0.4}{0.6} & \BAMPaperZoom{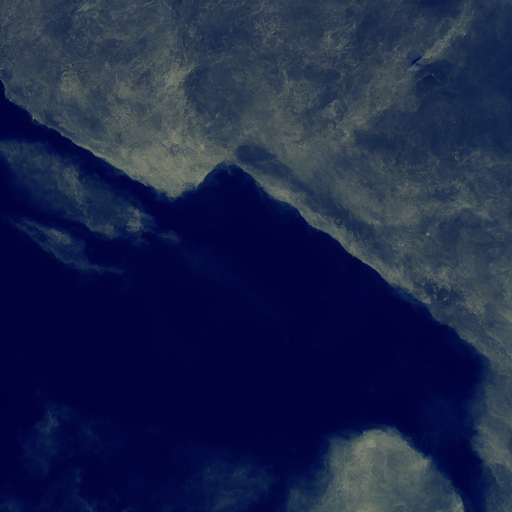}{0.4}{0.6} & \BAMPaperZoom{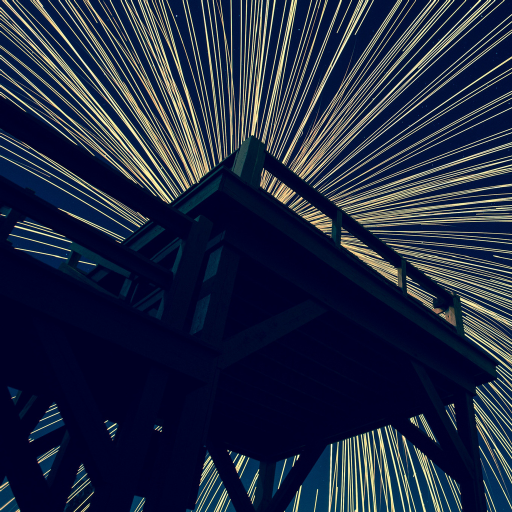}{0.4}{0.6} \\[3pt]
\multicolumn{10}{@{}l}{\textbf{DIV2K --- Gaussian deblurring}, $\sigma_y=0.05$} \\
\bottomrule
\end{tabular}
\endgroup
\caption{\textbf{DIV2K restorations.} One example per problem, repeated at both noise levels. BAM and \BAMFT use three steps.
For compressed sensing, the observation column shows $A^\dagger y$. Matched yellow boxes show regions magnified $4\times$.}
\label{fig:compact-DIV2K}
\end{figure}

\clearpage
\begin{figure}[H]
\centering
\begingroup
\setlength{\BAMPaperImageWidth}{.080\linewidth}
\setlength{\tabcolsep}{1pt}
\scriptsize
\begin{tabular}{@{}cccccccccc@{}}
\toprule
\textbf{Obs.} & \textbf{BAM} & \textbf{\BAMFT} & \textbf{RAM} & \textbf{\RAMFT} & \textbf{LATINO} & \textbf{L-PRO} & \textbf{TReg} & \textbf{SILO} & \textbf{GT} \\
\midrule
\multicolumn{10}{@{}l}{\textbf{FFHQ --- Gaussian deblurring}, $\sigma_y=0.025$} \\[1pt]
\BAMPaperZoom{Figures/FFHQ/deb/sigma_0.025/BAM/obs/y_00062.pdf}{0.27}{0.45} & \BAMPaperZoom{Figures/FFHQ/deb/sigma_0.025/BAM/3-steps/x_pred_3step_00062.pdf}{0.27}{0.45} & \BAMPaperZoom{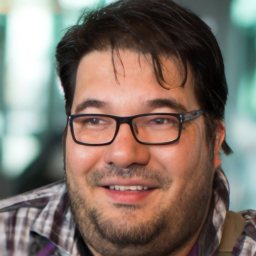}{0.27}{0.45} & \BAMPaperZoom{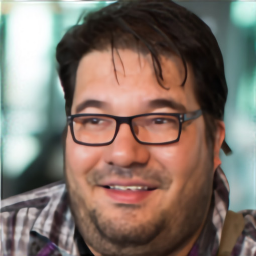}{0.27}{0.45} & \BAMPaperZoom{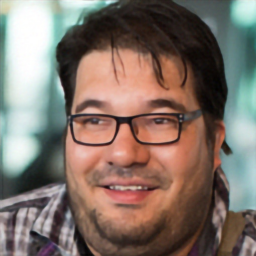}{0.27}{0.45} & \BAMPaperZoom{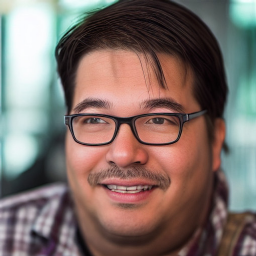}{0.27}{0.45} & \BAMPaperZoom{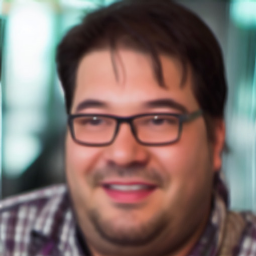}{0.27}{0.45} & \BAMPaperZoom{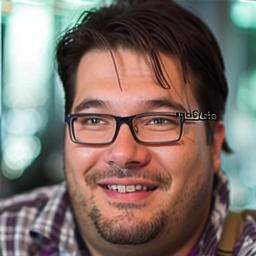}{0.27}{0.45} & \BAMPaperZoom{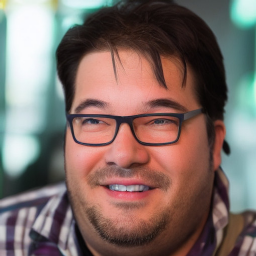}{0.27}{0.45} & \BAMPaperZoom{Figures/FFHQ/deb/sigma_0.025/BAM/clean/x_clean_00062.pdf}{0.27}{0.45} \\[3pt]
\multicolumn{10}{@{}l}{\textbf{FFHQ --- Gaussian deblurring}, $\sigma_y=0.05$} \\
\bottomrule
\end{tabular}
\endgroup
\caption{\textbf{FFHQ restorations.} One example per problem, repeated at both noise levels. BAM and \BAMFT use three steps. -- indicates an unavailable result. For compressed sensing, the observation column shows $A^\dagger y$. Matched yellow boxes show regions magnified $4\times$.}
\label{fig:compact-FFHQ}
\end{figure}

\clearpage
\begin{figure}[H]
\centering
\begingroup
\setlength{\BAMPaperImageWidth}{.12\linewidth}
\setlength{\tabcolsep}{1pt}
\scriptsize
\begin{tabular}{@{}cccccccc@{}}
\toprule
\textbf{Observation} & \textbf{BAM} & \textbf{\BAMFT} & \textbf{RAM} & \textbf{RAM$^\star$} & \textbf{CoSIGN} & \textbf{UD2M} & \textbf{GT} \\
\midrule
\multicolumn{8}{@{}l}{\textbf{LSUN --- Gaussian deblurring}, $\sigma_y=0.05$} \\[1pt]
\BAMPaperZoom{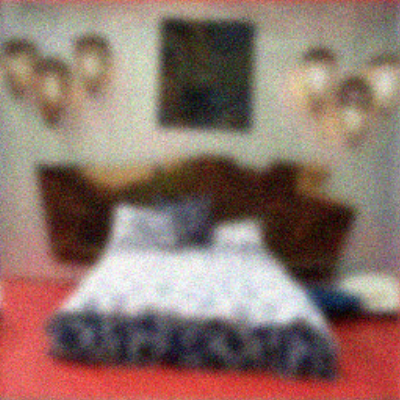}{0.26}{0.23} & \BAMPaperZoom{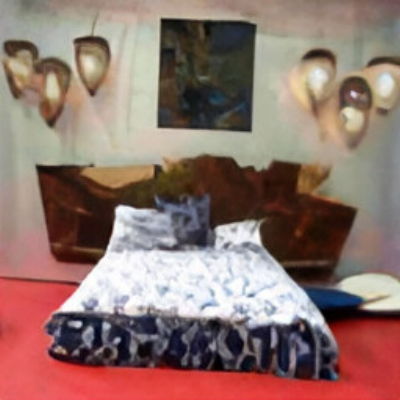}{0.26}{0.23} & \BAMPaperZoom{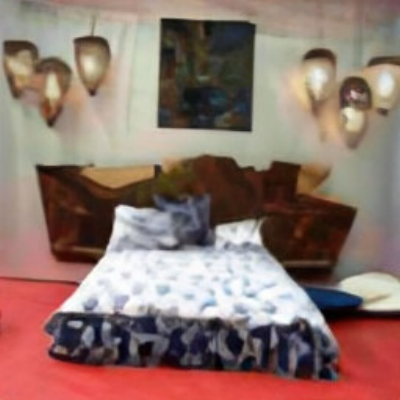}{0.26}{0.23} & \BAMPaperZoom{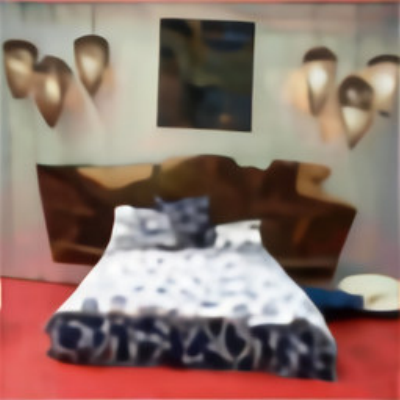}{0.26}{0.23} & \BAMPaperZoom{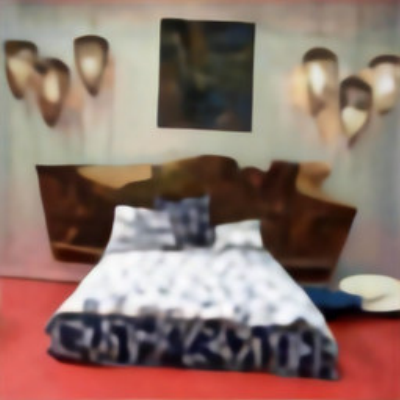}{0.26}{0.23} & \BAMPaperZoom{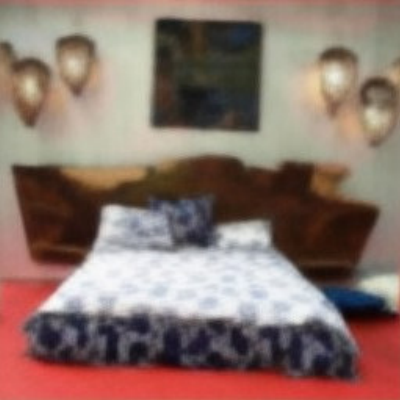}{0.26}{0.23} &
\BAMPaperZoom{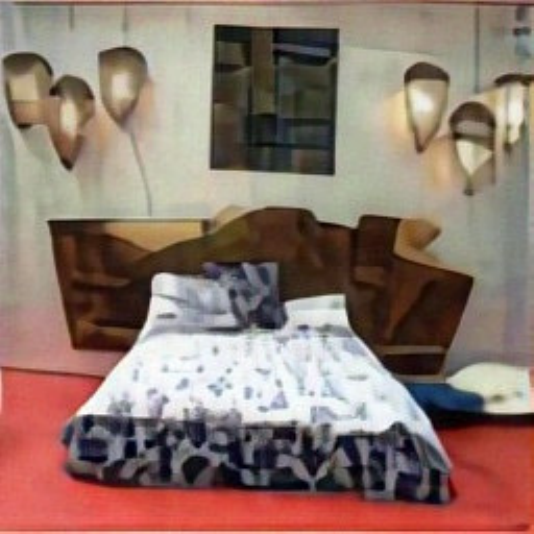}{0.26}{0.23} & \BAMPaperZoom{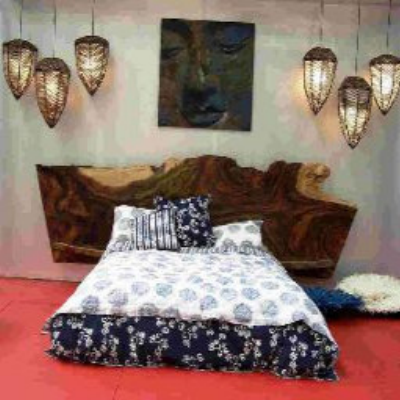}{0.26}{0.23} \\[3pt]
\multicolumn{8}{@{}l}{\textbf{LSUN --- SR $\times4$}, $\sigma_y=0.05$} \\[1pt]
\BAMPaperZoom{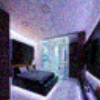}{0.42}{0.42} & \BAMPaperZoom{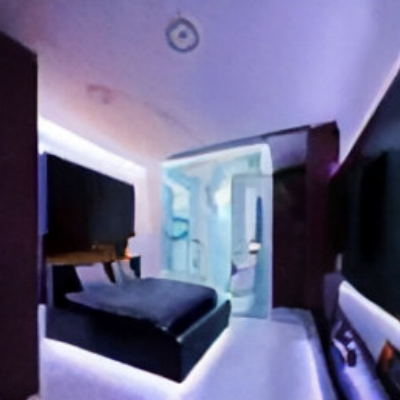}{0.42}{0.42} & \BAMPaperZoom{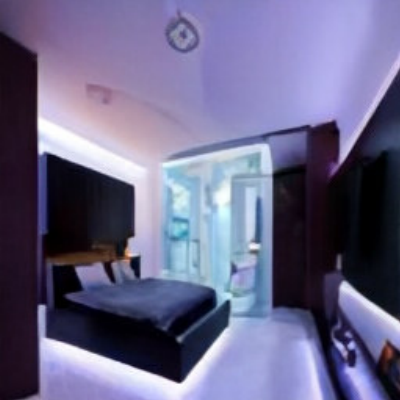}{0.42}{0.42} & \BAMPaperZoom{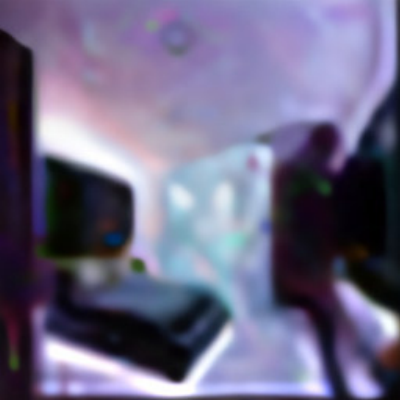}{0.42}{0.42} & \BAMPaperZoom{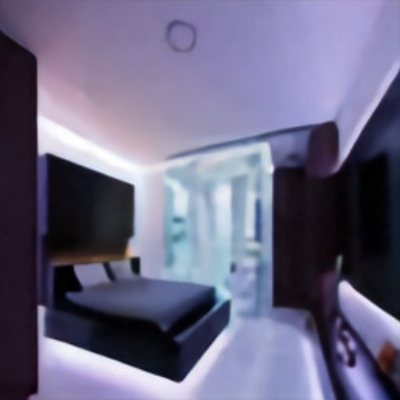}{0.42}{0.42} & \BAMPaperZoom{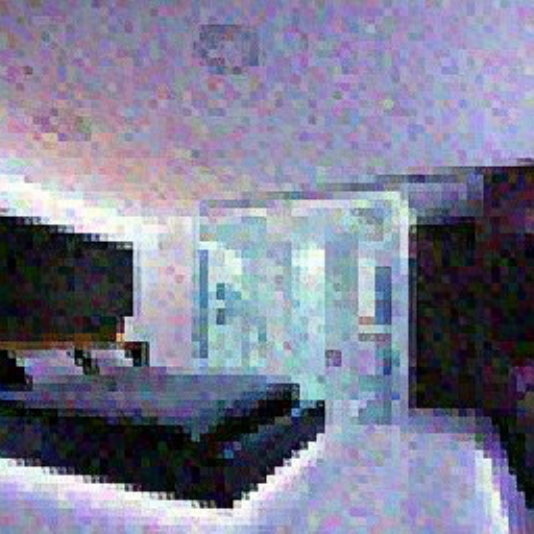}{0.42}{0.42} &
\BAMPaperZoom{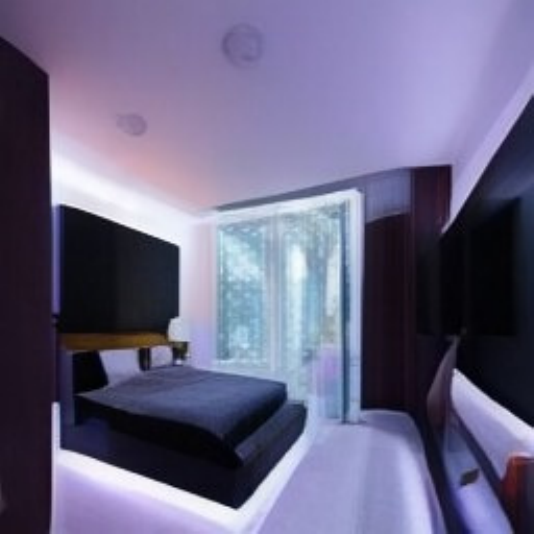}{0.42}{0.42} & \BAMPaperZoom{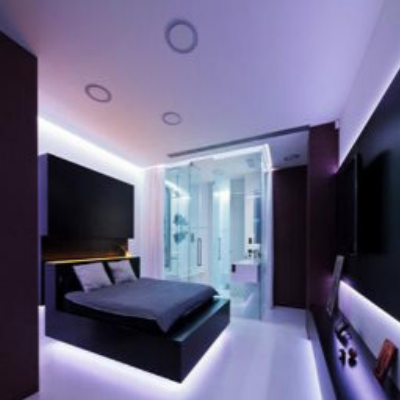}{0.42}{0.42} \\[3pt]
\multicolumn{8}{@{}l}{\textbf{LSUN --- Inpainting}, $\sigma_y=0.05$} \\
\bottomrule
\end{tabular}
\endgroup
\caption{\textbf{LSUN restorations.} One example per problem, at $\sigma_y=0.05$. BAM and \BAMFT use three steps; RAM$^\star$ denotes dataset-finetuned RAM. -- indicates an unavailable result. For compressed sensing, the observation column shows $A^\dagger y$. Matched yellow boxes show regions magnified $4\times$.}
\label{fig:compact-LSUN}
\end{figure}

\clearpage
\clearpage
\subsection{Sparse-view computed tomography}\label{app:ct}
\paragraph{Acquisition and data.}
We follow the Gaussian sparse-view CT evaluation setting of
\citet{Terris2025ReconstructAM} on LIDC-IDRI chest scans.
We use single-channel $512\times512$ images and a random slice split of
$95\%/4\%/1\%$ for training, validation, and testing.
The observation model is
\begin{equation}
  \vy=\mathcal A\vx+\sigma_y\vn,\qquad
  \vn\sim\mathcal N(0,I),\qquad \sigma_y=10^{-4},
  \label{eq:ct-observation}
\end{equation}
where $\mathcal A$ is a parallel-beam Radon transform with 51 uniformly spaced
projection angles in $[0,180^{\circ})$.
We use the DeepInv~\citep{Tachella2025DeepInverseAP} discretization with full-square image support and
spectral-norm normalization.

\begin{figure}[H]
\centering
\begingroup
\small
\begin{tabular}{@{}ccc@{}}
Observation $\vy$ & BAM standard deviation ($N=64$) & $|\mathrm{GT}-\mathrm{mean}|$ \\
\includegraphics[width=.285\linewidth]{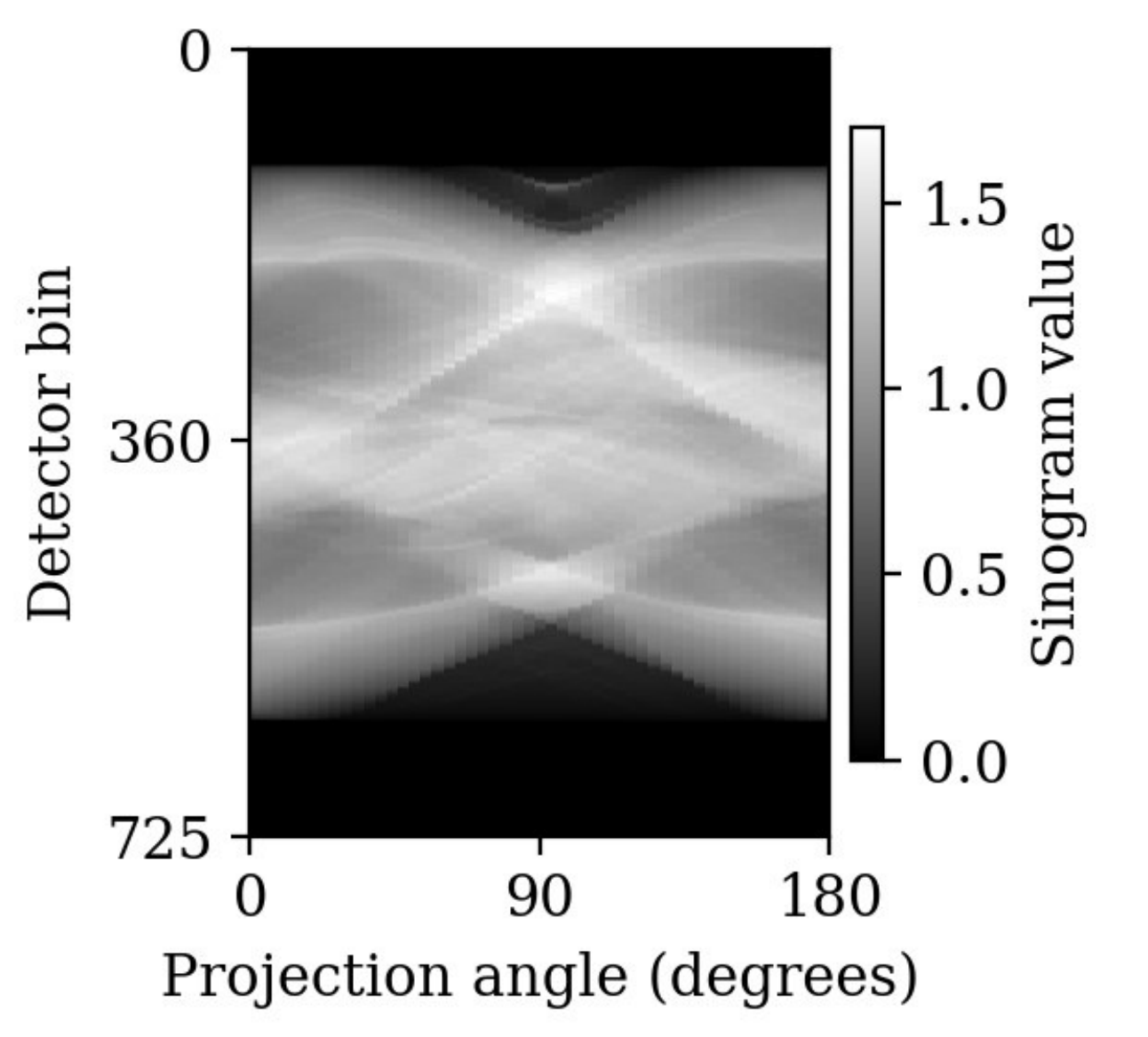} &
\includegraphics[width=.285\linewidth]{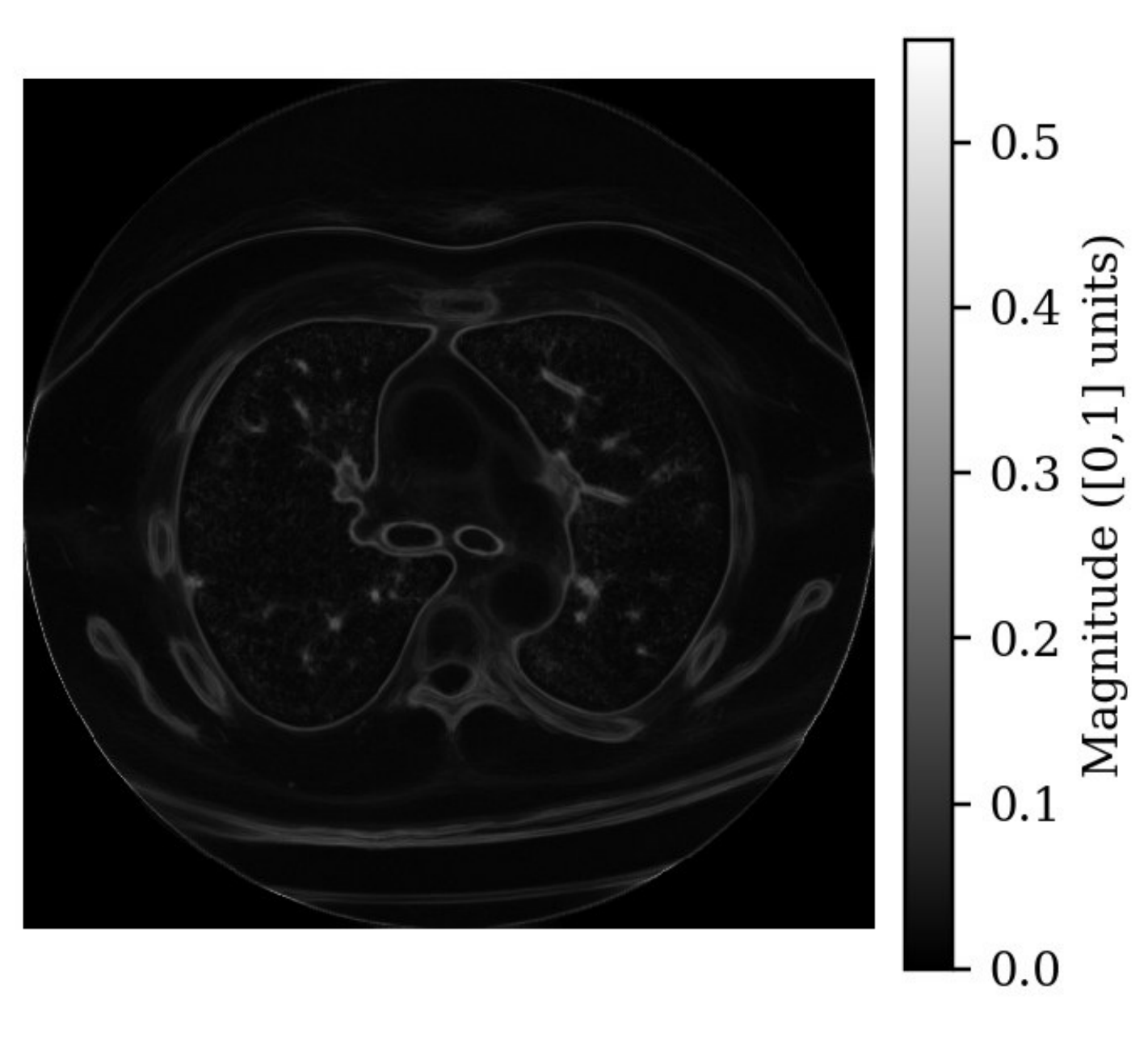} &
\includegraphics[width=.285\linewidth]{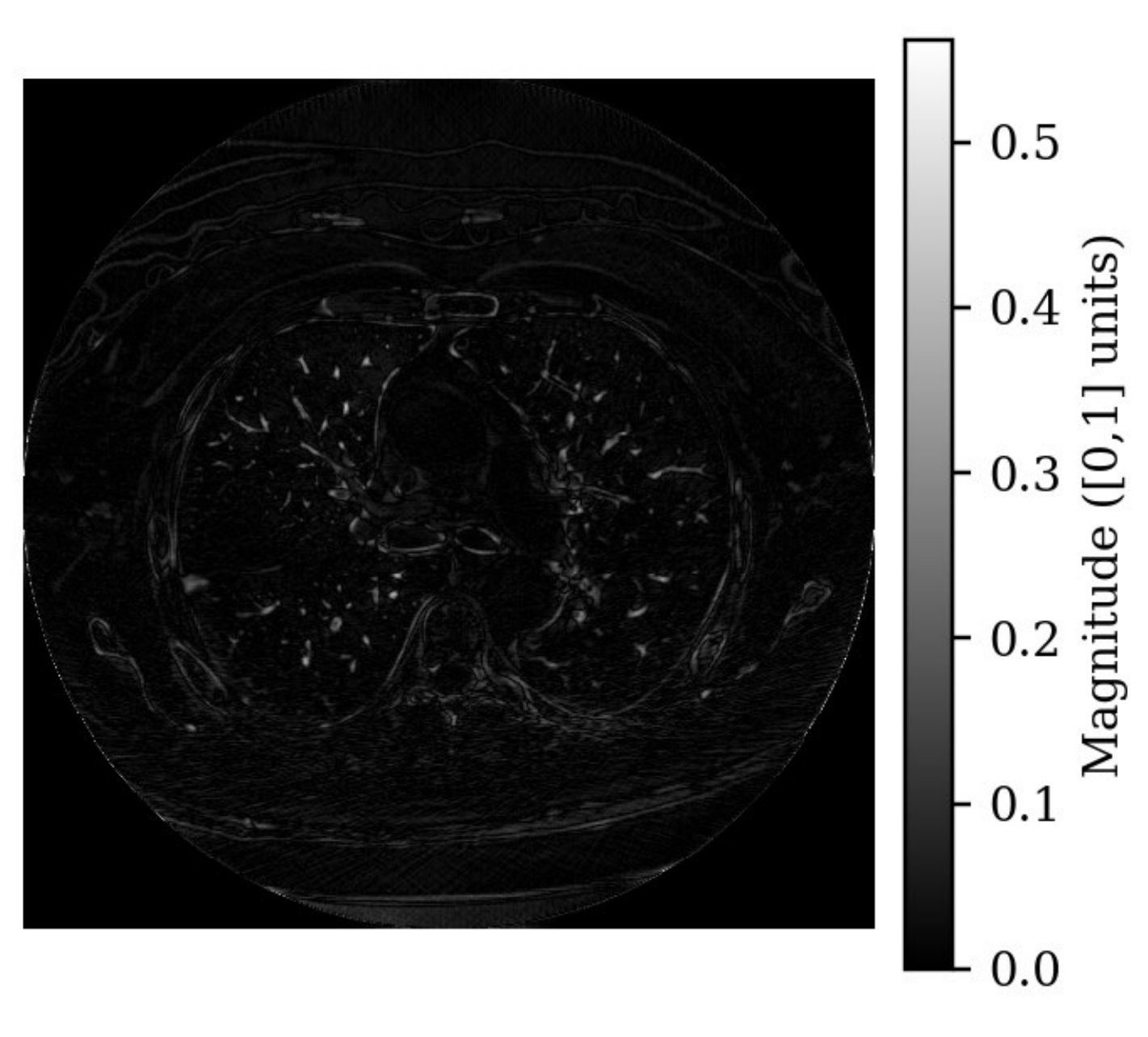}
\end{tabular}

\vspace{0.7em}
\scriptsize
\setlength{\tabcolsep}{1.2pt}
\begin{tabular}{@{}ccccc@{}}
$\mathcal A^{\dagger}\vy$ (FBP) & BAM sample & BAM mean ($N=64$) & RAM & Ground truth \\[2pt]
\includegraphics[width=.193\linewidth]{Figures/CT/fbp_panel.pdf} &
\includegraphics[width=.193\linewidth]{Figures/CT/bam_sample_panel.pdf} &
\includegraphics[width=.193\linewidth]{Figures/CT/bam_mean_panel.pdf} &
\includegraphics[width=.193\linewidth]{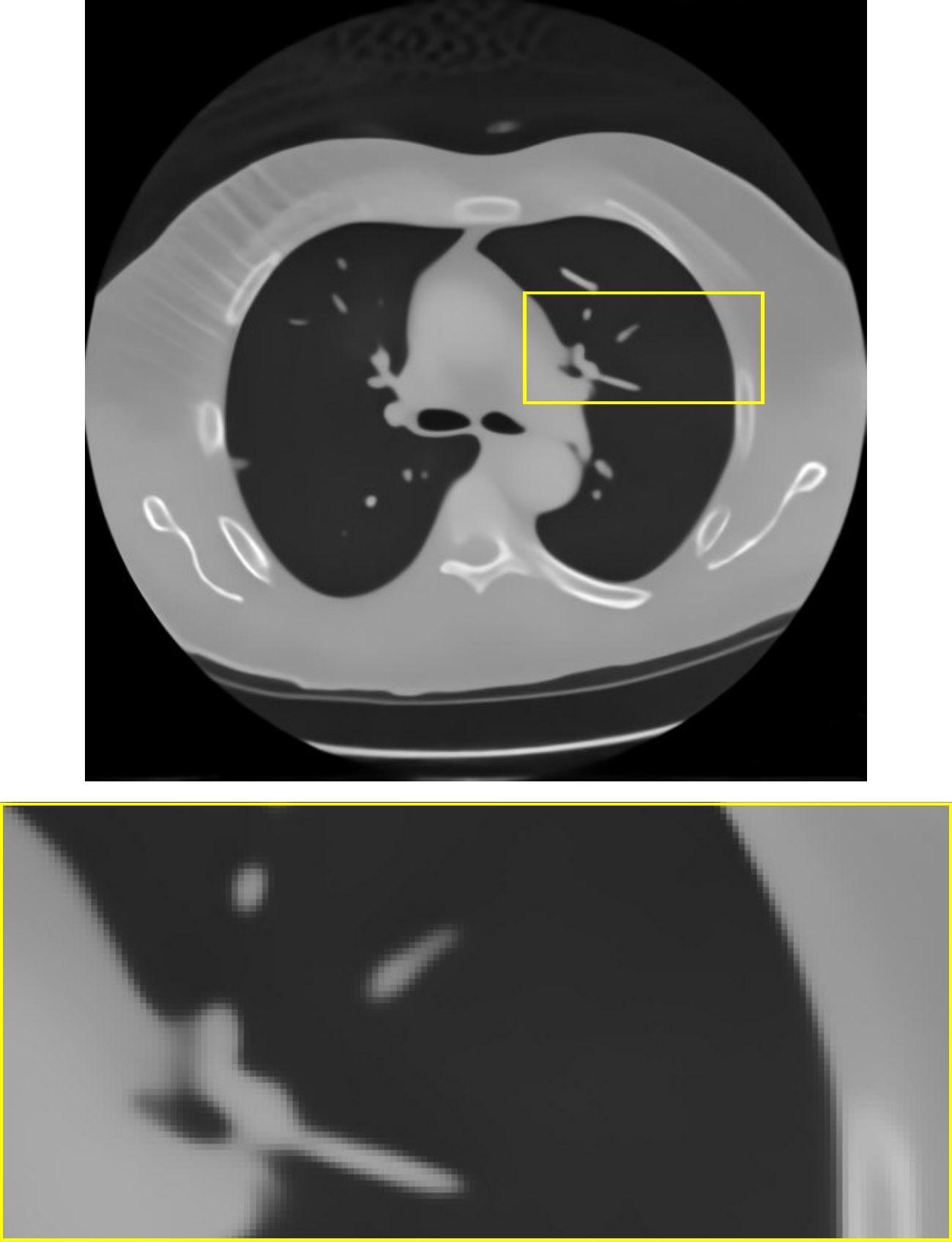} &
\includegraphics[width=.193\linewidth]{Figures/CT/gt_panel.pdf}
\end{tabular}
\endgroup
\caption{\textbf{Gaussian sparse-view CT on LIDC-IDRI.}
The upper panels show the sinogram, the pixelwise standard deviation of
64 BAM draws, and the absolute difference between ground truth and the empirical mean. The lower panels compare filtered backprojection, one BAM
draw, the BAM mean, RAM, and the reference slice.
Yellow rectangles select the same lung region; the strips immediately below
show that region at $4\times$ linear magnification.
}
\label{fig:ct-comparison}
\end{figure}

\paragraph{Reconstructions and sample variability.}
Figure~\ref{fig:ct-comparison} compares a 10h CT-finetuned BAM draw, its empirical posterior mean, the original pretrained RAM output, and ground truth for a test slice. BAM uses three reconstruction steps, as in the other experiments throughout the paper. The mean and pixelwise standard deviation are computed from $N=64$ repeated
draws for this slice with a fixed observation:
\begin{equation}
  \bar x_i=\frac{1}{N}\sum_{k=1}^{N}x_i^{(k)},\qquad
  s_i=\left(\frac{1}{N}\sum_{k=1}^{N}
  \bigl(x_i^{(k)}-\bar x_i\bigr)^2\right)^{1/2}.
\end{equation}
The standard deviation is displayed in $[0,1]$ image units and measures
variation among the generated samples.
For the linear reconstruction baseline, we display ramp-filtered
backprojection, denoted $\mathcal A^{\dagger}\vy$ (FBP). We observe that the single-sample increased details yield an empirical mean that reveals otherwise hidden structures in the magnified lung region (compared to RAM).

\begin{figure}[H]
\centering
\textbf{$4\times$ downsampling: $4\times4$ block averages ($128\times128$)}\\[3pt]
\includegraphics[width=.78\linewidth]{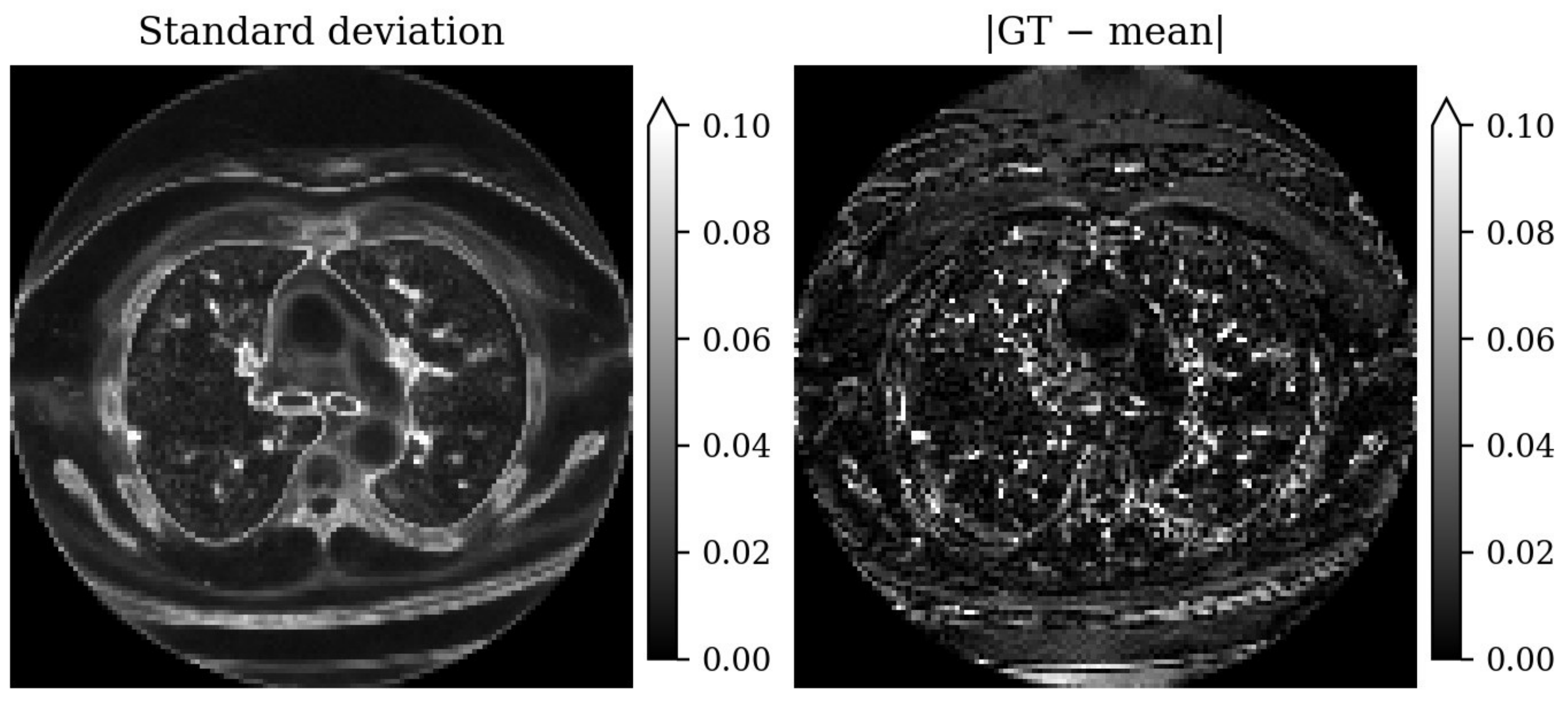}

\vspace{1em}
\textbf{$8\times$ downsampling: $8\times8$ block averages ($64\times64$)}\\[3pt]
\includegraphics[width=.78\linewidth]{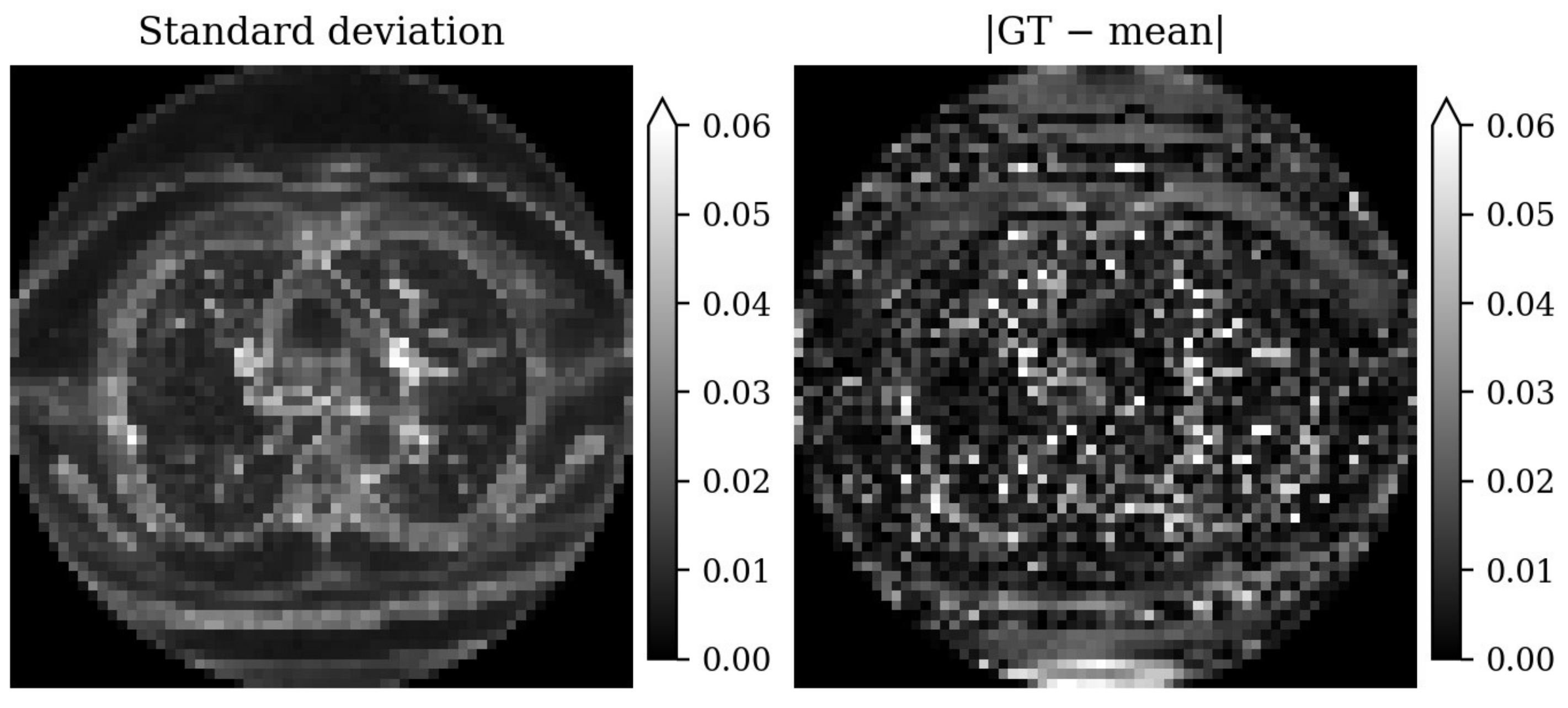}
\caption{\textbf{Multiscale CT variability and absolute mean error.}
Each row compares the population standard deviation (left) with
$|\mathrm{GT}-\mathrm{mean}|$ (right), recomputed after block averaging each sample and the ground truth.
All maps are in $[0,1]$ image units. Black denotes zero and brighter values indicate larger variability or
absolute error. These display ranges enhance contrast at each scale.}
\label{fig:ct-multiscale}
\end{figure}

\clearpage
\paragraph{Multiscale variability and reconstruction error.}
We compare sample variability and the absolute mean error after spatial averaging
in Figure~\ref{fig:ct-multiscale}. Let $D_k$ denote non-overlapping $k\times k$
block averaging, with $k\in\{4,8\}$. We first downsample each of the 64 draws
and the ground truth, then compute
\begin{equation}
 \bar x^{[k]}=\frac{1}{N}\sum_{j=1}^{N}D_kx^{(j)},\qquad
 s^{[k]}=\left[\frac{1}{N}\sum_{j=1}^{N}
 (D_kx^{(j)}-\bar x^{[k]})^2\right]^{1/2},\qquad
 e^{[k]}=\left|D_kx_{\mathrm{GT}}-\bar x^{[k]}\right|.
\end{equation}
All operations in the variance expression are pixelwise. In particular,
$s^{[k]}$ is recomputed across downsampled draws, rather than obtained by
averaging the original standard-deviation map.

\raggedbottom
\subsection{Motion deblurring experiments}\label{app:motion}
We consider three settings: \emph{(i)} a known spatially invariant motion blur kernel, and two blind cases, namely \emph{(ii)} synthetic spatially varying blur with an estimated operator, and \emph{(iii)} real blurred photographs from the~\citet{kohler2012recording} dataset.

\paragraph{\emph{(i)} Known, spatially invariant motion blur.}
Motion blur is included in the mixed-operator training problem, alongside Gaussian blur.
We draw a normalized $61\times61$ kernel using the random-trajectory generator with intensity $0.5$: random step lengths and turning angles define a polyline, which is rasterized, smoothed, downsampled, and normalized to unit sum.
The forward operator is circular convolution with this kernel.
The sampled kernel is known and held fixed during reconstruction; the saved evaluation runs draw a fresh kernel for each image, rather than sharing one realization across the entire dataset.
Synthetic observations use additive Gaussian noise with $\sigma_y=0.025$.

\paragraph{\emph{(ii)} Synthetic spatially varying motion blur: blind reconstruction.}
We draw 25 independent $33\times33$ motion kernels $h_k$.
Each kernel results from a 2D trajectory of 1000 points, with independent horizontal and vertical coordinates drawn from the DeepInv Mat\'ern Gaussian-process generator (length scale $0.3$, standard deviation $0.25$). This process simulates camera-shake motion blur.
Centered trajectories are histogrammed into kernels and normalized to unit sum.
The actual blur operator linearly combines these kernels with pixel dependent weights $w_k$:
\begin{equation}
  \mathcal A\vx=\sum_{k=1}^{25}w_k\odot(h_k*\vx),\qquad \sum_{k=1}^{25}w_k=\mathbf 1.
  \label{eq:motion-spatial}
\end{equation}
The weights $w_k$ are the bilinear (tent) interpolation functions associated with the nodes of a uniform $5\times5$ grid spanning the image, so that $w_k\ge 0$, $\sum_k w_k=\mathbf 1$, and each pixel's effective kernel is a convex combination of at most four neighbouring $h_k$.

The true kernels generate the observations but are not supplied to either reconstruction method.
Instead, we use the pretrained kernel prediction network of \citet{carbajal2023blind}, which estimates image-dependent basis kernels and spatial mixing coefficients from the observation.
Its 25 estimated $33\times33$ kernels and weights define a frozen linear operator $\widehat{\mathcal A}$ of the form~\eqref{eq:motion-spatial}, with adjoint $\widehat{\mathcal A}^{*}z=\sum_k\widehat h_k^{*}*(\widehat w_k\odot z)$.
Both BAM and RAM use this estimated operator; no ground-truth kernel enters reconstruction.

\paragraph{\emph{(iii)} Real K\"ohler observations.}
We also apply the same kernel prediction network to the captured blurry images in the~\citet{kohler2012recording} dataset, without synthesizing additional blur or observation noise.

\begin{figure}%
\centering
\includegraphics[width=.82\linewidth]{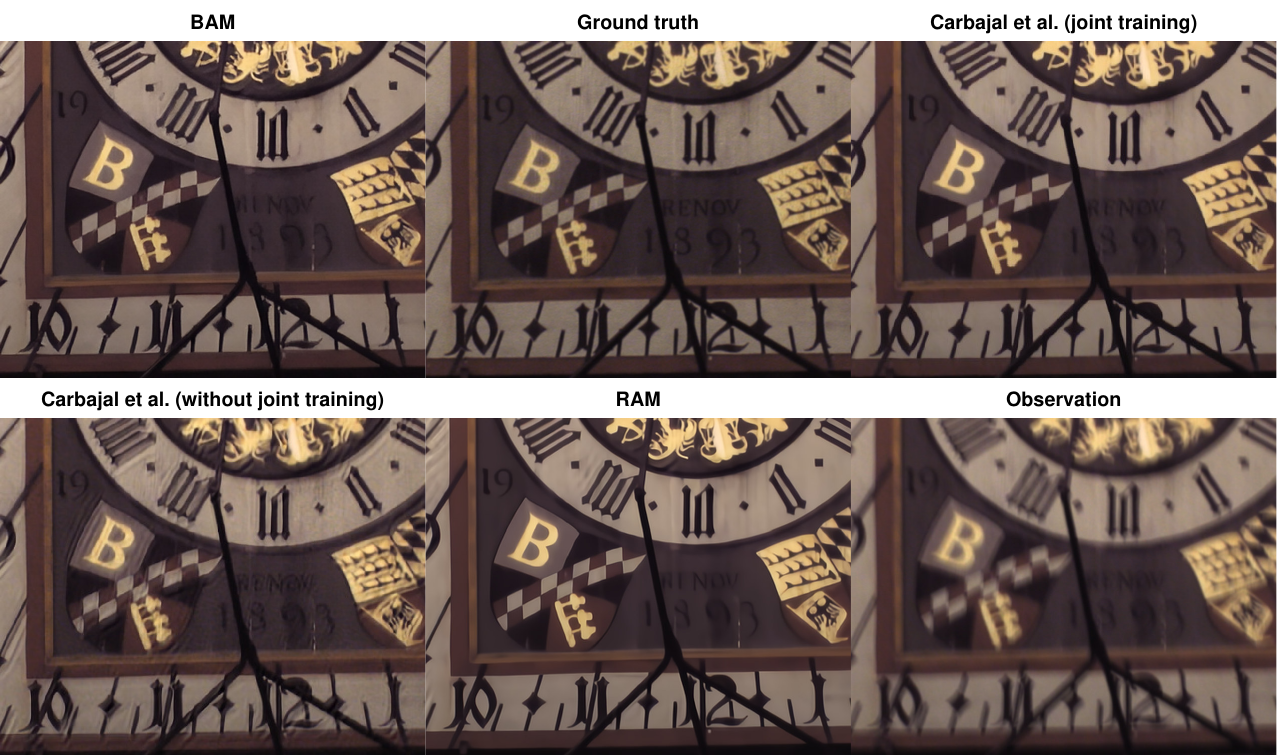}
\caption{\small\textbf{K\"ohler clock detail.} Top row, left to right: BAM, ground truth, and Carbajal et al.\ with joint training. Bottom row: Carbajal et al.\ without joint training, RAM, and the observation. Joint training refers to the kernel-estimation/deconvolution pipeline of \citet{carbajal2023blind}.}
\label{fig:kohler-carbajal}
\end{figure}

For both blind settings (\emph{(ii)} and \emph{(iii)}), the blur estimator receives the observation clipped to $[0,1]$, with $\gamma=1$ (no inverse-gamma correction), and its output remains fixed during reconstruction.
Figure~\ref{fig:motion-three-settings} compares BAM and RAM in all three settings.
Figure~\ref{fig:kohler-carbajal} shows the supplied clock crop together with the jointly trained and separately trained variants of \citet{carbajal2023blind}.

\begin{wraptable}{r}{0.52\textwidth}
\vspace{-1.8em}
\centering
\label{tab:motion-div2k}
\begingroup
\fontsize{7}{8}\selectfont
\setlength{\tabcolsep}{2.2pt}
\renewcommand{\arraystretch}{1.0}
\begin{tabular}{@{}lrrrrrrrr@{}}
\toprule
& \multicolumn{4}{c}{Known kernel}
& \multicolumn{4}{c}{Spatially varying} \\
\cmidrule(lr){2-5} \cmidrule(lr){6-9}
Method
& PSNR & LPIPS & CMMD & FID
& PSNR & LPIPS & CMMD & FID \\
\midrule
\rowcolor{BAMRow} BAM & 25.81 & \textbf{0.329} & \textbf{0.05} & \textbf{40.56} & \textbf{23.37} & \textbf{0.340} & \textbf{0.20} & \textbf{40.32} \\
RAM & \textbf{27.49} & 0.355 & 0.25 & 45.84 & 22.85 & 0.388 & 0.50 & 50.80 \\
\bottomrule
\end{tabular}
\caption{\textbf{DIV2K motion deblurring.} PSNR (dB; $\uparrow$),
LPIPS ($\downarrow$), CMMD ($\downarrow$), and FID ($\downarrow$), over 64 images.
BAM uses three reconstruction steps.}
\endgroup
\vspace{-0.8em}
\end{wraptable}

\paragraph{Quantitative comparison.}
Table~\ref{tab:motion-div2k} reports the quantitative results on the two
synthetic motion-deblurring settings. Note that in the third setting (Real Köhler observations), no ground truth is available, so we can only perform a qualitative evaluation as shown in Figures~\ref{fig:motion-three-settings}~and~\ref{fig:kohler-carbajal}.

\begin{figure}%
\centering
\begingroup
\scriptsize
\setlength{\BAMPaperImageWidth}{.18\linewidth}
\setlength{\tabcolsep}{1pt}
\begin{tabular}{@{}cccc@{}}
\toprule
\textbf{Observation} & \textbf{BAM (3 steps)} & \textbf{RAM} & \textbf{GT} \\
\midrule
\multicolumn{4}{@{}l}{\textbf{Known motion kernel --- DIV2K}} \\[1pt]
\BAMPaperZoom{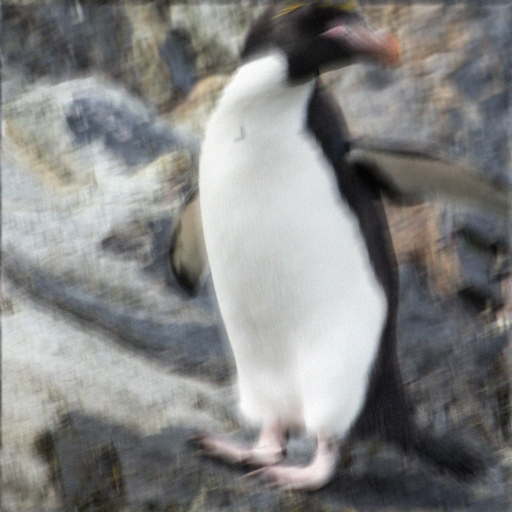}{0.47}{0.73} & \BAMPaperZoom{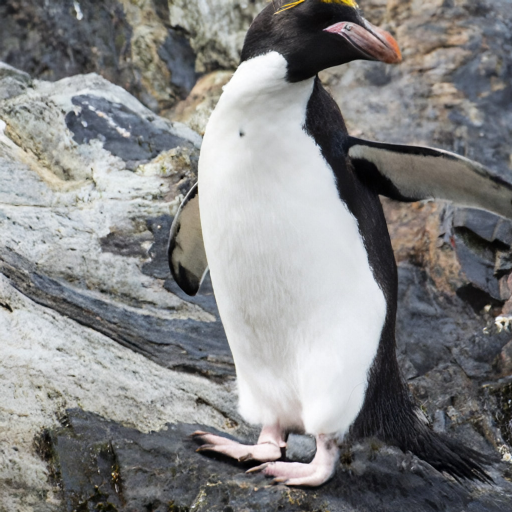}{0.47}{0.73} & \BAMPaperZoom{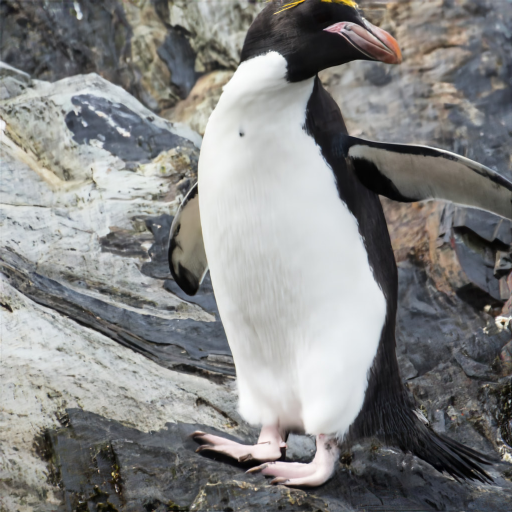}{0.47}{0.73} & \BAMPaperZoom{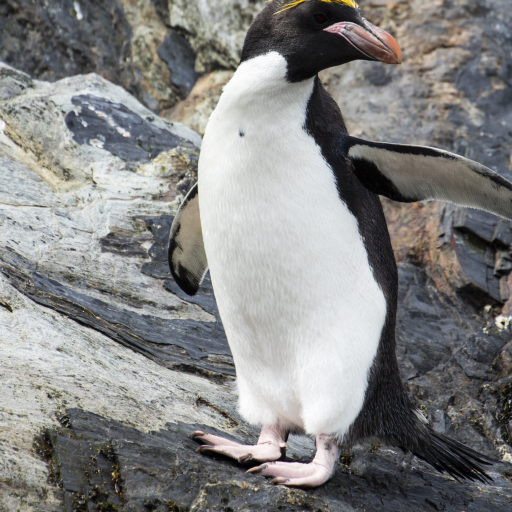}{0.47}{0.73} \\[3pt]
\multicolumn{4}{@{}l}{\textbf{Blind spatially varying motion blur --- DIV2K}} \\[1pt]
\BAMPaperZoom{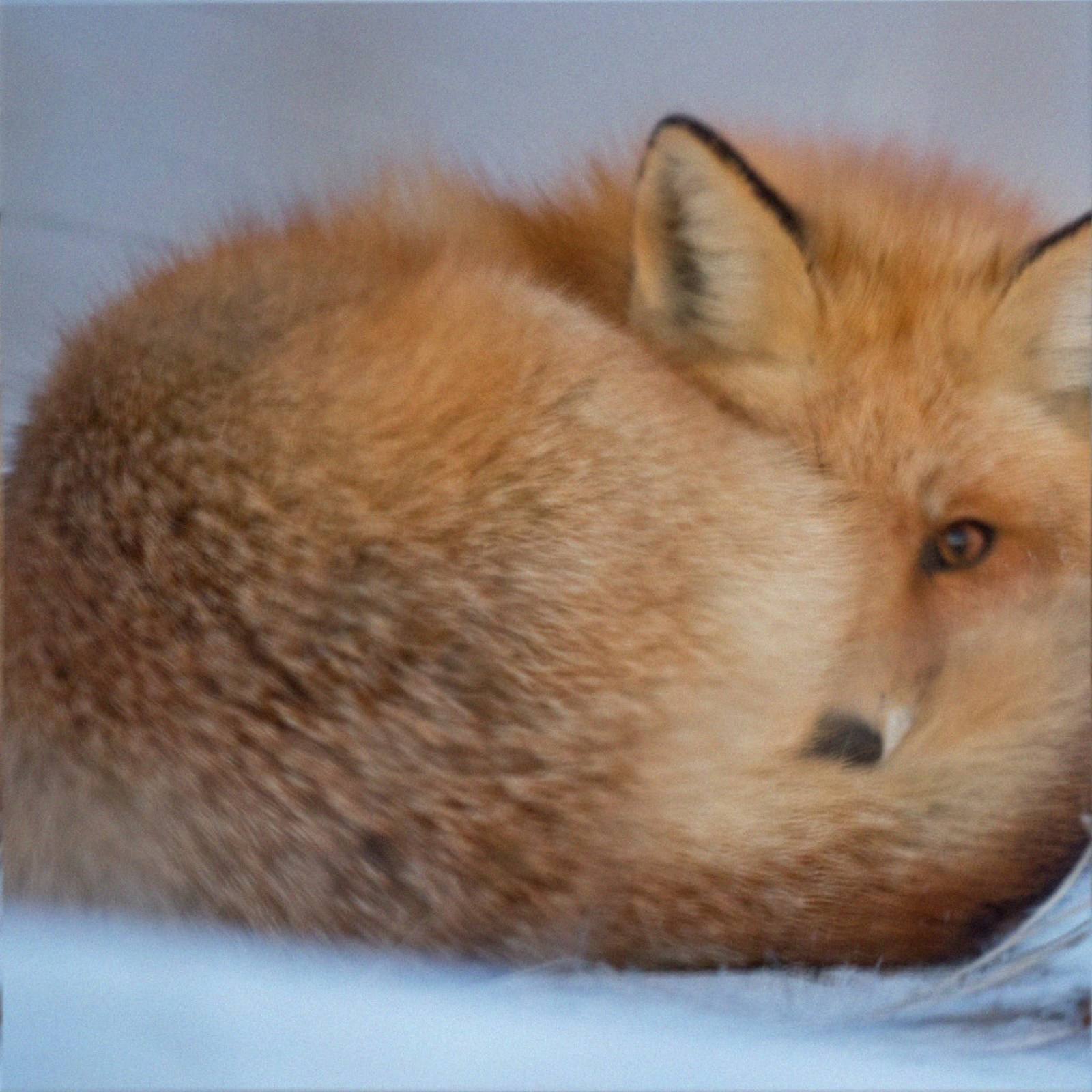}{0.73}{0.36} & \BAMPaperZoom{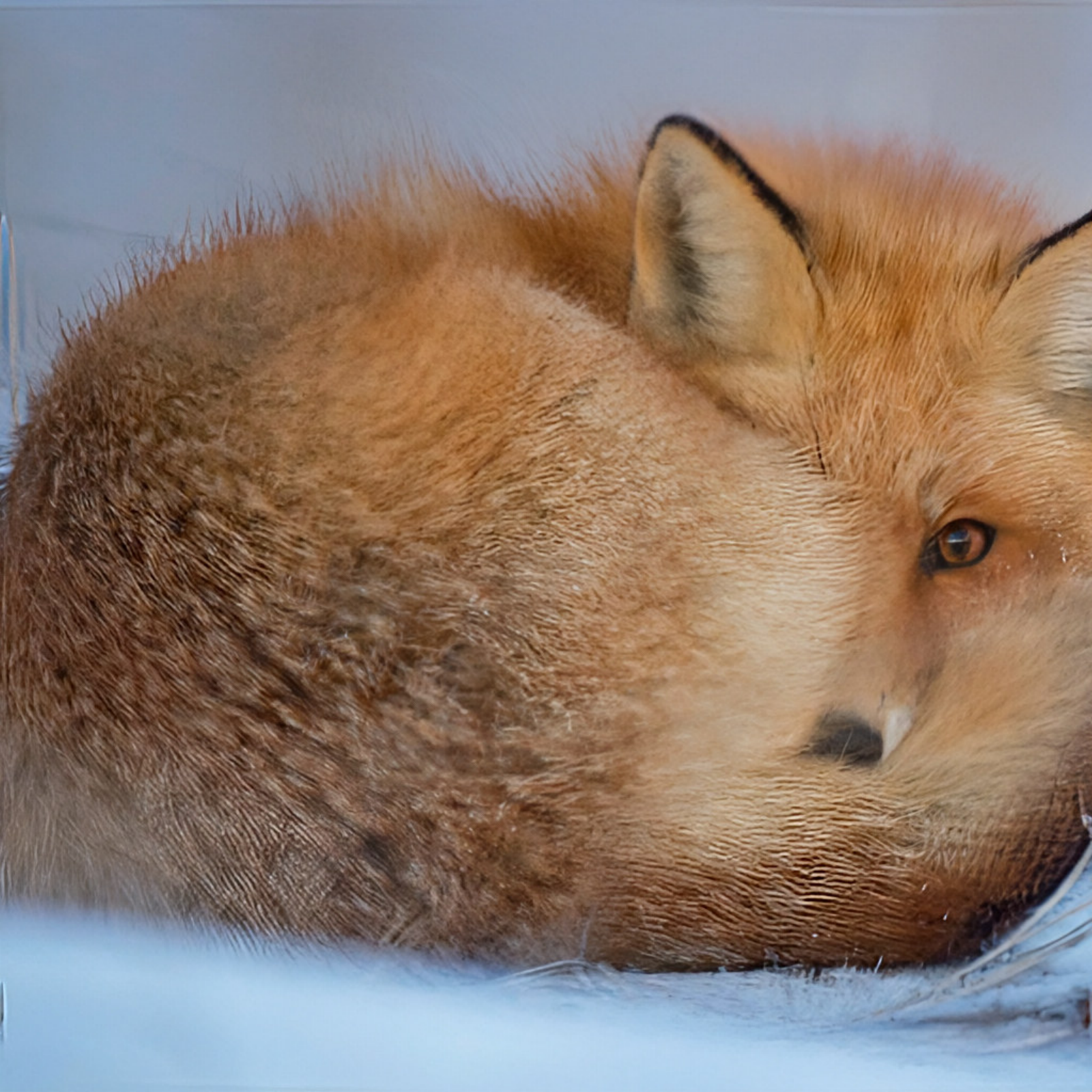}{0.73}{0.36} & \BAMPaperZoom{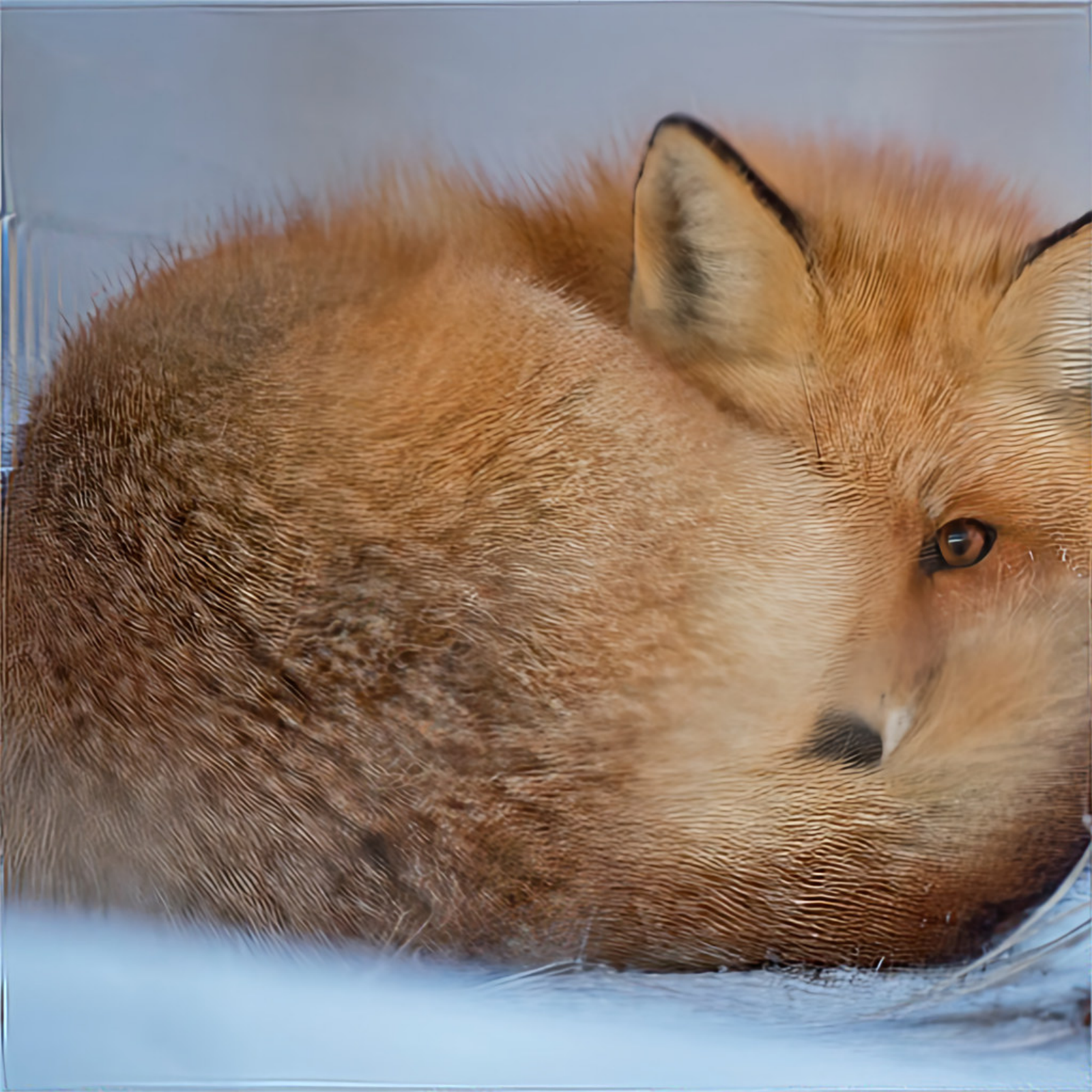}{0.73}{0.36} & \BAMPaperZoom{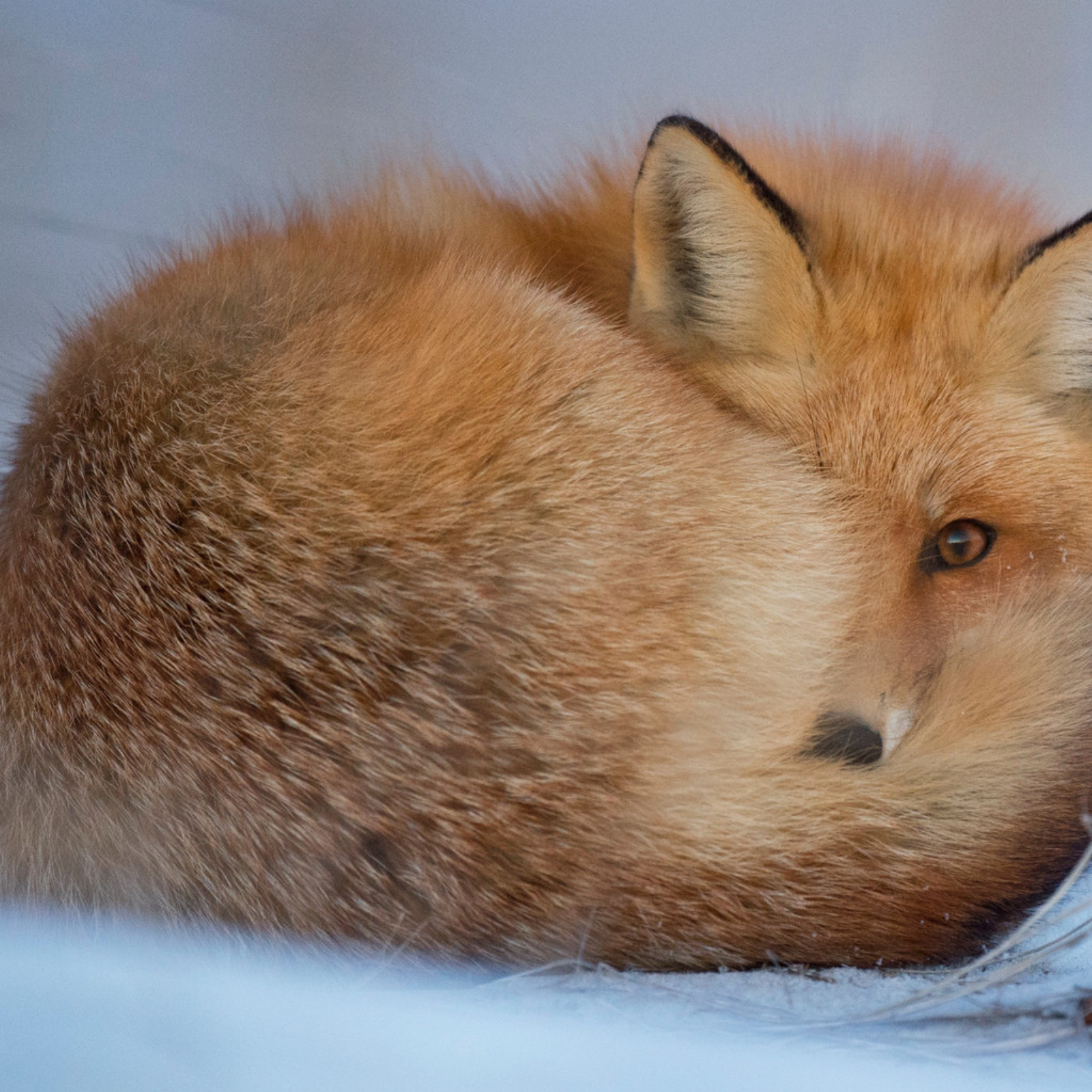}{0.73}{0.36} \\[3pt]
\multicolumn{4}{@{}l}{\textbf{Blind real motion blur --- K\"ohler}} \\[1pt]
\BAMPaperZoom{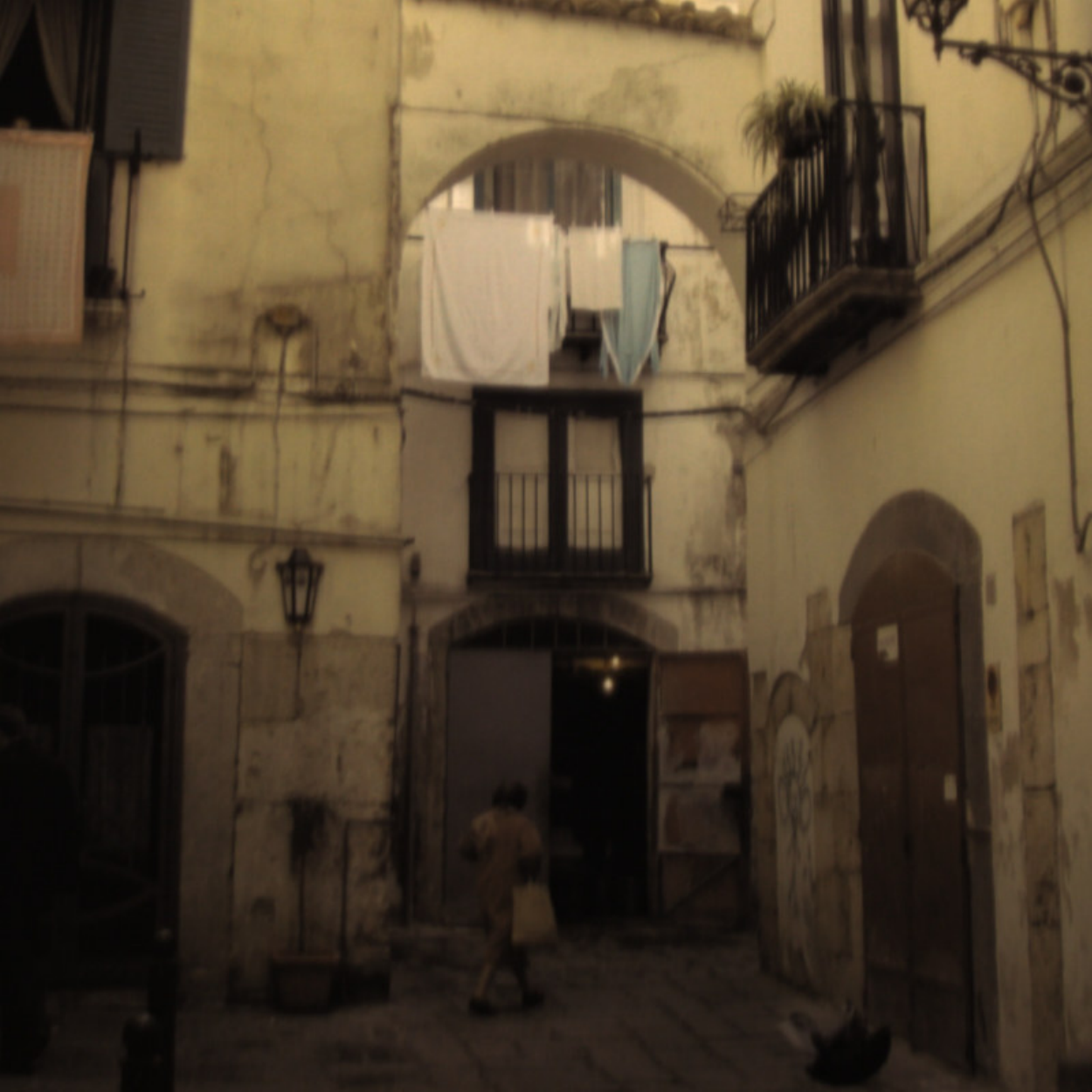}{0.39}{0.54} & \BAMPaperZoom{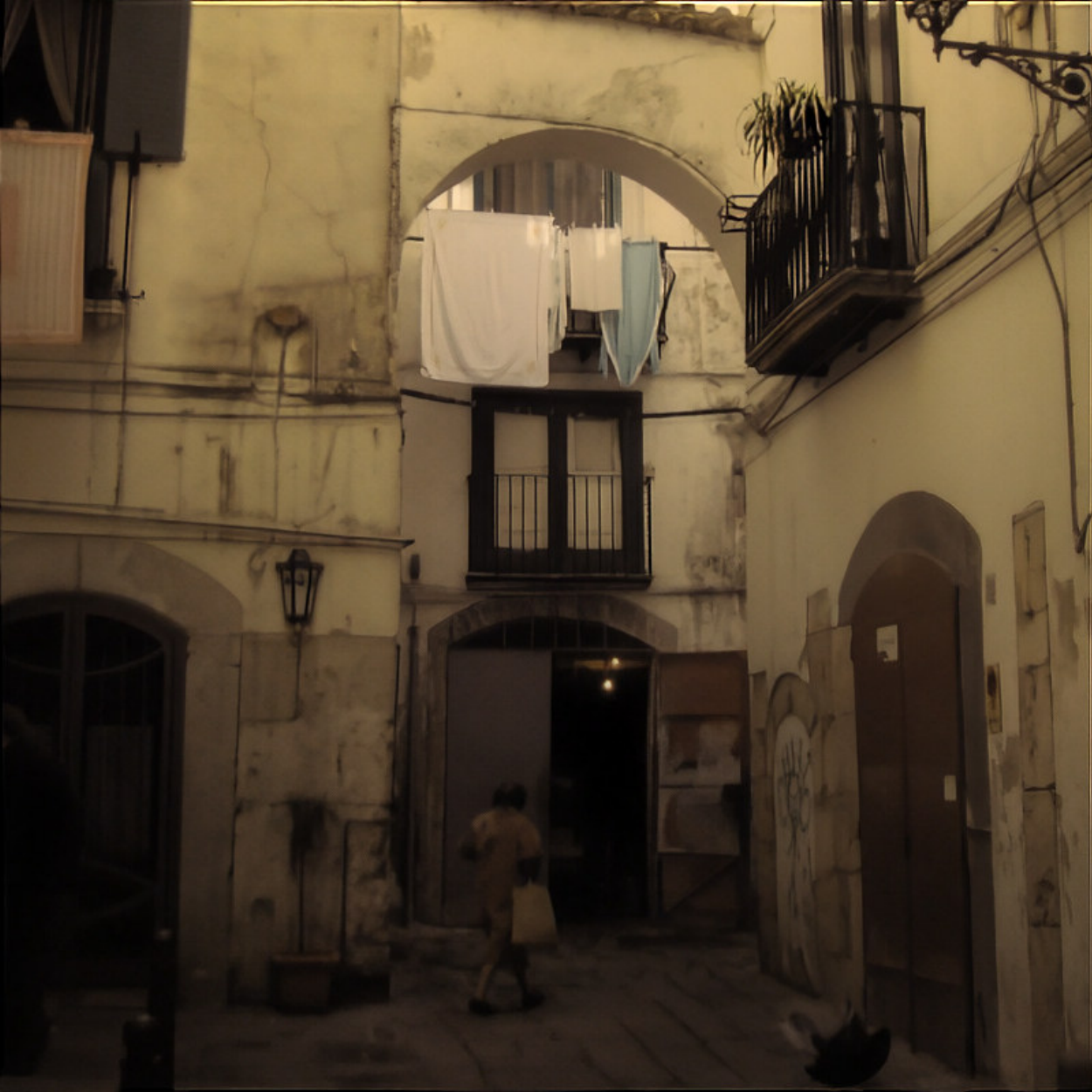}{0.39}{0.54} & \BAMPaperZoom{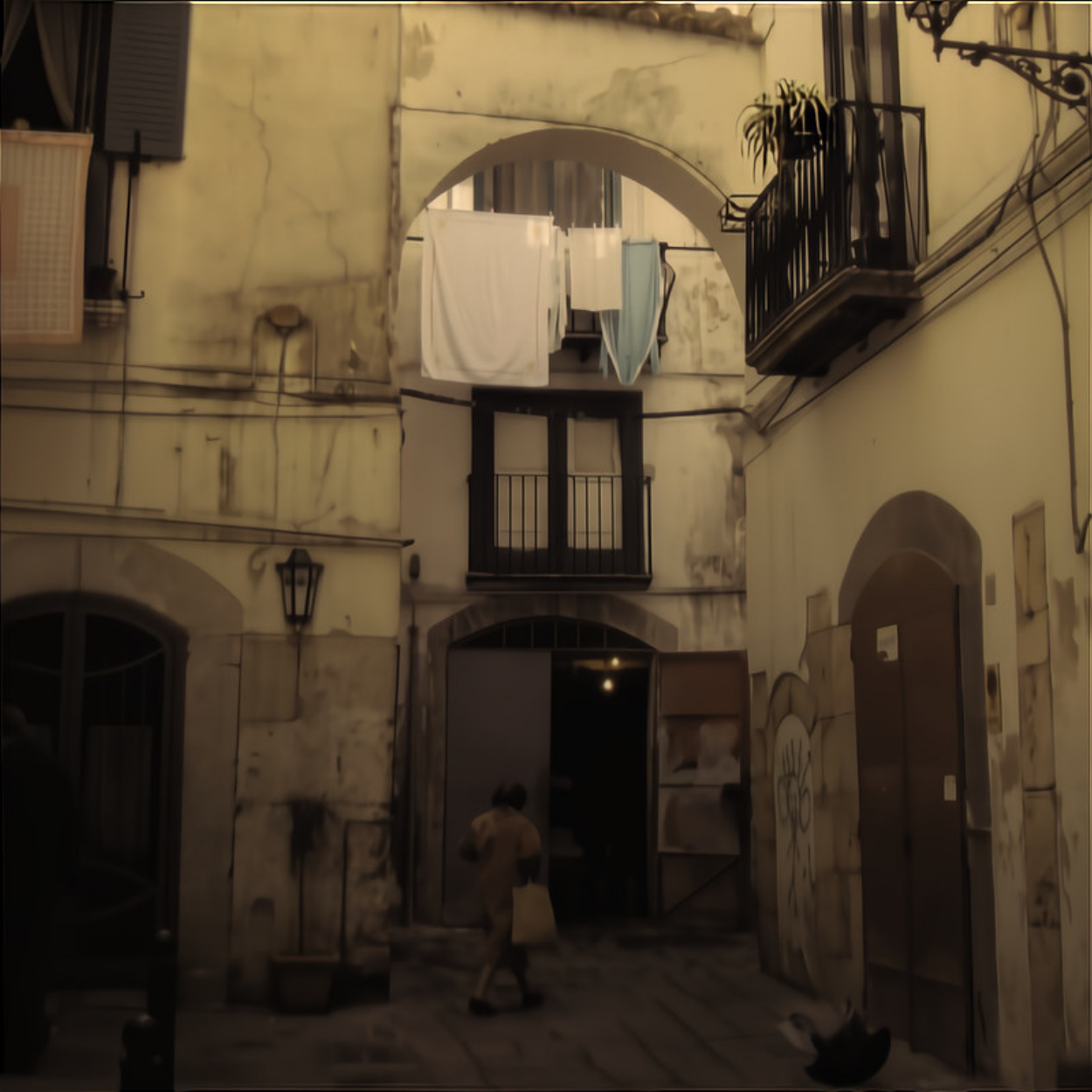}{0.39}{0.54} & \BAMPaperZoom{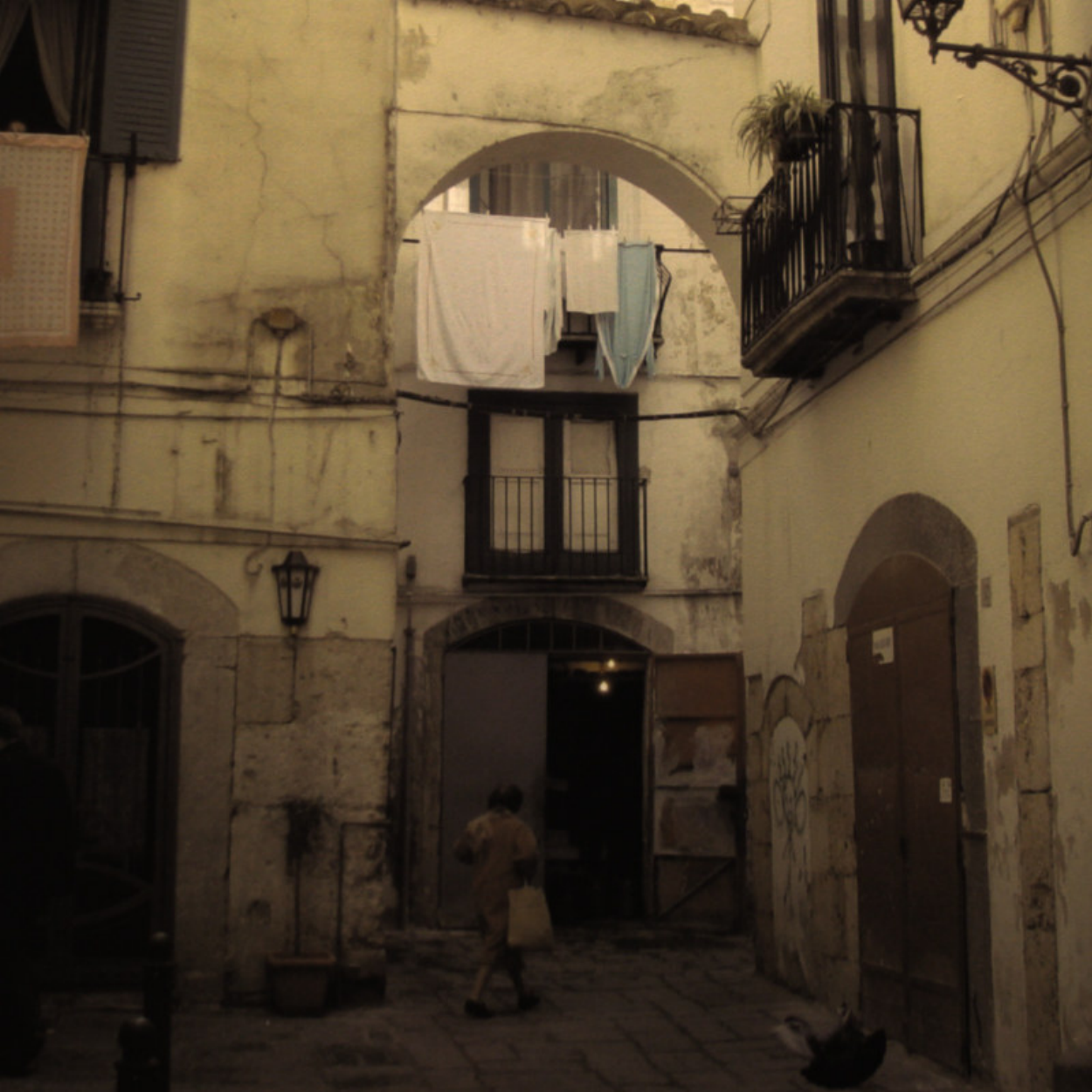}{0.39}{0.54} \\[3pt]
\bottomrule
\end{tabular}
\endgroup
\caption{\small\textbf{Motion deblurring in three settings.} Matched yellow boxes show regions magnified $4\times$.}
\label{fig:motion-three-settings}
\end{figure}

\clearpage

\subsection{JPEG restoration}\label{app:jpeg}
\paragraph{Observation model.}
We study FFHQ restoration after noisy JPEG compression at quality factor $Q=10$.
For images represented in $[0,1]$, the observation is
\begin{equation}
  \vy=\mathcal J_{10}(\vx+\epsilon) + \sigma_y \vn,\qquad
  \epsilon\sim\mathcal N(0,\sigma_\text{JPEG}^2\Id),
  \label{eq:jpeg-observation}
\end{equation}
where $\mathcal J_{10}$ denotes compression followed by decoding.
The Gaussian noise is added \emph{before} quantization; its standard deviation is $\sigma_\text{JPEG} = 0.01$.
The codec converts RGB to YCbCr, retains luminance at full resolution, point-samples chroma at 4:2:0 resolution, and applies an orthonormal $8\times8$ block DCT.
Coefficients are divided by the quality-scaled standard luminance/chrominance tables, rounded, dequantized, and inverse-transformed; chroma is reconstructed by nearest-neighbor upsampling before conversion back to RGB.
No additional pixel clipping is applied inside the codec.

\paragraph{Operator approximation.}
The discrete JPEG map is nonlinear because of coefficient rounding.
For operator conditioning and the linear solves in the reconstruction network, we replace rounding by the identity.
The DCT and quantization scalings then cancel, leaving the linear color/chroma operator
\begin{equation}
  \widetilde{\mathcal A}=C_{\mathrm{YCC}\to\mathrm{RGB}}
  \operatorname{diag}(I,UD,UD)C_{\mathrm{RGB}\to\mathrm{YCC}},
  \label{eq:jpeg-linear}
\end{equation}
where $D$ selects one chroma pixel per $2\times2$ block and $U$ repeats it over that block.
Affine offsets are excluded from this linear operator.
Its true adjoint transposes the color matrices, sums the four repeated chroma contributions, and inserts zeros at unsampled locations; it is not an inverse JPEG decoder.
Thus the observation retains quantization artifacts while the network is conditioned on a rounding-relaxed approximation.
The saved sampler setting $\sigma_y=0.05$ is distinct from the codec's pre-compression noise level $0.01$.

\begin{wraptable}{r}{0.48\textwidth}
\vspace{-3em}
\centering
\label{tab:jpeg-ffhq}
\begingroup
\fontsize{7}{8}\selectfont
\setlength{\tabcolsep}{3pt}
\renewcommand{\arraystretch}{1.0}
\begin{tabular}{@{}lrrrr@{}}
\toprule
Method & PSNR $\uparrow$ & LPIPS $\downarrow$ & CMMD $\downarrow$ & FID $\downarrow$ \\
\midrule
\rowcolor{BAMRow} BAM & 16.14 & 0.746 & 2.11 & 312.88 \\
\rowcolor{BAMRow} \BAMFT & \textbf{30.95} & \textbf{0.236} & \textbf{0.12} & \textbf{43.10} \\
RAM & 8.96 & 0.826 & 2.41 & 377.62 \\
\RAMFT & \underline{30.44} & 0.424 & 0.63 & 103.79 \\
SILO & 23.54 & 0.446 & 0.83 & 110.87 \\
UD2M & \textbf{30.95} & \underline{0.283} & \underline{0.42} & \underline{55.71} \\
\bottomrule
\end{tabular}
\caption{\textbf{FFHQ JPEG restoration.} Results over 64 images at JPEG quality factor $Q=10$. FT denotes finetuning.}
\endgroup
\vspace{-3em}
\end{wraptable}

\begin{figure}[h]
\centering
\begingroup
\scriptsize
\setlength{\BAMPaperImageWidth}{.120\linewidth}
\setlength{\tabcolsep}{1pt}
\begin{tabular}{@{}cccccccc@{}}
\toprule
\textbf{Observation} & \textbf{BAM} & \textbf{\BAMFT} & \textbf{RAM} & \textbf{\RAMFT} & \textbf{SILO} & \textbf{UD2M} & \textbf{GT} \\
\midrule
\multicolumn{8}{@{}l}{\textbf{FFHQ JPEG, $Q=10$}} \\[1pt]
\BAMPaperZoom{Figures/FFHQ/jpeg/quality_10/BAM_finetuned/observation_00010.pdf}{0.39}{0.46} & \BAMPaperZoom{Figures/FFHQ/jpeg/quality_10/BAM/restored_00010.pdf}{0.39}{0.46} & \BAMPaperZoom{Figures/FFHQ/jpeg/quality_10/BAM_finetuned/restored_00010.pdf}{0.39}{0.46} & \BAMPaperZoom{Figures/FFHQ/jpeg/quality_10/RAM/restored_00010.pdf}{0.39}{0.46} & \BAMPaperZoom{Figures/FFHQ/jpeg/quality_10/RAM_finetuned/restored_00010.pdf}{0.39}{0.46} & \BAMPaperZoom{Figures/FFHQ/jpeg/quality_10/SILO/restored_00010.pdf}{0.39}{0.46} & \BAMPaperZoom{Figures/FFHQ/jpeg/quality_10/UD2M/restored_00010.pdf}{0.39}{0.46} & \BAMPaperZoom{Figures/FFHQ/jpeg/quality_10/BAM_finetuned/gt_00010.pdf}{0.39}{0.46} \\[3pt]
\multicolumn{8}{@{}l}{\textbf{FFHQ JPEG, $Q=10$}} \\[1pt]
\BAMPaperZoom{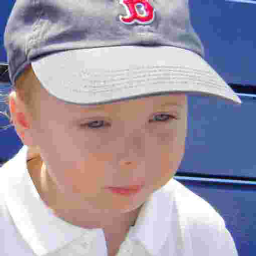}{0.3}{0.47} & \BAMPaperZoom{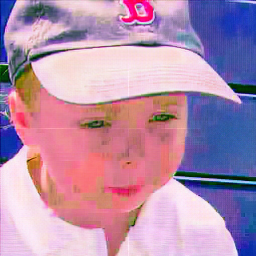}{0.3}{0.47} & \BAMPaperZoom{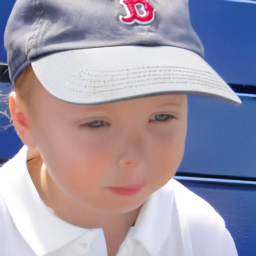}{0.3}{0.47} & \BAMPaperZoom{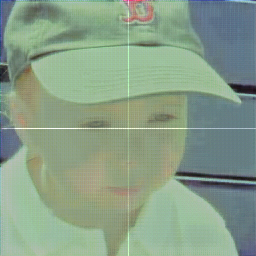}{0.3}{0.47} & \BAMPaperZoom{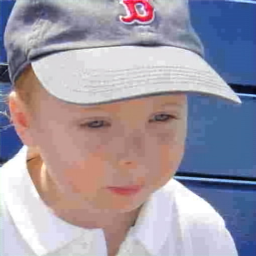}{0.3}{0.47} & \BAMPaperZoom{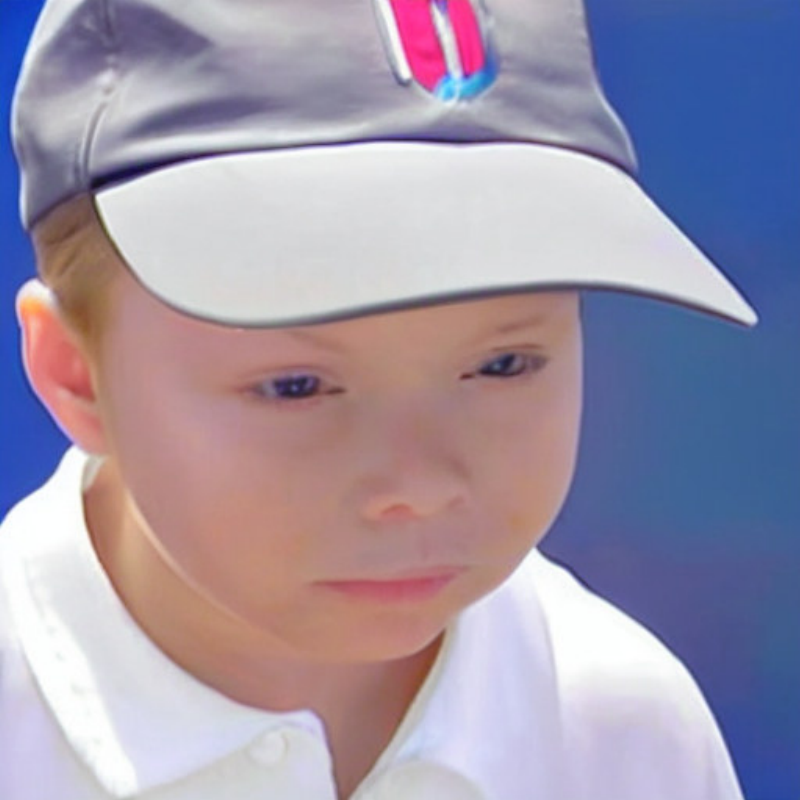}{0.3}{0.47} & \BAMPaperZoom{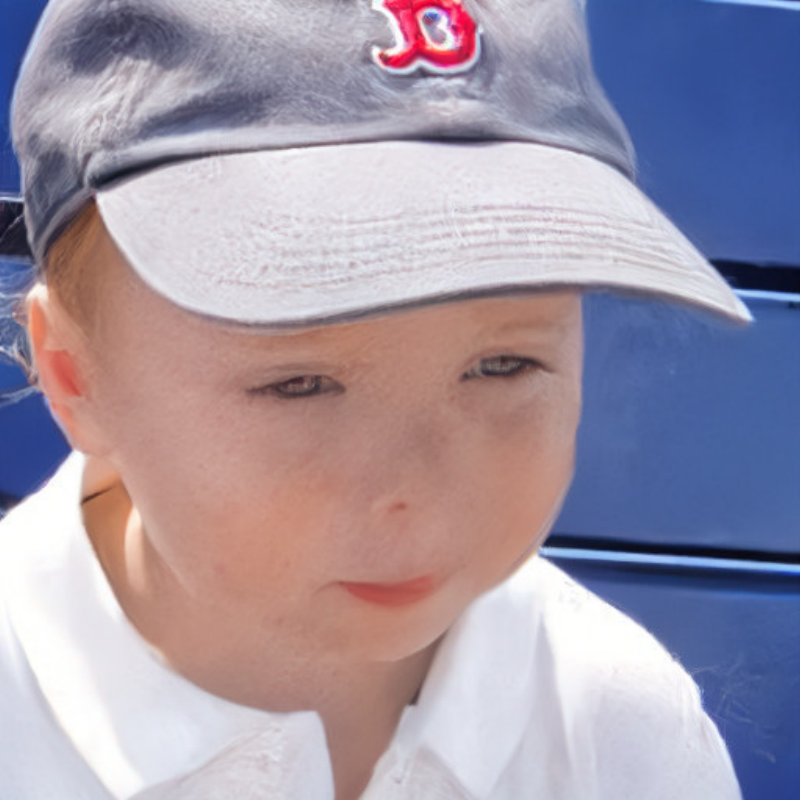}{0.3}{0.47} & \BAMPaperZoom{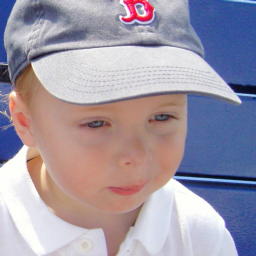}{0.3}{0.47} \\[3pt]
\bottomrule
\end{tabular}
\endgroup
\caption{\textbf{FFHQ JPEG restoration.} Quality factor 10, with Gaussian noise of standard deviation $0.01$ in $[0,1]$ added before compression. All columns refer to the same clean image. Matched yellow boxes show regions magnified $4\times$.}
\label{fig:jpeg-ffhq}
\end{figure}

\paragraph{Qualitative comparison.}
Figure~\ref{fig:jpeg-ffhq} compares BAM, BAM finetuned with the contrast penalty, RAM, RAM finetuned on all problems, SILO, and UD2M. Both BAM variants use three steps.

\paragraph{Quantitative comparison.}
Table~\ref{tab:jpeg-ffhq} reports the quantitative restoration results
for JPEG compression at quality factor $Q=10$.
\vspace{1em}

\subsection{Single problem}\label{app:single-problem}
We examine repeated posterior draws for a fixed FFHQ inverse problem.
Table~\ref{tab:single-ffhq-psnr} distinguishes the PSNR of the average of 64 draws from the average PSNR of individual draws, using all 64 evaluated images for each problem.
Here the noise level is $\sigma_y=0.025$.
The posterior mean has higher PSNR than the average individual draw for every problem and exceeds the problem-specific \RAMFT result on four of the five problems.
\vspace{-1em}
\begin{table}[!htbp]
\centering
\caption{\textbf{Single-problem FFHQ results.} PSNR (dB; $\uparrow$) over 64 images and 64 draws per image at $\sigma_y = 0.025$.
PSNR(mean) evaluates the mean of the 64 draws; mean PSNR averages the PSNR of those draws. Bold values mark the best in each row.}
\vspace{0.5em}
\label{tab:single-ffhq-psnr}
\small
\setlength{\tabcolsep}{5pt}
\begin{tabularx}{\linewidth}{@{}l>{\raggedleft\arraybackslash}X>{\raggedleft\arraybackslash}X>{\raggedleft\arraybackslash}X@{}}
\toprule
Problem & BAM PSNR(mean) & BAM mean PSNR & \RAMFT \\
\midrule
Deblurring & \textbf{32.49} & 30.61 & 32.31 \\
Super-resolution & 31.57 & 29.97 & \textbf{32.36} \\
Inpainting & \textbf{34.20} & 32.90 & 33.36 \\
Demosaicing & \textbf{38.10} & 36.44 & 37.36 \\
Compressed sensing & \textbf{36.18} & 34.58 & 33.74 \\
\bottomrule
\end{tabularx}
\end{table}
\vspace{-1em}
\begin{figure}[H]
\centering
\begingroup
\setlength{\BAMPaperImageWidth}{.137\linewidth}
\setlength{\tabcolsep}{1pt}
\scriptsize
\begin{tabular}{@{}ccccccc@{}}
\toprule
\textbf{Observation} & \textbf{Draw 1} & \textbf{Draw 2} & \textbf{Draw 3} & \textbf{Mean (64)} & \textbf{\RAMFT} & \textbf{GT} \\
\midrule
\multicolumn{7}{@{}l}{\textbf{Gaussian deblurring}}\\[1pt]
\BAMPaperZoom{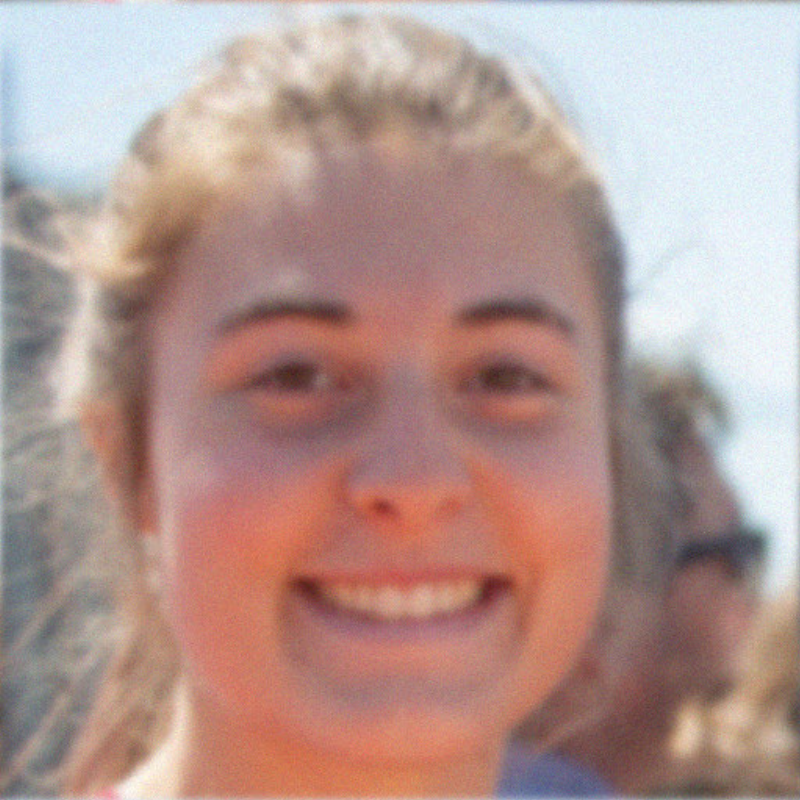}{0.24}{0.48} & \BAMPaperZoom{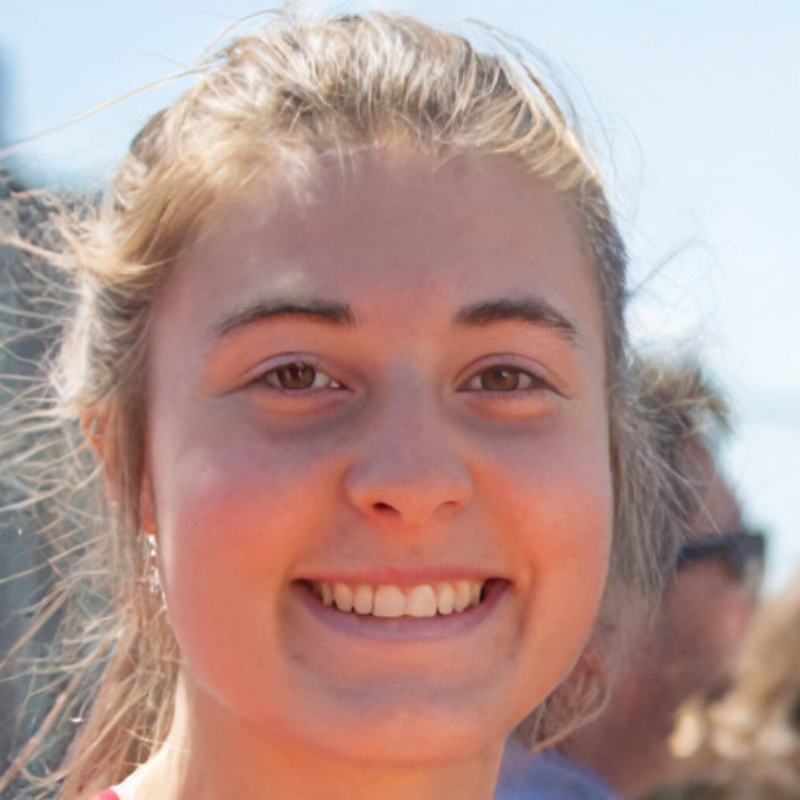}{0.24}{0.48} & \BAMPaperZoom{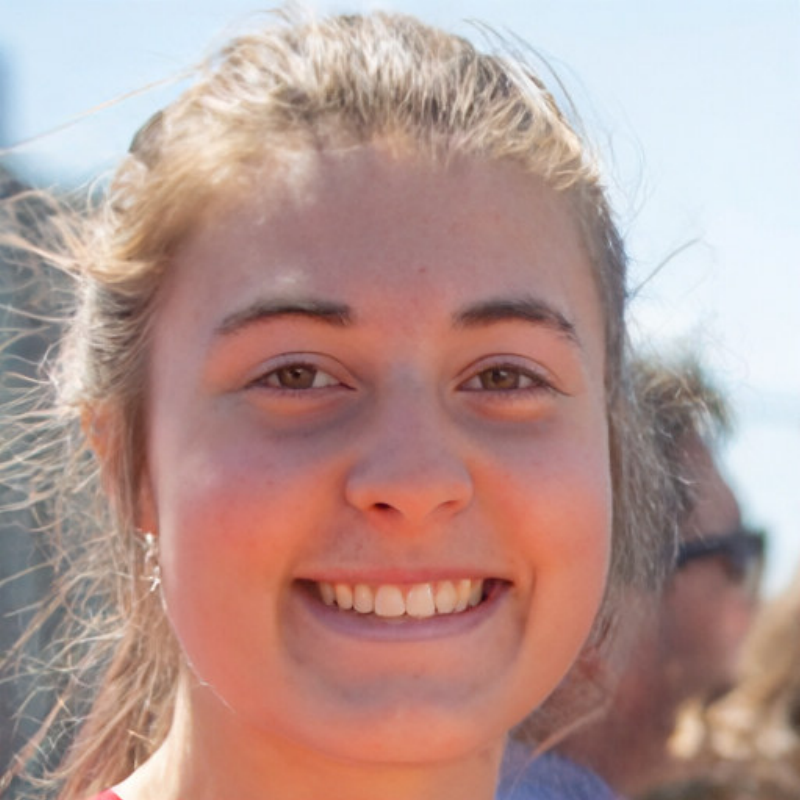}{0.24}{0.48} & \BAMPaperZoom{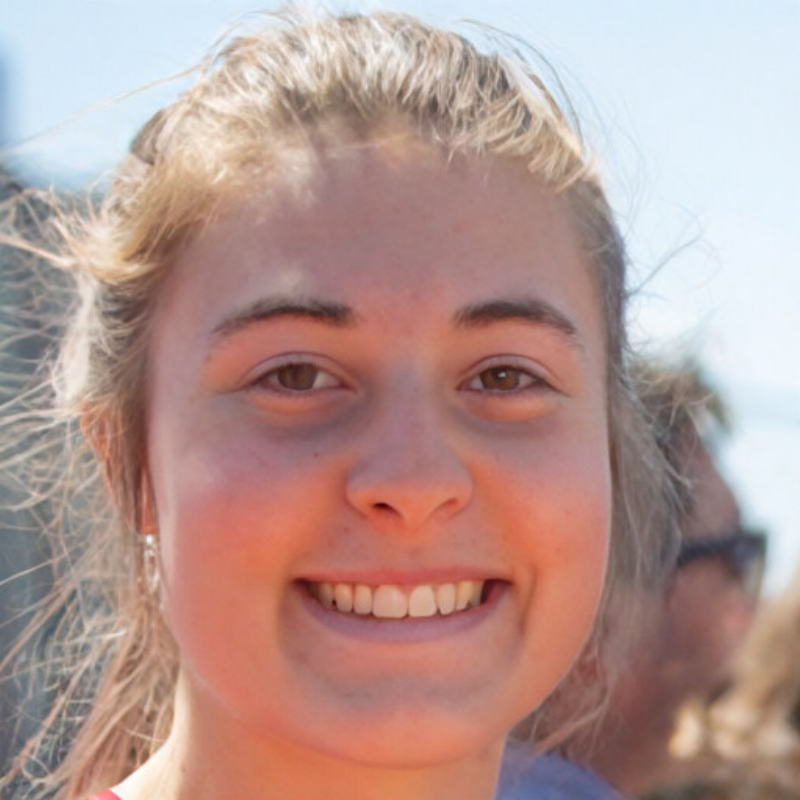}{0.24}{0.48} & \BAMPaperZoom{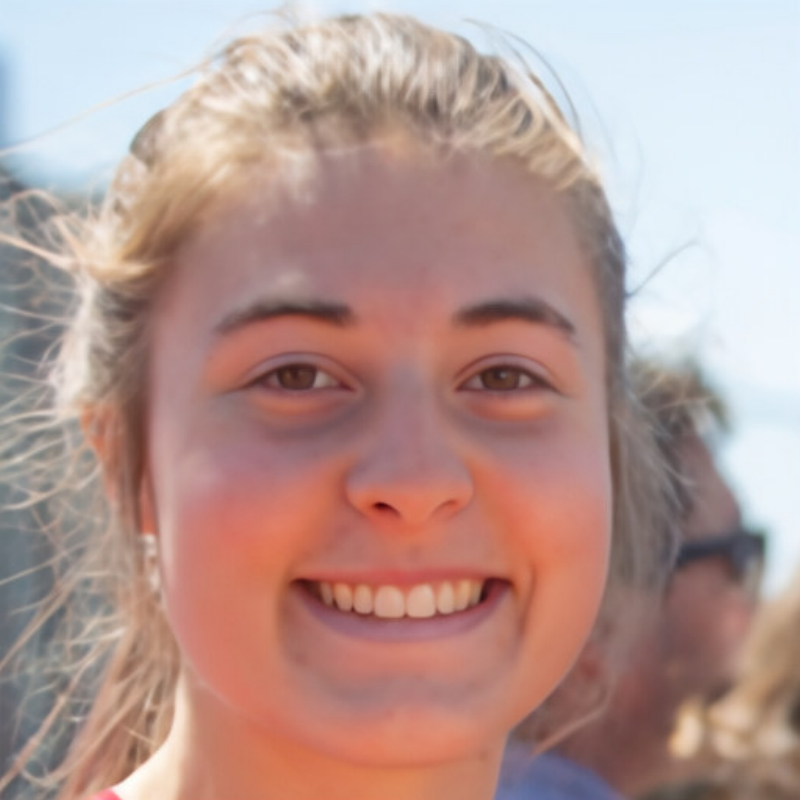}{0.24}{0.48} & \BAMPaperZoom{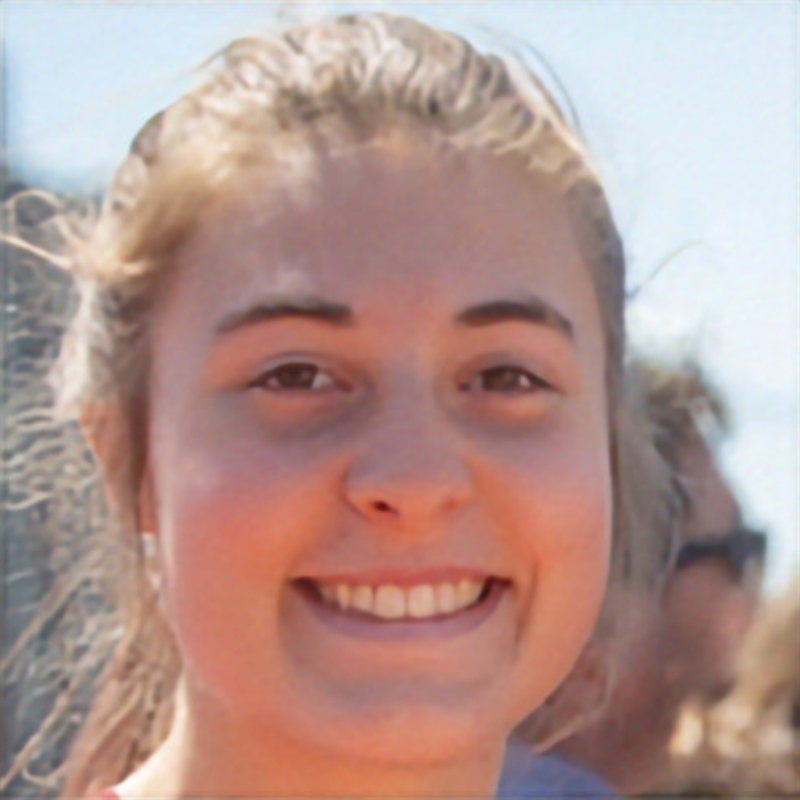}{0.24}{0.48} & \BAMPaperZoom{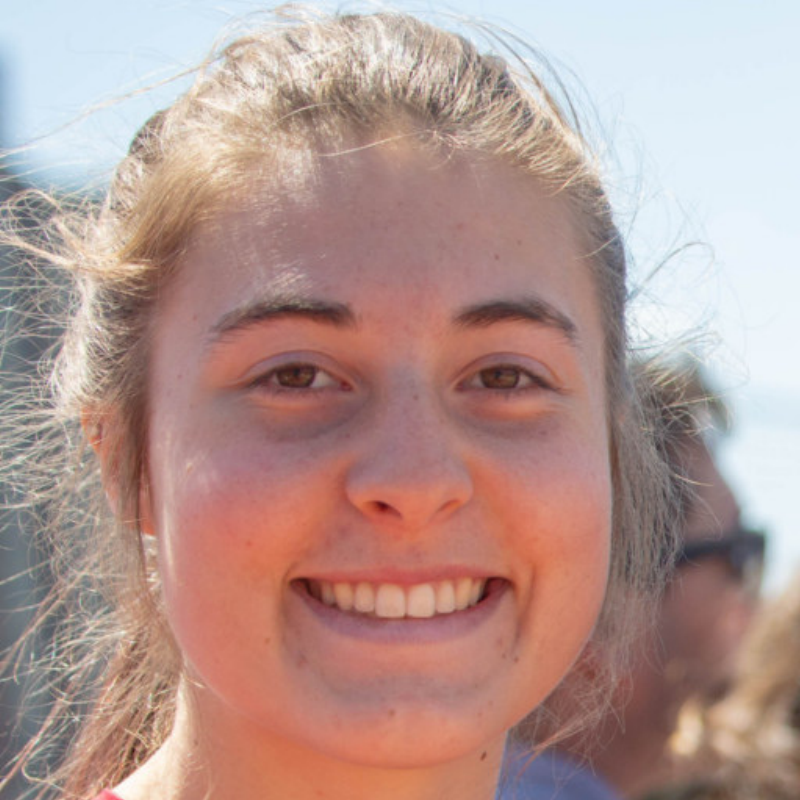}{0.24}{0.48} \\[3pt]
\multicolumn{7}{@{}l}{\textbf{SR $\times4$}}\\[1pt]
\BAMPaperZoom{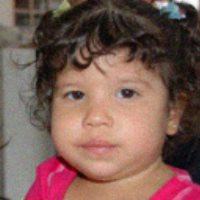}{0.28}{0.46} & \BAMPaperZoom{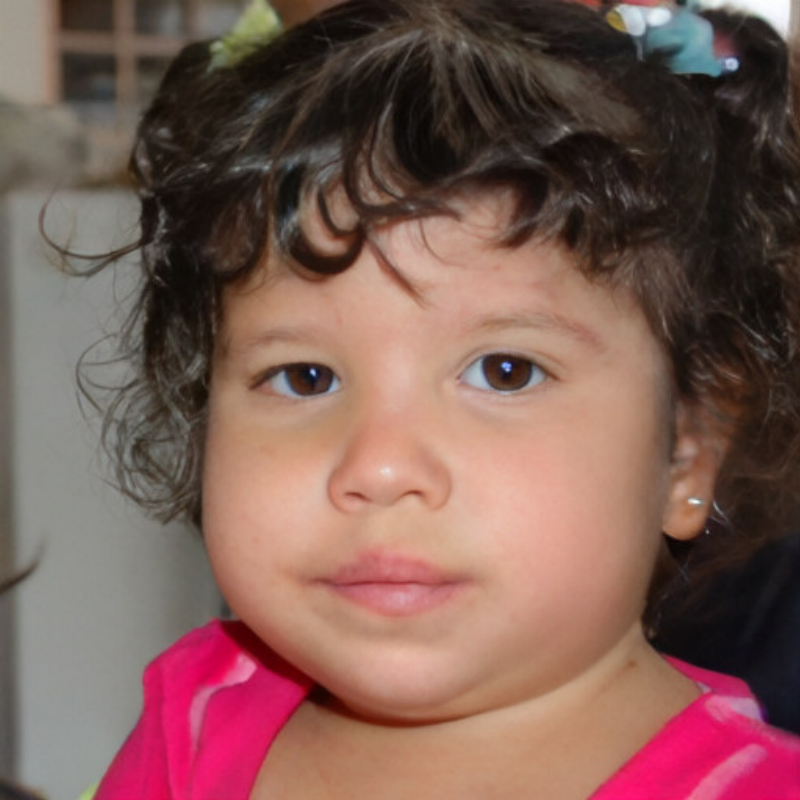}{0.28}{0.46} & \BAMPaperZoom{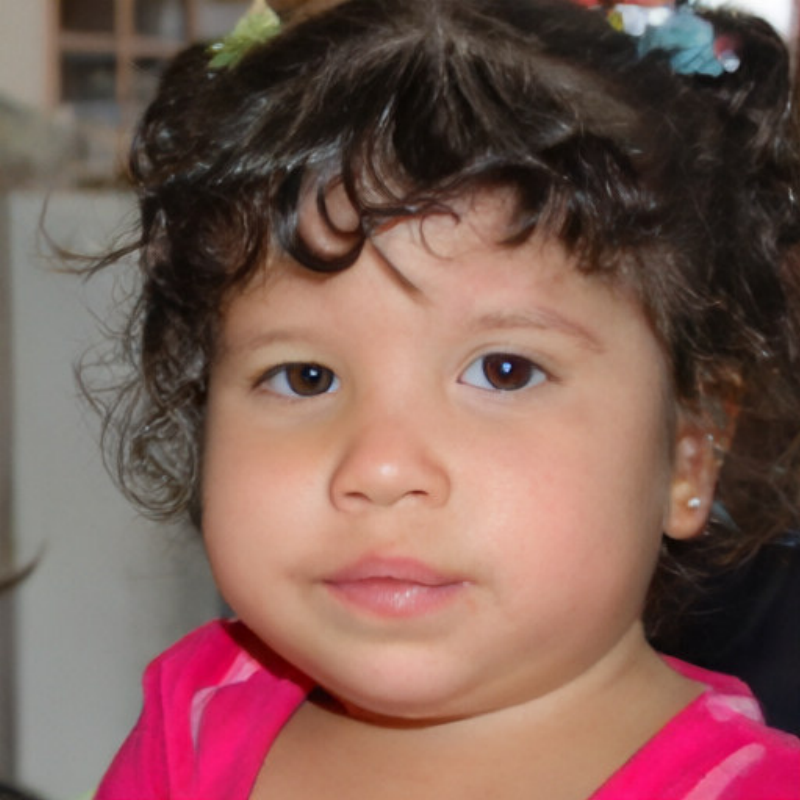}{0.28}{0.46} & \BAMPaperZoom{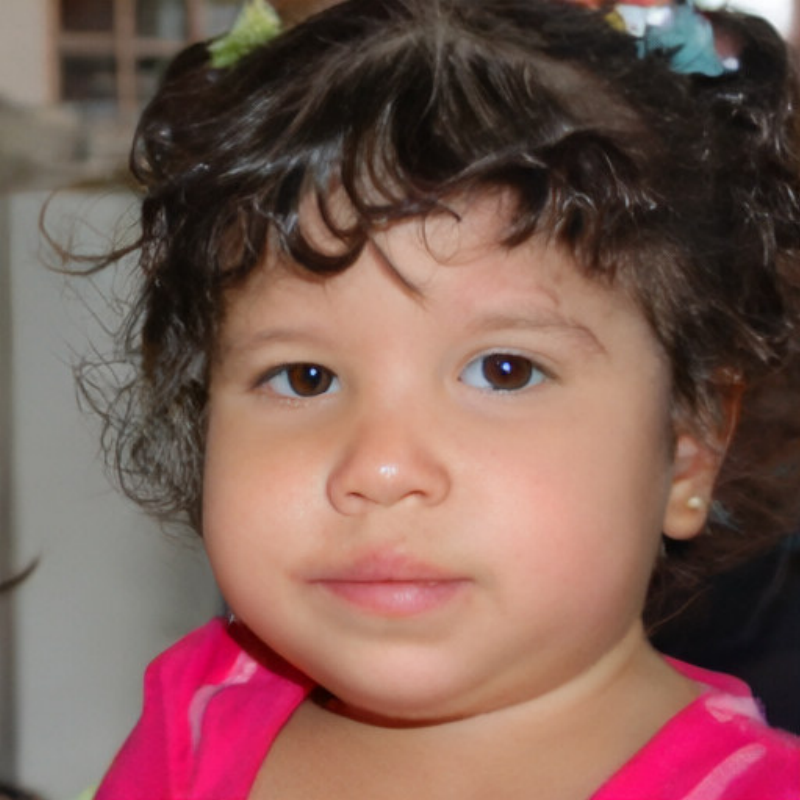}{0.28}{0.46} & \BAMPaperZoom{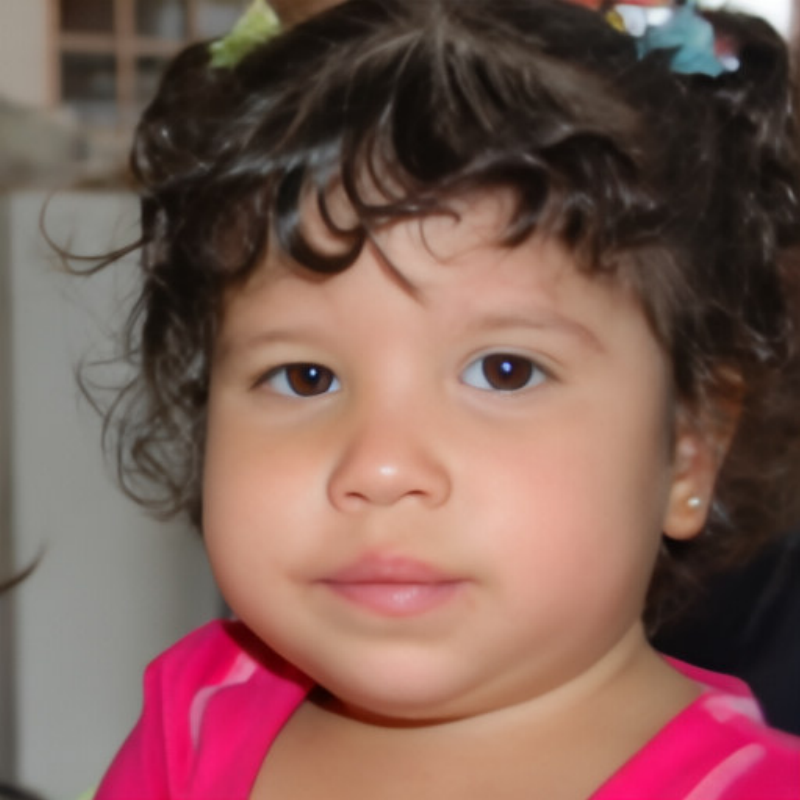}{0.28}{0.46} & \BAMPaperZoom{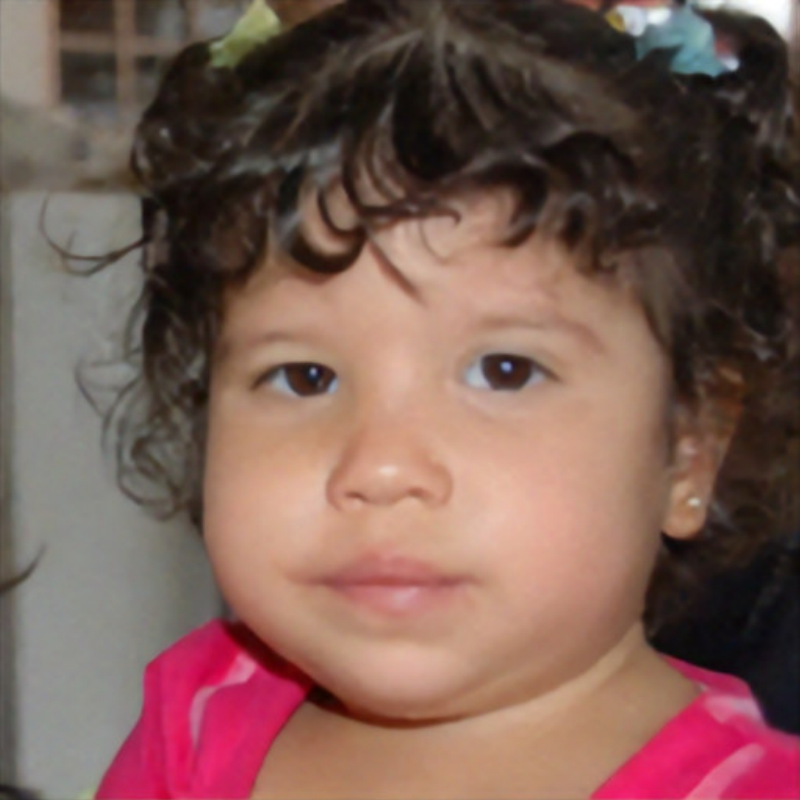}{0.28}{0.46} & \BAMPaperZoom{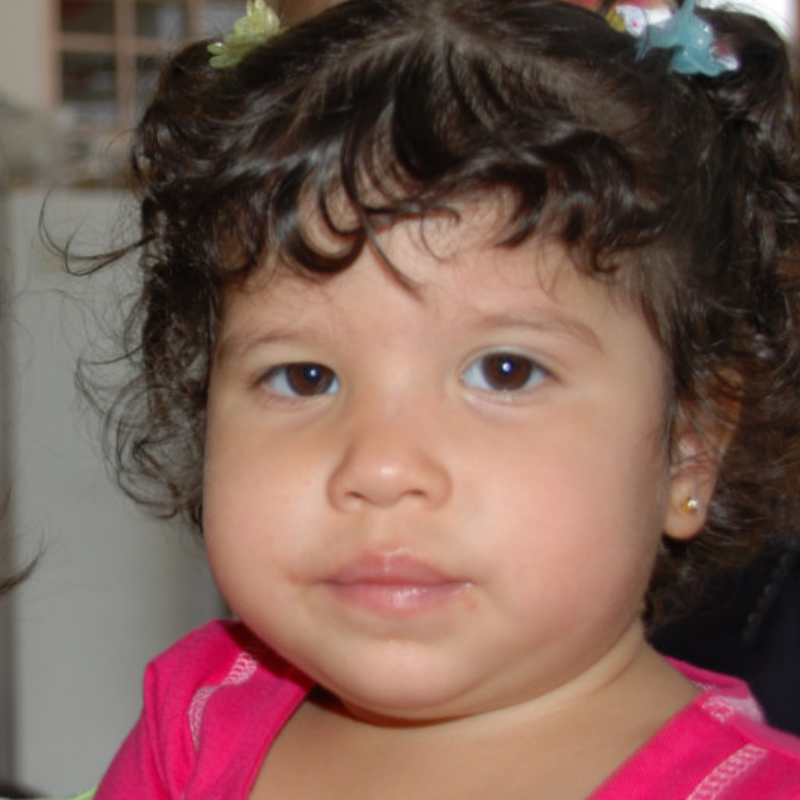}{0.28}{0.46} \\[3pt]
\multicolumn{7}{@{}l}{\textbf{Inpainting}}\\[1pt]
\BAMPaperZoom{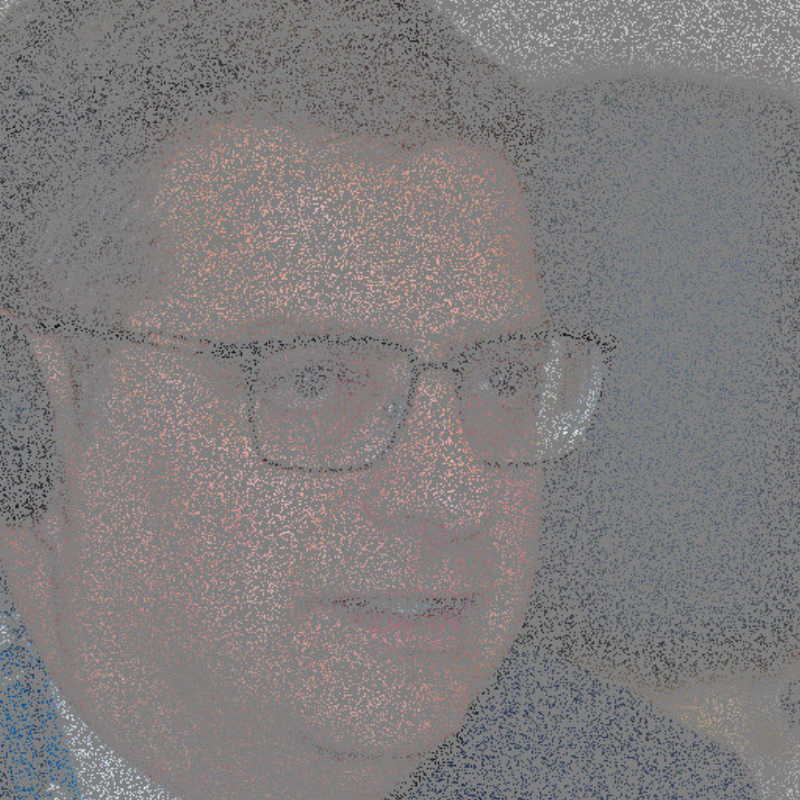}{0.34}{0.46} & \BAMPaperZoom{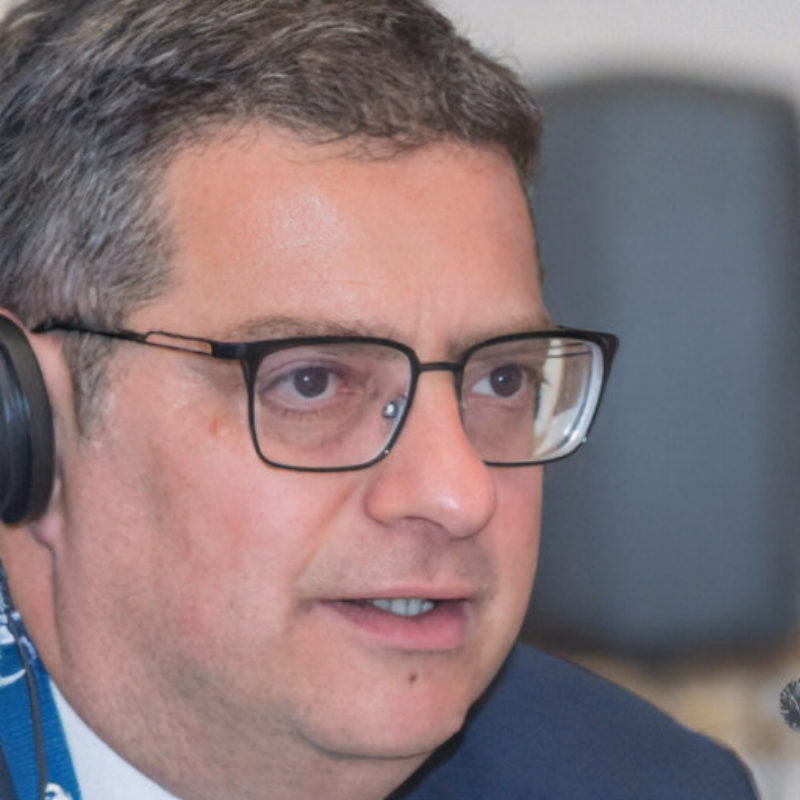}{0.34}{0.46} & \BAMPaperZoom{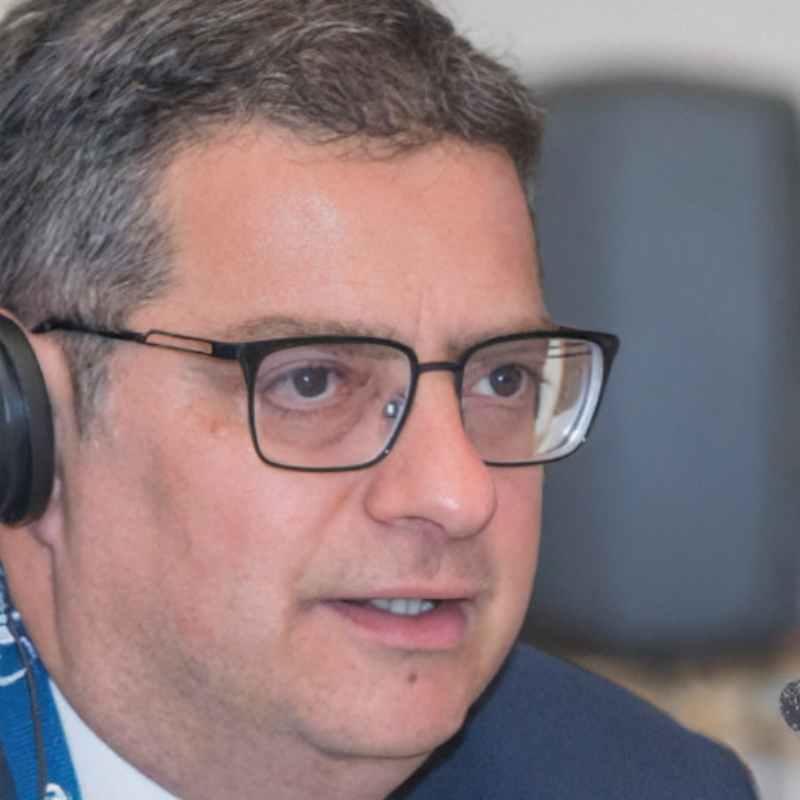}{0.34}{0.46} & \BAMPaperZoom{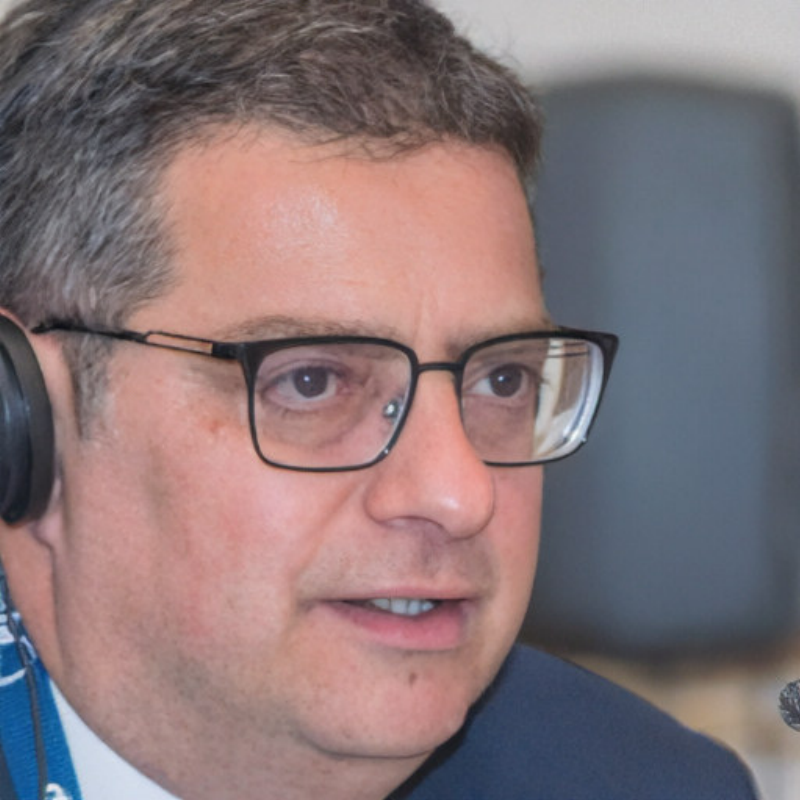}{0.34}{0.46} & \BAMPaperZoom{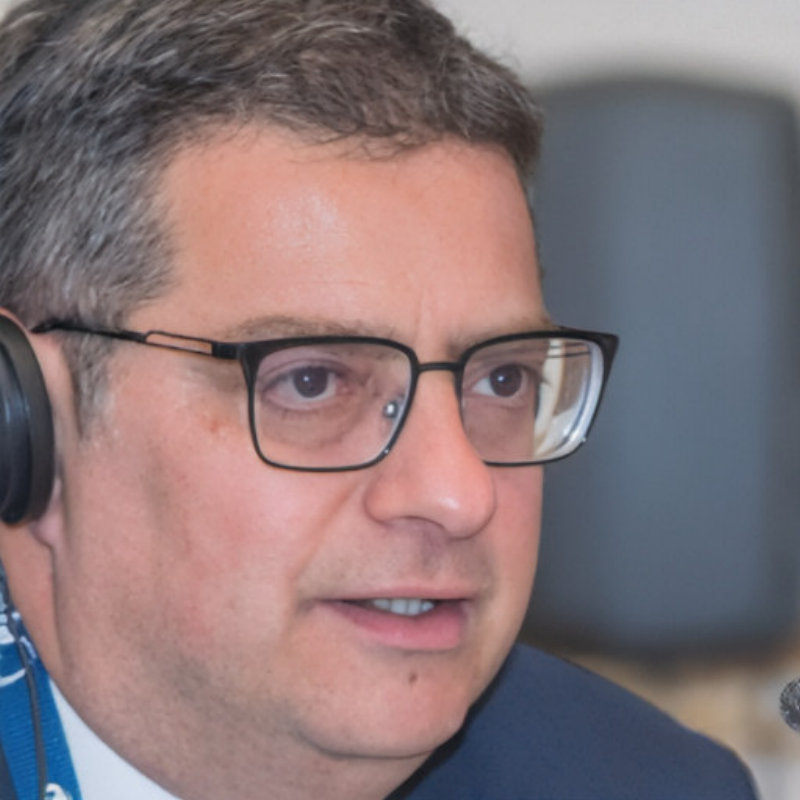}{0.34}{0.46} & \BAMPaperZoom{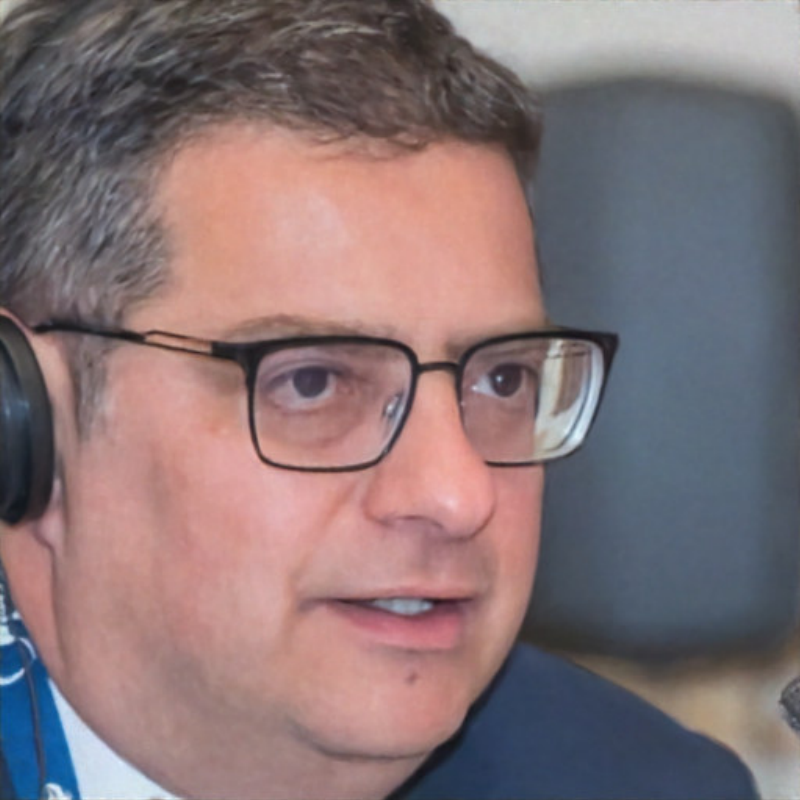}{0.34}{0.46} & \BAMPaperZoom{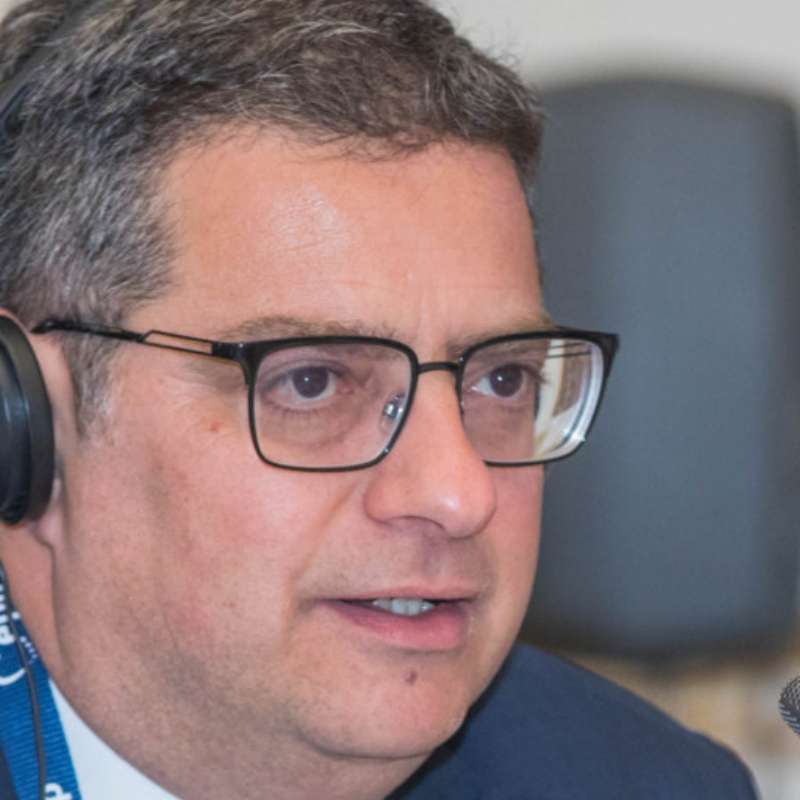}{0.34}{0.46} \\[3pt]
\bottomrule
\end{tabular}
\endgroup
\caption{\textbf{Posterior draws for a single problem.} One FFHQ image per problem at $\sigma_y=0.025$.
The three displayed draws illustrate diversity rather than typical random draws. The mean uses all 64 draws. Matched yellow boxes show regions magnified $4\times$.}
\label{fig:single-problem}
\end{figure}

\section{Robustness to Int8 quantization}\label{app:int8}
In this section, we evaluate the robustness and computational efficiency of BAM under INT8 quantization for TensorRT deployment. To achieve this, we convert the model from floating-point execution to INT8 where supported. In contrast, we retained higher-precision operations that are sensitive to reduced precision or are not efficiently supported in INT8. All experiments are performed on $512\times 512$ images using an NVIDIA A40 GPU.

\paragraph{Optimization process.}
The optimization combines INT8 quantization with a more efficient implementation of the physics-aware operations. Profiling revealed that operations within the \texttt{PhysicsBlock}, particularly convolution and transposed-convolution layers, accounted for a substantial fraction of the overall inference latency. These operations were therefore reformulated to reduce redundant computation and improve their execution under TensorRT.

\begin{figure}[H]
\centering
\begingroup
\small
\begin{tabular}{@{}cc@{}}
Baseline & INT8-Optimized \\
\includegraphics[width=.485\linewidth]{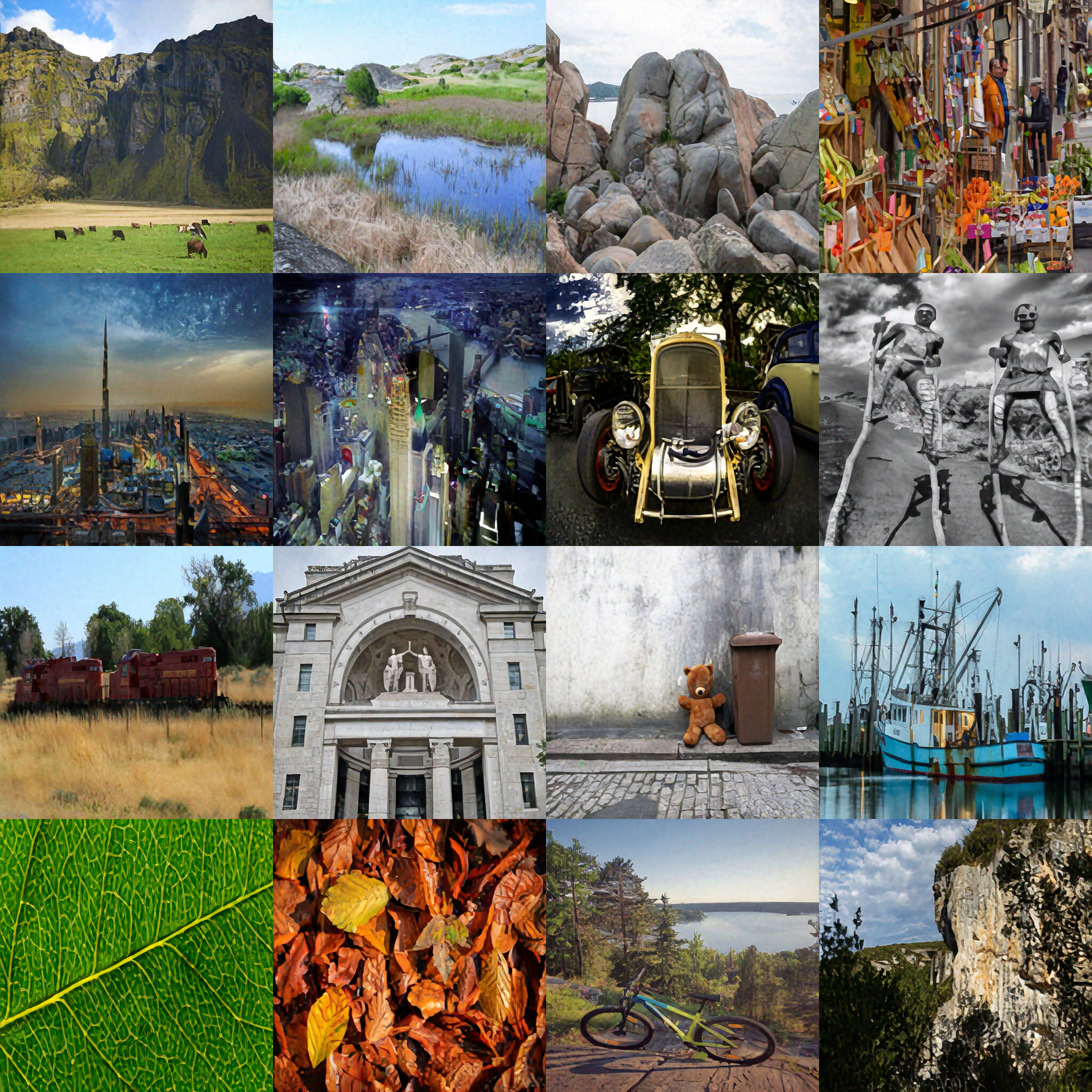} &
\includegraphics[width=.485\linewidth]{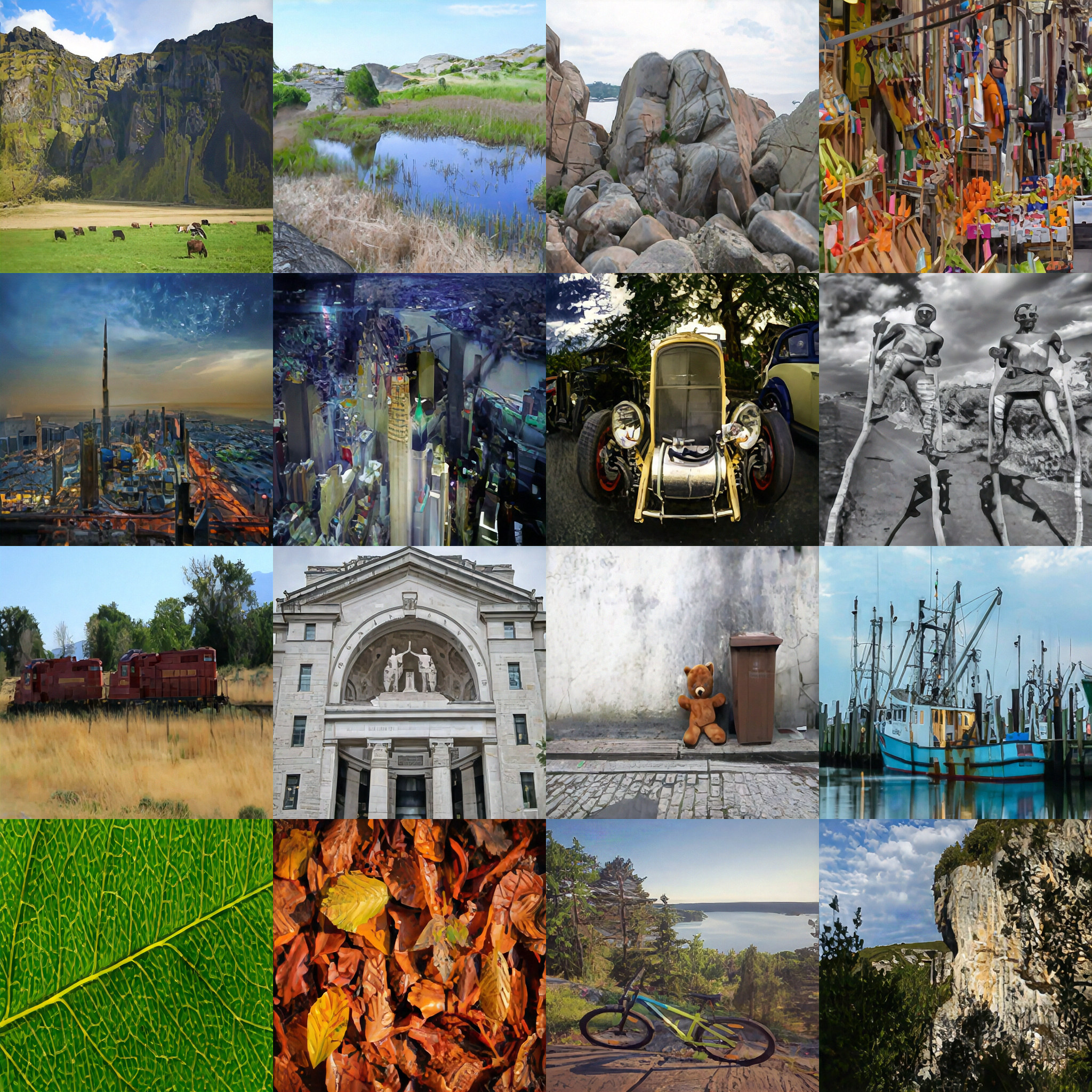}
\end{tabular}
\endgroup
\caption{\textbf{Qualitative comparison of the baseline and INT8-optimized models on Gaussian deblurring.}
Reconstruction results are shown for the original baseline (left) and the INT8-optimized model (right) on test images from the LSDIR dataset. The close visual agreement between both outputs indicates that the proposed INT8 optimization substantially improves inference efficiency while preserving the reconstruction quality of the baseline model.}
\label{fig:int8-comparison}
\end{figure}

\paragraph{End-to-end performance.}
In \Cref{tab:compression-int8}, we compare the end-to-end performance before and after quantization. The baseline model requires $280.56$ ms per inference, whereas the INT8-optimized model reduces the latency to $79.91$ ms. This corresponds to a $3.51\times$ end-to-end speedup and a $71.5\%$ reduction in inference latency. This reduction in latency is directly reflected in the inference throughput, which increases from $3.56$ to $12.51$ queries per second (qps). This represents an absolute gain of $8.95$ qps. The acceleration comes with a moderate degradation in reconstruction quality, with PSNR decreasing from $22.49$ to $21.20$ dB, SSIM from $0.570$ to $0.498$, and LPIPS increasing from $0.400$ to $0.427$. In \Cref{fig:int8-comparison}, we present qualitative results from both models, showing that the INT8-optimized model preserves reconstruction quality comparable to the baseline despite the substantial reduction in inference cost. \par

\begin{table}[!htbp]
\centering
\caption{\textbf{Effect of INT8 quantization on inference efficiency and reconstruction quality.}
We report latency, throughput, PSNR, and LPIPS for the baseline and INT8-optimized models. Bold values mark the best in each column.}
\vspace{0.5em}
\label{tab:compression-int8}
\small
\setlength{\tabcolsep}{5pt}
\begin{tabularx}{\linewidth}{@{}l
    >{\raggedleft\arraybackslash}X
    >{\raggedleft\arraybackslash}X
    >{\raggedleft\arraybackslash}X
    >{\raggedleft\arraybackslash}X@{}}
\toprule
Model &
Latency (ms) &
Throughput (qps) &
PSNR $\uparrow$ &
LPIPS $\downarrow$ \\
\midrule
Baseline
& 280.56
& 3.56
& \textbf{22.49}
& \textbf{0.400} \\
INT8-Optimized
& \textbf{79.91}
& \textbf{12.51}
& 21.20
& 0.427 \\
\bottomrule
\end{tabularx}
\end{table}

These results demonstrate that BAM can be efficiently deployed under INT8 quantization. More importantly, the gains arise from combining reduced-precision inference with a TensorRT-friendly reformulation of the physics-aware computations, resulting in substantially lower latency and higher throughput. These results are obtained without fine-tuning. Quantization-aware fine-tuning could further improve performance in the quantized regime and is left for future work.

\end{document}